\documentclass[11pt,letterpaper]{article}
\ifdefined\pdfsuppressptexinfo\pdfsuppressptexinfo=15\relax\fi
\usepackage[margin=1.05in,top=0.9in,bottom=0.95in]{geometry}
\usepackage[T1]{fontenc}
\usepackage{lmodern}
\usepackage{natbib}
\setcitestyle{authoryear,round,citesep={;},aysep={,},yysep={;}}
\usepackage{amsmath,amssymb,amsthm,mathtools,bm}
\usepackage{booktabs,tabularx,array,multirow,longtable}
\usepackage{float}
\usepackage{wrapfig}
\usepackage{placeins}
\usepackage{microtype}
\usepackage{enumitem}
\usepackage{xcolor}
\usepackage{graphicx}
\usepackage{etoolbox}
\usepackage{tikz}
\usetikzlibrary{arrows.meta,backgrounds,calc,fit,positioning}
\usepackage{hyperref}
\usepackage{url}
\usepackage{aliascnt}
\usepackage[capitalize,noabbrev]{cleveref}

\definecolor{graphblue}{RGB}{34,103,172}
\definecolor{pathorange}{RGB}{225,116,45}
\definecolor{pairgreen}{RGB}{45,138,105}
\definecolor{pairpurple}{RGB}{119,83,164}
\definecolor{queryred}{RGB}{196,57,62}
\definecolor{softgray}{RGB}{245,246,248}
\hypersetup{hidelinks,
  pdftitle={StrikeTwice: Reusable Graph Representations for Path Prediction},
  pdfauthor={Gwenn Guichaoua and Elodie Laine}}

\newtheorem{theorem}{Theorem}
\newaliascnt{proposition}{theorem}
\newtheorem{proposition}[proposition]{Proposition}
\aliascntresetthe{proposition}
\newaliascnt{lemma}{theorem}
\newtheorem{lemma}[lemma]{Lemma}
\aliascntresetthe{lemma}
\newaliascnt{corollary}{theorem}
\newtheorem{corollary}[corollary]{Corollary}
\aliascntresetthe{corollary}
\theoremstyle{definition}
\newaliascnt{definition}{theorem}
\newtheorem{definition}[definition]{Definition}
\aliascntresetthe{definition}
\newaliascnt{assumption}{theorem}

\aliascntresetthe{assumption}
\theoremstyle{remark}
\newaliascnt{remark}{theorem}
\newtheorem{remark}[remark]{Remark}
\aliascntresetthe{remark}
\crefname{proposition}{Proposition}{Propositions}
\crefname{lemma}{Lemma}{Lemmas}
\crefname{corollary}{Corollary}{Corollaries}
\crefname{definition}{Definition}{Definitions}
\crefname{assumption}{Assumption}{Assumptions}
\crefname{remark}{Remark}{Remarks}

\newcommand{\R}{\mathbb R}
\newcommand{\one}{\mathbf 1}

\newcommand{\MLP}{\operatorname{MLP}}
\newcommand{\Att}{\operatorname{Att}}
\newcommand{\TriAtt}{\operatorname{TriAtt}}
\newcommand{\AlMLP}{\operatorname{AlMLP}}
\newcommand{\EVT}{\operatorname{EVT}}
\newcommand{\ET}{\operatorname{ET}}
\newcommand{\GraphEnc}{\mathrm{GraphEnc}}
\newcommand{\PathProc}{\mathrm{PathProc}}
\newcommand{\PathGraphNet}{\mathrm{PathGraphNet}}
\newcommand{\Aff}{\mathrm{Aff}}
\newcommand{\Sub}{\mathrm{Sub}}
\newcommand{\Mark}{\mathrm{Mark}}
\newcommand{\STarch}{\mathrm{ST}}

\newcommand{\PMarch}{\mathrm{PM}}
\newcommand{\PMEVT}{\texorpdfstring{\ensuremath{\PMarch(\EVT)}}{PM(EVT)}}
\newcommand{\STEE}{\ensuremath{\mathrm{ST}[\EVT,\EVT]}}

\newcommand{\Norm}{\operatorname{Norm}}
\newcommand{\LN}{\operatorname{LN}}
\newcommand{\RMSNorm}{\operatorname{RMSNorm}}

\newcommand{\Flat}{\operatorname{flat}}
\newcommand{\Unflat}{\operatorname{unflat}}
\newcommand{\Ker}{\operatorname{Ker}}
\newcommand{\concat}{\mathbin\vert}
\newcommand{\multiset}[1]{\left\{\!\left\{#1\right\}\!\right\}}
\newcommand{\cstar}{c^{\star}}
\newcolumntype{Y}{>{\raggedright\arraybackslash}X}

\setlist[itemize]{leftmargin=*,topsep=2pt,itemsep=1pt,parsep=0pt}
\setlist[enumerate]{leftmargin=*,topsep=2pt,itemsep=1pt,parsep=0pt}

\title{StrikeTwice: Reusable Graph Representations\\for Path Prediction}
\author{%
  Gwenn Guichaoua\textsuperscript{1}
  \enspace\textperiodcentered\enspace
  Elodie Laine\textsuperscript{1,2}\\[4pt]
  {\normalfont\small\textsuperscript{1} Department of Computational, Quantitative, and Synthetic Biology (CQSB), UMR 7238}\\ {\normalfont\small IBPS, Sorbonne Université, CNRS, Paris, 75005, France}\\
  {\normalfont\small\textsuperscript{2} Institut Universitaire de France (IUF)}
}

\date{}
\begin{document}
\raggedbottom
\maketitle
\begin{abstract}
Graph--path learning predicts properties of ordered paths in shared graphs, with applications in network routing and biology.
In this article, we advocate for caching graph representations to amortize their encoding cost across many path queries.
We introduce the Edge–Vertex Transformer (EVT), a graph encoder that couples vertex and pair states through configurable interactions and uses direction-aware triangular attention to compose pairwise relations. For graph--path learning, we introduce StrikeTwice: a modular architecture connecting reusable graph encodings to path processors through ordered selection of vertex and pair states, preserving path order and repeated visits.
Our main theoretical contribution is relative universality for EVT in graph learning and for StrikeTwice with EVT at both stages in graph–path learning. 
We show that, on compact domains, EVT uniformly approximates exactly the continuous vertex- and pair-level predictions determined by their stable bidirectional
two-dimensional folklore Weisfeiler--Leman (2-FWL) colors.
With EVT at both stages, StrikeTwice uniformly approximates, on compact
domains, exactly the continuous predictions for path occurrences and their
ordered pairs that are determined by graph size, path length, and the
ordered table of pair colors along the path.
For a given target, compact domain, and accuracy, one parameter set
serves all admissible graph and path sizes.
At the level of stable \(2\)-FWL profiles, reading each graph vertex's
relations to all path positions jointly gives strictly greater
distinguishability than ordered selection. Marking the path positions before
graph encoding gives a further strict increase.
To study our architectural contributions in light of our theoretical results, 
we build a coherent benchmarking suite combining controlled synthetic tasks, ThorAxe-Combo, a curated biological benchmark of splicing tasks, and temporal routing measurements from RIPE Atlas. Through controlled ablations and real-data
comparisons, this suite enables a systematic study of our StrikeTwice family of modular two-stage architectures. We examine how the graph encoder, path processor and query interface affect predictive performance at finite model capacity and training budgets. We also measure their computational costs and the savings from reusing graph encodings across queries, including the initial encoding cost.

\end{abstract}

\section{Introduction}
\label{sec:introduction}

Many prediction problems concern an ordered path within a larger graph. Internet measurements attach delays to the routes that packets follow through the network
\citep{ripencc2015atlas,ripeatlasmeshes}. In genomics, pangenome graphs represent each individual genome as a path \citep{garrison2018variation,liao2023pangenome}, and evolutionary splicing graphs represent each protein isoform as a path through exon segments aligned across species \citep{zea2021thoraxe}. Because splicing choices at distant genomic positions are often coordinated \citep{tilgner2015comprehensive,santos2026isoform}, whether an isoform is produced can depend on its full combination of exon segments. Each choice is also defined by the alternatives the isoform skips, which lie outside its path. Learning should therefore
exploit both the relations along each path and the graph context outside it. Models must also preserve path order and repeated visits, and be invariant to vertex relabeling, and, since many paths share a graph, answer many queries efficiently.

In this article, we study graph--path learning: predicting properties of ordered paths in a collection of graphs, with multiple path queries per graph.
Predictions may depend on both the path and its surrounding graph context. Encoding the whole graph separately for every query can dominate inference cost, making reuse central to scalability.
This raises four questions:  
How can graph encodings be reused while preserving path order and repeated visits?
What information remains accessible after selection from reusable graph states, through ordered selection or joint reading of the same cache,  and what does marking the path before encoding add?
Which of these relations do finite models learn to exploit? 
How do architectural choices affect predictive performance, and when does graph reuse reduce inference cost?
We address these questions by introducing the Edge--Vertex Transformer (EVT), a configurable vertex–pair graph encoder, and StrikeTwice (ST), a modular architecture for reusing such encodings across ordered path queries (\cref{fig:query-interfaces}b,c).

\paragraph{Graph encoders and expressivity.}
For graph-to-path prediction, the graph encoder determines how graph context is represented for subsequent path processing.
Message-passing Graph Neural Networks (GNNs) aggregate neighborhood information, with edge-aware variants such as GINE incorporating edge attributes.
Their distinguishing power is bounded by edge-attributed 1-dimensional Weisfeiler-Leman (1-WL) refinement \citep{xu2019gin,morris2019weisfeiler,hu2019strategies}.
To exchange information beyond local neighborhoods, graph transformers such as Graphormer \citep{ying2021graphormer} and GraphGPS \citep{rampasek2022graphgps} combine
global attention with structural or positional encodings that can overcome some limitations of 1-WL, with expressivity depending on the chosen encodings.
A complementary approach represents vertex pairs explicitly, allowing relations to be composed through an intermediate vertex:
PPGN composes pair states through matrix multiplication \citep{maron2019ppgn}, while Edge Transformers (ET) use triangular attention \citep{bergen2021}.
AlphaFold also uses related triangle-attention and multiplicative updates within its residue-pair representation \citep{jumper2021alphafold}.
With suitable initialization, ET matches two-dimensional folklore Weisfeiler--Leman (2-FWL) refinement, but this guarantee is established at a fixed graph size \citep{muller2024}, whereas a trained model is applied across graphs of varying size. Moreover, ET maintains a single tensor of pair states, whose diagonal can be used for vertex readout. The standard ET layer applies the same update operators to diagonal and off-diagonal states, limiting how independently vertex and pair computations can be configured.
Approximation theory further connects a model family's distinguishing power to the continuous functions it can uniformly approximate on compact domains \citep{azizian2021}.
Our configurable EVT (\cref{sec:architecture}) addresses these points. It extends ET with explicit vertex states and triangular pair attention in four orientations, covering every way of composing two pairs through an intermediate vertex. Separate streams allow vertex attention and pair composition to be configured independently or jointly.
We characterize the information available at individual vertex and pair outputs and the resulting equivariant approximation class, with a single set of parameters shared across all graphs up to a bounded size.

\paragraph{Graph-to-path models and reusable encodings.}
Graph-to-path models differ in when the path enters the computation.
Query-conditioned encoders (\cref{fig:query-interfaces}a) recompute graph representations for each query path. As a consequence, the encoder can attend to the targets but must be rerun per query.
Target-labeling methods follow this principle by marking target vertices or ordered structures before graph updates
\citep{wang2025labeling,zhou2023relational,xu2024higherorder}.
Query-independent encoders (\cref{fig:query-interfaces}b) instead encode the graph once and reuse this encoding across all query paths; only the per-path computation is repeated.
For vertex-pair and higher-order queries, SUREL reuses precomputed random
walks \citep{yin2022surel}, and SUREL+ reuses vertex sets with structural
features \citep{yin2023surelplus}.
Graph encoding followed by path processing appears in GNN--Transformer architectures:
START encodes road segments with graph attention, then runs a time-aware Transformer over their trajectory sequence 
\citep{jiang2023start}.
SP4LP applies a GNN followed by a Transformer or recurrent encoder to shortest paths for link prediction \citep{ferrini2025sp4lp}.
Other models use recurrent path encoders. GTS combines graph convolutions over road-network neighborhoods
with a Long Short-Term Memory (LSTM) over visited locations \citep{zhou2023gts}.
STDGNN encodes intersections and road segments through coupled spatio-temporal graph convolutions, then models their
route sequences with separate Gated Recurrent Units (GRUs) \citep{jin2024stdgnn}.
MM-STGED instead treats the trajectory itself as a complete graph, jointly updating vertex and pair states initialized from spatial and temporal features, and incorporating road-network context for trajectory recovery
\citep{wei2024mmstged}.
None of these works characterizes exactly which graph information a path model can still access once the graph encoding is restricted to the path.
StrikeTwice formalizes this restriction. It first computes vertex and pair states on the full graph. It then gathers the states of the query path’s vertices and of the pairs they form into a complete graph with one vertex per path position, on which a second network, the path processor, operates (\cref{sec:architecture}).
Because these states were computed on the full graph, they can encode context from vertices outside the path.
For StrikeTwice with EVT at both stages, we characterize the information the path processor can still access through the selected states. %
Furthermore, we establish that, for stable \(2\)-FWL profiles, a joint reader of the same cache can distinguish
inputs with identical selected path states by combining each graph vertex's
relations to all path positions. Marking the path occurrences before graph
encoding can distinguish inputs with identical joint profiles.

\paragraph{Benchmarks for graph--path prediction and reuse.}
Existing benchmarks evaluate graph–path models through end-task performance.
START tests the usefulness of trajectory representations for travel-time estimation, classification, and similarity search \citep{jiang2023start}.
SP4LP evaluates whether shortest-path representations improve link prediction \citep{ferrini2025sp4lp}, 
while RouteNet predicts network
performance across topologies, routing configurations,
and traffic conditions \citep{rusek2019routenet}.
Gretel evaluates the prediction of path continuations \citep{cordonnier2019gretel}, 
and LRGB tests graph learning on tasks requiring long-range interactions \citep{dwivedi2022lrgb}.
Most of these benchmarks measure predictive performance without controlling which information a task requires. None of them systematically isolates whether a model succeeds by composing relations along the path or by using information outside it.
We introduce controlled tasks that  separate the need to compose relations
along the query path from the need to access information outside it.
Comparing the same trained model with and without graph reuse then measures the computational benefit of reuse.
Communication networks and comparative biology provide two applications with multiple path queries per graph
across a collection of graphs.
For biology, ThorAxe constructs gene-specific evolutionary splicing graphs whose vertices represent exonic segments aligned between species and whose paths represent protein isoforms
\citep{zea2021thoraxe}.
ThorAxe-Combo turns these graphs into a path classification task. We recombine alternative blocks, {\it i.e.} subpaths of consecutive vertices and edges,  from two isoforms observed outside the held-out clade to form candidate paths absent from the visible annotations. Each alternative block must be observed in the held-out clade. The task is to predict whether the complete path is itself annotated in the clade: since every block is individually observed, the answer depends on how they combine. Evaluation uses genes held out from training. 
For Internet routing, we curate RIPE Atlas measurements into routing graphs with many observed paths and round-trip-time
targets \citep{ripencc2015atlas,ripeatlasmeshes}.
The task is to predict latency along these paths at later dates, with the monthly network topology available.

\paragraph{Contributions.}
Our contributions connect architecture, expressivity, and empirical design:
\begin{enumerate}
\item \textbf{EVT: configurable vertex--pair encoding.}
We introduce the Edge--Vertex Transformer (EVT), which couples
explicit vertex and pair states with independently configurable
vertex attention and triangular pair composition
(\cref{sec:architecture}).
Using stable \(2\)-FWL refinement, we characterize distinguishability
at individual vertex and pair outputs and the resulting equivariant
uniform approximation class on compact domains. 
Unlike existing guarantees for Edge Transformers (ET)\citep{muller2024}, which hold at a fixed graph size, ours use one parameter set shared across all graphs up to a bounded size
(\cref{thm:evt-expressivity,prop:evt-et}).

\item \textbf{StrikeTwice: reusable graph-to-path learning.}
We introduce a modular two-stage architecture in which
a graph encoder computes vertex and pair states once per graph and a path processor operates on the states selected along each query path, preserving path order and repeated visits
(\cref{sec:architecture}).
For StrikeTwice with EVT at both stages, a path-indexed
\(2\)-FWL profile exactly characterizes which inputs the model cannot distinguish and which continuous predictions it can uniformly approximate
on compact domains (\cref{thm:main-expressivity}), with 
one parameter set shared across graphs
with \(n\leq N\) vertices and paths with \(k\leq K\) vertex occurrences.
We establish a strict hierarchy of stable access profiles for selection,
joint reading, and marked encoding. Disabling path-stage triangular
attention preserves the selected-profile approximation class when the
other EVT operations and occurrence-position codes are retained
(\cref{sec:theory}). This separates information limits from finite-resource
operator choices.

\item \textbf{Controlled and application benchmarks.}
We provide synthetic tasks that separate the need to compose relations along the query path from the need to access information outside it. Task~1 tests triangular attention for path-edge composition, Task~2 tests graph-context encoding, and the Latin-square controls
test whether finite models learn distinctions predicted by the theory.
From existing measurements and annotations, we curate
two complementary application benchmarks with many path queries per graph. 
On held-out genes, ThorAxe-Combo tests whether a complete candidate path matches an annotated isoform in a reserved clade, given that each of its alternative blocks is observed
there. RIPE Atlas tests temporal latency prediction along observed
Internet routes at later dates.
The comparisons reveal task- and stage-dependent benefits of attention.
Inference measurements on synthetic, ThorAxe-Combo, and Atlas workloads
quantify the gains from graph-cache reuse, holding model
weights and physical batch sizes fixed and including initial
cache construction (\cref{sec:experiments}).

\end{enumerate}

\section{EVT and Graph-to-Path Models}
\label{sec:architecture}

We first define graphs and path queries and describe the EVT encoder
(\cref{fig:query-interfaces}c). We then present StrikeTwice
(\cref{fig:query-interfaces}b) and compare its ordered selection with a joint
reading of the same graph cache. PathMarked (\cref{fig:query-interfaces}a), which marks the path occurrences before encoding the graph for each query, 
serves as a query-conditioned reference for both forms of cache access.

\begin{figure}[t]
\centering
\resizebox{\linewidth}{!}{\begin{tikzpicture}[
  x=1cm,y=1cm,font=\fontsize{9}{10.5}\selectfont,
  box/.style={draw=black!40,rounded corners=2pt,fill=white,
    align=center,inner xsep=3pt,inner ysep=2pt,minimum height=.62cm},
  graphstage/.style={box,draw=graphblue!75,fill=graphblue!7},
  selection/.style={box,draw=pairpurple!85,fill=pairpurple!9},
  pathstage/.style={box,draw=red!65!black,fill=red!5},
  head/.style={box,draw=black!55,fill=black!9,text width=1.35cm,
    font=\fontsize{6.7}{7.7}\selectfont},
  input/.style={inner sep=0pt},
  heading/.style={anchor=west,inner sep=0pt,
    font=\fontsize{8.8}{10}\selectfont},
  blocklabel/.style={anchor=north,align=center,inner sep=0pt,
    text width=2.10cm,font=\fontsize{7}{8}\selectfont},
  flow/.style={-{Stealth[length=1.5mm]},line width=.45pt,draw=black!65,
    shorten <=1pt,shorten >=1pt},
  queryflow/.style={-{Stealth[length=1.6mm]},line width=1pt,draw=black!80,
    shorten <=3pt,shorten >=2pt},
  kvflow/.style={flow,densely dashed,draw=black!45}]

\node[heading] at (0,3.30)
  {\textbf{(a) Query-conditioned encoding} ---
   $\mathrm{PM}(\mathrm{PathGraphNet})$};
\node[input] (ainxy) at (.42,2.25) {$(X,Y,$};
\node[input] (ainp) at (.99,2.25) {$P$};
\node[input] at (1.12,2.25) {$)$};
\node[graphstage,text width=1.15cm,anchor=west] (mark) at (1.80,2.25)
  {$\mathrm{Mark}_P$};
\node[blocklabel,text=graphblue] at ($(mark.south)+(0,-.10)$)
  {Path marking};
\node[graphstage,text width=2.50cm,right=.65cm of mark] (marked)
  {$\mathrm{PathGraphNet}_{\theta_{\mathrm M}}$};
\node[blocklabel,text=graphblue] at ($(marked.south)+(0,-.10)$)
  {Graph encoder};
\node[selection,text width=1.15cm,right=.65cm of marked] (msub)
  {$\mathrm{Sub}_P$};
\node[blocklabel,text=pairpurple] at ($(msub.south)+(0,-.10)$)
  {Selection + tags};
\node[pathstage,text width=1.15cm,right=.65cm of msub] (mlp)
  {$\mathrm{MLP}_{\theta_{\mathrm R}}$};
\node[blocklabel,text=red!65!black] at ($(mlp.south)+(0,-.10)$)
  {Readout};
\node[input,right=.65cm of mlp] (aout) {$(\widehat X,\widehat Y)$};
\node[head,right=.65cm of aout] (ahead) {Task-specific\\head};
\node[input,font=\fontsize{7.2}{8.2}\selectfont,right=.65cm of ahead] (apred)
  {Prediction};
\draw[flow] (ainxy.south)--(.42,1.75)--(1.35,1.75)--(1.35,2.25)--(mark.west);
\foreach \a/\b in {mark/marked,marked/msub,msub/mlp,mlp/aout,aout/ahead,ahead/apred}
  \draw[flow] (\a.east)--(\b.west);
\draw[queryflow] (ainp.north)--(.99,2.91)--(msub.north |- .99,2.91)--(msub.north);
\draw[queryflow,shorten <=0pt] (mark.north |- .99,2.91)--(mark.north);

\node[heading] at (0,1.10)
  {\textbf{(b) Query-independent encoding} ---
   $\mathrm{ST}[\mathrm{GraphEnc},\mathrm{PathProc}]$};
\node[input] (binxy) at (.42,.05) {$(X,Y,$};
\node[input] (binp) at (.99,.05) {$P$};
\node[input] at (1.12,.05) {$)$};
\node[graphstage,text width=1.80cm,anchor=west] (graph) at (1.80,.05)
  {$\mathrm{GraphEnc}_{\theta_{\mathrm G}}$};
\node[blocklabel,text=graphblue] at ($(graph.south)+(0,-.10)$)
  {Graph encoder};
\node[graphstage,draw=none,fill=none,text=black,
  text width=1.20cm,right=.42cm of graph] (cache)
  {};
\node[input] (cachestates) at (cache.center) {$(\widetilde X,\widetilde Y)$};
\node[blocklabel,text=graphblue] at ($(cache.south)+(0,-.10)$)
  {Reusable graph cache};
\node[selection,text width=1.10cm,right=.42cm of cache] (sub)
  {$\mathrm{Sub}_P$};
\node[blocklabel,text=pairpurple] at ($(sub.south)+(0,-.10)$)
  {Selection + tags};
\node[pathstage,text width=.90cm,right=.42cm of sub] (aff)
  {$\mathrm{Aff}_{\theta_{\mathrm{aff}}}$};
\node[blocklabel,text=red!65!black] at ($(aff.south)+(0,-.10)$)
  {Feature\\projection};
\node[pathstage,text width=1.70cm,right=.42cm of aff] (path)
  {$\mathrm{PathProc}_{\theta_{\mathrm P}}$};
\node[blocklabel,text=red!65!black] at ($(path.south)+(0,-.10)$)
  {Path processor};
\node[input,right=.42cm of path] (bout) {$(\widehat X,\widehat Y)$};
\node[head,right=.42cm of bout] (bhead) {Task-specific\\head};
\node[input,font=\fontsize{7.2}{8.2}\selectfont,right=.42cm of bhead] (bpred)
  {Prediction};
\draw[flow] (binxy.south)--(.42,-.45)--(1.35,-.45)--(1.35,.05)--(graph.west);
\foreach \a/\b in {graph/cachestates,cachestates/sub,sub/aff,aff/path,path/bout,bout/bhead,bhead/bpred}
  \draw[flow] (\a.east)--(\b.west);
\draw[queryflow] (binp.north)--(.99,.71)--(sub.north |- .99,.71)--(sub.north);

\node[heading] at (0,-1.12) {\textbf{(c) One EVT layer}};
\filldraw[rounded corners=3pt,fill=green!6,draw=green!25!black!25]
  (1.38,-3.30) rectangle (7.12,-1.47);
\filldraw[rounded corners=3pt,fill=pathorange!7,draw=pathorange!25]
  (7.62,-3.30) rectangle (14.14,-1.47);
\node[font=\fontsize{8}{9}\selectfont,text=green!35!black]
  at (4.25,-1.66) {Self-attention stage};
\node[font=\fontsize{8}{9}\selectfont,text=pathorange!80!black]
  at (10.88,-1.66) {Cross-attention stage};
\node[input] (x) at (.45,-2.08) {$X^\ell$};
\node[input] (y) at (.45,-2.86) {$Y^\ell$};
\node[box,text width=1.25cm,minimum height=.56cm] (va) at (2.95,-2.08)
  {$\mathrm{Att}$};
\node[box,text width=1.25cm,minimum height=.56cm] (pa) at (2.95,-2.86)
  {$\mathrm{TriAtt}$};
\node[box,text width=1.55cm,minimum height=1.16cm] (mix1)
  at (6.05,-2.47) {$\mathrm{AlMLP}_{\theta_1}$};
\node[box,text width=1.55cm,minimum height=.56cm] (vx) at (9.16,-2.08)
  {$\mathrm{Att}_{\mathcal Y\to\mathcal X}$};
\node[box,text width=1.55cm,minimum height=.56cm] (py) at (9.16,-2.86)
  {$\mathrm{Att}_{\mathcal X\to\mathcal Y}$};
\node[box,text width=1.55cm,minimum height=1.16cm] (mix2)
  at (12.88,-2.47) {$\mathrm{AlMLP}_{\theta_2}$};
\node[input] (xo) at (15.65,-2.08) {$X^{\ell+1}$};
\node[input] (yo) at (15.65,-2.86) {$Y^{\ell+1}$};
\foreach \a/\b in {x/va,y/pa}
  \draw[flow] (\a.east)--(\b.west);
\foreach \a in {va,pa}
  \draw[flow] (\a.east)--(mix1.west |- \a.east);
\coordinate (hx) at (mix1.east |- vx.west);
\coordinate (hy) at (mix1.east |- py.west);
\draw[flow] (hx)--(vx.west);
\draw[flow] (hy)--(py.west);
\draw[kvflow] (hx)--(py.west);
\draw[kvflow] (hy)--(vx.west);
\foreach \a in {vx,py}
  \draw[flow] (\a.east)--(mix2.west |- \a.east);
\foreach \b in {xo,yo}
  \draw[flow] (mix2.east |- \b.west)--(\b.west);
\end{tikzpicture}}
\input{figures/figure1-caption.tex}
\label{fig:query-interfaces}
\end{figure}

\paragraph{Graphs and path queries.}
A graph has vertex set \([n]=\{1,\ldots,n\}\), vertex features
\(X=(x_i)\in\R^{n\times d_{\mathcal X}^{\mathrm{in}}}\), and ordered-pair features
\(Y=(y_{ij})\in\R^{n\times n\times d_{\mathcal Y}^{\mathrm{in}}}\).
The tensor \(Y\) includes diagonal and nonadjacent pairs, encoding
adjacency and any pair attributes.
A query path is an ordered sequence \(P=(p_1,\ldots,p_k)\in[n]^k\),
with repetitions allowed: \(s\in[k]\) indexes an occurrence and \(p_s\)
its graph vertex. The architecture accepts any such sequence;
applications may require consecutive vertices to be adjacent.
Further conventions appear in \cref{app:notation,app:domains}.

\paragraph{A flexible graph neural network: Edge--Vertex Transformer (EVT).}
EVT maintains a state for each vertex and for each ordered pair of vertices. 
It combines triangular attention among pairs, self-attention among vertices and cross-attention between vertex and pair states,
in a permutation-equivariant architecture (\cref{fig:query-interfaces}c). Updates within vertex and pair
states and exchanges between them are separate modules, such that each
mechanism  can be enabled independently, as explored in our benchmarks.
The EVT core is a stack of residual layers that preserves state dimensions:
\begin{equation}
 \EVT_{\theta_{\mathrm{EVT}}}:
 \R^{n\times d_{\mathcal{X}}}\times\R^{n\times n\times d_{\mathcal{Y}}}
 \longrightarrow\R^{n\times d_{\mathcal{X}}}\times\R^{n\times n\times d_{\mathcal{Y}}}.
 \label{eq:evt-core}
\end{equation}
An input affine layer maps raw vertex and pair features to these dimensions; it can be the identity when the widths already match. In the architecture descriptions below, \((X,Y)\) denotes the
features after this input layer.
For graph-to-graph prediction, the model studied in \cref{sec:evt-expressivity} uses a separate affine output map
to produce vertex and pair predictions (\cref{eq:tokenwise-affine}).
In StrikeTwice, the graph encoder instead returns reusable states; a task-specific head makes predictions after path processing. We keep these affine layer maps outside the core because they adapt feature widths to the data and the prediction task, whereas the residual layers preserve the internal state widths, and \cref{sec:evt-expressivity} specifies their theoretical roles. 

EVT is permutation equivariant.
For a vertex permutation \(\sigma\), let
\((\sigma\cdot X)_{\sigma(i)}=x_i\) and
\((\sigma\cdot Y)_{\sigma(i)\sigma(j)}=y_{ij}\).
Equivariance means
\(\EVT_{\theta_{\mathrm{EVT}}}(\sigma\cdot X,\sigma\cdot Y)
=\sigma\cdot\EVT_{\theta_{\mathrm{EVT}}}(X,Y)\), with the same action on the output vertex and pair tensors.

All internal attention and MLP updates are residual and pre-normalized.
Attention normalizes each token; aligned MLPs normalize their complete concatenated inputs.
We fix either LayerNorm or RMSNorm \citep{zhang2019rmsnorm} throughout each model. While our theory covers both choices, we use LayerNorm in
the EVT experiments. Normalization and residual additions are omitted below for readability.
Each EVT layer has two stages (\cref{fig:query-interfaces}c).

\emph{Self-attention stage.}
Vertices attend to one another and pairs undergo triangular attention in parallel.
Define \(y^\rightarrow_{ij}:=y_{ij}\) and
\(y^\leftarrow_{ij}:=y_{ji}\). A triangular head \(h\) with orientation
\((\alpha,\beta)\in\{\leftarrow,\rightarrow\}^2\) computes
\begin{equation}
 \TriAtt_h^{\alpha,\beta}(Y)_{ij}
 =\sum_{r=1}^n\operatorname{softmax}_r\!\left(
 \frac{\langle W_Qy^\alpha_{ir},W_Ky^\beta_{rj}\rangle}{\sqrt{d_a}}
 \right)(W_Ly^\alpha_{ir}\odot W_Ry^\beta_{rj}),
 \label{eq:oriented-triatt-main}
\end{equation}
where \(\odot\) is the element-wise product, \(d_a\) is the query/key width,
\(W_Q,W_K\in\R^{d_a\times d_{\mathcal Y}}\), and
\(W_L,W_R\in\R^{d_v\times d_{\mathcal Y}}\) are learned projections.
Each head returns a \(d_v\)-vector; concatenation and a learned output
projection return the pair width \(d_{\mathcal Y}\) before residual addition.
We denote this multi-head increment by \(\TriAtt\); head indices on the
projection matrices are omitted here.
Each head has its own projections and softmax
over \(r\), and its orientation is chosen independently of the other heads.
For intuition, ignore normalization and interpret matrix products channelwise.
With \(W_Q=0\) and \(W_L=W_R=\mathrm{Id}\), the four
orientations \((\rightarrow,\rightarrow)\), \((\rightarrow,\leftarrow)\),
\((\leftarrow,\rightarrow)\), and \((\leftarrow,\leftarrow)\) give
\(Y^2/n\), \(YY^\top/n\), \(Y^\top Y/n\), and \((Y^\top)^2/n\), respectively.
The first is the chain composition of ET
\citep{bergen2021}; the middle two are the bilinear cores of AlphaFold's
triangle multiplication \citep{jumper2021alphafold}.

Aligned MLPs then exchange information between the streams:
\begin{equation}
 \AlMLP(X,Y):=\left(
 (\MLP_{\mathcal X}(x_i,y_{ii}))_i,\,
 (\MLP_{\mathcal Y}(y_{ij},x_i,x_j;\one_{\{i=j\}}))_{ij}
 \right).
 \label{eq:almlp}
\end{equation}
Arguments are concatenated. The two GELU MLPs have separate
parameters, shared over vertices and ordered pairs, respectively.
Their input/output widths are
\(d_{\mathcal X}+d_{\mathcal Y}\to d_{\mathcal X}\) for vertices and
\(d_{\mathcal Y}+2d_{\mathcal X}+1\to d_{\mathcal Y}\) for pairs.
Each MLP has two affine layers.
The pair MLP does not directly receive \(y_{ji}\);
oriented triangular attention provides access to reversed pair states.

\emph{Cross-attention stage.}
Each ordered pair attends to all vertices. For one head, its increment is
\begin{equation}
 (\Att_{\mathcal{X}\to\mathcal{Y},h}(X,Y))_{ij}
 =\sum_{r=1}^n\operatorname{softmax}_r\!\left(
 \frac{\langle W_Qy_{ij},W_Kx_r\rangle}{\sqrt{d_a}}\right)W_Vx_r.
 \label{eq:cross-att-main}
\end{equation}
Thus the arrow denotes information flow, from vertex values to a pair update.
Here \(W_Q\in\R^{d_a\times d_{\mathcal Y}}\),
\(W_K\in\R^{d_a\times d_{\mathcal X}}\), and
\(W_V\in\R^{d_v\times d_{\mathcal X}}\); the concatenated heads are
projected to the receiver width \(d_{\mathcal Y}\).
The reverse update uses each \(x_i\) as a query and all \(n^2\)
pair states as keys and values. Both directions read the same preceding
states, then a second \(\AlMLP\) updates the two streams.
Flattening only enumerates pair tokens and adds no ordering information.
The complete definitions appear in \cref{eq:y-to-x,eq:x-to-y,eq:attention-detail}.

\paragraph{Reusable graph-to-path learning: StrikeTwice.}
StrikeTwice separates graph encoding from query-specific path processing.
A graph encoder computes vertex and pair states independently of the
queried path. Ordered selection then extracts states for the path
occurrences and their pairs, which a second network processes
(\cref{fig:query-interfaces}b).
We write \(\STarch[\GraphEnc,\PathProc]\), where the graph encoder
\(\GraphEnc\) and path processor \(\PathProc\) are permutation-equivariant
networks returning vertex and pair states. Their feature widths are
\(d_{\mathcal X},d_{\mathcal Y}\) and
\(q_{\mathcal X},q_{\mathcal Y}\), respectively.
Each stage preserves its own widths; an intermediate affine layer
changes dimensions between them.

The first network computes query-independent vertex and pair states,
which preserve the widths \(d_{\mathcal X},d_{\mathcal Y}\):
\begin{equation}
 (\widetilde X,\widetilde Y)=\GraphEnc_{\theta_{\mathrm G}}(X,Y).
 \label{eq:graph-stage}
\end{equation}
At inference, they can be cached across path queries while the graph
inputs and encoder weights remain fixed. Changing either requires
rebuilding the cache.

Next, for a path \(P=(p_1,\ldots,p_k)\), the path-selection stage
\(\Aff\circ\Sub_P\) first selects and tags the corresponding
vertex and pair states, then adapts their feature dimensions
for the path processor.
Let \(\phi:\mathbb N_{>0}\to\R^{d_\phi}\) be a fixed positional encoding,
injective on the supported positions, and let \(\concat\) denote
concatenation. Define
\begin{equation}
 \Sub_P(\widetilde X,\widetilde Y):=\left(
 (\widetilde x_{p_s}\concat\phi(s))_s,\,
 (\widetilde y_{p_sp_t}\concat\phi(s)\concat\phi(t)
       \concat\one_{\{p_s=p_t\}})_{s,t}\right).
 \label{eq:sub-operator}
\end{equation}
This constructs a complete occurrence graph with one vertex per
path position and all \(k^2\) ordered pairs, including self-pairs
and nonconsecutive occurrences. Repeated visits receive distinct position codes,
while the equality tag identifies occurrences of the same graph vertex.
The selected states may already contain information from outside
the path, computed during graph encoding.

An affine interface then adapts the tagged states to the path
processor's widths. For vertex--pair states
\((U,V)=((u_i)_i,(v_{ij})_{ij})\), define
\begin{equation}
 \Aff_{\theta_{\mathrm{aff}}}(U,V)
 :=\left((A_{\mathcal X}u_i+b_{\mathcal X})_i,\,
         (A_{\mathcal Y}v_{ij}+b_{\mathcal Y})_{ij}\right),
 \label{eq:tokenwise-affine}
\end{equation}
where \(\theta_{\mathrm{aff}}\) collects the learned projection
parameters. The vertex and pair maps have independent parameters,
shared over their respective indices.
They adapt the tagged features to the path-processor widths
\(q_{\mathcal X},q_{\mathcal Y}\).
This allows a smaller path processor or, at sufficient width,
an information-preserving embedding of the selected states.
Detailed dimensions are summarized in
\cref{tab:interface-dimensions}.

The path processor updates the resulting occurrence graph.
With
\(\theta_{\mathrm{ST}}=
(\theta_{\mathrm G},\theta_{\mathrm{aff}},\theta_{\mathrm P})\),
the complete architecture is
\begin{equation}
 (\widehat X,\widehat Y)
 =\STarch_{\theta_{\mathrm{ST}}}(X,Y,P)
 =\bigl(\PathProc_{\theta_{\mathrm P}}
 \circ\Aff_{\theta_{\mathrm{aff}}}
 \circ\Sub_P
 \circ\GraphEnc_{\theta_{\mathrm G}}\bigr)(X,Y).
 \label{eq:path-stage}
\end{equation}
Its outputs
\(\widehat X\in\R^{k\times q_{\mathcal X}}\) and
\(\widehat Y\in\R^{k\times k\times q_{\mathcal Y}}\)
are indexed by path occurrences and occurrence pairs.
A task-specific head reads vertex states, pair states, or both
to produce the required prediction; these heads are specified
in \cref{sec:experiments}.

Consistent relabeling of the graph and path leaves these
occurrence-indexed outputs unchanged. For
\((\sigma P)_s=\sigma(p_s)\),
\begin{equation}
 \STarch_{\theta_{\mathrm{ST}}}
 (\sigma\cdot X,\sigma\cdot Y,\sigma P)
 =\STarch_{\theta_{\mathrm{ST}}}(X,Y,P).
 \label{eq:striketwice-invariance}
\end{equation}
This invariance concerns graph vertex names, not path order:
occurrence positions remain fixed under graph relabeling,
and invariance to their reordering is not imposed.

EVT can instantiate both stages. Our experiments compare
\(\STarch[\EVT,\EVT]\),
\(\STarch[\EVT,\MLP]\), and
\(\STarch[\ET,\MLP]\), among other controls.
Here \(\MLP\) denotes an attention-free path processor that updates
the selected pair states, computes a mean for each row, and updates
the corresponding vertex states. We evaluate two variants of this processor in
\cref{sec:experiments}.
In our \(\ET\) configuration, vertex states are read from diagonal pair states.
The theory in \cref{sec:exact-profile} characterizes
\(\STarch[\EVT,\EVT]\), with one-hot positional encodings. It adds
affine input and output interfaces, from the raw feature widths and
to the prediction widths; the output interface takes the place of the
task-specific head.

\paragraph{Selection and joint reading of the same graph cache.}
The selector \(\Sub_P\) in \cref{eq:sub-operator} reads the ordered
pair block \(\widetilde Y[P,P]=(\widetilde y_{p_sp_t})_{s,t\in[k]}\).
Writing \(V=[n]\), we define a second access operator on the same cache:
\begin{equation}
 \Sub_P^{\mathrm{joint}}(\widetilde X,\widetilde Y)
 := \left(
   (\widetilde x_{p_s}\concat\phi(s))_{s\in[k]},\,
   (\widetilde y_{p_su}\concat\phi(s)\concat\one_{\{p_s=u\}})
      _{s\in[k],\,u\in V}
 \right).
 \label{eq:cache-pair-access}
\end{equation}
The rows follow path order, with one row per occurrence, including
repeated visits. For each graph vertex \(u\), the joint reader processes
the ordered tuple \((\widetilde y_{p_su})_{s\in[k]}\) and then pools
the resulting representations over \(u\in V\)
(\cref{sec:access-hierarchy}).

\paragraph{Path-conditioned graph learning: PathMarked.}
PathMarked introduces the query before graph encoding.
It first marks the graph with the path occurrences, then applies a
permutation-equivariant graph network \(\PathGraphNet\).
The resulting graph states depend on the supplied path and are
generally recomputed for each query. Therefore, it does not offer StrikeTwice's scalability through graph-cache reuse.

For \(P=(p_1,\ldots,p_k)\), define the occurrence mark
\(m_P(i)=\sum_{s:p_s=i}\phi(s)\), with zero for unvisited vertices.
The marking operator appends these marks to vertices and pair endpoints:
\begin{equation}
 \Mark_P(X,Y):=\left(
 (x_i\concat m_P(i))_i,\,
 (y_{ij}\concat m_P(i)\concat m_P(j))_{ij}\right).
 \label{eq:mark-operator}
\end{equation}
Injectivity of individual position codes does not alone guarantee distinct sums. The theory uses one-hot codes such that different occurrence sets have distinct marks. 
Marking both endpoints makes this information directly available to pairs. An EVT used as \(\PathGraphNet\) therefore operates at widths \(d_{\mathcal X}+d_\phi,d_{\mathcal Y}+2d_\phi\). %

After graph encoding, the same operator \(\Sub_P\) as in StrikeTwice selects vertex and pair states and appends occurrence tags. Shared MLPs then produce the occurrence-indexed outputs.
With \(\theta_{\mathrm{PM}}=(\theta_{\mathrm M},\theta_{\mathrm R})\),
the architecture is
\begin{equation}
 (\widehat X,\widehat Y)
 =\PMarch_{\theta_{\mathrm{PM}}}(X,Y,P)
 =\bigl(\MLP_{\theta_{\mathrm R}}\circ\Sub_P\circ
 \PathGraphNet_{\theta_{\mathrm M}}\circ\Mark_P\bigr)(X,Y).
 \label{eq:pathmarked-composition}
\end{equation}
For selected tagged states \((\widehat U,\widehat V)\), the readout is
\[
 \MLP_{\theta_{\mathrm R}}(\widehat U,\widehat V)
 =\left((\rho_{\mathcal{X}}(\widehat u_s))_s,\,
 (\rho_{\mathcal{Y}}(\widehat v_{st},\widehat u_s,\widehat u_t))_{s,t}\right),
\]
For an EVT core,
\(\widehat u_s\in\R^{d_{\mathcal X}+2d_\phi}\) and
\(\widehat v_{st}\in\R^{d_{\mathcal Y}+4d_\phi+1}\):
\(\Sub_P\) appends occurrence tags to the query-conditioned graph states.
The maps \(\rho_{\mathcal{X}},\rho_{\mathcal{Y}}\) are shared, biased two-layer GELU
MLPs with output widths \(q_{\mathcal{X}},q_{\mathcal{Y}}\). Unlike internal EVT updates, these terminal maps use neither normalization nor residual addition.
Occurrence tags distinguish repeated visits to the same graph vertex. The nonlinear readout allows differences between their outputs to depend on graph context, whereas a shared affine readout of the state and position code cannot.
An additional task-specific layer can then be appended.
The architecture satisfies the same graph-relabeling invariance as
StrikeTwice in \cref{eq:striketwice-invariance}.
We use PathMarked with EVT (\PMEVT{}) as our query-conditioned reference for the
expressivity comparison in \cref{sec:access-hierarchy};
the formal model and readout conventions are detailed in
\cref{app:marked-model}. 

\paragraph{Computational complexity of StrikeTwice versus PathMarked.}
For the dense EVT instances
\(\STarch[\EVT,\EVT]\) and \(\PMEVT{}\), let \(L_G,L_P,L_M\) be the graph-encoder,
path-processor and marked-network depths, and let \(d\) bound the hidden
token and total attention widths, including appended mark coordinates.
Consider one graph with at most \(N\) vertices and \(\mathcal P\geq1\)
query paths, each of length at most \(K\).
The leading dense attention costs at inference are bounded by
\[
 \STarch:\ O(L_GN^3d+\mathcal P L_PK^3d),
 \qquad
 \PMarch:\ O(\mathcal P L_MN^3d),
\]
up to feature-mixing, selection and readout costs. With fixed graph inputs
and encoder weights, StrikeTwice encodes the graph once and performs
\(\mathcal P\) path-processor computations on occurrence graphs of size at most \(K\).
PathMarked instead performs \(\mathcal P\) query-dependent full-graph
encodings. The amortized StrikeTwice attention cost per path is
\(O(L_GN^3d/\mathcal P+L_PK^3d)\): many short paths sharing a graph
favor reuse, whereas long paths can make path processing expensive.
The reusable cache stores only the final graph-encoder vertex and pair
states, requiring \(O(N^2d)\) memory independently of \(\mathcal P\).
It need not retain intermediate graph layers; attention workspace and
path-processor activations are separate memory costs.
Measured reuse appears in \cref{sec:reuse}. Optional positional and
streaming-memory extensions appear in \cref{app:evt-rope,app:flash-triatt}.

We call a model \emph{cache-reusing} when its graph encoding is
independent of the query, so that the graph states can be computed
once and shared across path queries; all \(\STarch\) variants are
cache-reusing, whereas \(\PMarch\) is not.

\section{Expressivity of EVT and StrikeTwice}
\label{sec:theory}

We characterize which inputs EVT, \(\STarch[\EVT,\EVT]\), and the
query-conditioned reference \(\PMEVT{}\) can distinguish and which
continuous predictions they can uniformly approximate. These characterizations
hold with either LayerNorm or RMSNorm throughout the internal normalization
layers. We then compare ordered selection and joint reading of the
same unmarked graph cache with path-conditioned encoding
(\cref{sec:access-hierarchy}). Two experiments on Latin-square graphs
test whether finite models learn to use path marks and to read cached
relations jointly (\cref{sec:latin-square}).

\subsection{Distinguishability and uniform approximation}
\label{sec:def_k_H}

We fix maximum graph and path sizes \(N,K\geq1\), raw input widths
\(d_{\mathcal X}^{\mathrm{in}},d_{\mathcal Y}^{\mathrm{in}}\geq1\),
and prediction widths
\(d_{\mathcal X}^{\mathrm{out}},d_{\mathcal Y}^{\mathrm{out}}\geq1\).
Graphs have \(1\leq n\leq N\) vertices, and paths have
\(1\leq k\leq K\) vertex occurrences, including repeated visits.
Each network has finite capacity and uses one parameter set across
all admissible sizes; the model families allow depth and hidden
widths to grow. 
To simplify notation, we omit the bounds \(N,K\) in the model and profile notation
(\cref{tab:notation}).

For a compact subset \(S\) of the relevant input domain
\(\mathcal D\), let \(\mathcal C(S)\) denote continuous maps with
the prescribed output shapes and feature widths.
Write \(f=(f_{\mathcal X},f_{\mathcal Y})\) for the vertex and pair
components, indexed by graph vertices for EVT and by path
occurrences for graph--path models.
We measure approximation using
\(\|f\|_{\infty,S}:=\sup_{z\in S}\|f(z)\|_{\mathrm F}\), where
\(\|f(z)\|_{\mathrm F}^2:=\|f_{\mathcal X}(z)\|_{\mathrm F}^2
+\|f_{\mathcal Y}(z)\|_{\mathrm F}^2\).
%where
%\(\|(U,V)\|_{\mathrm F}^2=\|U\|_{\mathrm F}^2+\|V\|_{\mathrm F}^2\).
For a network family \(\mathcal H\), its uniform closure
\(\overline{\mathcal H|_S}\) consists of the maps
\(f\in\mathcal C(S)\) such that, for every \(\varepsilon>0\),
there exists \(h\in\mathcal H\) satisfying
\(\|f-h|_S\|_{\infty,S}<\varepsilon\).
Formal domains, output spaces and topologies are specified in
\cref{app:domains}.

EVT is equivariant to vertex relabeling, whereas StrikeTwice and PathMarked are invariant to consistent relabeling of the graph and path.

\paragraph{Invariant graph--path predictions.}
For a family \(\mathcal H\) of graph--path networks on an input
domain \(\mathcal D\), we compare complete occurrence-indexed
outputs. Define
\(\ker(f):=\{(z,z')\in\mathcal D^2:f(z)=f(z')\}\) and
\(\Ker(\mathcal H):=\bigcap_{f\in\mathcal H}\ker(f)\).
Thus, \(\Ker(\mathcal H)\) contains exactly the input pairs that
no network in \(\mathcal H\) can distinguish. We seek a coloring profile \(\Psi\) characterizing this common kernel:
\(\Ker(\mathcal H)=\ker(\Psi)\).
The profile includes the output index-set size, and thus, equal profiles
have matching output shapes.
Any uniformly approximable target \(f\) must then satisfy
\(\ker(\Psi|_S)\subseteq\ker(f)\). This motivates the kernel-compatible class
\[
 \mathcal C_\Psi(S)
 :=\{f\in\mathcal C(S):\ker(\Psi|_S)\subseteq\ker(f)\}.
\]

\paragraph{Equivariant graph predictions.}
For graph-to-graph equivariant networks such as EVT, these definitions
must be refined to compare individual output locations. For
\(f=(f_{\mathcal X},f_{\mathcal Y})\), define the rooted evaluations
\begin{equation}
 f_{\mathcal X}^{\bullet}(X,Y;i):=f_{\mathcal X}(X,Y)_i,
 \qquad
 f_{\mathcal Y}^{\bullet}(X,Y;i,j):=f_{\mathcal Y}(X,Y)_{ij},
 \label{eq:rooted-evaluations}
\end{equation}
and the equivariant kernel
\begin{equation}
 \ker^{\mathrm{eq}}(f):=
 \bigl(\ker(f_{\mathcal X}^{\bullet}),
       \ker(f_{\mathcal Y}^{\bullet})\bigr).
 \label{eq:equivariant-kernel}
\end{equation}
Inclusions and intersections act componentwise; in particular,
\(\Ker^{\mathrm{eq}}(\mathcal H):=
\bigcap_{f\in\mathcal H}\ker^{\mathrm{eq}}(f)\).
The indices select outputs without marking the network input.
This kernel compares vertices with vertices and pairs with pairs,
including across different graphs, rather than comparing only
complete output arrays.
For a graph profile \(\Psi(X,Y)=(\psi_{ij}(X,Y))_{ij}\), use
\(\Psi_{\mathcal X}^{\bullet}(X,Y;i):=\psi_{ii}(X,Y)\) and
\(\Psi_{\mathcal Y}^{\bullet}(X,Y;i,j):=\psi_{ij}(X,Y)\) to define
\(\ker^{\mathrm{eq}}(\Psi)\) by \cref{eq:equivariant-kernel}. Then
\begin{equation}
 \mathcal C_\Psi^{\mathrm{eq}}(S)
 :=\{f\in\mathcal C(S):
 \ker^{\mathrm{eq}}(\Psi|_S)\subseteq\ker^{\mathrm{eq}}(f)\}
 \subseteq\mathcal C_\Psi(S).
 \label{eq:evt-entrywise-compatible}
\end{equation}
Equal diagonal colors must therefore give equal vertex outputs,
and equal pair colors must give equal pair outputs.
For an equivariant profile and a relabeling-stable domain \(S\),
this condition implies equivariance and additionally constrains
outputs at locations sharing the same color.
The rooted domains are formalized in \cref{app:evt-array-proof}.

\emph{Goals of the theory.} %
First, characterize distinguishing power by one coloring profile:
\(\Ker(\mathcal H)=\ker(\Psi)\) in the invariant setting, and
\(\Ker^{\mathrm{eq}}(\mathcal H)=\ker^{\mathrm{eq}}(\Psi)\) in the equivariant setting.
Second, prove that the corresponding compatibility condition is sufficient
for uniform approximation on every compact \(S\):
\(\overline{\mathcal H|_S}=\mathcal C_\Psi(S)\) for invariant graph--path predictions,
and \(\overline{\mathcal H|_S}=\mathcal C_\Psi^{\mathrm{eq}}(S)\)
for equivariant graph predictions.

\subsection{Expressivity of EVT}
\label{sec:evt-expressivity}

Set \(\mathcal G_n:=\R^{n\times d_{\mathcal{X}}^{\mathrm{in}}}\times
\R^{n\times n\times d_{\mathcal{Y}}^{\mathrm{in}}}\) and
\(\mathcal G_{\leq N}:=\bigsqcup_{n=1}^N\mathcal G_n\).
The family \(\mathcal H_{\mathrm{EVT}}\) consists of graph-shaped maps
\begin{equation}
 f_\theta
 =\Aff_{\theta_{\mathrm{out}}}\circ
   \EVT_{\theta_{\mathrm{EVT}}}\circ\Aff_{\theta_{\mathrm{in}}},
 \qquad
 \theta=(\theta_{\mathrm{in}},\theta_{\mathrm{EVT}},\theta_{\mathrm{out}}).
 \label{eq:evt-family}
\end{equation}
The EVT core and affine interfaces follow
\cref{eq:evt-core,eq:tokenwise-affine}.
Independent tokenwise affine maps adapt feature widths before
and after the core, without normalization or residual addition.
Their parameters are shared across token indices; both interfaces
and the core are permutation equivariant.
Depth, internal and MLP widths, head counts and head dimensions
may grow within the family; each model uses one parameter set
across all \(n\leq N\).

Following the folklore WL and pair-network analyses
\citep{maron2019ppgn,azizian2021,muller2024}, define bidirectionally initialized
\(2\)-FWL colors by
\begin{equation}
 c^0_{ij} =(y_{ij},y_{ji},x_i,x_j,\one_{\{i=j\}}), \qquad
 c^{r+1}_{ij} =\left(c^r_{ij}, \multiset{(c^r_{iu},c^r_{uj}):u\in[n]}\right).
\label{eq:2fwl-step}
\end{equation}
Colors are formal tuples and multisets retaining exact features and multiplicities, and, as such, they can be compared across
graphs and sizes.
Set \(c^\star_{ij}:=c^{R_N}_{ij}\), where \(R_N=2N^2\) suffices for joint stabilization on any two graphs of size at most \(N\) (\cref{lem:joint-stabilization}).
Define the graph profile
\begin{equation}
 \Psi_{\mathrm{EVT}}(X,Y):=(c^\star_{ij})_{i,j\in[n]}.
 \label{eq:stable-2fwl-map}
\end{equation}
The rooted profiles are \(c^\star_{ii}\) and \(c^\star_{ij}\). 
\cref{eq:evt-entrywise-compatible} therefore requires equal stable colors to give
equal outputs at any two vertices or any two pairs, even when the complete graph profiles differ (\cref{app:evt-array-proof}).

\begin{theorem}[Expressivity of EVT]
\label{thm:evt-expressivity}
For every compact set \(S\subseteq\mathcal G_{\leq N}\),
\[
 \Ker^{\mathrm{eq}}(\mathcal H_{\mathrm{EVT}})
   =\ker^{\mathrm{eq}}(\Psi_{\mathrm{EVT}}),
 \qquad
 \overline{\mathcal H_{\mathrm{EVT}}|_S}
   =\mathcal C^{\mathrm{eq}}_{\Psi_{\mathrm{EVT}}}(S).
\]
\end{theorem}
Each network keeps its hidden widths fixed across layers.
These widths may be chosen according to the target function
and desired approximation accuracy, but the same network
and parameters apply to all graph sizes represented in \(S\).

\begin{proof}[Proof sketch]
We view each output location as a rooted input \((X,Y;i)\) or
\((X,Y;i,j)\). Its stable color determines the colors and multisets
used by the EVT updates at that location. Equal stable colors
therefore give equal outputs for every EVT, yielding the first
kernel inclusion. Continuity and compatibility with the profile
are preserved under uniform limits, giving the first closure
inclusion.

To separate distinct colors, the auxiliary construction reproduces the
refinement rounds. A triangular head with uniform weights forms a
normalized histogram of color pairs through the same intermediate vertex.
The diagonal indicator singles out one contribution of weight \(1/n\),
allowing a shared MLP to recover the counts and encode the next color.
Iterating this update separates the two rooted inputs with a single set
of parameters, even when their graph sizes differ. The simulation argument
below preserves this separation in a normalized EVT, giving the kernel
equality.

For uniform approximation, affine input widening allows auxiliary EVT computations to run in parallel groups of coordinates. Their scalar
outputs at the same rooted location are then available to a shared
MLP, which can approximate their sum and product. Together with the
constant functions, this gives a unital algebra after uniform
closure. The separators constructed above distinguish points of the
quotient obtained by identifying rooted inputs with the same stable
color. Stone--Weierstrass then gives uniform approximation of every
continuous target compatible with the stable colors, following
\citet{azizian2021}.
The parallel construction is first carried out without normalization,
since normalizing the full state couples the coordinate groups. Constant
anchors make the normalization scale nearly constant on compact sets, and
a zero reference cancels centering in LayerNorm. This allows a fixed
linear map to recover the auxiliary states to arbitrary uniform accuracy
after RMSNorm or LayerNorm. Combining the vertex and pair computations
before simulation gives one normalized GELU network that approximates
both arrays simultaneously.
Full proofs are in \cref{app:evt-et-proof,app:evt-array-proof,app:unified-simulation}.
\end{proof}

\paragraph{Relation to Edge Transformers.}
ET embeds pair features and both endpoint features into a single pair
tensor, then alternates triangular attention and shared pair MLPs
\citep{bergen2021,muller2024}. We prove the same rooted characterization
for its endpoint-initialized, direction-mixed variant (\cref{prop:evt-et}).
The ET theorem of \citet{muller2024} establishes WL distinguishing power
at fixed graph size; our result additionally characterizes uniform
approximation for normalized triangular-attention networks, with shared
parameters across bounded sizes. Bounded-size approximation is known for
other GNN families \citep{keriven2019universal,azizian2021}; here it also extends through the ordered graph-to-path interfaces below.

\subsection{Expressivity of StrikeTwice}
\label{sec:exact-profile}

We now consider \((X,Y,P)\in\mathcal G_n\times[n]^k\) and predictions in
\(\R^{k\times d_{\mathcal X}^{\mathrm{out}}}\times
\R^{k\times k\times d_{\mathcal Y}^{\mathrm{out}}}\), invariant under
consistent graph relabeling as in \cref{eq:striketwice-invariance}.
The family \(\mathcal H_{\mathrm{ST}}\) uses
\(\STarch[\EVT,\EVT]\), with both components equal to the EVT core:
\begin{equation}
 \begin{aligned}
 f_\theta(X,Y,P)
   &=\Aff_{\theta_{\mathrm{out}}}
       \left(\STarch_{\theta_{\mathrm{ST}}}
       \bigl(\Aff_{\theta_{\mathrm{in}}}(X,Y),P\bigr)\right),\\
 \theta&=(\theta_{\mathrm{in}},\theta_{\mathrm{ST}},\theta_{\mathrm{out}}),
 \qquad
 \theta_{\mathrm{ST}}=(\theta_{\mathrm G},\theta_{\mathrm{aff}},\theta_{\mathrm P}).
 \end{aligned}
 \label{eq:striketwice-family}
\end{equation}
Here \(\STarch_{\theta_{\mathrm{ST}}}\) is
\cref{eq:path-stage}, and every affine map has the tokenwise form
\cref{eq:tokenwise-affine}. The outer input map changes only graph
features, from the raw widths to \(d_{\mathcal X},d_{\mathcal Y}\);
it leaves \(P\) unchanged. The graph encoder and its cache have these
widths. The selector \(\Sub_P\)
(\cref{eq:sub-operator}) reads the cached vertex and pair states
in path-occurrence order. With one-hot codes \(\phi(s)\in\R^K\), selection gives widths
\(d_{\mathcal X}+K\) and \(d_{\mathcal Y}+2K+1\).
The extra pair coordinate is the indicator \(\one_{\{p_s=p_t\}}\)
in \cref{eq:sub-operator}, which records whether the two path
occurrences visit the same graph vertex.
The bridge \(\Aff_{\theta_{\mathrm{aff}}}\) maps these to the independently
chosen path widths \(q_{\mathcal X},q_{\mathcal Y}\), which the path processor
preserves. Finally, \(\Aff_{\theta_{\mathrm{out}}}\) maps them to the
prescribed prediction widths. All three affine parameter blocks and
both EVT parameter blocks are independent.
The internal graph and path widths, MLP widths, depths, head counts and head dimensions are free;
one parameter set acts on all \(n\leq N,k\leq K\)
(\cref{def:striketwice-family}).

\begin{definition}[Selected path profile]
\label{def:profile}
For \(P=(p_1,\ldots,p_k)\), define
\begin{equation}
 \Psi_{\mathrm{ST}}(X,Y,P)
 :=\left(n,k,
   \Psi_{\mathrm{EVT}}(X,Y)[P,P]\right)
 =\left(n,k,(c^\star_{p_sp_t})_{s,t\in[k]}\right).
 \label{eq:path-profile}
\end{equation}
\end{definition}
This profile retains the ordered colors of all occurrence pairs, including equality between repeated visits.
\begin{theorem}[Expressivity of StrikeTwice]
\label{thm:main-expressivity}
For every compact set \(S\) of graph--path inputs,
\[
 \Ker(\mathcal H_{\mathrm{ST}})
   =\ker(\Psi_{\mathrm{ST}}),
 \qquad
 \overline{\mathcal H_{\mathrm{ST}}|_S}
   =\mathcal C_{\Psi_{\mathrm{ST}}}(S).
\]
\end{theorem}

\begin{proof}[Proof sketch]
Each graph-stage vertex or pair state depends only on its stable
color. Equal selected profiles thus give identical inputs to the
second EVT and identical outputs for every StrikeTwice model. Hence
\(\ker(\Psi_{\mathrm{ST}})\subseteq\operatorname{Ker}(\mathcal H_{\mathrm{ST}})\).
Every StrikeTwice model is continuous, and both continuity and
compatibility with the selected profile are preserved under uniform
limits. Thus
\(\overline{\mathcal H_{\mathrm{ST}}|_S}\subseteq\mathcal C_{\Psi_{\mathrm{ST}}}(S)\).

Conversely, to separate two inputs with different selected profiles,
we first work without normalization. The graph encoder EVT assigns
distinct codes to the stable colors occurring in the two inputs,
using the construction from the proof of \cref{thm:evt-expressivity}.
The path processor EVT then uses the occurrence-position codes to
gather the columns of the selected color matrix in two refinement
rounds, encoding the full profile at each output occurrence
(\cref{lem:rooted-recovery}). This
gives a separator for the two profiles. The simulation argument
below preserves this separation in a normalized StrikeTwice model,
giving the reverse inclusion and hence the kernel equality.

For uniform approximation, fix a continuous profile-compatible
target and a tolerance. By compactness, finitely many separators
suffice to distinguish every pair of inputs of the same path length
whose target outputs differ by at least that tolerance. Inputs with
identical combined separator outputs therefore have target values
within the tolerance. Each
separator can be evaluated in its own group of coordinates within
a single wider StrikeTwice model without normalization. The path
processor uses the position codes to gather the separators' outputs
at each occurrence. A shared MLP approximates a continuous readout
of these outputs, giving uniform approximation for vertex outputs.
For joint vertex--pair predictions, the same construction also
stores at each occurrence \(s\) its row of pair predictions.
The aligned pair MLP uses the position code at \(t\) to select the
\((s,t)\)-entry and write it into the corresponding pair state
(\cref{app:joint-output-extension}). The simulation theorem
transfers these approximants to normalized GELU networks
(\cref{thm:ln-gelu-simulation}), giving the reverse inclusion and
hence the uniform closure equality.

Full proofs are given in \cref{app:profile-proof,app:universality-proof}.
\end{proof}

\begin{remark}[No triangular path attention is needed for expressivity]
\label{rem:path-attention-expressivity}
Both equalities in \cref{thm:main-expressivity} continue to hold when
triangular attention is disabled at the path stage, with the other EVT
operations and occurrence-position codes retained
(\cref{cor:path-decoder-kernel}). The construction uses auxiliary
coordinates to encode the full ordered pair profile
(\cref{app:path-decoder}). Task~1 measures the benefit of triangular path
attention at the tested widths and training budgets (\cref{sec:t1}).
\end{remark}

\subsection{Expressivity of PathMarked}
\label{sec:pathmarked-expressivity}
\label{sec:pathmarked-comparison}
We use the same graph--path input/output spaces and graph-relabeling
invariance. The family \(\mathcal H_{\mathrm{PM}}\) uses
\(\PMEVT{}\), with the same external affine interfaces:
\begin{equation}
 \begin{aligned}
 f_\theta(X,Y,P)
   &=\Aff_{\theta_{\mathrm{out}}}
       \left(\PMarch_{\theta_{\mathrm{PM}}}
       \bigl(\Aff_{\theta_{\mathrm{in}}}(X,Y),P\bigr)\right),\\
 \theta&=(\theta_{\mathrm{in}},\theta_{\mathrm{PM}},\theta_{\mathrm{out}}),
 \qquad
 \theta_{\mathrm{PM}}=(\theta_{\mathrm M},\theta_{\mathrm R}).
 \end{aligned}
 \label{eq:pathmarked-family}
\end{equation}
The core is \cref{eq:pathmarked-composition}, with
\(\PathGraphNet=\EVT\), and the affine maps follow
\cref{eq:tokenwise-affine}. The input map sends the raw feature widths
to \(d_{\mathcal X},d_{\mathcal Y}\), leaving \(P\) unchanged.
The marking operator in \cref{eq:mark-operator} appends the
\(K\)-dimensional occurrence mark \(m_P(i)\) to vertex \(i\) and the
two endpoint marks \(m_P(i)\) and \(m_P(j)\) to pair \((i,j)\).
The resulting widths \(d_{\mathcal X}+K\) and \(d_{\mathcal Y}+2K\)
are preserved by the marked EVT.
After selection, the occurrence-tagged MLPs produce
\(q_{\mathcal X},q_{\mathcal Y}\), and the outer affine map sets
\(d_{\mathcal X}^{\mathrm{out}},d_{\mathcal Y}^{\mathrm{out}}\).
It may be absorbed into the final affine maps of these MLPs.
All parameter blocks are independent; internal state and MLP widths,
depths, head counts and head dimensions range over finite choices,
and the same parameters act across all bounded sizes.
The input lift can reserve constant coordinates and workspace before
marking; \cref{app:marked-model} explains how this placement retains
the marked-color approximation guarantees.
For one-hot occurrence codes \(\phi(s)\in\R^K\), recall
$ m_P(i):=\sum_{s\in[k]:p_s=i}\phi(s).$
Define marked colors on the raw input by \cref{eq:2fwl-step}, initialized as
\(
 \bar c^0_{ij}
 :=(y_{ij},y_{ji},x_i,x_j,\one_{\{i=j\}},m_P(i),m_P(j)).
\)
Writing \(\bar c^\star_{ij}:=\bar c^{R_N}_{ij}\), set
\[
 \Psi_{\mathrm{PM}}(X,Y,P)
 :=\left(n,k,(\bar c^\star_{p_sp_s})_{s\in[k]}\right).
\]
At stabilization, occurrence marks allow selected diagonal
colors to determine all selected pair colors
(\cref{lem:marked-rooted-recovery}).
The profile therefore suffices for joint vertex--pair outputs.

\begin{theorem}[Expressivity of PathMarked and comparison with StrikeTwice]
\label{thm:pathmarked-expressivity}
For every compact set \(S\) of graph--path inputs,
\[
 \Ker(\mathcal H_{\mathrm{PM}})
   =\ker(\Psi_{\mathrm{PM}}),
 \qquad
 \overline{\mathcal H_{\mathrm{PM}}|_S}
   =\mathcal C_{\Psi_{\mathrm{PM}}}(S).
\]
Moreover,
\begin{equation}
\Ker(\mathcal H_{\mathrm{PM}})
\subseteq\Ker(\mathcal H_{\mathrm{ST}}),\qquad
 \mathcal C_{\Psi_{\mathrm{ST}}}(S)
   \subseteq\mathcal C_{\Psi_{\mathrm{PM}}}(S).
\label{eq:pathmarked-selected-hierarchy}
\end{equation}
For \(N\geq16\) and \(K\geq1\), the kernel inclusion is
strict, and the approximation classes are strictly nested
on suitable finite input sets.
\end{theorem}
\begin{proof}[Proof sketch]
%The stable diagonal color \(\bar c^\star_{p_sp_s}\) contains the multiset
%of pairs \((\bar c^\star_{p_su},\bar c^\star_{up_s})\) over intermediate
%vertices \(u\). The mark \(m_P(u)\) carried by \(\bar c^\star_{p_su}\)
%identifies the entries with \(u=p_t\), such that this diagonal yields the whole
%row \((\bar c^\star_{p_sp_t})_t\).
In the multiset recovered from a selected marked diagonal, the mark
for each path position identifies its intermediate vertex. 
Each such pair color determines the diagonal at the other endpoint.
These diagonals give the full marked profile. Applying the same
step at each diagonal gives all selected marked pair colors
(\cref{lem:marked-rooted-recovery}). Since EVT states factor through
their marked colors, equal marked profiles give identical inputs
to the occurrence-tagged readouts. This gives the first kernel
inclusion. Continuity and compatibility with the marked profile
are preserved under uniform limits, giving the first closure
inclusion.

For each fixed occurrence index \(s\), the vertex prediction and its row of
pair predictions form one rooted target at \(p_s\). This target
is compatible with the marked diagonal color because that color
determines the full profile. Rooted EVT approximation and the
occurrence-tagged readouts therefore approximate every continuous
profile-compatible target. The tags select the prediction for
occurrence \(s\) and the entry for pair \((s,t)\), including when two
occurrences visit the same vertex (\cref{app:marked-approximation}).
This gives the reverse closure inclusion. On a two-input set
with distinct marked profiles, choosing a target with distinct
outputs also gives a separator, proving the reverse kernel
inclusion.

Forgetting the marks in the recovered pair colors recovers the selected
StrikeTwice profile from the marked profile. This gives the kernel
and approximation-class inclusions in
\cref{eq:pathmarked-selected-hierarchy}. 
The strict witnesses are given in
\cref{app:pathmarked-hierarchy-proof,cor:marked-small-strict}.
\end{proof}

\begin{remark}[Marking only the vertex inputs]
\label{rem:vertex-only-marking}
The conclusions of \cref{thm:pathmarked-expressivity} also hold when
occurrence marks are appended only to the vertex inputs of \PMEVT{}.
With width and depth allowed to vary, the first aligned pair MLP can
transfer the two endpoint marks to the pair stream before subsequent
triangular attention (\cref{app:pathmarked-hierarchy-proof}).
\end{remark}

\subsection{Selection, joint cache reading, and path marking: an expressivity hierarchy}
\label{sec:access-hierarchy}

The preceding theorems characterize StrikeTwice by its selected profile
\(\Psi_{\mathrm{ST}}\) and PathMarked by its marked profile
\(\Psi_{\mathrm{PM}}\). The joint selector \(\Sub_P^{\mathrm{joint}}\)
in \cref{eq:cache-pair-access} reads the same query-independent cache as
\(\Sub_P\). It selects one tagged vertex state per path occurrence and
one tagged pair state for each occurrence and graph vertex. The joint
reader processes each column in path order and pools over graph vertices
in a permutation-invariant way. At the level of stable colors, this gives
a multiset of ordered columns from \(\Psi_{\mathrm{EVT}}(X,Y)\).

\begin{definition}[Joint path-to-vertex profile]
For \(V=[n]\) and \(P=(p_1,\ldots,p_k)\), define
\begin{equation}
 \Gamma(X,Y,P)
 :=\left(n,k,\multiset{(c^\star_{p_su})_{s\in[k]}:u\in V}\right).
 \label{eq:joint-access-profile}
\end{equation}
Each tuple lists the relations from the path occurrences to a single
graph vertex in path order, including repeated visits. The multiset
retains the multiplicities of identical columns.
\end{definition}

The coordinate \(s\) determines the position code \(\phi(s)\), and
\(c^\star_{p_su}\) retains the indicator \(\one_{\{p_s=u\}}\) from its
initial color (\cref{eq:2fwl-step}). For each \(s\), the unique diagonal
color in coordinate \(s\) is \(c^\star_{p_sp_s}\). For a fixed EVT
encoder, this color determines \(\widetilde x_{p_s}\), and each pair
color \(c^\star_{p_su}\) determines \(\widetilde y_{p_su}\)
(\cref{thm:evt-expressivity}). Thus \(\Gamma\) determines the input to
the joint reader up to a permutation of the graph-vertex columns.

\begin{theorem}[Expressivity hierarchy of access profiles]
\label{prop:access-profile-hierarchy}
Use stable bidirectionally initialized \(2\)-FWL colors and
occurrence-identifying path marks. For every compact set \(S\) of
graph--path inputs,
\begin{equation}
 \begin{aligned}
  \ker(\Psi_{\mathrm{PM}})&\subseteq\ker(\Gamma)
                           \subseteq\ker(\Psi_{\mathrm{ST}}),\\
  \mathcal C_{\Psi_{\mathrm{ST}}}(S)&\subseteq\mathcal C_\Gamma(S)
                                     \subseteq\mathcal C_{\Psi_{\mathrm{PM}}}(S).
 \end{aligned}
 \label{eq:access-profile-hierarchy}
\end{equation}
For \(N\geq16\) and \(K\geq3\), both kernel inclusions are strict,
already on simple three-position paths in undirected binary graphs with
constant vertex features. Both continuous-class inclusions are strict
on suitable finite sets of these inputs.
On queries with \(k\leq2\), \(\Psi_{\mathrm{ST}}\) and \(\Gamma\)
determine each other. For \(N\geq16\), there are single-occurrence
queries with equal selected and joint profiles and different marked
profiles.
\end{theorem}

\begin{proof}[Proof sketch]
For each occurrence \(t\), exactly one tuple in \(\Gamma\) has a diagonal
pair color in coordinate \(t\). It corresponds to \(u=p_t\) and gives
column \(t\) of \(\Psi_{\mathrm{ST}}\), including when \(p_t\) is visited
more than once. A stable marked diagonal recovers the tuples of marked
path-to-vertex colors with their multiplicities. Forgetting the marks in
each tuple recovers \(\Gamma\). These reconstructions give the kernel
inclusions and hence the inclusions of the continuous compatible classes.

For \(k\leq2\), stable pair-color reconstruction also recovers \(\Gamma\)
from \(\Psi_{\mathrm{ST}}\). On the rook and Shrikhande graphs, a pair of
triangles has equal selected profiles and different joint profiles, and
a pair of open paths has equal joint profiles and different marked
profiles. The separating labels are continuous on these finite witness
sets. The same graphs also give the single-occurrence separation under
marking. Details are given in \cref{app:access-hierarchy-proof}.
\end{proof}

By \cref{thm:main-expressivity,thm:pathmarked-expressivity}, the outer
profiles characterize the kernels of the two-EVT StrikeTwice and
\PMEVT{} families. The corresponding continuous classes are their
uniform closures on \(S\). The same marked-profile characterization holds
for the ET variant in \cref{prop:marked-strict}. The middle class
\(\mathcal C_\Gamma(S)\) consists of continuous predictions compatible
with joint access to the stable colors.

\begin{remark}[Cost of joint cache reading]
Let \(d\) bound both cached state widths and state widths after tagging.
Selection and joint reading reuse the same \(O(n^2d)\) graph cache.
Selection exposes \(k^2\) pair slots per query and joint
reading exposes \(kn\). Materializing the corresponding arrays takes
\(O(k^2d)\) and \(O(knd)\) space, respectively. Total work and peak
working memory depend on how the reader processes these entries.
\end{remark}

\subsection{Sharpness of the expressivity inclusions}
\label{sec:latin-square}
\label{sec:cache-access}

\paragraph{Theoretical separation.}
The witnesses in \cref{prop:access-profile-hierarchy} establish
strictness of both inclusions. For the classification task, we use
graphs whose \(49\) vertices are the cells of a Latin square of order
seven. Two distinct cells are adjacent when they share a row, column,
or symbol \citep{nakasora2006mutually}.
These graphs have degree \(18\). Adjacent vertices have \(7\) common
neighbors and distinct nonadjacent vertices have \(6\).
We use constant vertex features \(X=\one_{49\times1}\) and binary
adjacency as the only pair feature. Zero-padding gives the same
construction at larger input widths. For an ordered triangle
\(P=(p_1,p_2,p_3)\) of distinct vertices, the target is the number
of neighbors common to all three vertices,
\[
 t_G(P):=\bigl|N_G(p_1)\cap N_G(p_2)\cap N_G(p_3)\bigr|,
\]
where \(N_G(v)\) is the neighbor set of \(v\).

\begin{proposition}[Certified selection and aggregation separation]
\label{prop:cache-aggregation-separation}
Let \(G\) be a simple undirected graph with \(49\) vertices, degree \(18\)
at every vertex, and \(7\) common neighbors for adjacent pairs and \(6\)
for distinct nonadjacent pairs. Give all vertices the same feature and
use binary adjacency as the only pair feature. Let \(\mathcal Q\) be a
nonempty set of ordered triangles \(P=(p_1,p_2,p_3)\) with distinct
vertices, balanced between \(t_G(P)=0\) and \(t_G(P)=1\).

For \(N\geq49\) and \(K\geq3\), the selected profile
\(\Psi_{\mathrm{ST}}\) is constant on \(\mathcal Q\). The same holds for
each row histogram \(\multiset{c^\star_{p_su}:u\in V}\), \(s\in[3]\).
For each \(1\leq s<t\leq3\), the histogram of pairs of colors to the
same graph vertex is also constant on \(\mathcal Q\),
\[
 \multiset{(c^\star_{p_su},c^\star_{p_tu}):u\in V}.
\]
Every deterministic binary classifier based on these summaries has
accuracy exactly \(50\%\) on \(\mathcal Q\). The joint profile \(\Gamma\)
determines \(t_G(P)\), giving a classifier with \(100\%\) accuracy.
\end{proposition}

\begin{proof}[Proof sketch]
As in an order-seven Latin-square graph, \(G\) has 49 vertices and degree 18.
Adjacent pairs have 7 common neighbors, and distinct nonadjacent
pairs have 6. These counts fix the multiset of relation pairs
through an intermediate vertex for each pair type.  Thus, at every refinement
round, diagonal, adjacent, and distinct nonadjacent pairs form
three color classes. 
Both profiles use the stable colors of the same unmarked graph.
The selected profile of \(\STarch[\EVT,\EVT]\) contains the
\(3\times3\) block of colors between the three path occurrences.
For every query triangle, this block has the diagonal color on the
diagonal and the adjacent-pair color off the diagonal. The selected
profile is therefore constant and \(\STarch[\EVT,\EVT]\) gives
identical outputs for all query triangles. Any deterministic binary
readout of these outputs achieves exactly \(50\%\) accuracy on the
balanced query set. The degree and common-neighbor counts also fix
the separate row and two-row histograms, giving the same accuracy
for classifiers based on these summaries.

For each graph vertex \(u\), the joint profile groups the three colors
\((c^\star_{p_1u},c^\star_{p_2u},c^\star_{p_3u})\) into an ordered triple.
This triple consists entirely of adjacent-pair colors exactly when
\(u\) is adjacent to all three triangle
vertices. Counting such triples gives \(t_G(P)\). A classifier
based on the joint profile therefore achieves \(100\%\) accuracy
(\cref{app:cache-aggregation-proof}).
\end{proof}

\begin{remark}[Recovery by path marking]
By \cref{prop:access-profile-hierarchy,prop:cache-aggregation-separation},
the marked profile determines \(\Gamma\) and therefore \(t_G(P)\).
Two marked refinement rounds suffice. The first recovers the adjacency
of \(u\) to the three triangle vertices in the color of \((p_s,u)\).
The second collects these colors at a selected diagonal, which determines
the number of vertices adjacent to all three.
For \(N\geq49\), \(K\geq3\), and scalar vertex and pair outputs, any
nonempty finite balanced set \(\mathcal Q\) of these queries therefore satisfies
\(\mathcal C_{\Psi_{\mathrm{ST}}}(\mathcal Q)
\subsetneq\mathcal C_{\Psi_{\mathrm{PM}}}(\mathcal Q)\).
A \PMEVT{} model can classify these queries exactly
(\cref{app:latin-marked-proof}).
The two open paths in \cref{eq:joint-access-marked-witness} have the
same joint profile \(\Gamma\) and different marked profiles
\(\Psi_{\mathrm{PM}}\).
\end{remark}

\paragraph{Numerical realization.}
Both experiments use queries with \(t_G(P)\in\{0,1\}\) and balance the
two labels within each graph. The first compares aggregation rules
on a shared frozen cache. The second trains StrikeTwice and PathMarked
on a fixed graph split to test the effect of auxiliary supervision.

\emph{Joint reading of a frozen cache.}
At each of three seeds, we freeze an EVT encoder at initialization.
Both heads receive the same cached pair states
\((\widetilde y_{p_su})_{s\in[3],\,u\in V}\) accessed by
\(\Sub_P^{\mathrm{joint}}\) in \cref{eq:cache-pair-access}. The pairwise
head pools over graph vertices separately for each pair of rows before
combining the three summaries. The joint head combines all three
relations to the same vertex before pooling. The heads differ in
parameter and operation counts. We train only the heads, using eight
queries from one graph and 1,000 AdamW updates at learning rate \(10^{-3}\).

After fixing the protocol and final-step checkpoints, we generate \(20\)
new test graphs with \(16\) queries per label on each. On the resulting
\(640\) queries, the joint head reaches \(100\%\) accuracy and the
pairwise head reaches \(50\%\) at each seed (\cref{tab:cache-access}).
Each test graph satisfies the hypotheses of
\cref{prop:cache-aggregation-separation}, such that the certified \(50\%\)
accuracy for \(\STarch[\EVT,\EVT]\) holds graph by graph and therefore
on the pooled test set. 
The two head scores are measured on the frozen cache.
This tests transfer to new graph structures with the same label-specific
joint relation histograms as the training graph
(\cref{app:certified-cache-access}). The numerical results recover the theoretical separation between
pairwise pooling and joint aggregation on the same frozen cache (\cref{prop:cache-aggregation-separation}).
The target makes the extra information from joint reading useful for
prediction by assigning different labels to queries with identical
selected profiles.

\emph{Learning with path marks.}
In addition, we test whether fixed-capacity models realize the expressivity gap
between ordered selection and path marking. We construct one
graph from each of the \(147\) order-seven main-class representatives
in McKay's catalogue and exclude the two graphs that contain no
triangle of label \(1\) (\cref{app:latin-numerical}). Of the \(145\)
remaining graphs, \(87\) are used for training, \(29\) for validation,
and \(29\) for testing (\(928\) test queries). We train both the encoders and readers of \(\STarch[\EVT,\EVT]\) and PathMarked with EVT (\PMEVT{}) from scratch.
In this experiment, \PMEVT{} appends occurrence marks only to the
vertex inputs (\cref{rem:vertex-only-marking}).
From the first EVT layer, the aligned pair MLP can access
these marks through the hidden states of both endpoints. A shared MLP
reads each selected vertex state together with its occurrence code, and
the query prediction averages the resulting logits.
At each of three seeds, both models reach \(50\%\) test accuracy with
only the query labels. With a training-only auxiliary loss that predicts
adjacency between graph vertices and path occurrences, \PMEVT{} reaches
\(100\%\) and StrikeTwice stays at \(50\%\)
(\cref{tab:latin-realization}). At width \(24\), the marked model
with auxiliary supervision reaches \(100\%\) at depths three and five and
\(50\%\) at depth one (\cref{app:latin-capacity}). The results at depths
three and five with auxiliary supervision recover the theoretical separation
between path marking and ordered selection (\cref{app:latin-marked-proof}).

\section{Synthetic and Real-Data Benchmarking}
\label{sec:experiments}

Our benchmarks evaluate graph and path processing, predictive
performance, and the computational benefits of graph reuse. Synthetic datasets test triangular attention for path-edge composition (Task~1) and graph-context encoding (Task~2). ThorAxe-Combo tests generalization to unobserved combinations of splicing choices; an Internet-routing benchmark tests temporal prediction and computational reuse.

We compare \STEE{}, \(\STarch[\EVT,\MLP_{\mathrm{add}}]\),
\(\STarch[\ET,\MLP_{\mathrm{add}}]\), and
\(\STarch[\EVT,\MLP_{\mathrm{concat}}]\), among other configurations.
In StrikeTwice model names, the graph encoder comes first and the path
processor second.
Both MLP path processors update the selected pair states with a pair MLP
and compute a mean for each row. The additive processor
\(\MLP_{\mathrm{add}}\) follows the update order of a simplified EVT block,
with vertex self-attention, triangular pair attention, and vertex-to-pair
attention disabled. After a residual pair MLP, the row mean replaces the
learned pair-to-vertex attention update and is added to the corresponding
vertex state before a residual vertex MLP. The concatenating processor
\(\MLP_{\mathrm{concat}}\) concatenates the row mean and the selected vertex
state as input to the vertex MLP.
A separate task-specific prediction head reads the resulting states.
Our EVT implementations use pre-LayerNorm and GELU MLPs.
PathTransformer processes input path features without a learned graph
encoder. EVT-Transformer and GINE-Transformer apply a path Transformer
to selected graph-vertex states. In the predictive benchmarks,
PathMarked with EVT (\PMEVT{}) adds occurrence marks to the vertex inputs
and encodes the graph separately for each path. A path Transformer
then processes the selected vertex states. Task-specific inputs and
prediction heads are detailed in \cref{app:model-input-access}.

\subsection{Synthetic datasets}
\paragraph{Shared synthetic protocol.}
For Tasks~1 and~2, we generate weighted layered directed acyclic graphs
(DAGs) with $n=28$ vertices and four query paths of $k=7$ vertices each.
Inputs comprise eight Gaussian nuisance vertex features and four pair
features encoding adjacency, positive edge weights, and two adjacency-masked
Gaussian nuisance variables (\cref{app:synthetic-inputs}). Graph-disjoint splits contain 512 optimization, 128 validation and
256 test graphs. Targets use training-only standardization; paired seeds
and validation-selected checkpoints are detailed in
\cref{app:synthetic-experiments}. We train models
independently with paired seeds and select checkpoints on validation.
We report held-out $R^2$; higher is better, with zero corresponding to
prediction by the mean target.

\subsubsection{Task 1. Learning relational composition along the path}
\label{sec:t1}
Task~1 evaluates how graph--path models learn multiplicative interactions
between consecutive path edges and whether triangular attention helps with
this computation. Let $A_{ij}$ denote adjacency and $w_{ij}$ the edge
weight. The weighted-adjacency channel of $Y$ is $B_{ij}=A_{ij}w_{ij}$.
For an ordered path $P=(p_1,\ldots,p_k)$, the target is the mean product
over windows of three consecutive edges,
\[
 s_1(Y,P)=\frac1{k-3}\sum_{\ell=1}^{k-3}
 B_{p_\ell,p_{\ell+1}}B_{p_{\ell+1},p_{\ell+2}}B_{p_{\ell+2},p_{\ell+3}}.
\]
All weights required by this target lie on the supplied path. Off-path graph information is therefore not needed to determine its value.

\begin{wraptable}{R}{0.46\textwidth}
\centering\small
\setlength{\abovecaptionskip}{0pt}
\caption{Task~1 test $R^2$ at the 3,200-update budget, reported as mean
$\pm$ sample standard deviation across ten paired seeds. Bold marks
the highest mean.}
\label{tab:task1-model-comparison}
\small\setlength{\tabcolsep}{6pt}
\begin{tabular}{@{}lc@{}}
\toprule
Model & Mean test $R^2$ $\pm$ SD\\
\midrule
\(\mathrm{ST}[\EVT,\EVT]\) & $\mathbf{0.982\pm0.006}$\\
\(\STarch[\EVT,\MLP_{\mathrm{add}}]\) & $0.867\pm0.054$\\
\(\STarch[\ET,\MLP_{\mathrm{add}}]\) & $0.466\pm0.078$\\
\bottomrule
\end{tabular}

\end{wraptable}
\Cref{tab:task1-model-comparison} compares the three StrikeTwice configurations on the same
queries and splits at an initial learning rate of \(3\times10^{-4}\).
The models have similar total parameter counts, with different
allocations across the two stages (\cref{app:synthetic-flexibility}).
\STEE{} scores above \(\STarch[\EVT,\MLP_{\mathrm{add}}]\), which in turn
scores above \(\STarch[\ET,\MLP_{\mathrm{add}}]\) on every paired seed.
The ranking of the three configurations in \cref{tab:task1-model-comparison}
is unchanged when the initial learning rate is selected separately for
each model on validation (\cref{tab:synthetic-flexibility-selected}).
With an EVT graph encoder, the additive and concatenating MLP
configurations give similar mean test \(R^2\), and the full two-stage
EVT scores higher than both (\cref{tab:t1-mlp-readers}).
Then, on this composition task, the tested configurations favor EVT over ET
at the graph stage and a full EVT over either MLP at the path stage. %
We train a small local MLP to test learnability with the target's local
structure supplied directly. It is a control rather than a generic
graph--path model, bypassing graph encoding and path processing.
It reads each window of three consecutive edge weights and averages the
four outputs. Under the same training protocol, it reaches a mean test \(R^2\) of \(0.9817\), close to \STEE{}'s \(0.9820\) (\cref{app:t1-mechanism}).

\paragraph{Path-stage triangular attention.}
EVT's modular design allows triangular pair attention to be added to the
MLP path processor or removed from the full path EVT. Adding this operator
to \(\STarch[\EVT,\MLP_{\mathrm{add}}]\) raises its mean test $R^2$ from
$0.867$ to $0.949$. Removing
it from \STEE{} reduces the mean from $0.982$ to $0.908$. Both effects hold on all ten seeds:
triangular attention contributes strongly to composing the selected
relations, beyond the MLP updates alone. These controls keep widths fixed but change parameter counts (\cref{tab:synthetic-flexibility-fixed}).

In a separate end-to-end control, we replace path-stage triangular attention with dense attention over all pair tokens.
Using the same graphs, queries and splits as \cref{tab:task1-model-comparison},
triangular attention leads on all ten paired seeds at 800 updates.
The advantage persists on all three paired seeds in a separate
3,200-update experiment on the same data, with mean test \(R^2\) of
\(0.984\) versus \(0.895\) for dense attention (\cref{tab:t1-budget}).
At 800 updates, triangular attention also leads on all three paired
seeds in each of two resamplings from the same graph generator
(\cref{tab:t1-graph-generations}).
At the tested training budgets, triangular attention learns to compose
consecutive edge weights more effectively than dense attention on Task~1.

In addition, to isolate the path processor, we train three reader variants on the same selected states from one frozen graph encoder trained with \(\STarch[\EVT,\MLP_{\mathrm{add}}]\).
We compare the full EVT reader with two controls that remove triangular
attention. One also widens the feed-forward layers to bring its total
parameter count within \(0.14\%\) of the full reader and its
output-connected parameter count within \(1.15\%\)
(\cref{tab:t1-additive-frozen-config}).  The full reader
scores higher than both controls on all three paired reader seeds, at
widths 32 and 64 and budgets of 800 and 3,200 updates. At 3,200 updates,
the mean test \(R^2\) gains over the widened control are \(0.0614\) and
\(0.0619\) at the two widths, respectively
(\cref{tab:t1-additive-frozen-cache}).  A separate control on a frozen graph
encoder trained with \STEE{} shows the same ordering
(\cref{tab:t1-frozen-cache}). These comparisons support a benefit from
triangular attention in the path reader on fixed graph states.

Thus, although triangular path attention is not needed for the limiting expressivity result (\cref{rem:path-attention-expressivity}), it contributes strongly to composing selected relations at the tested widths and training budgets. This supports our hypothesis that it helps when the reader still needs to compose relations across path positions after graph encoding. 

\paragraph{Sensitivity to vertex-feature noise.}
To test how much the trained models rely on noisy features,  Task~1 includes eight Gaussian vertex features as nuisance inputs, since the target depends only on the path's edge weights. At inference, we redraw these vertex features while
holding the topology, pair features, path, and target fixed.
We apply three paired redraws to the two MLP configurations
in \cref{tab:task1-model-comparison}, using one query per graph on all 128 validation
graphs. The root mean square change in predictions averages \(0.523\)
for \(\STarch[\ET,\MLP_{\mathrm{add}}]\) and \(0.289\) for
\(\STarch[\EVT,\MLP_{\mathrm{add}}]\). The EVT configuration is more robust to vertex-feature resampling, with smaller changes in predictions on all ten seeds. Against the unchanged targets,
mean squared errors rise from \(0.532\) to \(0.546\) and from \(0.145\) to
\(0.151\), respectively (\cref{tab:t1-nuisance-sensitivity}).

\subsubsection{Task 2. Learning graph context with reusable encodings}
\label{sec:t2}
Task 2 is motivated by variation-graph bubbles, which provide alternative
routes between shared endpoints
\citep{garrison2018variation,liao2023pangenome,sibbesen2023pantranscriptome}.
For a supplied path $P=(p_1,\ldots,p_k)$, let $N_G(u,v)$ denote the number
of directed paths from $u$ to $v$ in the full graph. The target averages
their log-counts over nonconsecutive path positions,
\[
 s_2(Y,P)=\frac{2}{(k-1)(k-2)}
 \sum_{s=1}^{k-2}\sum_{t=s+2}^{k}\log N_G(p_s,p_t).
\]
The prescribed segment contributes one route to each count, and thus, each term
equals $\log(1+\text{number of alternatives})$.
Exact dynamic programming supplies the labels
(\cref{app:t2-path-alternatives}).
Alternative routes may traverse vertices outside the supplied path.
On DAGs, their counts are nevertheless recoverable in principle from the
stable selected profile (\cref{app:t2-profile-compatibility}).
Task 2 therefore tests whether finite trained models learn to encode this
graph context and extract it through the reusable interface.

We compare the five StrikeTwice configurations with EVT-Transformer and
GINE-Transformer+raw (GINE; \citealp{hu2019strategies}) as graph--Transformer
references based on selected vertex states; +raw adds a raw-vertex-feature
bypass. PathTransformer reads raw path-vertex features. \PMEVT{}
provides the query-conditioned reference and applies a path Transformer
to the selected vertex states.
All nine models share graphs, queries, splits and paired evaluation
seeds. Validation MSE selects checkpoints within a budget of 2,000
updates. All models have parameter counts within 5\% of \STEE{}'s
and select their initial learning rates on three tuning seeds
(\cref{app:model-input-access,app:t2-common-lr,app:t2-additive-mlp-lr}).
\begin{wraptable}{R}{0.50\textwidth}
\centering\footnotesize
\setlength{\abovecaptionskip}{0pt}
\caption{Task~2 test $R^2$ (mean $\pm$ sample SD over five paired seeds)
with Gaussian vertex inputs and validation-selected checkpoints.
Bold marks the highest mean. The best cache-reusing mean is underlined.}
\label{tab:task2-model-comparison}
\resizebox{\linewidth}{!}{%
\small\setlength{\tabcolsep}{6pt}
\begin{tabular}{@{}lc@{}}
\toprule
Model & Mean test $R^2$ $\pm$ SD\\
\midrule
\(\STarch[\EVT,\MLP_{\mathrm{add}}]\) & $\underline{0.661\pm0.058}$\\
\STEE{} & $0.589\pm0.102$\\
\(\STarch[\EVT,\MLP_{\mathrm{concat}}]\) & $0.525\pm0.091$\\
\(\STarch[\ET,\MLP_{\mathrm{add}}]\) & $-0.009\pm0.004$\\
\(\mathrm{ST}[\ET,\ET]\) & $-0.009\pm0.010$\\
EVT-Transformer & $0.409\pm0.121$\\
GINE-Transformer+raw & $0.191\pm0.047$\\
PathTransformer & $-0.007\pm0.004$\\
\midrule
\PMEVT{} & $\mathbf{0.696\pm0.039}$\\
\bottomrule
\end{tabular}
}
\end{wraptable}

In \cref{tab:task2-model-comparison},
\(\STarch[\EVT,\MLP_{\mathrm{add}}]\) has the highest mean test \(R^2\)
among cache-reusing models. It exceeds \STEE{} on three of five paired seeds and \(\STarch[\EVT,\MLP_{\mathrm{concat}}]\) on all five paired seeds. 
The full EVT reader has a higher mean than the concatenating MLP reader. 
The score gap between the two MLPs comes from an end-to-end comparison in which
both the path processor and the graph encoder differ (\cref{app:t2-common-lr,app:t2-additive-mlp-lr}).
Thus, a full second EVT does not improve mean test $R^2$ over the additive MLP path processor on this task at the tested training budget.
All three EVT-based StrikeTwice configurations also exceed the graph--Transformer
references in mean test \(R^2\). This suggests that a path-processing stage with direct access to both selected vertex and pair states helps extract the graph context needed for Task~2. 
A test \(R^2\) near zero is expected for PathTransformer because it  sees only the  features of the  vertices and their fixed occurrence positions and these inputs are independent of the
graph's route counts.

\paragraph{Sensitivity to vertex inputs.}
\(\STarch[\EVT,\MLP_{\mathrm{add}}]\) outperforms
\(\STarch[\ET,\MLP_{\mathrm{add}}]\) on all five paired seeds,
and the ET-based model scores near zero
(\cref{tab:task2-model-comparison}).
Task~2 predicts alternative-route counts from graph and path structure.
We run a separate
paired control at a fixed initial learning rate of \(3\times10^{-4}\) that tests sensitivity to irrelevant vertex features.
Retraining and evaluating with zero vertex features raises mean test
\(R^2\) from \(-0.026\) to \(0.984\) for
\(\STarch[\ET,\MLP_{\mathrm{add}}]\) and from \(0.661\) to \(0.824\) for
\(\STarch[\EVT,\MLP_{\mathrm{add}}]\), with gains on all five paired
seeds for both models (\cref{tab:t2-vertex-input-ablation}).
Other ET- and EVT-based configurations show a similar pattern
(\cref{tab:t2-attribute-controls}). Together with the smaller changes in
the EVT-based model's predictions when vertex features are redrawn at
inference in Task~1 (\cref{tab:t1-nuisance-sensitivity}), this suggests
greater robustness to irrelevant vertex inputs for the tested EVT-based
configurations. One possible explanation is that EVT can process vertex
features in a separate stream before exchanging information with pair
states.

\paragraph{Encoding graph context with triangular attention.}
We now test the specific contribution of triangular attention within the EVT graph encoder.
At the graph stage, EVT combines triangular pair attention with explicit
vertex states, vertex attention and cross-stream exchanges. At a fixed
rate of \(3\times10^{-4}\), removing graph-stage triangular attention
decreases test \(R^2\) on every paired seed for
\(\STarch[\EVT,\MLP_{\mathrm{add}}]\) and \STEE{}. With the operator
removed, their means are \(-0.012\) and \(-0.003\), respectively.
These controls fix widths and reduce parameter counts
(\cref{tab:synthetic-flexibility-fixed}).
The drop to near-zero scores suggests that graph-stage triangular
attention helps strongly these EVT configurations to encode information about
alternative routes in states that the path processor can use. 

\paragraph{Selected and joint reading of a frozen graph cache.}
To compare two ways of reading the same frozen cache, we freeze the graph
encoders of three \(\STarch[\EVT,\MLP_{\mathrm{add}}]\) models trained on
Gaussian inputs. For each encoder, two readers receive the same ordered
path-vertex and path-pair states. Their additional branch combines the relations
from all path occurrences to each candidate vertex. The selected reader
pools over distinct vertices visited by the path, and the joint reader
pools over all graph vertices. The two readers have identical parameter
counts and paired initialization and minibatch order.
At the 800-update budget, mean test \(R^2\) is \(0.684\) for joint
reading and \(0.683\) for selection, with a paired difference of
\(+0.0008\pm0.0077\) across the three encoders
(\cref{tab:t2-joint-frozen-readers}). At the tested capacity and training
budget, the two readers achieve virtually identical test scores on Task~2.
This is consistent with the stable selected profile already determining
the route counts used in the target (\cref{app:t2-profile-compatibility}).

\paragraph{Inference with reuse.}
\PMEVT{} achieves the highest mean test \(R^2\) in
\cref{tab:task2-model-comparison} and re-encodes the graph for each path.
We time the selected checkpoints on 12 test graphs, processing four
queries per graph sequentially on CPU. With the same trained weights,
retaining the graph cache gives speedups of \(2.40\times\) for \STEE{}
and \(3.22\times\) for \(\STarch[\EVT,\MLP_{\mathrm{add}}]\) over rebuilding
the encoding for every query.
With reuse, the two models are \(2.31\times\) and \(3.72\times\) as fast
as \PMEVT{}, respectively. These timings include initial graph encoding.
At 16 queries and the same batch size of one, the within-model speedups
rise to \(3.78\times\) and \(7.42\times\)
(\cref{tab:t2-transformer-readout-latencies}).

\subsection{ThorAxe-Combo: predicting combinations of splicing choices}
\label{sec:biology}
Motivated by coordinated splicing choices
\citep{tilgner2015comprehensive,santos2026isoform}, we construct ThorAxe-Combo to test whether a candidate path assembled from individually supported splice alternatives is itself annotated as a complete isoform in the reserved clade.

\begin{figure}[H]
\centering
\resizebox{\linewidth}{!}{\input{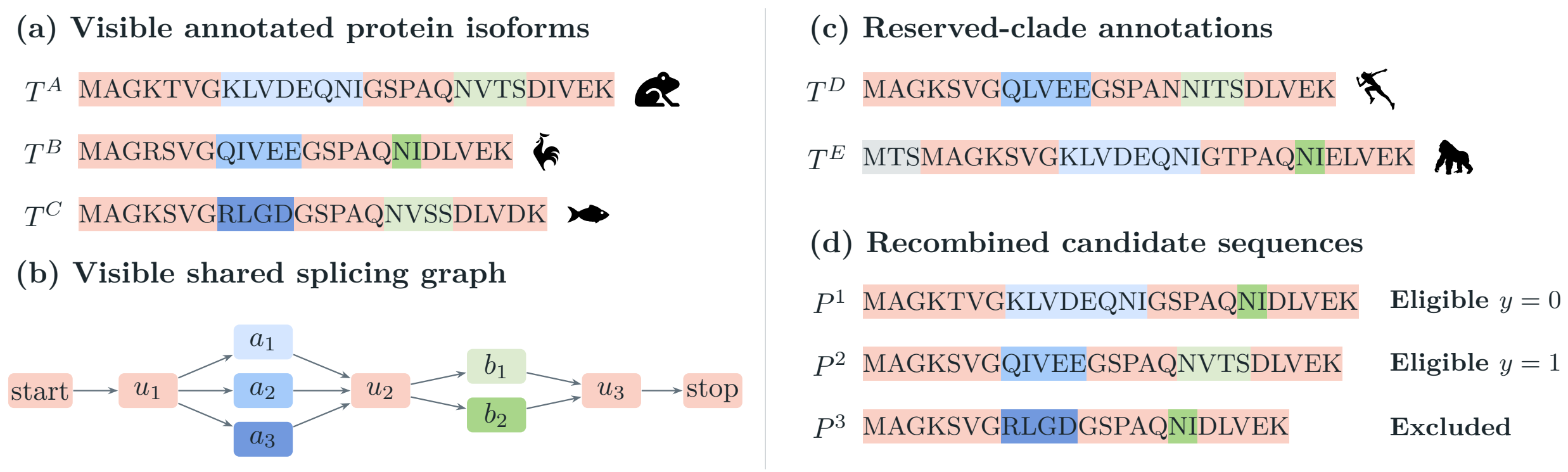}}
\caption{ThorAxe-Combo construction, with primates as the reserved clade.
\textbf{(a--b)} Visible isoforms define the splicing graph.
\textbf{(c--d)} Reserved-clade annotations support individual alternatives
and label complete paths: $P^2$ is annotated, $P^1$ is absent,
and $P^3$ is excluded because $a_3$ lacks support.
Illustrative amino-acid sequences may differ within a shared s-exon;
the gray prefix denotes an additional s-exon.}
\label{fig:thoraxe-combo-construction}
\end{figure}

For each gene, ThorAxe integrates annotated protein isoforms across species into a shared evolutionary splicing graph whose vertices are aligned amino-acid segments, called s-exons \citep{zea2021thoraxe}.
Isoforms from species outside the reserved clade define the visible graph (\cref{fig:thoraxe-combo-construction}a--b).
We recombine their subpaths into candidates absent from visible annotations, requiring each selected alternative to be supported
by reserved-clade annotations. A candidate is positive if its complete path matches an annotated isoform from that clade, and negative otherwise
(\cref{fig:thoraxe-combo-construction}c--d).
A negative label records a complete candidate path absent from the reserved-clade annotations, not established biological incompatibility. 
The benchmark contains 16,544 candidate paths (3,120 positives) across 1,596 gene–clade graphs derived from 1,072 genes, using four reserved clades.
Construction rules, filtering and cohort statistics are detailed
in \cref{app:biology-data}. ThorAxe segmentation is fixed upstream; sequence embeddings and annotation-derived path support supplied to the model come from
visible species.

Five outer folds hold out entire genes, keeping their groups and queries together.
Learning rates are selected by three-fold inner validation.
Inner validation selects the readout
combining focal and mean path-vertex states, mean consecutive-pair states,
and clade identity for all four StrikeTwice configurations. GINE-Transformer
uses a vertex-only readout. The references use only path features, with
PathTransformer+J also receiving raw attributes of consecutive pairs
(\cref{app:biology-models,app:biology-provenance}). All models use fixed embeddings of s-exon amino-acid sequences
from the pretrained ESM Cambrian (ESM-C) protein language model
\citep{scale2024esm}.
Evaluation uses AUROC and average precision (AP), computed within
gene--clade groups and averaged with equal group weights across
outer tests. We report means and sample SDs over five training seeds
(\cref{app:biology-evaluation}).
The benchmark tests unseen genes within the eligible population,
without enforcing clade or homologous-family holdout
(\cref{app:biology-scope}).

In \cref{tab:thoraxe-main}, with an EVT graph encoder and the consecutive-pair readout, both MLP configurations achieve higher mean AUROC and AP than \STEE{}. The concatenating configuration has the highest mean on both metrics. Among the additive configurations, the EVT
graph encoder gives slightly higher means and lower variability across
seeds. The ET graph encoder has higher AUROC on four of the five paired
seeds.
GINE-Transformer and both PathTransformer variants achieve means close to
\(\mathrm{ST}[\EVT,\EVT]\). PathTransformer improves on the embedding-only
reference in mean score, with a further gain from junction attributes.
\begin{table}[!htbp]
\centering\small
\setlength{\abovecaptionskip}{0pt}
\caption{ThorAxe-Combo: test mean $\pm$ SD over five seeds.
Bold: highest mean among models with graph encoding (left).
+J: raw junction attributes; ESM-C f+m: focal and mean embeddings.
Random: expected within-group ranking scores, averaged equally across groups.}
\label{tab:thoraxe-main}
\setlength{\tabcolsep}{2.5pt}
\begin{tabular*}{\linewidth}{@{\extracolsep{\fill}}lcc@{\hspace{8pt}}|@{\hspace{8pt}}lcc@{}}
\toprule
Graph--path model & AUROC & AP & Reference model & AUROC & AP\\
\midrule
\(\STarch[\EVT,\MLP_{\mathrm{concat}}]\) & $\mathbf{.702\pm.018}$ & $\mathbf{.676\pm.013}$ & PathTransformer & $.632\pm.004$ & $.618\pm.003$\\
\(\STarch[\EVT,\MLP_{\mathrm{add}}]\) & $.698\pm.012$ & $.672\pm.012$ & PathTransformer+J & $.641\pm.006$ & $.626\pm.003$\\
\(\STarch[\ET,\MLP_{\mathrm{add}}]\) & $.689\pm.031$ & $.668\pm.028$ & ESM-C f+m & $.575\pm.009$ & $.577\pm.005$\\
GINE-Transformer & $.643\pm.001$ & $.627\pm.003$ & Random & $.500$ & $.523$\\
\cmidrule(l){4-6}
\(\mathrm{ST}[\EVT,\EVT]\) & $.635\pm.023$ & $.621\pm.019$ & \PMEVT{} & $.636\pm.014$ & $.620\pm.010$\\
\bottomrule
\end{tabular*}

\end{table}

\paragraph{Pair states in the readout.}
We compare a vertex-only readout with two readouts that add pair
information. The focal-pair readout uses both directed pair states
between the positions associated with the first and last differing
blocks. The consecutive-pair readout averages the pair states at
consecutive positions along the candidate path. In all six
configurations tested with all three readouts, the consecutive-pair
readout raises mean AUROC and the focal-pair readout lowers it,
relative to the vertex-only readout (\cref{tab:biology-heads}).
One possible explanation is that the consecutive-pair readout captures how alternative s-exon blocks connect to the surrounding sequence.
These local transitions may help distinguish full combinations of
splicing choices that a single focal pair summarizes only coarsely. With the consecutive-pair readout, the additive and concatenating
MLPs achieve similar mean AUROC for either graph encoder. These independently trained comparisons also change readout projections and selected learning rates. Therefore, the gain cannot be attributed to junction access alone.

\paragraph{Attention at the path and graph stages.}
\begin{wraptable}{R}{0.50\textwidth}
\centering\small
\setlength{\abovecaptionskip}{0pt}
\setlength{\tabcolsep}{3pt}
\caption{ThorAxe-Combo path-stage attention. Test mean $\pm$ SD over five
seeds. Bold marks the highest mean.}
\label{tab:biology-path-attention}
\begin{tabular}{@{}lcc@{}}
\toprule
Added attention & AUROC & AP\\
\midrule
None (reference) & $.698\pm.012$ & $.672\pm.012$\\
Vertex attention & $\mathbf{.705\pm.007}$ & $\mathbf{.683\pm.004}$\\
Triangular attention & $.690\pm.014$ & $.667\pm.012$\\
Both attentions & $.687\pm.020$ & $.664\pm.018$\\
\bottomrule
\end{tabular}

\end{wraptable}
We keep the consecutive-pair readout fixed and test whether adding attention to the MLP path processor can improve \(\STarch[\EVT,\MLP_{\mathrm{add}}]\). We use the additive MLP as the reference for the path-stage attention
study because it is a simplified EVT block with separate vertex and pair MLP updates. We add vertex attention, triangular attention, or both, keeping the MLP
updates, row aggregation, consecutive-pair readout, graph-encoder
architecture, and hidden widths fixed. The attention additions change
the parameter count.
Adding vertex attention improves both AUROC and AP on four of the five
paired seeds and yields the highest means in
\cref{tab:biology-path-attention}. The mean scores fall below the
additive reference when triangular attention is added alone or with
vertex attention. All four variants also retain separate vertex and pair
MLP updates (\cref{app:biology-path-attention}). Because splicing choices at distant positions can be coordinated \citep{tilgner2015comprehensive}, one possible explanation for the gain from vertex attention is that it can relate s-exon states across the path, complementing the local junction information in the readout.

At the graph stage, removing triangular attention from
\(\STarch[\EVT,\MLP_{\mathrm{add}}]\) and \STEE{} produces small changes
in opposite directions on ThorAxe-Combo (\cref{app:biology-flexibility}).
On Task~2, the same ablation brings both configurations close to zero \(R^2\).
Under the evaluated protocols, the benefit of attention depends on the
task and the processing stage.

\paragraph{Training cost and inference with reuse.}
\PMEVT{} re-encodes the graph for each candidate and uses a path
Transformer with a vertex-only readout. Its mean AUROC and AP are close
to those of \STEE{} with the consecutive-pair readout
(\cref{tab:thoraxe-main}). Across 25 final fits per model, mean wall
time is 38.56 minutes for \PMEVT{} and 13.76 minutes for \STEE{}
(\(2.80\times\)), including training, validation and final prediction.
Readouts, batch sizes and stopping times differ
(\cref{app:thoraxe-marked-application}).

We time warm inference on all 16,544 candidates on an A100 with 80\,GB
of memory, with eight queries per batch. With the same initialized
\STEE{} weights in both modes, retaining the encoding across batches
reduces time from \(3.339\) to \(2.538\) ms per query (\(1.32\times\)).
The timings start with packed inputs on the GPU and include initial
graph encoding. The speedup is \(1.62\times\) for the 488 gene--clade
groups that span multiple batches and \(1.00\times\) for the 1,108
groups that fit in one. On the same workload, \PMEVT{} with trained
checkpoints takes \(3.577\) ms per query
(\cref{app:thoraxe-marked-application}).
\FloatBarrier

\subsection{Atlas routing benchmark: assessing temporal prediction and graph reuse}
\label{sec:atlas}
\label{sec:reuse}
We curate RIPE Atlas Internet-routing measurements
\citep{ripencc2015atlas,ripeatlasmeshes} into round-trip-time (RTT) prediction
along observed responding-interface paths. Graphs contain up to 256 vertices and paths up to 32 occurrences. A shared prior summarizes February--March 2026 RTTs
(\cref{app:atlas-validation}). Training uses April--July, validation August and the reserved test September. Unlabelled monthly topology is available: this is temporal
prediction on known topology, not prediction of unseen network topology.
We report mean absolute error (MAE) in milliseconds, averaging errors within
each path and then across paths.
PathTransformer and the StrikeTwice models predict corrections to the
same prior. Their input access is detailed in \cref{app:profile-input-access}.
Models receive the same three-rate search on August validation,
followed by five evaluation seeds at 8,000 updates; parameter counts
approximately match except for \PMEVT{}
(\cref{app:atlas-common-comparison,app:atlas-additive-mlp-protocol}).

\begin{table}[!htbp]
\centering\small
\setlength{\tabcolsep}{4pt}
\caption{Atlas September MAE (ms), mean $\pm$ SD over five seeds.
Rates are validation-selected, except for the separate \STEE{} control at $3\times10^{-4}$.
24k uses the same rate as 8k within each row. --: not run.
Bold marks the lowest mean per budget.}
\label{tab:atlas-main}
\begin{tabular*}{\linewidth}{@{\extracolsep{\fill}}lccc@{}}
\toprule
Model & Learning rate & 8,000-update budget & 24,000-update budget\\
\midrule
\STEE{} & \(10^{-4}\) & \(11.089\pm0.093\) & ---\\
\STEE{} & \(3\times10^{-4}\) & \(11.011\pm0.145\) & \(\mathbf{10.396\pm0.052}\)\\
\(\STarch[\EVT,\MLP_{\mathrm{add}}]\) & \(3\times10^{-4}\) & \(11.204\pm0.191\) & \(10.628\pm0.057\)\\
\(\STarch[\EVT,\MLP_{\mathrm{concat}}]\) & \(3\times10^{-4}\) & \(\mathbf{11.004\pm0.106}\) & \(10.460\pm0.068\)\\
PathTransformer & \(10^{-4}\) & \(11.159\pm0.077\) & \(10.947\pm0.073\)\\
\PMEVT{} & \(3\times10^{-4}\) & \(11.059\pm0.080\) & ---\\
RouteNet-ReLU & \(3\times10^{-4}\) & \(11.236\pm0.096\) & ---\\
RouteNet-SELU & \(3\times10^{-4}\) & \(11.458\pm0.095\) & \(10.886\pm0.110\)\\
\bottomrule
\end{tabular*}
\end{table}

At 8,000 updates, \(\STarch[\EVT,\MLP_{\mathrm{concat}}]\) has the lowest mean September MAE under the common learning-rate screen.
\PMEVT{}, all StrikeTwice configurations and PathTransformer achieve lower mean
MAE than either RouteNet adaptation with ReLU or official-style SELU
readouts (\cref{app:atlas-common-comparison}). 
Both RouteNet adaptations retain path order; available inputs do not describe interactions between simultaneous flows.

The separate \STEE{} control at \(3\times10^{-4}\) improves August
validation over the screen-selected rate on all five seeds. At 8,000
updates, its mean September MAE is close to that of
\(\STarch[\EVT,\MLP_{\mathrm{concat}}]\) at the same rate, with \STEE{}
giving lower MAE on three of the five paired seeds.
Based on the August validation results, we use this rate for \STEE{}
at 24,000 updates. Both MLP configurations and PathTransformer retain their
validation-selected rates (\cref{app:atlas-long-budget,app:atlas-additive-mlp-protocol}).
At 24,000 updates, \STEE{} has the lowest mean MAE in \cref{tab:atlas-main}
and outperforms the concatenating and additive MLP configurations
on four and five paired seeds, respectively. Both MLP configurations
outperform PathTransformer on all five paired seeds. All three StrikeTwice
configurations achieve lower mean MAE than RouteNet-SELU
(\cref{tab:atlas-main,app:atlas-additive-mlp-protocol}).

\paragraph{Path-stage triangular attention on a frozen cache.}
To isolate the contribution of path-stage triangular attention, we freeze
the graph encoder from the \STEE{} run at the screen-selected learning
rate of \(10^{-4}\), using its August-selected checkpoint
(seed~1, update 8,000). The full EVT reader has lower mean MAE than both
triangular-attention deletion controls at widths 32 and 64 and budgets of
2,000 and 8,000 updates. At width 64 and 8,000 updates, mean MAE over three
reader seeds is \(10.610\)~ms for the full reader, \(10.684\)~ms for
deletion and \(10.728\)~ms for deletion with wider feed-forward layers.
As in Task~1, path-stage triangular attention helps on the same frozen representation (\cref{app:atlas-triangle-control}). 

\paragraph{Selected and joint reading of a frozen cache.}
To test whether access to off-path vertices improves RTT prediction, we
train two readers matched in capacity and optimization on each of three
frozen EVT caches from \(\STarch[\EVT,\MLP_{\mathrm{concat}}]\) fits (August-selected
checkpoints, seeds~0--2). Both read the ordered path states and use a
second branch to pool relations to candidate vertices. The selected
reader pools over distinct visited vertices and the joint reader pools
over all graph vertices. With checkpoints selected on August within
8,000 updates, mean September MAE is \(10.956\)~ms for selection and
\(11.050\)~ms for joint reading. Selection has slightly lower MAE on each
of the three caches (\cref{app:atlas-readers}). As in Task~2, the theoretical
expressivity advantage of joint reading (\cref{prop:access-profile-hierarchy})
yields no clear predictive gain on the tested caches.

\paragraph{Training cost and inference with reuse.}
\PMEVT{} marks each path before encoding the graph, and \STEE{} can
share one path-independent graph encoding across queries.
In the common 8,000-update comparison, \PMEVT{}'s final training runs
take \(3.34\times\) as long as \STEE{}'s on average (9.22 versus 2.76
hours, including validation), with microbatches of two paths versus
eight (\cref{app:marked-applications}).

To isolate the inference savings from caching across batches, we time
32 queries on one graph in four batches of eight with identical trained
weights and inputs. We compare encoding the graph once for all four
batches with encoding it once per batch. Both timings include the
initial graph encoding. On an A100-40GB, the median device time for
all 32 paths with the \(\STarch[\EVT,\MLP_{\mathrm{add}}]\) model from
\cref{tab:atlas-main}
falls from \(733\)\,ms with one encoding per batch to \(186\)\,ms with
one encoding shared across batches. The median paired speedup over
nine August graph groups is \(3.94\times\), with identical RTT predictions
(\cref{app:atlas-additive-mlp-protocol}).
On TPU, a separate study using the configurations in
\cref{tab:application-configurations} gives speedups of
\(2.32\times\) for \(\STarch[\EVT,\MLP_{\mathrm{add}}]\), \(1.84\times\) for
\(\STarch[\ET,\MLP_{\mathrm{add}}]\), and \(1.58\times\) for \STEE{}. The MLP and
two-EVT profiles use different graph cohorts (\cref{app:atlas-reuse}).
Keeping the cache across batches saves time beyond the reuse already
available within each batch.

\paragraph{Larger graphs and amortization.}
Our implementation avoids materializing the $O(n^3)$ triangular
score tensor through row/channel tiling (\cref{app:flash-triatt};
\citealp{dao2022flash}) and reduces training memory through activation
checkpointing \citep{chen2016sublinear}. Cubic arithmetic cost remains.
With unchanged architecture and FP32 on an A100-40GB, the aggressively
tiled \STEE{} uses 10.16\,GiB of peak allocated memory and takes 71.45\,s
on average for a forward and backward pass on a synthetic graph with
1,024 vertices and 16 length-16 paths, without an optimizer step.
In a separate inference experiment with
length-eight paths, encoding contributes at most 10\% of total time after
an estimated 8,946 queries. Reuse can save time below this threshold. These synthetic profiles demonstrate that \STEE{} can run at this scale
on a single GPU with tiled attention and activation checkpointing
(\cref{app:large-graph-profiling}).

\FloatBarrier

\section{Conclusion and Open Questions}
\label{sec:conclusion}

We introduce EVT, a graph encoder coupling vertex and direction-aware
pair states, and StrikeTwice, a modular graph-to-path architecture that
reuses graph states across ordered queries. StrikeTwice selects cached states in path-occurrence order. Position codes distinguish repeated visits, and equality tags identify occurrences of the same vertex.

We characterize EVT's equivariant approximation class.
For StrikeTwice with EVT at both stages,
a path-indexed \(2\)-FWL profile exactly characterizes
distinguishing power and uniform approximation on compact
domains, with one parameter set shared across bounded graph and path sizes.
Selected states can already encode off-path context. Joint reading
of the same cache can distinguish additional inputs, and marking
path occurrences before encoding yields a further separation
(\cref{prop:access-profile-hierarchy}).

Latin-square queries provide an explicit example in which both joint
reading of the same graph cache and path-marked encoding distinguish
inputs with identical selected profiles.
With capacity allowed to grow, disabling path-stage triangular attention preserves this approximation class when the remaining EVT operations and occurrence-position codes are retained.
This equality does not imply equal learning efficiency at finite capacity.

Our benchmarks connect these results to finite-capacity learning.
On Latin-square queries, joint reading outperforms separate pairwise
pooling on the same frozen cache. In a separate control, \PMEVT{} learns
the distinguishing function with training-only auxiliary supervision,
and \STEE{} remains constrained by its certified information limit.
In comparison, selection and joint reading achieve similar predictive
performance on the tested caches in Task~2 and Atlas.

Two levels of modularity make it possible to test individual operations
at either stage. StrikeTwice separates the choice of graph encoder from
that of path processor, and EVT makes attention operations configurable
within each stage. The resulting comparisons show that the best-performing
configuration depends on the task.

In each main benchmark comparison, the best-performing reusable
configuration uses EVT as its graph encoder. On ThorAxe-Combo, both
EVT--MLP configurations achieve higher mean AUROC and AP than \STEE{}.
On Atlas, \(\STarch[\EVT,\MLP_{\mathrm{concat}}]\) has the lowest mean
error at 8,000 updates, with a close mean for \STEE{} at the same
learning rate. \STEE{} has the lowest mean error in the 24,000-update
column of \cref{tab:atlas-main}. The synthetic controls further show that the tested EVT-based models are less sensitive to irrelevant vertex features than their ET-based
counterparts.

Within the EVT-based configurations, the ablations locate the benefit
of triangular attention at different stages. On Task~2, it helps the
graph encoder represent alternative routes outside the query path before
selection. At the path stage, it helps compose consecutive edge weights
on Task~1 and lowers mean RTT error in the Atlas frozen-encoder control.
On ThorAxe-Combo, adding vertex attention to the residual MLP gives the
best scores among the tested path-stage variants.

Finally, a shared graph encoding can preserve enough context for accurate path
prediction. Indeed, on ThorAxe-Combo and Atlas, \STEE{} achieves predictive
performance close to \PMEVT{}'s with shorter final training runs under the
tested protocols. In addition, retaining graph encodings across batches
reduces inference time on Task~2, ThorAxe-Combo and Atlas, with weights
and batch sizes held fixed.

Two questions follow.
Can reusable graph-to-path models retain predictive performance with subcubic graph-encoding cost and compressed caches?
How do a task's demands for relational composition and off-path context guide the choice of graph representation and path processor under a fixed computational budget?

\section*{Data availability}
The ThorAxe-Combo inputs are available through \href{https://figshare.com/s/40e9fcd4db13faffe419}{Figshare}.
The appendices document dataset construction, splits and model selection.
The Task~1 frozen-reader study and the Task~2 feature and cache-access
studies reuse the same graph splits as the corresponding benchmarks. The
appendices specify the source models, validation rules and test protocols (\cref{app:t1-additive-frozen-cache,app:t2-attribute-controls,app:t2-joint-frozen-readers}).

\section*{Code availability}
The PyTorch implementations of StrikeTwice and the Edge--Vertex Transformer are
available in the \href{https://github.com/Guichaoua/StrikeTwice}{\texttt{Guichaoua/StrikeTwice} GitHub repository}.
Install the package from \href{https://pypi.org/project/StrikeTwice/}{PyPI} with
\texttt{pip install StrikeTwice}.
The \href{https://striketwice.readthedocs.io/en/latest/}{Read the Docs documentation}
provides installation instructions, runnable examples, and a Python API reference.
See also the \href{https://figshare.com/s/40e9fcd4db13faffe419}{ThorAxe-Combo inputs on Figshare}.

\section*{AI use statement}
Generative AI tools, including ChatGPT/Codex, assisted with literature synthesis,
research and experimental design, mathematical formulation and proof development,
implementation, result interpretation, and manuscript drafting. Synthetic datasets
and numerical results came from executed experiments and archived predictions or
metrics. The authors
remain responsible for verifying claims, proofs, citations, code, protocols,
numerical outputs, and the final content.

\section*{Acknowledgments}
This work was granted access to the HPC resources of the MeSU platform at Sorbonne Université. 
The MeSU platform and its supercomputing resources are supported by the Ile-de-France region, the 
French Ministère de l'Enseignement Supérieur et de la Recherche, and Sorbonne Université.
This work has been funded by the European Union (ERC, PROMISE, 101087830). Views and opinions expressed are however those of the author(s) only and do not
necessarily reflect those of the European Union or the European Research Council. Neither the European Union nor the granting authority can be held responsible for them. This work has also been co-funded by the French National Agency for Research (ANR, France 2030, PostGenAI@Paris, ANR-23-IACL-0007).

\section{Conflict of interest statement}

None declared.

\bibliography{references}
\bibliographystyle{plainnat}

\clearpage
\appendix
\numberwithin{table}{section}
\numberwithin{figure}{section}
\numberwithin{theorem}{section}
\aliascntresetthe{proposition}
\aliascntresetthe{lemma}
\aliascntresetthe{corollary}
\aliascntresetthe{definition}
\aliascntresetthe{assumption}
\aliascntresetthe{remark}
\renewcommand{\theHproposition}{\theHtheorem}
\renewcommand{\theHlemma}{\theHtheorem}
\renewcommand{\theHcorollary}{\theHtheorem}
\renewcommand{\theHdefinition}{\theHtheorem}
\renewcommand{\theHassumption}{\theHtheorem}
\renewcommand{\theHremark}{\theHtheorem}

\section{Notation}
\label{app:notation}

\Cref{tab:notation} collects the main symbols. Where a symbol has several
uses, its context is specified explicitly. Graph vertices and path
occurrences are distinct index sets, even when the path has no repeated visits.

The dimensions in \cref{sec:architecture} describe architecture
representations; the universality results prescribe prediction dimensions
separately. For StrikeTwice, the outer input map takes
\((d_{\mathcal{X}}^{\mathrm{in}},d_{\mathcal{Y}}^{\mathrm{in}})\)
to \((d_{\mathcal{X}},d_{\mathcal{Y}})\), the graph encoder preserves
those widths, and selection appends tags to produce
\((d_{\mathcal{X}}+d_\phi,d_{\mathcal{Y}}+2d_\phi+1)\).
The core bridge maps these to \((q_{\mathcal{X}},q_{\mathcal{Y}})\),
the path processor preserves them, and the outer output map produces
\((d_{\mathcal{X}}^{\mathrm{out}},d_{\mathcal{Y}}^{\mathrm{out}})\).
For PathMarked, \(d_{\mathcal{X}},d_{\mathcal{Y}}\) are the widths
before marking: marking follows the outer input lift and gives graph
widths \((d_{\mathcal{X}}+d_\phi,d_{\mathcal{Y}}+2d_\phi)\),
which its EVT core preserves. Selection appends occurrence tags again,
giving widths \((d_{\mathcal{X}}+2d_\phi,d_{\mathcal{Y}}+4d_\phi+1)\).
Its terminal MLPs produce the free widths
\((q_{\mathcal{X}},q_{\mathcal{Y}})\) before the same outer output map.
In the bounded-size proofs, \(d_\phi=K\).
\Cref{tab:interface-dimensions} separates the number of indices from the
feature widths: vertex and pair tensors have shapes
\(n\times\text{width}\) and \(n\times n\times\text{width}\)
on a graph, and the analogous shapes with \(k\) on an occurrence graph.
Selection changes the index sets; affine maps change only feature widths.

\begingroup
\small
\setlength{\tabcolsep}{4pt}
\setlength{\LTleft}{0pt}
\setlength{\LTright}{0pt}
\renewcommand{\arraystretch}{1.13}
\begin{longtable}{@{}>{\raggedright\arraybackslash}p{0.37\textwidth}
                    >{\centering\arraybackslash}p{0.13\textwidth}
                    >{\centering\arraybackslash}p{0.20\textwidth}
                    >{\centering\arraybackslash}p{0.24\textwidth}@{}}
\caption{Dimensions along the theoretical interfaces. The last two columns
are per-token feature widths; \(d_\phi=K\) in the proofs.
All states of an EVT core retain its input widths.}
\label{tab:interface-dimensions}\\
\toprule
State or operation just completed & Index sets & Vertex width & Pair width\\
\midrule
Raw graph input & \([n],[n]^2\) & \(d_{\mathcal{X}}^{\mathrm{in}}\) & \(d_{\mathcal{Y}}^{\mathrm{in}}\)\\
Outer input \(\Aff_{\theta_{\mathrm{in}}}\) & \([n],[n]^2\) & \(d_{\mathcal{X}}\) & \(d_{\mathcal{Y}}\)\\
EVT core; ST graph cache & \([n],[n]^2\) & \(d_{\mathcal{X}}\) & \(d_{\mathcal{Y}}\)\\
ST \(\Sub_P\) & \([k],[k]^2\) & \(d_{\mathcal{X}}+d_\phi\) & \(d_{\mathcal{Y}}+2d_\phi+1\)\\
ST bridge and path EVT & \([k],[k]^2\) & \(q_{\mathcal{X}}\) & \(q_{\mathcal{Y}}\)\\
PM \(\Mark_P\) and graph EVT & \([n],[n]^2\) & \(d_{\mathcal{X}}+d_\phi\) & \(d_{\mathcal{Y}}+2d_\phi\)\\
PM \(\Sub_P\) & \([k],[k]^2\) & \(d_{\mathcal{X}}+2d_\phi\) & \(d_{\mathcal{Y}}+4d_\phi+1\)\\
PM terminal MLPs & \([k],[k]^2\) & \(q_{\mathcal{X}}\) & \(q_{\mathcal{Y}}\)\\
EVT outer output \(\Aff_{\theta_{\mathrm{out}}}\) & \([n],[n]^2\) & \(d_{\mathcal{X}}^{\mathrm{out}}\) & \(d_{\mathcal{Y}}^{\mathrm{out}}\)\\
ST/PM outer output & \([k],[k]^2\) & \(d_{\mathcal{X}}^{\mathrm{out}}\) & \(d_{\mathcal{Y}}^{\mathrm{out}}\)\\
\bottomrule
\end{longtable}
\endgroup

\begingroup
\small
\setlength{\tabcolsep}{5pt}
\setlength{\LTleft}{0pt}
\setlength{\LTright}{0pt}
\renewcommand{\arraystretch}{1.12}
\begin{longtable}{@{}>{\raggedright\arraybackslash}p{0.28\textwidth}
                    >{\raggedright\arraybackslash}p{0.69\textwidth}@{}}
\caption{Main notation for the reusable interface, its information profiles,
and the experiments.}\label{tab:notation}\\
\toprule
Symbol & Meaning\\
\midrule
\endfirsthead
\multicolumn{2}{l}{\tablename\ \thetable\ (continued)}\\
\toprule
Symbol & Meaning\\
\midrule
\endhead
\midrule
\multicolumn{2}{r}{Continued on the next page}\\
\endfoot
\bottomrule
\endlastfoot

\multicolumn{2}{@{}l}{\textbf{Graph inputs and ordered paths}}\\*
\(G,\ V=[n]\) & Input graph and its vertex set;
\([n]=\{1,\ldots,n\}\).\\
\(n,N\) & Number of vertices in one input and the maximum number allowed
in the bounded-size input domain.\\
\(P=(p_1,\ldots,p_k),\ k,K\) & Ordered query sequence in \(V^k\),
its number of vertex occurrences, and the maximum allowed number of
positions. Each \(p_s\in V\); repeated vertices are allowed.\\
\(i,j,u;\ s,t\) & Graph-vertex indices; path-occurrence indices.
The vertex visited at position \(s\) is \(p_s\). Some proofs also use
\(u=(X,Y,P)\) for a complete input.\\
\(X=(x_i),\ Y=(y_{ij})\) & Vertex and ordered-pair features or states.
In a raw theoretical input their widths are
\(d_{\mathcal{X}}^{\mathrm{in}},d_{\mathcal{Y}}^{\mathrm{in}}\);
when describing an EVT core they are generic current states of its fixed
widths. Pair states include self-pairs and pairs without an edge.\\
\(d_{\mathcal{X}}^{\mathrm{in}},d_{\mathcal{Y}}^{\mathrm{in}}\) & Raw input feature dimensions.\\
\(d_{\mathcal{X}},d_{\mathcal{Y}}\) & Free vertex and pair widths after
the outer input lift. EVT and the ST graph core preserve them;
PM appends marks before its core, which instead preserves
\(d_{\mathcal{X}}+d_\phi,d_{\mathcal{Y}}+2d_\phi\).\\
\(q_{\mathcal{X}},q_{\mathcal{Y}}\) & Free path representation widths:
preserved by the ST path EVT, or produced by the PM
terminal MLPs. They are not prescribed prediction dimensions.\\
\(d_{\mathcal{X}}^{\mathrm{out}},d_{\mathcal{Y}}^{\mathrm{out}};\ q\) &
Fixed final prediction widths after the outer affine output map;
bare \(q\) is a separately scoped vector width in rooted or auxiliary
vertex-only statements.\\
\(P\times P\) & Shorthand for occurrence-indexed pair-state access:
\(\widetilde Y[P,P]=(\widetilde y_{p_sp_t})_{s,t\in[k]}\),
with \(k^2\) positions, preserving order and repeated visits
(\cref{sec:architecture}).\\
\(\one_{\{p_s=p_t\}},\ \one_{\{s=t\}}\) & Indicators of visiting the
same graph vertex and of being the same path occurrence, respectively.
The path-stage aligned MLP uses the occurrence-diagonal indicator
\(\one_{\{s=t\}}\).\\

\addlinespace[4pt]
\multicolumn{2}{@{}l}{\textbf{Encoded states, operators, and dimensions}}\\*
\(\widetilde X,\widetilde Y\) & Encoded graph states before ordered
selection; in ST these are query-independent and reusable.
\(\widetilde X[P],\widetilde Y[P,P]\) denote their ordered selection.\\
\(\widehat X^0,\widehat Y^0\) & Path-processor input states after ordered
selection, tagging, and the core bridge \(\Aff_{\theta_{\mathrm{aff}}}\);
their widths are \(q_{\mathcal{X}},q_{\mathcal{Y}}\).\\
\(\phi(s)\in\R^{d_\phi}\) & Fixed occurrence-position code, injective on
the supported positions; the theory uses
\(\phi(s)=(\one_{\{r=s\}})_{r=1}^K\in\R^K\).
Pairs receive the ordered concatenation \(\phi(s)\concat\phi(t)\).\\
\(\widehat X,\widehat Y;\ Z,R\) & Occurrence-indexed vertex and pair
representations returned by the architecture core, of widths
\(q_{\mathcal{X}},q_{\mathcal{Y}}\). Experimental \(Z,R\) denote these
features before the task-specific layer; theoretical predictions are
\(f_\theta=\Aff_{\theta_{\mathrm{out}}}(\widehat X,\widehat Y)\).\\
\(\GraphEnc,\PathProc\),\newline\(\PathGraphNet\) & Equivariant graph networks
parameterizing ST and PM.\\
\(\STarch[\GraphEnc,\)\newline\(\PathProc]\),\newline\(\PMarch(\PathGraphNet)\) & Architecture families with a graph
encoder and path processor, or a query-marked graph network.
Their core maps are \(\STarch_{\theta_{\mathrm{ST}}}\) and
\(\PMarch_{\theta_{\mathrm{PM}}}\).\\
\(\Sub_P,\Mark_P,\Aff\) & Ordered selection with occurrence tags;
graph marking with endpoint marks; independent shared affine maps on
vertex and pair tokens.\\
\(\theta_{\mathrm G},\theta_{\mathrm P},\theta_{\mathrm M};\ L_G,L_P,L_M\) & Independent
graph-encoder, path-processor and marked-network parameters and depths.\\
\(\theta_{\mathrm{aff}},\theta_{\mathrm R}\) & Parameters of the core
graph-to-path affine bridge and terminal marked readout, respectively.\\
\(\theta_{\mathrm{in}},\theta_{\mathrm{out}}\) & Parameters of the
outer affine maps from raw features to graph representations and from
core output representations to prescribed predictions. These are separate
from the architecture-core parameters.\\
\(\theta_{\mathrm{ST}},\theta_{\mathrm{PM}}\) & Core parameter tuples
\((\theta_{\mathrm G},\theta_{\mathrm{aff}},\theta_{\mathrm P})\) and
\((\theta_{\mathrm M},\theta_{\mathrm R})\), respectively.\\
\(m\) & Auxiliary matrix size in the contraction and streaming-cost
formulas; equal to \(n\) for graph states or \(k\) for path states.\\
\(d;\ d_a,d_v\) & Upper bound on hidden token and total attention widths
in complexity bounds, including marks; query/key
and value widths per attention head. In local normalization formulas,
\(d\) denotes the dimension of the token being normalized; its separate
color-code meaning is specified below.\\
\(W_Q,W_K;\ W_L,W_R\) & Learned query/key projections and the two
value projections of a triangular attention head, respectively.\\
\(\Att,\TriAtt^{\alpha,\beta}\) & Multi-head standard attention and triangular
pair attention; a subscript \(h\) denotes one unprojected head.
Each \((\alpha,\beta)\in\{\rightarrow,\leftarrow\}^2\)
chooses two oriented pair views, with
\(y^\rightarrow_{ij}=y_{ij}\) and \(y^\leftarrow_{ij}=y_{ji}\).\\
\(\AlMLP,\ \Flat,\ \Unflat\) & Aligned vertex--pair MLP exchange;
reshaping between an \(n\times n\) pair array and its \(n^2\)-token list
(replace \(n\) by \(k\) on a path).\\
\(\RMSNorm,\LN,\Norm\) & RMSNorm, LayerNorm, and the chosen
normalization before attention projections and on complete concatenated
aligned-MLP inputs. Position and equality
coordinates are included in the normalized states
(\cref{def:striketwice-family}).\\
\(\concat,\ \odot,\ \multiset{\cdot}\) & Concatenation,
element-wise product, and a multiset retaining multiplicities.\\

\addlinespace[4pt]
\multicolumn{2}{@{}l}{\textbf{Colors, information profiles, and function classes}}\\*
\(c^r_{ij},\ R_N,\ \cstar_{ij}\) & Bidirectionally initialized
\(2\)-FWL pair color after \(r\) rounds; common terminal depth
\(R_N=2N^2\); terminal color \(\cstar_{ij}=c^{R_N}_{ij}\).\\
\(\mathfrak C_N\) & Common formal color space across graph comparisons.\\
\(\Psi_{\mathrm{EVT}}\) & Complete ordered-pair color array
\((c^\star_{ij})_{ij}\), depending only on \((X,Y)\).
It is equivariant under graph relabeling, not an invariant graph hash.\\
\(\Psi_{\mathrm{ST}}=\Psi_{N,K}\) & Selected profile
\(\bigl(n,k,(\cstar_{p_sp_t})_{s,t\in[k]}\bigr)\), retaining the
ordered path-pair color matrix.\\
\(m_P(i),\ \bar c^\star_{ij}\) & Occurrence mark
\(m_P(i)=\sum_{s\in[k]:p_s=i}\phi(s)\) assigned before graph encoding;
terminal pair color after marked refinement.\\
\(\Psi_{\mathrm{PM}}=\Psi^{\mathrm{mark}}_{N,K}\) & Marked profile
\(\bigl(n,k,(\bar c^\star_{p_sp_s})_{s\in[k]}\bigr)\), using the
ordered selected marked diagonals.\\
\(\mathcal D_{n,k},\ \mathcal D_{\leq N,\leq K}\) & Input triples
with fixed sizes \((n,k)\), and their disjoint union over
\(n\leq N,\ k\leq K\).\\
\(\mathcal O_K,\ S\) & Output space
\(\bigsqcup_{k=1}^K(\R^{k\times d_{\mathcal{X}}^{\mathrm{out}}}\times\R^{k\times k\times d_{\mathcal{Y}}^{\mathrm{out}}})\);
a finite or compact input set,
as specified in each result.\\
\(\ker f,\ \Ker(\mathcal F)\) & Pairs of inputs receiving the same
output under \(f\); common indistinguishability relation
\(\bigcap_{f\in\mathcal F}\ker f\) for a function family.\\
\(f_{\mathcal X}^{\bullet},\ f_{\mathcal Y}^{\bullet}\) & Rooted output evaluations
\((X,Y;i)\mapsto f_{\mathcal X}(X,Y)_i\) and
\((X,Y;i,j)\mapsto f_{\mathcal Y}(X,Y)_{ij}\).
The indices select outputs and do not mark encoder inputs.\\
\(\ker^{\mathrm{eq}}f,\ \Ker^{\mathrm{eq}}(\mathcal F)\) & Equivariant kernel:
the pair of ordinary kernels of the vertex and pair rooted evaluations;
the common kernel intersects these componentwise over \(f\in\mathcal F\)
(\cref{eq:equivariant-kernel}).\\
\(\mathcal G_n,\mathcal G_{\leq N}\) & Graph-feature input space of size
\(n\), and its disjoint union over \(n\leq N\).\\
\(\mathcal H_{\mathrm{EVT}}\) & Variable-capacity equivariant EVT maps
with joint vertex--pair predictions: the dimension-preserving EVT core
surrounded by outer affine input and output maps.\\
\(\mathcal H_{\mathrm{ST}}=\mathcal H_{N,K}\) & ST family at
variable finite capacity in \cref{def:striketwice-family}.\\
\(\mathcal H^{\mathrm{proc}}_{N,K}\) & Its subfamily with all path-stage
triangular residual branches disabled.\\
\(\mathcal H_{\mathrm{PM}}=\mathcal H^{\mathrm{mark}}_{N,K}\) &
PM with EVT and occurrence-tagged vertex--pair GELU readouts,
surrounded by outer affine maps
(\cref{app:marked-model}); \(\mathcal H^{\mathrm{mark,ET}}_{N,K}\) uses
endpoint-initialized, direction-mixed ET instead.\\
\(\mathcal C(S),\mathcal C_\Psi(S)\) & Continuous predictions with the prescribed
output shape; its subclass constant on the fibers of \(\Psi|_S\).
Appendix subscripts \(\Psi\) and \(\Psi^{\mathrm{mark}}\) abbreviate the
selected and marked profiles.\\
\(\mathcal C(S;E)\) & Continuous maps from \(S\) to an explicitly
specified target space \(E\); used when the output space is not implicit.\\
\(\mathcal C^{\mathrm{eq}}_{\Psi_{\mathrm{EVT}}}(S)\) & Continuous graph outputs
whose equivariant kernel contains that of \(\Psi_{\mathrm{EVT}}|_S\):
equal stable colors give equal outputs at any two vertices or any two pairs.
This is stronger than whole-array kernel compatibility
(\cref{eq:evt-entrywise-compatible}).\\
\(\gamma_N^{\mathcal{X}},\gamma_N^{\mathcal{Y}}\) & Diagonal and ordered-pair
colors on rooted input domains in the proofs (\cref{app:evt-et-comparison}).\\
\(\mathcal F|_S,\ \|\cdot\|_{\infty,S}\) & Family restricted to \(S\); uniform output
error on \(S\), using the joint Frobenius norm
\(\|(U,V)\|_{\mathrm F}^2=\|U\|_{\mathrm F}^2+\|V\|_{\mathrm F}^2\).\\
\(\eta(c)\in\R^d,\ U_{st},\ B\) & Color code of dimension \(d\),
explicit \(K^2(1+d)\)-coordinate auxiliary vector, and its globally
broadcast mean in \cref{eq:path-decoder-blocks,eq:indexed-pair-broadcast}.
This \(d\) is chosen for the proof's color codes and is not fixed by an
experimental hidden width.\\

\addlinespace[4pt]
\multicolumn{2}{@{}l}{\textbf{Experimental notation and contextual uses}}\\*
\(A,w,B\) & Adjacency, positive edge weights, and weighted adjacency
\(B_{ij}=A_{ij}w_{ij}\) in the synthetic tasks.\\
\(\mathcal P;\ B\) & In cost statements, number of query paths sharing
one graph (\cref{sec:architecture}); \(B\) denotes the
experimental physical batch size.\\
\(\mathrm D,\mathrm A,\mathrm N;\ t_G(P),t\) & Diagonal, adjacent, and
non-adjacent relation colors in the Latin-graph experiment; common-neighbor
count \(t=t_G(P)\in\{0,1\}\) for the three anchors in the retained queries.\\
\((n,d,\lambda,\mu)\) & Strongly regular graph parameters: vertex
count, degree, and common-neighbor counts for adjacent and distinct
non-adjacent vertices. Here \(d\) denotes degree.\\
\(g_R\) & Mean consecutive-pair state
\((k-1)^{-1}\sum_{s=1}^{k-1}R_{s,s+1}\) used by biological path-edge
readouts, in addition to focal vertex states and path context.\\
\(\EVT_{\mathrm D}\) & EVT variant that retains its vertex stream and exchanges while
replacing triangular pair attention by dense pair self-attention.\\
\end{longtable}
\endgroup

\section{Inputs and Access in the Evaluated Models}
\label{app:model-input-access}

The additive and concatenating MLP path processors use the same raw
observation rules. Task-specific configurations for the additive processor
are detailed in
\cref{app:synthetic-flexibility,app:biology-additive-mlp,app:application-protocol}.

The tables separate the raw observations given to each model from the encoded
states read by its path processor and prediction head. Selected states can
already contain off-path information. For example, an MLP path processor
can read states computed by EVT on the full graph.
Here $X[P]$ denotes raw path-vertex features; $\widetilde X[P]$ and
$\widetilde Y[P,P]$ denote states selected from the graph cache.
$Z,R$ denote vertex and pair outputs of the path processor, when present.
All means exclude padding. Positions refer to occurrence indices; repeated-vertex indicators
refer to equality of the underlying graph vertices.

\subsection{Synthetic comparisons and certified readers}
\label{app:synthetic-input-access}
\begin{table}[H]
\centering\small\setlength{\tabcolsep}{3pt}
\begin{tabular}{@{}>{\raggedright\arraybackslash}p{.26\linewidth}>{\raggedright\arraybackslash}p{.21\linewidth}>{\raggedright\arraybackslash}p{.28\linewidth}>{\raggedright\arraybackslash}p{.19\linewidth}@{}}
\toprule
System & Graph observations & Query access & Query information\\
\midrule
PathTransformer (Task 2) & None beyond $X[P]$ & Raw path vertices & Ordered positions\\
GINE-Transformer + raw (Task 2) & $X$; directed edges and their attributes & Selected $\widetilde X[P]$ and raw $X[P]$ bypass & Ordered positions\\
Edge Transformer, faithful / direction-mixed & $X,Y$ via endpoint lifts & Self-pair proxies, then path Transformer & Ordered positions\\
EVT-Transformer & $X,Y$ & Selected $\widetilde X[P]$ & Ordered positions\\
\(\STarch[\EVT,\MLP_{\mathrm{add}}]\), \(\STarch[\ET,\MLP_{\mathrm{add}}]\) & $X,Y$ & Complete $P\times P$; residual MLPs and normalized row aggregation & Positions, vertex equality\\
\(\STarch[\EVT,\MLP_{\mathrm{concat}}]\), \(\STarch[\ET,\MLP_{\mathrm{concat}}]\) & $X,Y$ & Complete $P\times P$; pair MLPs and row means concatenated with vertex states & Positions, vertex equality\\
\STEE{}, TD, DT, DD; ET-Twice & $X,Y$ & Complete $P\times P$; second pair processor & Positions, vertex equality\\
\PMEVT{} (Task 2 main comparison) & $X,Y$ and path-position marks & Query-conditioned graph states, then path Transformer & Marks before encoding; positions\\
\PMEVT{} (Task 2 readout control) & $X,Y$ and path-position marks & Masked mean of selected vertex states, then MLP or affine readout & Marks before encoding; no extra readout tag\\
\PMEVT{} (Latin control) & Constant vertices, adjacency, occurrence marks & Tagged occurrence MLPs, then mean logits & Marks before encoding and occurrence readout tags\\
Local MLP (Task 1 control) & No graph encoding & Windows of three raw path-edge weights & Window order\\
\bottomrule
\end{tabular}
\caption{Synthetic input access. Graph encoders can use off-path vertices
and relations. PathTransformer and the local MLP access only the query
path. Models grouped in one row share access rules. The task sections
specify their architectures and training settings.}
\label{tab:synthetic-input-access}
\end{table}

The Task 2 MLP/affine marked readouts are specified in
\cref{app:t2-marked-readout}; the separate tagged Latin readout is specified
in \cref{app:latin-numerical}. These are different empirical adapters.

In Task 1, the \STEE{}--TD intervention retains
the same graph-encoder architecture and selected observations, with separately
trained weights and a small FFN adjustment. The factorial controls vary both
operators. Canonical Task 2 rankings instead compare complete systems with
different widths and readout interfaces (\cref{app:experimental-interfaces}).

The Task 2 frozen-cache readers both retain the complete ordered selected array.
Their added branch combines $(\widetilde y_{p_su})_{s\in[k]}$ for each common graph vertex $u$
before summation. The selection control restricts $u$ to distinct vertices
visited by the path; the joint reader uses all valid graph vertices. Both
receive graph and path counts. Their equal-sized heads are trained from
scratch on the same frozen cache within each seed.

The certified collision readers use constant vertex features and binary
adjacency, a frozen random EVT cache, and ordered anchor relations. Their
pairwise and joint aggregations retain different numbers of anchor relations
before pooling over graph vertices. Exact color profiles certify the examples
and are not network inputs. These readers differ from the Task 2 readers and have
different parameter counts (\cref{app:certified-cache-access}).

\subsection{ThorAxe-Combo}
\label{app:biological-input-access}
\begin{table}[H]
\centering\small\setlength{\tabcolsep}{3pt}
\begin{tabular}{@{}>{\raggedright\arraybackslash}p{.23\linewidth}>{\raggedright\arraybackslash}p{.26\linewidth}>{\raggedright\arraybackslash}p{.24\linewidth}>{\raggedright\arraybackslash}p{.21\linewidth}@{}}
\toprule
System & Observations & Learned graph context & Final prediction inputs\\
\midrule
StrikeTwice with EVT or ET & 964 vertex channels; seven pair channels & All visible vertices and pairs & Focal $Z$, mean $Z$, clade; optional focal-pair or path-edge $R$\\
GINE-Transformer & Same vertex/pair channels; preserved directed adjacency & Directed message passing on visible edges & Focal $Z$, mean $Z$, clade; no $R$ branch\\
\PMEVT{} & Same vertex/pair channels; occurrence marks on vertices & Marked graph encoded for each query, then path Transformer on selected vertex states & Focal $Z$, mean $Z$, clade\\
ESM-C focal + mean & 960 sequence channels at path vertices & No learned graph encoder & Focal embeddings, path mean, clade\\
PathTransformer, with/without raw junctions & 964 path-vertex channels; optional seven adjacent pair channels & Path attention only & Focal $Z$, mean $Z$, clade; optional mean of locally projected junctions\\
\bottomrule
\end{tabular}
\caption{Inputs for the retained biological comparisons. Focal positions are
supplied to the head. Query-independent graph encoders use unmarked inputs.
Graph-based systems use the 8/4-channel stem, and local references use
their own projections. Readout
choices and widths are specified in \cref{app:biology-models}.}
\label{tab:biological-input-access}
\end{table}

All graph models receive visible-species and visible-path support attributes
through their pair input. GINE uses those attributes on
present edges, while preserving the discrete adjacency channel through its
stem. Its vertex-only final head therefore still contains learned graph
context. The ESM-C reference uses embeddings averaged over visible instances
and species during preprocessing. All systems receive the same focal indices
and clade identity, and predict the candidate label directly.
Within each pair-state graph-model family, the three readout conditions are trained
independently from input projections to prediction head.

\subsection{Atlas profile models}
\label{app:profile-input-access}
For Atlas, PathTransformer receives raw
vertex features and locally transformed adjacent junction attributes. The
StrikeTwice models receive the same graph observations and encode the whole
graph before selection. At vertex-level targets, the final head reads an occurrence state
and its normalized ordinal coordinate. These models differ from the synthetic and biological
implementations carrying similar model names.

Every profile model predicts a correction to a shared context prior:
\[
 \widehat t_{qs}=b_{qs}+\alpha f_\theta(G,P_q)_s,
 \qquad
 \alpha=10\text{ ms}.
\]
The prior $b$ is computed from February--March 2026 RTT observations
(\cref{app:atlas-validation}) and included among the raw features.
PathTransformer reads path features that include collective snapshot
context and the prior. Missing labels and padding are masked.
There are no biological focal marks or clade embeddings in these adapters.
On Atlas, \(\MLP_{\mathrm{add}}\) uses residual pair and vertex updates
with normalized source-row aggregation before the occurrence-level task
head. \(\MLP_{\mathrm{concat}}\) uses the same cache access and feeds the
row mean concatenated with the selected vertex state into its vertex
MLP. ET graph encoders supply diagonal vertex proxies.
The main model comparison evaluates complete, independently trained systems
selected on validation (\cref{app:atlas}). The matched-reader control instead
holds EVT graph states fixed and varies which vertices its context branch
can read, with identical inputs, trainable capacity, and selection rules
(\cref{app:atlas-readers}).

\section{Certified Cache Selection and Aggregation}
\label{app:certified-cache-access}

Latin-square graphs provide queries with identical selected \(2\)-FWL profiles
and different labels. Their structural diversity lets us test the readouts
on new, non-isomorphic graphs.

\paragraph{Data and exact summaries.}
We use graphs constructed from Latin squares of order seven
\citep{nakasora2006mutually}. Each graph is a strongly regular graph with parameters $(n,d,\lambda,\mu)=(49,18,7,6)$: it has 49 vertices, every vertex has 18 neighbors, adjacent vertices have exactly 7 common neighbors, and distinct non-adjacent vertices have exactly 6 common neighbors.
Vertices have constant scalar features, and binary adjacency is the only pair attribute.
For an ordered triangle $P=(p_1,p_2,p_3)$, the target $t$ is
the number of vertices adjacent to all three anchors:
\[
t = |N(p_1)\cap N(p_2)\cap N(p_3)|,
\]
where $N(v)$ denotes the set of neighbors of vertex $v$.
We retain 32 queries per graph, balanced between 16 triangle paths with $t=1$ and 16 with $t=0$.
The confirmation cohort contains 20 new test graphs and 640 queries. The earlier 48-graph cohort is used for development; only eight queries from one of its graphs are used to fit the readouts.

Joint unmarked \(2\)-FWL refinement over all 68 development and confirmation graphs yields a stable partition into three common color classes:
diagonal \(\mathrm D\), adjacent \(\mathrm A\), and non-adjacent \(\mathrm N\).
For each graph, write \(\cstar_{uv}\in\{\mathrm D,\mathrm A,\mathrm N\}\)
for the resulting stable color of pair \((u,v)\).

We compare four ways of retaining the colors associated with a query $P$.
$\mathsf S$ keeps the selected pair colors between its three anchors.
$\mathsf U$ counts colors separately for each anchor over all graph vertices.
$\mathsf B$ counts pairs of colors that share the same graph vertex, separately for each pair of anchors.
$\mathsf T$ counts joint color triples for all three anchors before discarding that shared vertex index:
\begin{align*}
 \mathsf S(P)&=(\cstar_{p_ip_j})_{i,j=1}^3,\\
 \mathsf U(P)&=\left(\multiset{\cstar_{p_i u}:u\in V}\right)_{i=1}^3,\\
 \mathsf B(P)&=\left(\multiset{(\cstar_{p_i u},\cstar_{p_j u}):u\in V}\right)_{i<j},\\
 \mathsf T(P)&=\multiset{(\cstar_{p_1u},\cstar_{p_2u},\cstar_{p_3u}):u\in V}.
\end{align*}

\paragraph{Detailed proof of \cref{prop:cache-aggregation-separation}.}
\label{app:cache-aggregation-proof}
The main-text proposition applies separately within every graph in these
balanced cohorts. In the notation above, the optimal deterministic binary
classification accuracies for functions of \(\mathsf S,\mathsf U,\mathsf B,\mathsf T\)
are respectively \(1/2,1/2,1/2,1\).
\begin{proof}
The degree and common-neighbor counts make every refinement round depend
only on whether a pair is diagonal, adjacent, or nonadjacent. Thus the
stable colors are precisely the three classes
\(\mathrm D,\mathrm A,\mathrm N\) used above.

For every query triangle, the selected $3\times3$ block has $\mathrm D$ on the diagonal and $\mathrm A$ off the diagonal, so $\mathsf S$ is constant.

Regularity gives the same row histogram for every anchor $p_i$: the row $(\cstar_{p_i u})_{u\in V}$ has multiplicities $(\mathrm D:\!1,\mathrm A:\!18,\mathrm N:\!30)$,  so $\mathsf U$ is constant.

The strongly regular intersection counts give the same pairwise histogram for every adjacent anchor pair.
Indeed, for adjacent anchors, $\mathrm{DA}$  and $\mathrm{AD}$  each occur once, $\mathrm{AA}$  occurs
$\lambda=7$ times, $\mathrm{AN}$  and $\mathrm{NA}$  each occur $d-1-\lambda=10$
times, and $\mathrm{NN}$  occurs $n-2d+\lambda=20$ times. Thus $\mathsf B$ is constant.
The pairwise counts and the three-anchor counts derived below are summarized here:
\begin{center}
\small
\setlength{\tabcolsep}{3pt}
\begin{tabular}[t]{@{}lrrrrrr@{}}
\toprule
Two-row pattern & $\mathrm{DA}$ & $\mathrm{AD}$ & $\mathrm{AA}$ & $\mathrm{AN}$ & $\mathrm{NA}$ & $\mathrm{NN}$\\
Multiplicity & 1 & 1 & 7 & 10 & 10 & 20\\
\bottomrule
\end{tabular}
\quad
\begin{tabular}[t]{@{}lr@{}}
\toprule
Three-row pattern & Each multiplicity\\
$\mathrm{DAA, ADA, AAD}$ & $1$\\
$\mathrm{AAA}$ & $t$\\
$\mathrm{AAN, ANA, NAA}$ & $6-t$\\
$\mathrm{ANN, NAN, NNA}$ & $4+t$\\
$\mathrm{NNN}$ & $16-t$\\
\bottomrule
\end{tabular}
\end{center}
For three anchors, $\mathrm{DAA}$, $\mathrm{ADA}$, and $\mathrm{AAD}$ each occur once,
while $\mathrm{AAA}$ occurs $t$ times. Each pair has $\lambda$ common neighbors, including the third anchor and the $t$ neighbors
shared by all three. Thus $\mathrm{AAN, ANA, NAA}$ each occur  $\lambda-1-t=6-t$ times.
The one-adjacency and zero-adjacency counts are respectively $d-2\lambda+t=4+t$ and $n-3d+3\lambda-t=16-t$.
Thus $\mathsf T$ distinguishes the two labels because its $\mathrm{AAA}$  count recovers $t$, whereas $\mathsf S$, $\mathsf U$, and $\mathsf B$ are constant across the retained queries.
Since labels are balanced within every graph, any deterministic classifier based on these summaries has accuracy $1/2$.
In contrast, $\mathsf T$ gives optimal accuracy $1$.
\end{proof}

The bound applies to each of the three summaries that discard joint
three-anchor information. Joint aggregation can use binary operations,
provided they retain the shared vertex index. For example, one may first compute
$r(u)=A_{p_1u}A_{p_2u}$ and then
$t=\sum_u r(u)A_{p_3u}$. The limitation therefore concerns aggregation schemes that discard the shared index $u$ before the information from all three anchors has been combined.

The full set of 1,029 triangles in each graph is not assumed to be balanced, so its majority-class baseline need not be $1/2$.

For the reported experiment, we independently recomputed targets from
adjacency and verified exact equality of the complete tuple signatures. The signatures
certify the dataset; the networks
receive only the graph features described above.

\paragraph{Learned readouts of a frozen cache.}
For each seed, we randomly initialize an EVT encoder with two blocks, width 32, feed-forward network (FFN) width 64, four direction-balanced
heads, LayerNorm, and zero dropout. Its weights remain fixed, and both readouts use the same cached pair states $\widetilde Y$.
Training only the readouts isolates their aggregation schemes on identical
representations.

The joint readout combines the relations of all three anchors to each vertex $u$, then sums over vertices.
The pairwise readout first combines the relations of two anchors to each vertex, sums separately for each anchor pair,
and finally combines the three summaries. In terms of the cached states,
\begin{align*}
\ell_{\mathrm{joint}}(P)
&=\beta+\sum_{u\in V}
\phi([\widetilde y_{p_1u};\widetilde y_{p_2u};\widetilde y_{p_3u}]),\\
h_{ij}
&=\sum_{u\in V}f([\widetilde y_{p_i u};\widetilde y_{p_j u}]),
\qquad (i,j)\in\{(1,2),(1,3),(2,3)\},\\
\ell_{\mathrm{pair}}(P)
&=g([h_{12};h_{13};h_{23}]).
\end{align*}
Here $[\cdot;\cdot]$ denotes concatenation, $\beta$ is a
learned scalar bias, and $\phi$, $f$, and $g$ are MLPs
with rectified linear unit (ReLU) hidden activations. The logits $\ell$ are converted
to class probabilities by a sigmoid.

Since each pair state has 32 coordinates, $\phi$ has widths
$96\to64\to32\to1$, $f$ has widths $64\to64\to32$,
and $g$ has widths $96\to32\to1$.
The joint and pairwise readouts contain 8,322 and 9,377
parameters, respectively; neither parameter counts nor
operation counts are matched. Final affine outputs are
initialized at zero. Both readouts access the same three
full rows of pair states.

Only eight fixed training queries from a single graph are used for optimization.
Each step uses all eight queries and a graph permutation transporting the
frozen cache and anchors. Paired readouts share the permutation stream.
Training uses AdamW for 1,000 steps with learning rate $10^{-3}$, zero weight decay, and gradient clipping at 1.
The six final readout checkpoints and three encoders are fixed before
generating the confirmation cohort, which is used only for testing.

\paragraph{Fresh confirmation protocol.}
We fix the generation and evaluation protocol before drawing the test graphs.
Independent randomized row matchings generate reduced Latin squares, using a fixed seed and a limit of 10,000 attempts.
We retain the first 20 candidates that supply at least 16 queries of each label and have distinct structural invariants from all previously retained or excluded graphs.
The exclusion set contains 107 distinct structures considered during
development, including candidates rejected by the graph generator.
Generation stops after 365 attempts; sampling is non-uniform and conditioned on structural novelty and class coverage, with no model-dependent filtering.
Exact structural signatures certify non-isomorphism, and a common color identification verifies the profiles and labels before inference.
All queries and vertex permutations of the new graphs are used only for testing.

\paragraph{Results and interpretation.}
Across three seeds, the joint readout achieves 100\% accuracy on each of the 20 new graphs, while pairwise pooling remains at its certified
50\% ceiling.
\begin{table}[H]
\centering\small
\setlength{\tabcolsep}{6pt}
\setlength{\abovecaptionskip}{0pt}
\caption{Accuracy (\%) on 640 queries from 20 new graphs, for three
shared encoder seeds.}
\label{tab:cache-access}
\begin{tabular}{@{}lrr@{}}
\toprule
Seed & Pairwise pooling & Joint three-row\\
\midrule
17 & 50 & 100\\
29 & 50 & 100\\
43 & 50 & 100\\
\bottomrule
\end{tabular}
\end{table}
Re-encoding each test graph under three vertex permutations preserves both accuracies.
Raw accuracy at a zero-logit threshold and accuracy assigning half credit when $|\ell|\leq10^{-6}$ agree here; all pairwise-readout logits lie within this tolerance.
The results show that jointly reading $P\times V$ relations from frozen EVT representations recovers information unavailable
through $P\times P$ selection.
As the multiplicity formulas above show, the joint histogram $\mathsf T$ depends only on $t$: each label has the same joint relation histogram across development and confirmation graphs.
The confirmation therefore tests the same information-access contrast on
new, non-isomorphic structures of the same size.
The saved logits and labels support recalculation of the reported
accuracies. Repeating model inference and structural certification
requires separate access to the fitted weights and graph arrays,
respectively.

\section{Path-Marked Learning on a Fixed Latin-Square Benchmark}
\label{app:latin-marked-benchmark}

The experiment here is complementary to
\cref{app:certified-cache-access}: it trains PathMarked and StrikeTwice
encoders on the fixed catalogue split, instead of comparing two heads of
one frozen random cache. The two experiments have different graph cohorts,
training conditions and test interpretations and are reported separately.

Latin-square graphs provide queries with identical selected \(2\)-FWL profiles
and different labels, certifying the
\(\PMarch(\EVT)\)/\(\STarch[\EVT,\EVT]\) separation.
We prove this comparison, then test its numerical realization with four
training conditions on one fixed benchmark.

\paragraph{Graph inputs and labels.}
We use Latin-square graphs of order seven, with strongly regular parameters
\((n,d,\lambda,\mu)=(49,18,7,6)\) \citep{nakasora2006mutually}.
Every vertex has the same constant scalar feature, and binary adjacency
\(A\) is the only pair attribute. For an ordered triangle
\(P=(p_1,p_2,p_3)\), the label is
\(t_G(P)=|N_G(p_1)\cap N_G(p_2)\cap N_G(p_3)|\), restricted to zero or one.
Labels are balanced within every retained graph.
Exact joint unmarked \(2\)-FWL refinement on the benchmark yields three
common color classes: diagonal \(\mathrm D\), adjacent \(\mathrm A\),
and nonadjacent \(\mathrm N\). We use these symbols for the stable
pair colors below.

\subsection{Proof of the neural expressivity separation}
\label{app:latin-marked-proof}

\begin{proof}[Proof of the Latin-square separation]
\textbf{Constant selected profile.}
For constant vertex features and binary adjacency, the initial unmarked
pair colors have three classes: diagonal, adjacent, and nonadjacent.
In a strongly regular graph, the numbers of intermediate vertices in
each ordered pair of relation classes depend only on the receiver's
class and on \((n,d,\lambda,\mu)\).
The parameters \((49,18,7,6)\) therefore give the same three formal
color classes at every refinement round in every graph considered.
For a triangle query, the selected matrix has \(\mathrm D\) on its
diagonal and \(\mathrm A\) elsewhere, so
\begin{equation}
 \Psi_{N,K}(X,Y,P)
 =\left(49,3,
 \begin{pmatrix}
  \mathrm D&\mathrm A&\mathrm A\\
  \mathrm A&\mathrm D&\mathrm A\\
  \mathrm A&\mathrm A&\mathrm D
 \end{pmatrix}\right)
 \label{eq:latin-constant-profile}
\end{equation}
throughout \(\mathcal Q\).
By \cref{thm:main-expressivity}, every \(\STarch[\EVT,\EVT]\) model
has one constant output array on \(\mathcal Q\).
Every deterministic binary readout of that array is constant, giving
accuracy exactly \(1/2\) because the nonempty set is balanced.

\textbf{Recovery by marking.}
The distinct anchors have distinct occurrence marks.
In the refinement multiset at \((p_s,u)\), the entry with intermediate
vertex \(p_a\) is identified by its \(a\)-th mark coordinate.
Its second initial pair color contains \(A_{p_a u}\).
Hence \(\bar c^1_{p_su}\) determines
\(a(u):=(A_{p_1u},A_{p_2u},A_{p_3u})\).
The multiset in \(\bar c^2_{p_sp_s}\) then determines all \(a(u)\)
with their multiplicities, and therefore
\begin{equation}
 t_G(P)=\sum_{u\in V}\one_{\{a(u)=(1,1,1)\}}.
 \label{eq:latin-marked-label}
\end{equation}
The stable marked diagonal retains this information, so the marked
profile determines the label.

\textbf{Neural strictness.}
For \(d_{\mathcal{X}}^{\mathrm{out}}=d_{\mathcal{Y}}^{\mathrm{out}}=1\), set
\(f_t(X,Y,P)=((t_G(P),t_G(P),t_G(P))^\top,0_{3\times3\times1})\).
On the finite query set, \(f_t\) is continuous and length preserving.
It agrees on equal marked profiles, but takes different values on the
constant selected profile because both labels occur.
Thus
\(f_t\in\mathcal C_{\Psi^{\mathrm{mark}}}(\mathcal Q)
\setminus\mathcal C_\Psi(\mathcal Q)\).
The finite neural interpolation of \cref{prop:marked-strict} supplies
one \(\PMarch(\EVT)\) model realizing the three vertex
labels and zero pair outputs, with
shared parameters over the full bounded-size domain.
This proves strictness of the approximation-class inclusion and the
claimed neural separation.
\end{proof}

\subsection{Numerical realization on a fixed benchmark}
\label{app:latin-numerical}

\paragraph{Collection and fixed splits.}
We start from all 147 order-seven main-class representatives in McKay's
public catalogue.\footnote{\raggedright\url{https://users.cecs.anu.edu.au/~bdm/data/latin.html};
order-seven collection downloaded on 6 October 2026.}
Two representatives have no triangle of label one; all 145 eligible
representatives are retained. Distinct exact adjacency invariants certify
that their graphs are pairwise non-isomorphic. Each graph supplies up to
16 distinct triangles per label, with a sampled ordering for each triangle.
Four graphs have fewer available triangles of one label and supply
8, 4, 8 and 12 queries per label, respectively; no queries are duplicated
to reach the nominal budget. Graphs are assigned to fixed 60/20/20 splits
with data seed 20261006 (\cref{tab:latin-splits}). All queries and
vertex relabelings of a graph remain in its split. The models receive only
\((X,Y,P)\): Latin-square coordinates, symbols and graph identities are
retained solely as audit metadata.

\begin{table}[H]
\centering\small
\begin{tabular}{@{}lrrrr@{}}
\toprule
Split & Graphs & Queries & Label 0 & Label 1\\
\midrule
Train & 87 & 2,744 & 1,372 & 1,372\\
Validation & 29 & 904 & 452 & 452\\
Test & 29 & 928 & 464 & 464\\
\midrule
Total & 145 & 4,576 & 2,288 & 2,288\\
\bottomrule
\end{tabular}
\caption{Fixed order-seven benchmark. Graphs, rather than individual
queries or vertex relabelings, define the split unit.}
\label{tab:latin-splits}
\end{table}

All catalogue graphs are eligible for the split, including those used
in exploratory diagnostics. We fix hyperparameters after these diagnostics
and before training the twelve fits reported here. Each fit holds out
the test graphs.

\paragraph{Architectures and vertex readout.}
Both models use pre-LayerNorm, GELU EVT layers with hidden width 24,
four triangular heads (one per orientation), FFN width 48 and no dropout.
Vertex and pair inputs have separate learned affine projections; no input
channel is protected or restored after an update, and residual maps use
the library initialization.
\PMEVT{} applies three layers to the graph with occurrence marks.
A shared \(27\to48\to1\) GELU MLP reads
\(x^{\mathrm M}_{p_s}\concat\phi(s)\) and produces one scalar logit per path
occurrence. Here \(X^{\mathrm M}\) denotes the query-conditioned graph
states before selection. This control inserts marks through the vertex
input and uses a vertex-only task head.
\(\mathrm{ST}[\EVT,\EVT]\), the practical two-EVT implementation of StrikeTwice,
applies three layers to the unmarked graph and three to the selected
occurrence graph, using \(\widetilde X[P]\), \(\widetilde Y[P,P]\),
position codes and the repeated-vertex indicator. A shared affine task-specific layer
produces one scalar logit from each final occurrence vertex.
Both models average the three occurrence logits for the query prediction;
the main task-specific layers do not read pair states directly.
The two loss conditions within an architecture have identical models and
initial weights. Including their training-only auxiliary layers, \PMEVT{}
and \(\mathrm{ST}[\EVT,\EVT]\) contain 84,624 and 166,722 parameters, respectively;
the comparison across architectures is not matched in parameter count.

\paragraph{Auxiliary supervision without changing inference.}
The intermediate objective predicts the same 147 adjacency bits per query,
\(A_{u,p_s}\), for \(u\in[49]\) and \(s\in[3]\).
For \PMEVT{}, a \(24\to32\to3\) GELU MLP predicts the three bits from
each final marked vertex state \(x_u^{\mathrm M}\).
For \(\mathrm{ST}[\EVT,\EVT]\), a shared \(24\to32\to1\) GELU MLP predicts each bit from
\(\widetilde y_{u,p_s}\).
The auxiliary loss is mean binary cross-entropy (BCE) over these bits.
Each adjacency column contains 18 positive and 31 negative entries;
auxiliary evaluation reports balanced accuracy.
The targets are computed from input adjacency for the loss; no auxiliary
output enters the main task-specific layer. \(\mathrm{ST}[\EVT,\EVT]\)'s graph encoder
receives no path mark, and the additional pair-state access occurs only
in the training loss. Its main inference retains the ordered-selection
interface above.

\paragraph{Training and checkpoint selection.}
All four conditions use 5,000 updates with seeds 17, 29 and 43.
Main-loss-only fits use BCE throughout. Auxiliary fits use 200 updates of
auxiliary BCE alone, followed by 4,800 updates of main BCE plus auxiliary
BCE, with coefficient one.
All fits use AdamW with learning rate \(10^{-3}\), zero weight decay,
and gradient clipping at one. Each minibatch contains four pairs of
opposite-label queries (eight queries): a training graph is sampled
uniformly for each pair, then a query pair is sampled within that graph.
A random vertex relabeling is applied at every update.
At a fixed seed, all four conditions share the minibatch and relabeling
streams, and the two losses within an architecture share initial weights.
All models are initialized at random.

Validation main BCE is evaluated every 250 updates; its minimum selects
the checkpoint, with the earliest retained in an exact tie.
The initial checkpoint is eligible; auxiliary-only checkpoints are not.
Every fit completes its full update budget. The chosen weights are saved
before test evaluation, and test scores do not select checkpoints.
Fits run on CPU, with two concurrent workers and two threads per worker.

\paragraph{Final evaluation.}
Training uses float32; final evaluation uses float64 to avoid rounding
effects near the decision threshold.
The saved weights are cast to float64 and evaluated without further
optimization or checkpoint selection. Validation histories and checkpoint
selection retain their original float32 measurements.
Main accuracy is averaged equally over graphs, assigning half credit
when the absolute logit is at most \(10^{-6}\); this tolerance is fixed
before training. Means and sample standard deviations are over the three
training seeds on the same split. The seeds measure optimization
variability under this protocol rather than uncertainty over a graph
population. Structural certificates are used for dataset validation;
their colors and counts are not supplied to the neural models.

\paragraph{Recorded results.}
\Cref{tab:latin-realization} reports the selected checkpoints on the 29 test graphs.
\begin{table}[H]
\centering\small
\setlength{\tabcolsep}{5pt}
\begin{tabular}{@{}llrrrrr@{}}
\toprule
Model & Auxiliary loss & Seed 17 & Seed 29 & Seed 43 & Mean $\pm$ SD & Mean BCE\\
\midrule
\PMEVT{} & No & 50.00 & 50.00 & 50.00 & $50.00\pm0.00$ & 0.693147\\
\PMEVT{} & Yes & 100.00 & 100.00 & 100.00 & $100.00\pm0.00$ & 2.59952e-07\\
\(\mathrm{ST}[\EVT,\EVT]\) & No & 50.00 & 50.00 & 50.00 & $50.00\pm0.00$ & 0.693147\\
\(\mathrm{ST}[\EVT,\EVT]\) & Yes & 50.00 & 50.00 & 50.00 & $50.00\pm0.00$ & 0.693147\\
\bottomrule
\end{tabular}
\caption{Test accuracy (\%) and BCE after validation-based checkpoint selection.
All conditions use the fixed benchmark, three seeds and 5,000 updates per fit.}
\label{tab:latin-realization}
\end{table}

\PMEVT{} with auxiliary supervision has test adjacency balanced accuracy $100.00\pm0.00$\%.
\(\mathrm{ST}[\EVT,\EVT]\) with auxiliary supervision has test adjacency balanced accuracy $100.00\pm0.00$\%.
Auxiliary supervision enables \PMEVT{} to learn the task under this
protocol. \(\mathrm{ST}[\EVT,\EVT]\) learns the intermediate adjacency bits, but its
main inference still receives the constant selected profile and remains
at the certified 50\% ceiling. This distinguishes an optimization
difficulty for \PMEVT{} from StrikeTwice's information limit.
These measurements describe finite-capacity training on one fixed split;
they do not establish a ranking across training protocols.
Separate width and depth controls reuse these twelve fits and add 48
fits under the same scientific protocol (\cref{app:latin-capacity});
they do not replace the main comparison.

\begin{figure}[H]
\centering
\includegraphics[width=\linewidth]{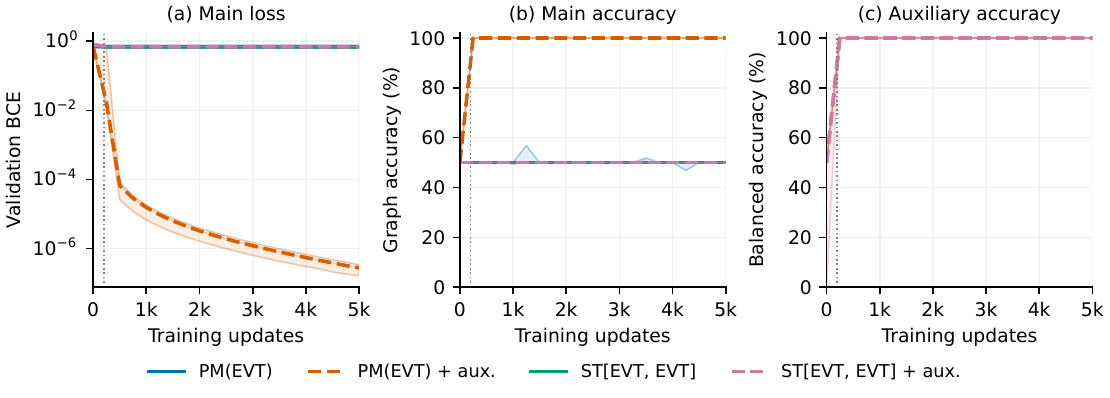}
\caption{Recorded float32 validation histories on the fixed Latin-square benchmark.
\textbf{(a)} Main BCE (log scale). \textbf{(b)} Graph-averaged main accuracy.
\textbf{(c)} Auxiliary adjacency balanced accuracy for supervised conditions.
Thin lines show seeds; thick lines their median, and shading their range.
The vertical dotted line marks the end of the 200-update auxiliary-only warmup.
Checkpoint selection uses the recorded float32 validation main BCE;
final checkpoint scores are evaluated in float64.}
\label{fig:latin-validation}
\end{figure}

\paragraph{Evaluation evidence.}
The protocol was fixed before training. We retain the exact structural
certificates, float64 query predictions and float32 validation histories
for the twelve evaluations. Each float64 evaluation uses the checkpoint
selected in float32, without retraining or checkpoint reselection.

\section{Latin-Square Capacity Controls}
\label{app:latin-capacity}

We extend the numerical realization in \cref{app:latin-numerical} along
one capacity axis at a time: widths $24,32,48$ at depth three, and depths
$1,3,5$ at width 24. Depth denotes the number of graph layers for
\PMEVT{} and the number of layers in \emph{each} of the two EVT stages in \(\mathrm{ST}[\EVT,\EVT]\).
The width-24, depth-three condition shares the twelve fits of
\cref{app:latin-numerical}. The other four settings contribute 48 fits,
for a total of 60 evaluated predictors.

\paragraph{Training and evaluation protocol.}
Every setting uses the same 145 graphs, graph-disjoint $87/29/29$ split,
ordered queries, seeds $17,29,43$, four loss/model conditions and
5,000-update budget as \cref{app:latin-numerical}.
Heads, input processing, auxiliary targets and optimizer settings are
unchanged except for the dimensions induced by width or depth; FFN width
remains twice hidden width. Validation main BCE selects the checkpoint
under the same rule. Training and selection use float32. The
selected arrays are subsequently cast to float64 for evaluation, without
optimization or checkpoint reselection. Means and sample SDs measure
three-seed optimization variability on this fixed split.
Each fit uses two CPU threads.

\begin{table}[H]
\centering\small
\setlength{\tabcolsep}{4pt}
\begin{tabular}{@{}rrrrrrrr@{}}
\toprule
& & \multicolumn{2}{c}{Parameters} & \multicolumn{2}{c}{Main test accuracy (\%)} & \multicolumn{2}{c}{Aux. balanced accuracy (\%)}\\
\cmidrule(lr){3-4}\cmidrule(lr){5-6}\cmidrule(l){7-8}
Width & Depth & \PMEVT{} & ST & \PMEVT{}+aux & ST+aux & \PMEVT{}+aux & ST+aux\\
\midrule
24 & 3 & 84,624 & 166,722 & 100 & 50 & 100 & 100\\
32 & 3 & 147,856 & 291,898 & 100 & 50 & 100 & 100\\
48 & 3 & 326,928 & 646,698 & 100 & 50 & 100 & 100\\
24 & 1 & 29,848 & 57,170 & 50 & 50 & 55.56 & 100\\
24 & 5 & 139,400 & 276,274 & 100 & 50 & 100 & 100\\
\bottomrule
\end{tabular}

\caption{One-axis Latin capacity controls. Here \(\mathrm{ST}\) denotes \(\mathrm{ST}[\EVT,\EVT]\). All displayed accuracies are
three-seed means with zero sample SD. Both main-loss-only models remain
at $50\%$ for every setting; the main train accuracies equal the test
accuracies shown. Auxiliary columns evaluate the training-only adjacency
objective at the main-validation-selected checkpoint. Parameter counts
include auxiliary task-specific layers; capacities are not parameter matched across
architectures. The first row uses the twelve fits of the reference condition.}
\label{tab:latin-capacity}
\end{table}

\paragraph{Interpretation and scope.}
Auxiliary-trained \PMEVT{} reaches $100\%$ main accuracy at depths three
and five for all tested widths. \(\mathrm{ST}[\EVT,\EVT]\) remains at $50\%$ with either loss at
every capacity, despite $100\%$ auxiliary balanced accuracy in all
auxiliary-trained \(\mathrm{ST}[\EVT,\EVT]\) fits. Thus additional width or depth does not remove
the selected-profile limitation in these implementations, and learning
the auxiliary task does not by itself overcome that limitation.
At depth one, \PMEVT{} remains at $50\%$ main accuracy and $55.56\%$
auxiliary balanced accuracy. This finite-budget failure does not establish
a universal minimum neural depth: the exact separation is proved in
\cref{app:latin-marked-proof}, whereas these controls test its learned
realization for five specific capacities. The three seeds and shared split
do not establish population confidence intervals or rule out other
optimization procedures.

\paragraph{Evaluation evidence.}
The 60 selected-checkpoint evaluations produce 274,560 float64 main
logits for independent rescoring. Repeating inference requires the
fitted weights.

\section{Synthetic Experiments}
\label{app:synthetic-experiments}
After the shared experimental setup, we describe the controls for
path composition in Task~1 and for path alternatives in Task~2.
The cache scaling study follows the task-specific results.

\subsection{Shared experimental setup}
\label{app:synthetic-common}
\label{app:synthetic-complete-results}
\label{app:experimental-details}
The two tasks use the same graph generator and query sampler. Their
targets and task-specific controls are described in
\cref{app:t1-mechanism,app:t2-path-alternatives}.

\subsubsection{Graph sampling and input channels}
\label{app:synthetic-inputs}
The generator has seven layers of four vertices. Edges can join only
consecutive layers, in the forward direction. Each candidate edge is initially
drawn independently with probability \(0.40\). Within each consecutive-layer
block, every empty outgoing row is repaired by adding an edge to a uniformly
chosen vertex in the next layer; every remaining empty incoming column is
then repaired with a uniformly chosen source. Proposals whose undirected
skeleton is disconnected are rejected. Thus \(0.40\) is the initial sampling
probability, not the final edge density.

Before task-specific query selection, all eight vertex channels are drawn
independently from \(\mathcal N(0,1)\). They are nuisance features, with no
layer labels or target-derived statistics. For every ordered pair, the four
input channels, in order, are
\[
 y_{ij}=\bigl(A_{ij},\ A_{ij}w_{ij},\ A_{ij}\eta^{(1)}_{ij},\
 A_{ij}\eta^{(2)}_{ij}\bigr),\qquad
 w_{ij}=\log(1+e^{g_{ij}})+0.01,
\]
where \(g_{ij},\eta^{(1)}_{ij},\eta^{(2)}_{ij}\) are independent standard
normal draws, independent of the vertex-feature draws and initial adjacency.
Thus the second channel is the positive edge weight \(B_{ij}=A_{ij}w_{ij}\);
the last two are Gaussian edge nuisance features. All four channels vanish
on nonedges, including self-pairs. These input arrays are shared by the
queries of a graph; position codes and query marks are added by the model
interfaces, not included in these channels.

For Task 1 and Task 2, four distinct full-length paths are sampled by random walks:
the starting vertex is uniform in the first layer, and each next vertex is
uniform among the current vertex's successors. Duplicate paths are rejected;
this is not uniform sampling over all full paths. Finally, a
seed-determined random permutation is applied consistently to vertices,
features and queries. Both tasks use noiseless labels computed from
the resulting graph and path.

\subsubsection{Data and optimization}
\label{app:hyperparameters}

The fixed-rate fits on the Task~1 reference dataset and the Task~2
comparison (\cref{tab:t1-graph-generations,tab:t2-fixed-rate-transformer}) use AdamW
with cosine learning-rate decay from \(3\times10^{-4}\) to \(0.05\) of that
value, weight decay \(10^{-4}\), and gradient clipping at \(5\). The
training and evaluation batch sizes are \(64\) and \(256\), respectively.
Checkpoint selection requires a validation MSE improvement of \(10^{-4}\).
The Task 1 and frozen-cache Task 2 controls specify their settings in
\cref{app:t1-mechanism,app:t2-frozen-cache}.

The nine configurations in the common screen in
\cref{tab:task2-model-comparison} use model-specific initial learning rates selected from
$\{10^{-4},3\times10^{-4},10^{-3}\}$ on three tuning seeds (100--102),
followed by five evaluation seeds (0--4). Every fit uses 2,000 updates,
80 warmup updates followed by cosine decay to 5\% of the initial rate,
training and evaluation batches of 64, and validation every 100 updates.
Checkpoints minimize validation MSE with no minimum-improvement threshold.
AdamW, weight decay and gradient clipping are unchanged.
The vertex-attribute and graph-attention controls use the same batch sizes
and checkpoint rule, at a fixed initial rate of $3\times10^{-4}$
(\cref{app:t2-common-lr,app:t2-attribute-controls,app:t2-graph-ablation}).

The Task 1 frozen-reader controls instead use training batches of 64 and
evaluation batches of 128, with 80 warmup updates followed by a constant
learning rate. Validation is evaluated every 100 updates; the checkpoint
with the lowest validation MSE within each budget (800 or 3,200 updates)
is selected, with no minimum-improvement threshold and the earlier
checkpoint retained on ties. Reader-specific rate selection and the
separate encoder fits are detailed in \cref{app:t1-frozen-cache}.

Target standardization uses training labels only. Each graph supplies four
queries of fixed length \(k=7\), using the sampler above.
The optimization, validation, and test sets contain \(512\), \(128\), and \(256\) graphs, respectively.
Task 2 uses a larger normalization pool, detailed in \cref{app:t2-path-alternatives}. The settings that differ by study are listed below; no test metric enters stopping or model selection.

\begin{table}[H]
\centering
\setlength{\abovecaptionskip}{0pt}
\caption{Fixed-rate optimization protocol for the Task 1 reference dataset and Task 2.
``Patience'' counts
validation evaluations after the minimum-update threshold.
MSE denotes mean squared error.}
\label{tab:appendix-hyperparameters}
\scriptsize
\begin{tabular}{@{}lrr@{}}
\toprule
Setting & Task 1 & Task 2\\
\midrule
Optimization seeds & \(10\) & \(5\)\\
Maximum updates & \(800\) & \(2000\)\\
Minimum updates & \(400\) & \(2000\)\\
Warm-up updates & \(80\) & \(80\)\\
Validation interval & \(40\) & \(100\)\\
Patience & \(10\) & \(100\)\\
Objective & scalar MSE & scalar MSE\\
\bottomrule
\end{tabular}
\end{table}

Every attention block uses four heads, linear input projections, two graph or pair layers, and two path/occurrence layers where that stream exists.  The practical EVT implementation uses LayerNorm and separate pair initialization; \STEE{} adds absolute endpoint one-hot codes and a scalar relative offset in its second stage.
The endpoint-aware rotary position embedding (RoPE) construction in \cref{app:evt-rope} is a theoretical extension and is not used by these checkpoints.
Capacity matching changes only the hidden/FFN width: PathTransformer \(112\), GINE-Path \(104\), the two
single-pass Edge Transformer path models \(80\), \(\STarch[\ET,\MLP_{\mathrm{concat}}]\) \(88\),
ET-Twice \(72\), EVT-Path and \PMEVT{} \(40\), and \STEE{} \(32\).

\subsubsection{Models and parameter matching}
\label{app:experimental-interfaces}
PathTransformer uses only the ordered path. GINE-Path \citep{hu2019strategies} applies a standard
message-passing graph encoder before the same path stack; the variant in \cref{tab:t2-fixed-rate-transformer}
concatenates the raw selected-vertex features to the GINE embeddings
before path projection, so message passing cannot irreversibly erase the raw
path signal.  The
faithful Edge
Transformer uses endpoint-lifted pair tokens and classical chain-oriented
triangular attention; its direction-mixed counterpart allocates heads across
all four contractions in \cref{eq:four-pair-contractions}. EVT-Path reads the
path from one graph EVT. \STEE{} uses complete ordered \(P\times P\)
selection and adds the positional fields before the second EVT.
\PMEVT{} marks the graph before EVT encoding, which is repeated for each
new query.
\(\STarch[\ET,\MLP_{\mathrm{concat}}]\) and ET-Twice use a balanced global Edge Transformer and the same
complete ordered selection. The final pair state at \((i,i)\)
supplies the vertex representation for \(i\). \(\STarch[\ET,\MLP_{\mathrm{concat}}]\) applies pointwise
pair MLPs with masked row means, and ET-Twice applies a second balanced
Edge Transformer to the selected pairs. Both retain ReLU in their ET blocks.
\(\STarch[\ET,\MLP_{\mathrm{concat}}]\) concatenates each selected vertex state with the row mean
of the updated pair states before a vertex MLP. It uses GELU in the pair and
vertex MLPs after selection. Its balanced, width-88 encoder differs from the
chain-only, width-80 biological \(\STarch[\ET,\MLP_{\mathrm{concat}}]\) configuration
(\cref{app:biology-models}).
\STEE{} uses the balanced allocation, with one of its four triangular
heads assigned to each contraction in
\cref{eq:four-pair-contractions}.
Widths and feed-forward dimensions are predeclared. The models in the
fixed-rate Task 2 comparison in \cref{tab:t2-fixed-rate-transformer} are
within \(5\%\) of \STEE{}'s parameter count. The matched Task 1 intervention and its
small FFN compensation are detailed in \cref{app:t1-mechanism}.

\subsubsection{Additive MLP configurations and attention controls}
\label{app:synthetic-flexibility}

\paragraph{Data and fit budgets.}
The comparisons use the \(512/128/256\) graph split in
\cref{app:hyperparameters} with four queries per graph. Task~1 uses
3,200 updates per fit and ten evaluation seeds (0--9). Task~2 uses
2,000 updates and five evaluation seeds (0--4). The reference fits in
\cref{tab:t1-main-statistics,tab:t2-fixed-rate-mlp-reference} supply
the unmodified models for the attention controls.

\paragraph{Architectures and optimization.}
\(\mathrm{ST}[\EVT,\MLP_{\mathrm{add}}]\) uses two graph EVT layers (width 32, FFN width 128) followed by an
EVT-derived MLP processor (FFN width 1,182): a residual pair MLP, normalized uniform
row aggregation and a residual vertex MLP. \(\mathrm{ST}[\EVT,\EVT]\) uses two full graph and two
full path EVT layers (width 32, FFN width 128). \(\mathrm{ST}[\ET,\MLP_{\mathrm{add}}]\) uses two ET-like graph
layers implemented in the shared EVT schedule (width 80, FFN width 279),
with endpoint lifting, chain-oriented triangular attention, pairwise ReLU
FFNs and diagonal vertex proxies, followed by the row-aggregation MLP
processor (FFN width 414). Thus \(\mathrm{ST}[\ET,\MLP_{\mathrm{add}}]\) differs from \(\mathrm{ST}[\EVT,\MLP_{\mathrm{add}}]\) in more than attention
orientation. All models give the mean path-vertex state to a scalar MLP and retain
absolute and relative path positions. Attention controls keep the other operations and widths fixed and vary
the parameter counts (\cref{tab:synthetic-flexibility-fixed}). A full EVT
path layer with triangular attention disabled retains its vertex--pair
exchanges. The additive reader uses a source-row mean for pair-to-vertex
aggregation (\cref{app:biology-additive-mlp}).

AdamW uses batches of 64, weight decay $10^{-4}$, gradient clipping at 5,
80 warm-up updates and cosine decay to 5\% of the initial learning rate.
Validation occurs every 40 updates on Task 1 and every 100 on Task 2;
evaluation uses batches of 64, and checkpoint selection minimizes validation
MSE over the full update budget with no minimum-improvement threshold.
The primary comparison fixes the initial LR at $3\times10^{-4}$.
A separate sensitivity analysis selects among
$\{10^{-5},3\times10^{-5},10^{-4},3\times10^{-4},10^{-3},3\times10^{-3}\}$
on the mean best validation MSE of three separate seeds (100--102), before
final training. A single fit supplies both estimates when the selected and fixed rates
coincide. The \(\mathrm{ST}[\EVT,\MLP_{\mathrm{add}}]\), \(\mathrm{ST}[\EVT,\EVT]\) and \(\mathrm{ST}[\ET,\MLP_{\mathrm{add}}]\) experiments contain respectively
206, 196 and 66 distinct fits, including their selection grids.
The ET-based fits use a different numerical software environment.
The two-seed ET checks in \cref{tab:synthetic-flexibility-diagnostics}
give similar validation scores in the numerical environment of the
EVT-based fits.

\begin{table}[t]
\centering\scriptsize
\caption{Synthetic reference configurations and attention controls at fixed
initial LR $3\times10^{-4}$.
Test $R^2$, mean $\pm$ sample SD over ten Task 1 or five Task 2 training
seeds. A dash denotes an untested task--configuration combination.
Vertex attention denotes vertex self-attention; triangular attention updates
pair states. Task 1 uses 3,200 updates and Task 2 uses 2,000.}
\label{tab:synthetic-flexibility-fixed}
\begin{tabular}{lrrr}
\toprule
Configuration & Parameters & Task 1 $R^2$ & Task 2 $R^2$ \\
\midrule
\(\mathrm{ST}[\EVT,\MLP_{\mathrm{add}}]\) & 312,485 & $0.867 \pm 0.054$ & $0.647 \pm 0.055$ \\
\(\mathrm{ST}[\EVT,\MLP_{\mathrm{add}}]\) + path vertex attention & 316,773 & $0.858 \pm 0.055$ & -- \\
\(\mathrm{ST}[\EVT,\MLP_{\mathrm{add}}]\) + path triangular attention & 317,829 & $0.949 \pm 0.012$ & -- \\
\(\mathrm{ST}[\EVT,\MLP_{\mathrm{add}}]\) + both path attentions & 322,117 & $0.947 \pm 0.009$ & -- \\
\(\mathrm{ST}[\EVT,\MLP_{\mathrm{add}}]\) -- graph vertex attention & 303,909 & -- & $0.616 \pm 0.045$ \\
\(\mathrm{ST}[\EVT,\MLP_{\mathrm{add}}]\) -- graph triangular attention & 301,797 & -- & $-0.012 \pm 0.017$ \\
\(\mathrm{ST}[\EVT,\EVT]\) & 312,753 & $0.982 \pm 0.006$ & $0.591 \pm 0.090$ \\
\(\mathrm{ST}[\EVT,\EVT]\) -- path vertex attention & 304,177 & $0.979 \pm 0.012$ & -- \\
\(\mathrm{ST}[\EVT,\EVT]\) -- path triangular attention & 302,065 & $0.908 \pm 0.008$ & -- \\
\(\mathrm{ST}[\EVT,\EVT]\) -- both path attentions & 293,489 & $0.904 \pm 0.011$ & -- \\
\(\mathrm{ST}[\EVT,\EVT]\) -- graph vertex attention & 304,177 & -- & $0.579 \pm 0.079$ \\
\(\mathrm{ST}[\EVT,\EVT]\) -- graph triangular attention & 302,065 & -- & $-0.003 \pm 0.003$ \\
\(\mathrm{ST}[\ET,\MLP_{\mathrm{add}}]\) & 312,507 & $0.466 \pm 0.078$ & $-0.026 \pm 0.018$ \\
\bottomrule
\end{tabular}

\end{table}

\paragraph{Location of the empirical benefit.}
On Task 1, adding path triangular attention to \(\mathrm{ST}[\EVT,\MLP_{\mathrm{add}}]\) improves test $R^2$
from $0.867$ to $0.949$, on all ten seeds. In \(\mathrm{ST}[\EVT,\EVT]\), removing it reduces the
mean from $0.982$ to $0.908$. On Task 2, removing graph triangular attention
reduces \(\mathrm{ST}[\EVT,\MLP_{\mathrm{add}}]\) and \(\mathrm{ST}[\EVT,\EVT]\) means from $0.647$ and $0.591$ to $-0.012$ and $-0.003$,
respectively, with decreases on all five seeds. Removing graph vertex
self-attention has a smaller effect. The effect of the attention intervention depends on the task, processing
stage, and training budget.

\begin{table}[t]
\centering\scriptsize
\caption{Validation-selected-LR sensitivity, separate from the fixed-LR
comparison. Test $R^2$, mean $\pm$ sample SD, with the same final seeds and
update budgets as \cref{tab:synthetic-flexibility-fixed}. Rates are selected
on validation using seeds 100--102. Coincident selected and fixed rates
share one fit.}
\label{tab:synthetic-flexibility-selected}
\begin{tabular}{lrrrr}
\toprule
 & \multicolumn{2}{c}{Task 1} & \multicolumn{2}{c}{Task 2} \\
Configuration & Selected LR & Test $R^2$ & Selected LR & Test $R^2$ \\
\midrule
\(\mathrm{ST}[\EVT,\MLP_{\mathrm{add}}]\) & $3\times10^{-4}$ & $0.867 \pm 0.054$ & $3\times10^{-4}$ & $0.647 \pm 0.055$ \\
\(\mathrm{ST}[\EVT,\MLP_{\mathrm{add}}]\) + path vertex attention & $3\times10^{-4}$ & $0.858 \pm 0.055$ & -- & -- \\
\(\mathrm{ST}[\EVT,\MLP_{\mathrm{add}}]\) + path triangular attention & $10^{-3}$ & $0.956 \pm 0.006$ & -- & -- \\
\(\mathrm{ST}[\EVT,\MLP_{\mathrm{add}}]\) + both path attentions & $10^{-3}$ & $0.948 \pm 0.017$ & -- & -- \\
\(\mathrm{ST}[\EVT,\MLP_{\mathrm{add}}]\) -- graph vertex attention & -- & -- & $3\times10^{-4}$ & $0.616 \pm 0.045$ \\
\(\mathrm{ST}[\EVT,\MLP_{\mathrm{add}}]\) -- graph triangular attention & -- & -- & $3\times10^{-5}$ & $-0.008 \pm 0.005$ \\
\(\mathrm{ST}[\EVT,\EVT]\) & $3\times10^{-4}$ & $0.982 \pm 0.006$ & $3\times10^{-4}$ & $0.591 \pm 0.090$ \\
\(\mathrm{ST}[\EVT,\EVT]\) -- path vertex attention & $10^{-3}$ & $0.912 \pm 0.124$ & -- & -- \\
\(\mathrm{ST}[\EVT,\EVT]\) -- path triangular attention & $3\times10^{-4}$ & $0.908 \pm 0.008$ & -- & -- \\
\(\mathrm{ST}[\EVT,\EVT]\) -- both path attentions & $3\times10^{-4}$ & $0.904 \pm 0.011$ & -- & -- \\
\(\mathrm{ST}[\EVT,\EVT]\) -- graph vertex attention & -- & -- & $3\times10^{-4}$ & $0.579 \pm 0.079$ \\
\(\mathrm{ST}[\EVT,\EVT]\) -- graph triangular attention & -- & -- & $10^{-3}$ & $-0.009 \pm 0.007$ \\
\(\mathrm{ST}[\ET,\MLP_{\mathrm{add}}]\) & $10^{-4}$ & $0.383 \pm 0.051$ & $10^{-4}$ & $-0.009 \pm 0.004$ \\
\bottomrule
\end{tabular}

\end{table}

\paragraph{Learning-rate sensitivity.}
The graph/path localization remains under validation-selected LRs
(\cref{tab:synthetic-flexibility-selected}). A model can have a higher test score with the fixed rate. For example, \(\mathrm{ST}[\EVT,\EVT]\) without path vertex self-attention selects $10^{-3}$ on
Task 1 and reaches $0.912\pm0.124$, versus $0.979\pm0.012$ at the fixed LR.
\(\mathrm{ST}[\ET,\MLP_{\mathrm{add}}]\) selects $10^{-4}$ but remains below \(\mathrm{ST}[\EVT,\MLP_{\mathrm{add}}]\) on Task 1 and near the
constant-prediction baseline on Task 2. These differences measure optimization and generalization at the tested
capacities.

\paragraph{Experimental records.}
Predictions and histories for the 480 fits in the three model families above
support recalculation of MSE, \(R^2\), selected learning rates and paired
differences. The inference-only nuisance analysis retains measurements by
feature draw. The fixed generation recipes preserve the graphs, queries,
splits and full Task~2 normalization pool. Repeating checkpoint inference
requires the fitted weights. Results from new fits can vary with the
numerical environment.

\FloatBarrier
\subsection{Task 1. Path composition and reader controls}
\label{app:t1-mechanism}

Task 1 tests how models compose selected path weights. Its target
(\cref{sec:t1}) averages four products of three consecutive edge
weights on a seven-occurrence path. All required weights are available after
selection.

\begin{table}[!htbp]\centering\small
\caption{Task 1 fixed-rate comparison: test $R^2$ over ten paired seeds at 3,200 updates.
Medians and IQRs complement the means and sample SDs in the main table.}
\label{tab:t1-main-statistics}
\small\setlength{\tabcolsep}{6pt}
\begin{tabular}{@{}lcc@{}}
\toprule
Model & Median test $R^2$ [IQR] & Mean $\pm$ SD\\
\midrule
\(\mathrm{ST}[\EVT,\EVT]\) & $\mathbf{0.984\;[0.979,0.986]}$ & $\mathbf{0.982\pm0.006}$\\
\(\mathrm{ST}[\EVT,\MLP_{\mathrm{add}}]\) & $0.881\;[0.858,0.898]$ & $0.867\pm0.054$\\
\(\mathrm{ST}[\ET,\MLP_{\mathrm{add}}]\) & $0.478\;[0.437,0.505]$ & $0.466\pm0.078$\\
\bottomrule
\end{tabular}

\end{table}

\paragraph{Additive and concatenating MLP path processors.}
We compare the two EVT-based MLP configurations on the same Task~1 graphs,
queries, and splits. Each uses ten training seeds (0--9), 3,200 updates,
an initial learning rate of \(3\times10^{-4}\), and checkpoints selected
by validation MSE. The configurations are trained end to end with different
graph-encoder widths and allocations of parameters across the two stages.

\begin{table}[htbp]
\centering\small
\caption{Task~1 comparison of MLP readers with an EVT graph encoder.
Mean and sample SD of test \(R^2\) over ten training seeds.}
\label{tab:t1-mlp-readers}
\begin{tabular}{@{}lrc@{}}
\toprule
Model & Graph width & Test \(R^2\) (mean \(\pm\) SD)\\
\midrule
\(\STarch[\EVT,\MLP_{\mathrm{add}}]\) & 32 & \(0.8666\pm0.0540\)\\
\(\STarch[\EVT,\MLP_{\mathrm{concat}}]\) & 44 & \(0.8701\pm0.0127\)\\
\bottomrule
\end{tabular}
\end{table}

The two MLP configurations have similar means. In each comparison,
\STEE{} reaches a mean near \(0.982\) with a higher score than the MLP
configuration on every paired seed.

\subsubsection{Triangular and dense path attention}
The three graph datasets in \cref{tab:t1-graph-generations} use CPU fits.
The factorial and longer-budget controls use separate float32 fits with
Apple Metal Performance Shaders (MPS).

\paragraph{Matched intervention.}
\STEE{} and its dense-path control both use width 32, four heads, two graph and two path layers, separate pair initialization, pre-LayerNorm and zero dropout.
Graph encoder, ordered selection, absolute and relative positions, cross-attention structure and scalar readout are retained.
Replacing triangular attention removes one value projection per path layer; the dense control widens the path FFN from 128 to 130 in compensation, giving 312,457 parameters against 312,753.
The comparison thus changes the attention operator together with a small FFN adjustment.

On the reference graph dataset, the two models use seeds \(0\)--\(9\), the same queries and sampling order,
and \(800\) completed updates under the shared protocol
(\cref{app:hyperparameters}). All ten paired gains are positive
(two-sided sign-test \(p=0.002\)). \Cref{tab:t1-graph-generations} reports means and sample
standard deviations across seeds within each dataset; gain statistics are
computed from the paired differences.

\paragraph{Factorial control.}
The graph and path pair operators are varied independently inside the same coupled vertex--pair architecture.
\STEE{} uses balanced triangular attention at both stages. In TD, DT and DD, T denotes balanced triangular attention and D dense self-attention over all pair tokens; the first letter specifies the graph stage and the second the path stage.
All four variants retain the matched intervention's width,
depth, positions, and readout architecture; each dense stage uses the same
FFN compensation.
D retains the EVT vertex stream and cross-attentions, which distinguishes it from the pair-only Edge Transformer baseline.
The two additional MLP-reader controls, PP\(_T\) and PP\(_D\), use the corresponding graph operator followed by pointwise pair MLPs and row means.
They use width \(44\) to match total capacity, with graph FFN width \(176\)
or \(179\), respectively. These controls match total parameter counts by
using wider states than the attention variants.

\begin{table}[htbp] \centering
\setlength{\abovecaptionskip}{0pt}
\caption{Task 1 factorial control: test \(R^2\) over seeds 0--9 at 800 updates and
common learning rate \(3\times10^{-4}\). These are float32 MPS fits;
\cref{tab:t1-graph-generations} reports the CPU comparisons.}
\label{tab:t1-factorial} \small
\begin{tabular}{@{}llllrrr@{}}
\toprule
Variant & Graph & Path & Width & Parameters & Mean \(R^2\) & IQR\\
\midrule
\STEE{} & Triangular & Triangular & 32 & 312,753 & \textbf{.952} & [0.944,0.960]\\
TD & Triangular & Dense & 32 & 312,457 & .884 & [0.867,0.898]\\
DT & Dense & Triangular & 32 & 312,457 & .948 & [0.931,0.961]\\
DD & Dense & Dense & 32 & 312,161 & .851 & [0.842,0.863]\\
PP\(_T\) & Triangular & MLPs / row means & 44 & 313,687 & .750 & [0.733,0.761]\\
PP\(_D\) & Dense & MLPs / row means & 44 & 313,459 & .573 & [0.609,0.720]\\
\bottomrule
\end{tabular}
\end{table}

These runs use the same dataset and optimizer settings as the CPU intervention,
but evaluation batches contain 64 examples instead of 256. Mean paired
differences are $+0.068$ for \STEE{}--TD and $+0.097$ for DT--DD, positive on all
ten seeds: the path stage carries a benefit under either graph operator.
\STEE{} and DT have similar means (difference $+0.004$).

Learning rates $\{5\times10^{-5},10^{-4},3\times10^{-4}\}$ were screened on
validation with seed 0 alone, with a smaller-rate tie-break and no test-based
selection; all six variants selected the predeclared common rate
$3\times10^{-4}$. The ten-seed summary includes the screening seed.

\paragraph{Fourfold training-budget control.}
The longer-budget comparison uses seeds \(1,2,3\), the same dataset,
and the common learning rate \(3\times10^{-4}\), without reselection.
All models are initialized afresh and trained for exactly \(3200\) updates.
The cosine schedule is stretched to the new horizon; warm-up remains
\(80\) updates and validation is still evaluated every \(40\) updates.
The intervention therefore changes both update count and learning-rate schedule.

\begin{table}[htbp] \centering
\setlength{\abovecaptionskip}{0pt}
\caption{Task 1 budget sensitivity on seeds 1--3 in both columns. Mean test \(R^2\)
from validation-selected checkpoints. The 800-update column uses the
corresponding three-seed subset of \cref{tab:t1-factorial}.
Bold marks the best score at each budget.}
\label{tab:t1-budget} \small
\begin{tabular}{@{}lrrr@{}}
\toprule
Variant & 800 updates & 3200 updates & Change\\
\midrule
\STEE{} & .956 & \textbf{.984} & $+.028$\\
TD & .870 & .895 & $+.025$\\
DT & \textbf{.961} & .982 & $+.021$\\
DD & .852 & .855 & $+.003$\\
PP\(_T\) & .727 & .873 & $+.146$\\
PP\(_D\) & .538 & .847 & $+.309$\\
\bottomrule
\end{tabular}
\end{table}

The two contrasts behave differently under a longer budget. The path-operator
contrast persists: \STEE{}--TD moves from $+0.086$ to $+0.089$ and DT--DD from
$+0.109$ to $+0.127$ on these same three seeds, positive throughout. The
gaps to the pointwise controls shrink: \STEE{}--PP\(_T\) falls from $+0.229$
to $+0.112$, while DD--PP\(_D\) almost closes, from $+0.313$ to $+0.008$.
The longer-budget result rests on three seeds; three positive signs give
a two-sided sign test at $p=0.25$.

\paragraph{Resources and statistical scope.}
Each comparison approximately matches parameter counts and uses equal update
counts. FLOPs, wall time, and peak memory can differ.
At fixed width \(d\), dense attention on \(k^2\) occurrence-pair tokens forms \(O(k^4d)\)
attention interactions, versus \(O(k^3d)\) for triangular attention.
Both operators also use \(O(k^2d^2)\) projections.
The corresponding graph-stage terms are \(O(n^4d)\) and \(O(n^3d)\).

The \STEE{}--TD prediction gain therefore accompanies a lower asymptotic
pair-attention cost. At the experimental \(n=28,k=7,d=32\), the forward
operation counts in \cref{tab:t1-compute} are also lower for \STEE{}: by
\(4.3\%\) for the complete path encoder and \(0.9\%\) for the full model
with four queries per graph. The shared graph encoder dominates the total.

\begin{table}[htbp]
\centering
\setlength{\abovecaptionskip}{0pt}
\caption{Task 1 \STEE{}--TD forward operation counts in MFLOPs per query.
The path column includes the two path-encoder layers and input projections.
Totals include graph encoding and the scalar head, with \(Q\) queries
sharing each graph. Parameters and FFN widths match the Task 1 intervention.}
\label{tab:t1-compute}
\small
\begin{tabular}{@{}lrrr@{}}
\toprule
Model & Path encoder & Total, \(Q=1\) & Total, \(Q=4\)\\
\midrule
\STEE{} & \textbf{9.641} & \textbf{163.052} & \textbf{47.995} \\
TD & 10.079 & 163.490 & 48.433\\
\bottomrule
\end{tabular}
\end{table}

The count covers linear projections and matrix products, with two FLOPs per
multiply--add. It excludes softmax, normalization, elementwise operations,
data movement, and backward computation. Random weights suffice because the
count depends on the tensor shapes.

The factorial and longer-budget comparisons share the reference graphs and queries and fixed absolute and
relative position codes. The following control varies the sampled graphs
while retaining these position codes.

\paragraph{Two additional graph generations and a local control.}
\label{app:t1-new-graphs}
The two resamplings in \cref{tab:t1-graph-generations} use the same weighted-DAG generator, each with
512 training, 128 validation, and 256 test graphs and four queries per graph.
Graph fingerprints have no overlap within or between these datasets or
with the reference dataset. All three datasets sample the same distribution.
\STEE{} and TD retain the matched architectures above. A local MLP applies a
shared $3\!\to\!64\!\to\!64\!\to\!1$ network with Gaussian error linear unit (GELU) activations to
each window of three consecutive raw path-edge weights and averages the
four outputs. It receives no products or off-path states. Its windowing and
averaging are informed by the target; its 4,481 parameters are not matched
to \STEE{}'s 312,753 or TD's 312,457.

On each resampling, all three models use seeds 0--2 and exactly 800 CPU updates at initial
learning rate $3\times10^{-4}$, with 80 warm-up updates and cosine decay
to 5\% of that rate. Other optimizer settings follow
\cref{app:hyperparameters}. This control uses one CPU thread, evaluation
batches of 32, validation every 40 updates, and selection on any strict
validation-MSE improvement (threshold zero). Each dataset's training labels
determine its target standardization. The reference dataset comparison uses
evaluation batches of 256 and a validation-improvement threshold of
\(10^{-4}\). We summarize the datasets separately to retain their variation
and avoid giving the ten-seed dataset more weight than either three-seed
resampling. At this budget, the local MLP was evaluated on the two resamplings.

\begin{table}[htbp]
\centering\small
\setlength{\abovecaptionskip}{0pt}
\caption{Task~1 test $R^2$ on three graph datasets at 800 updates:
mean $\pm$ sample SD across the seeds in each row. $\Delta$ is the
paired difference (triangular minus dense path attention), computed
before rounding. Bold: highest mean in each row; --: not evaluated.}
\label{tab:t1-graph-generations}
\setlength{\tabcolsep}{3pt}
\begin{tabular*}{\linewidth}{@{\extracolsep{\fill}}lrcccc@{}}
\toprule
Graph dataset & Seeds & $\mathrm{ST}[\EVT,\EVT]$ & $\mathrm{ST}[\EVT,\EVT_{\mathrm{D}}]$ & $\Delta$ & Local MLP\\
\midrule
Original & 10 & $\mathbf{.953\pm.011}$ & $.884\pm.019$ & $+.068\pm.023$ & --\\
Resampling 1 & 3 & $\mathbf{.954\pm.008}$ & $.872\pm.022$ & $+.082\pm.023$ & $.892\pm.023$\\
Resampling 2 & 3 & $\mathbf{.912\pm.009}$ & $.862\pm.009$ & $+.050\pm.009$ & $.865\pm.023$\\
\bottomrule
\end{tabular*}

\end{table}

\STEE{} exceeds TD on every paired seed in both datasets. The local MLP also
captures much of the target variation, showing that Task 1 can be learned with
a small model given the appropriate local decomposition. The \STEE{}--TD result
shows a benefit of triangular composition within the matched architecture
and training budget, while the local MLP offers another effective solution.
All these comparisons use the fixed position codes described above.

\paragraph{Local MLP on the reference dataset.}
We train the local MLP on the reference graphs and queries with
the training protocol of the StrikeTwice configurations in
\cref{tab:task1-model-comparison} (\cref{app:synthetic-flexibility}).
We retain the target standardization, use seeds \(0\)--\(9\) and
3,200 AdamW updates per seed with batches of 64, and select the first
checkpoint attaining the minimum validation MSE over evaluations
every 40 updates. The ten fits use the same 1,024 test queries and
targets as the StrikeTwice fits. The mean test \(R^2\) is \(0.9817\),
with a sample standard deviation of \(0.0019\) across the ten seeds.

\subsubsection{Sensitivity to vertex-feature resampling}
\label{app:t1-vertex-resampling}

We check layer parity with the reference ET implementation, checkpoint
compatibility, gradient finiteness, and repeatability of predictions.
On Task 1, inference-only resampling of the eight iid
normal vertex channels changes \(\mathrm{ST}[\ET,\MLP_{\mathrm{add}}]\) predictions more than \(\mathrm{ST}[\EVT,\MLP_{\mathrm{add}}]\) predictions
on all ten training seeds. Mean prediction-change RMSE is $0.603\pm0.036$
versus $0.328\pm0.071$ on 64 training graphs, and $0.523\pm0.047$ versus
$0.289\pm0.047$ on all 128 validation graphs. This uses query 0 of each graph,
three paired feature draws per checkpoint, fixed LR $3\times10^{-4}$ and
no refitting or test evaluation. It suggests greater sensitivity to nuisance
features at inference.

\begin{table}[!htbp]
\centering\footnotesize
\caption{Task~1 vertex-feature resampling over ten training seeds.
Prediction-change RMSE compares original and redrawn-$X$ predictions
(mean $\pm$ sample SD across seeds); the last two columns are mean MSE
against the fixed standardized targets. Each checkpoint receives three
paired feature draws using query~0 of each graph. The evaluation uses the
fixed weights on training and validation graphs.}
\label{tab:t1-nuisance-sensitivity}
\setlength{\tabcolsep}{4pt}
\begin{tabular}{@{}llrrr@{}}
\toprule
Partition & Model & Prediction-change RMSE & Original MSE & Redrawn-$X$ MSE\\
\midrule
Train (64 graphs) & $\mathrm{ST}[\ET,\MLP_{\mathrm{add}}]$ & $0.603\pm0.036$ & $0.0193$ & $0.3705$\\
Train (64 graphs) & $\mathrm{ST}[\EVT,\MLP_{\mathrm{add}}]$ & $0.328\pm0.071$ & $0.0053$ & $0.1183$\\
\midrule
Validation (128 graphs) & $\mathrm{ST}[\ET,\MLP_{\mathrm{add}}]$ & $0.523\pm0.047$ & $0.5322$ & $0.5464$\\
Validation (128 graphs) & $\mathrm{ST}[\EVT,\MLP_{\mathrm{add}}]$ & $0.289\pm0.047$ & $0.1447$ & $0.1514$\\
\bottomrule
\end{tabular}

\end{table}

The inference perturbation increases mean validation MSE from $0.5322$
to $0.5464$ for \(\STarch[\ET,\MLP_{\mathrm{add}}]\) and from $0.1447$
to $0.1514$ for \(\STarch[\EVT,\MLP_{\mathrm{add}}]\).
The change in predictions measures sensitivity to the nuisance features,
and the validation MSE measures error against the fixed target.

\paragraph{Vertex features during training.}
The five-seed Task~2 control in \cref{app:t2-vertex-input-ablation}
uses the same kind of nuisance channels with a fixed LR. Setting the
channels to zero during training and evaluation improves
\(\STarch[\ET,\MLP_{\mathrm{add}}]\) test scores. Task~1 measures
sensitivity of fixed weights to redrawn inputs. Task~2 measures sensitivity
when the input intervention also changes the optimization trajectory.

\subsubsection{Path attention on caches from two encoder configurations}
\label{app:t1-frozen-cache}
We refit two EVT graph encoders for 800 updates with seed~0, one at
width~44 with the concatenating MLP path processor and one at width~32
with the EVT path processor. For each fit we select a checkpoint by
validation MSE and freeze the graph encoder. For each source cache,
we train fresh two-layer path processors at widths 32 and 64: full EVT,
EVT with only the triangular-attention update removed (EVT-), and the latter
with wider feed-forward layers to approximately compensate for the
removed parameters. The remaining operations, prediction head, input
representations and minibatch order are unchanged within each comparison.

Each width and variant receives an independent search over learning rates
$\{10^{-4},3\times10^{-4},10^{-3}\}$, using tuning seed~100 and
800 updates. We select the rate by minimum validation MSE, then fit
three paired evaluation seeds (11--13) for 3,200 updates, with batch size
64, weight decay $10^{-4}$, gradient clipping at~5, and 80 warmup updates
followed by a constant rate. Validation is evaluated every 100 updates.
The two budgets select checkpoints within the first 800 or 3,200 updates
of the same trajectory. The graph-disjoint splits and targets follow \cref{app:hyperparameters}.

\Cref{tab:t1-frozen-config} gives the reader configurations and selected
rates, which are shared across final seeds and both budget prefixes.
Each reader has two EVT layers with four attention heads. Parameter
counts include its input projections and prediction head, but exclude
the frozen graph encoder; different source-cache widths change the
input-projection counts. The width-44 and width-32 checkpoints are
selected at updates 800 and 760, respectively.

\begin{table}[!htbp]
\color{black}
\centering\footnotesize
\setlength{\tabcolsep}{4pt}
\caption{Task 1 frozen-reader configurations. FFN is the feed-forward
hidden width; parameter counts include all registered reader parameters.
LR is selected independently for each source, reader width and variant.}
\label{tab:t1-frozen-config}
\begin{tabular}{@{}rlrrrrr@{}}
\toprule
& & & \multicolumn{2}{c}{Width-44 cache} & \multicolumn{2}{c}{Width-32 cache}\\
\cmidrule(lr){4-5}\cmidrule(l){6-7}
Width & Reader & FFN & Parameters & LR & Parameters & LR\\
\midrule
32 & Full EVT & 128 & 158,889 & $10^{-3}$ & 158,121 & $10^{-3}$\\
32 & EVT- & 128 & 148,201 & $10^{-3}$ & 147,433 & $10^{-3}$\\
32 & EVT- + wider FFNs & 140 & 159,097 & $10^{-3}$ & 158,329 & $10^{-3}$\\
\midrule
64 & Full EVT & 256 & 618,825 & $10^{-3}$ & 617,289 & $3\times10^{-4}$\\
64 & EVT- & 256 & 576,969 & $3\times10^{-4}$ & 575,433 & $3\times10^{-4}$\\
64 & EVT- + wider FFNs & 279 & 618,461 & $10^{-3}$ & 616,925 & $10^{-3}$\\
\bottomrule
\end{tabular}
\end{table}

\begin{table}[!htbp]
\color{black}
\centering\small
\setlength{\tabcolsep}{4pt}
\caption{Task 1 with fixed graph representations: test $R^2$, mean
$\pm$ SD over three paired reader seeds. ``Wider FFNs'' compensates
for removing triangular attention. Each source is one refitted, frozen
encoder; seed variation is conditional on that encoder.}
\label{tab:t1-frozen-cache}
\begin{tabular}{@{}lrrccc@{}}
\toprule
Encoder width & Reader width & Updates & Full EVT & EVT- & EVT- + wider FFNs\\
\midrule
44 & 32 & 800 & $\mathbf{.8935\pm.0058}$ & $.8425\pm.0115$ & $.8374\pm.0128$\\
44 & 32 & 3,200 & $\mathbf{.9065\pm.0091}$ & $.8493\pm.0022$ & $.8507\pm.0011$\\
44 & 64 & 800 & $\mathbf{.8945\pm.0120}$ & $.8417\pm.0127$ & $.8404\pm.0111$\\
44 & 64 & 3,200 & $\mathbf{.9018\pm.0067}$ & $.8556\pm.0032$ & $.8459\pm.0043$\\
\midrule
32 & 32 & 800 & $\mathbf{.9683\pm.0127}$ & $.8781\pm.0091$ & $.8605\pm.0244$\\
32 & 32 & 3,200 & $\mathbf{.9775\pm.0083}$ & $.8659\pm.0043$ & $.8732\pm.0120$\\
32 & 64 & 800 & $\mathbf{.9705\pm.0057}$ & $.8646\pm.0032$ & $.8557\pm.0093$\\
32 & 64 & 3,200 & $\mathbf{.9805\pm.0021}$ & $.8718\pm.0027$ & $.8629\pm.0022$\\
\bottomrule
\end{tabular}

\end{table}

Full EVT exceeds both controls on every paired seed at both widths and
budgets (\cref{tab:t1-frozen-cache}). At 3,200 updates, its mean gain
over the compensated control is about $0.056$ at either reader width
with the width-44 cache, and $0.104$ and $0.118$ with the width-32 cache.
Holding the encoder fixed isolates the benefit for the reader at these
training budgets.

\subsubsection{Path attention on a cache trained with the additive MLP}
\label{app:t1-additive-frozen-cache}

We fit one \(\STarch[\EVT,\MLP_{\mathrm{add}}]\) source model on the Task~1 graph split using
seed 0. We train on 512 graphs, validate on 128 and use four path queries
per graph. Validation selects the checkpoint at update 800. We freeze its
width-32 EVT graph encoder and cache its selected vertex and pair states
with their occurrence tags. All reader scores below use this one encoder.
The evaluation uses the same 256 test graphs as \cref{tab:task1-model-comparison}.

We train three fresh, two-layer, four-head EVT path readers on identical
cached inputs and minibatches. The full reader uses triangular path
attention. The first control removes this operation. The second also
widens the remaining feed-forward layers to bring the total reader
parameter count within \(0.14\%\) of the full reader. Parameters common to
each paired reader are initialized identically where their shapes match.

\Cref{tab:t1-additive-frozen-config} reports total and output-connected
parameter counts, including the input projections and prediction head
and excluding the frozen encoder. The second count includes only
parameters with a gradient path to the scalar prediction. The head reads
the final vertex states, so the terminal pair FFN and its normalization
are excluded from this count.
Relative to the full reader, the wider control has \(0.96\%\) fewer
output-connected parameters at width 32 and \(1.14\%\) fewer at width 64.

\begin{table}[!htbp]
\centering\footnotesize
\caption{Task 1 frozen-reader parameter counts. FFN denotes a
feed-forward network.}
\label{tab:t1-additive-frozen-config}
\begin{tabular}{@{}rlrr@{}}
\toprule
Width & Reader & Total & Output-connected\\
\midrule
32 & Full EVT & 158,121 & 141,255\\
32 & No triangular attention & 147,433 & 130,567\\
32 & No triangular attention, wider FFNs & 158,329 & 139,903\\
\midrule
64 & Full EVT & 617,289 & 550,791\\
64 & No triangular attention & 575,433 & 508,935\\
64 & No triangular attention, wider FFNs & 616,925 & 544,493\\
\bottomrule
\end{tabular}
\end{table}

For each width and reader, we screen rates
\(\{10^{-4},3\times10^{-4},10^{-3}\}\) using seed 100 and the minimum
validation MSE over 800 updates. All six configurations select \(10^{-3}\).
We then train each reader with seeds 11, 12 and 13 for 3,200 updates.
Readers use AdamW on CPU, batches of 64, weight decay \(10^{-4}\),
gradient clipping at 5, and an 80-update warm-up followed by a constant
rate. Validation runs every 100 updates. Each budget selects the earliest
checkpoint attaining the lowest validation MSE within the first 800 or
3,200 updates of the \emph{same} trajectory.

\begin{table}[!htbp]
\centering\footnotesize
\setlength{\tabcolsep}{3pt}
\caption{Task 1 on the frozen \(\STarch[\EVT,\MLP_{\mathrm{add}}]\) graph encoder. Test \(R^2\),
mean \(\pm\) sample SD over three paired reader seeds. Both controls remove
triangular attention. \emph{Wider FFNs} also reallocates capacity to the
feed-forward layers. The last column gives the within-seed gain of full EVT
over that wider control.}
\label{tab:t1-additive-frozen-cache}
\begin{tabular}{@{}rrcccc@{}}
\toprule
Width & Updates & Full EVT & No triangle & Wider FFNs & Paired gain\\
\midrule
32 & 800 & $\mathbf{0.9264\pm0.0170}$ & $0.8615\pm0.0057$ & $0.8722\pm0.0085$ & $+0.0542\pm0.0185$\\
32 & 3,200 & $\mathbf{0.9426\pm0.0093}$ & $0.8794\pm0.0005$ & $0.8812\pm0.0087$ & $+0.0614\pm0.0177$\\
\midrule
64 & 800 & $\mathbf{0.9401\pm0.0018}$ & $0.8750\pm0.0089$ & $0.8771\pm0.0098$ & $+0.0630\pm0.0110$\\
64 & 3,200 & $\mathbf{0.9432\pm0.0078}$ & $0.8752\pm0.0105$ & $0.8812\pm0.0110$ & $+0.0619\pm0.0033$\\
\bottomrule
\end{tabular}

\end{table}

The full EVT reader scores above both controls on each paired seed at
both widths and both budgets (\cref{tab:t1-additive-frozen-cache}).
The seed variation measures reader training conditional on this encoder.
The comparison uses one source model, 18 reader fits for rate selection,
and 18 final reader fits. The evaluation scores 36 validation-selected
checkpoints and 36 end-of-budget checkpoints on the same 1,024 test
queries. The table uses the validation-selected scores.

\FloatBarrier
\subsection{Task 2. Path alternatives and reuse}
\label{app:t2-path-alternatives}
\label{app:synthetic-tasks}

\paragraph{Target.}
Let \(N_G(u,v)\) count directed paths in the DAG. Since the prescribed segment
is one such path, \(N_G(p_s,p_t)-1\) counts alternatives. Dynamic programming
gives the exact label
\begin{equation}
 s_{2}(Y,P)=\frac{2}{(k-1)(k-2)}
 \sum_{s=1}^{k-2}\sum_{t=s+2}^{k}\log N_G(p_s,p_t).
 \label{eq:experimental-s4}
\end{equation}

\paragraph{Compatibility with the selected profile.}
\label{app:t2-profile-compatibility}
The adjacency channel is part of the initial pair color. Hence the stable
pair color $a=\cstar_{ij}$ determines $A_{ij}$. By
\cref{lem:stable-reconstruction}, it also determines the multiset
$M(a)=\multiset{(\cstar_{iu},\cstar_{uj}):u\in V}$, including multiplicities.
Set $f_1(\cstar_{ij})=A_{ij}$ and recursively define
\[
 f_{\ell+1}(a)=\sum_{(b,c)\in M(a)} f_\ell(b)f_1(c).
\]
Induction gives $f_\ell(\cstar_{ij})=(A^\ell)_{ij}$.
In a DAG, every directed walk is a path of length at most $n-1$. For
distinct $i,j$, this gives
$N_G(i,j)=\sum_{\ell=1}^{n-1}f_\ell(\cstar_{ij})$.
The selected profile contains $n$, $k$, and these ordered pair colors;
it therefore determines \cref{eq:experimental-s4}. The question for Task 2 is
how well a finite trained encoder and reader recover this information.

\paragraph{Target normalization.}
The Task~2 training pool contains 2048 graphs. Optimizer updates
use its fixed 512-graph subset, while normalization uses the full pool
(mean \(0.8260\), standard deviation \(0.3753\), rounded).
The additional 1536 graphs contribute training-label statistics only.
All models and training budgets use these same statistics; validation and
test labels remain excluded from preprocessing.

\subsubsection{End-to-end comparisons and learning-rate selection}
\label{app:t2-end-to-end-comparisons}
In the fixed-rate comparison in \cref{tab:t2-fixed-rate-transformer},
$\mathrm{ST}[\EVT,\EVT]$ has the highest median among reusable models, while
\PMEVT{} scores higher and recomputes its graph encoding for each query.
The models share graphs, queries, splits, a budget of 2,000 updates, and
five paired seeds. Parameter counts are within 5\% of the count for
$\mathrm{ST}[\EVT,\EVT]$, with different widths and readouts
(\cref{app:experimental-interfaces,tab:appendix-s4-s5}).
GINE without its raw-input bypass reaches median \(R^2=0.239\), versus
\(0.190\) for the bypass variant in the table.

\begin{table}[htbp]
\centering\small
\setlength{\abovecaptionskip}{0pt}
\caption{Task~2 fixed-rate comparison: median [interquartile range] and mean
test $R^2$ across seeds 0--4. Bold marks the best overall and underlining
marks the best reusable model. GINE+raw includes a raw path-vertex feature
bypass. ET-chain and ET-mix use chain-oriented and direction-mixed
Edge Transformers, respectively. \PMEVT{} marks the path before graph
encoding and uses a path Transformer on the selected vertex states.}
\label{tab:t2-fixed-rate-transformer}
\footnotesize\setlength{\tabcolsep}{2pt}
\begin{tabular*}{\linewidth}{@{\extracolsep{\fill}}lcc@{\hspace{8pt}}|@{\hspace{8pt}}lcc@{}}
\toprule
Model & Median [IQR] & Mean & Model & Median [IQR] & Mean\\
\midrule
PathTransformer & $-.013\;[-.018,-.011]$ & $-.016$ & $\mathrm{ST}[\ET,\ET]$ & $-.008\;[-.018,-.002]$ & $-.011$\\
GINE+raw & $.190\;[.184,.213]$ & $.182$ & EVT-Transformer & $.511\;[.476,.572]$ & $.538$\\
ET-chain & $-.015\;[-.022,-.005]$ & $-.014$ & $\mathrm{ST}[\EVT,\EVT]$ & $\underline{.637\;[.603,.643]}$ & $\underline{.593}$\\
ET-mix & $-.005\;[-.017,-.005]$ & $-.010$ & \PMEVT{} & $\mathbf{.700\;[.664,.702]}$ & $\mathbf{.688}$\\
$\mathrm{ST}[\ET,\MLP_{\mathrm{concat}}]$ & $-.006\;[-.006,-.002]$ & $-.006$ & & &\\
\bottomrule
\end{tabular*}

\end{table}

\begin{table}[!htbp]\centering\small
\caption{Task 2 fixed-rate reference comparison: test $R^2$ over five paired seeds at
2,000 updates. This panel uses the recorded per-model checkpoint rules
and includes a marked model with an MLP readout.}
\label{tab:t2-fixed-rate-mlp-reference}
\small\setlength{\tabcolsep}{6pt}
\begin{tabular}{@{}lcc@{}}
\toprule
Model & Median test $R^2$ [IQR] & Mean $\pm$ SD\\
\midrule
\(\mathrm{ST}[\EVT,\MLP_{\mathrm{add}}]\) & $\underline{0.668\;[0.641,0.681]}$ & $\underline{0.647\pm0.055}$\\
\(\mathrm{ST}[\EVT,\EVT]\) & $0.603\;[0.590,0.658]$ & $0.591\pm0.090$\\
\(\mathrm{ST}[\ET,\MLP_{\mathrm{add}}]\) & $-0.028\;[-0.038,-0.010]$ & $-0.026\pm0.018$\\
EVT-Transformer & $0.511\;[0.476,0.572]$ & $0.538\pm0.102$\\
GINE-Transformer+raw & $0.190\;[0.184,0.213]$ & $0.182\pm0.054$\\
\midrule
\PMEVT{} (MLP) & $\mathbf{0.688\;[0.670,0.702]}$ & $\mathbf{0.677\pm0.063}$\\
\bottomrule
\end{tabular}

\end{table}
\paragraph{Additional reference configurations.}
\label{app:t2-additional-baselines}
These fits use the same five seeds, graphs and queries, with the readouts
and checkpoint-selection rules indicated below.
\begin{table}[!htbp]\centering\small
\caption{Additional Task 2 references: validation-selected test $R^2$.
See \cref{app:t2-common-lr} for the common learning-rate screen.}
\label{tab:t2-additional-baselines}
\begin{tabular}{@{}lcc@{}}
\toprule
Model & Median test $R^2$ [IQR] & Mean $\pm$ SD\\
\midrule
PathTransformer & $-0.013\;[-0.018,-0.011]$ & $-0.016\pm0.008$\\
ET-chain & $-0.015\;[-0.022,-0.005]$ & $-0.014\pm0.011$\\
ET-mix & $-0.005\;[-0.017,-0.005]$ & $-0.010\pm0.008$\\
\(\mathrm{ST}[\ET,\ET]\) & $-0.008\;[-0.018,-0.002]$ & $-0.011\pm0.010$\\
\bottomrule
\end{tabular}

\end{table}

\paragraph{Learning-rate selection across model families.}
\label{app:t2-common-lr}
The screen uses Gaussian inputs and the same three initial rates
$\{10^{-4},3\times10^{-4},10^{-3}\}$ for every model. Each candidate is
trained for 2,000 updates on three tuning seeds (100--102), using the
schedule, splits and optimization settings in \cref{app:hyperparameters}.
For each model, selection minimizes the mean validation MSE of the
validation-selected checkpoints across these three seeds. Five distinct
evaluation seeds (0--4) are then trained at the selected rate and their
validation-selected checkpoints evaluated on the fixed test split. Rate selection uses validation scores only. The five \STEE{} fits
at the selected rate $3\times10^{-4}$ are the Gaussian-input fits in
\cref{tab:t2-attribute-controls}. The width-44 EVT graph encoder in
\(\STarch[\EVT,\MLP_{\mathrm{concat}}]\) is trained end to end with
the concatenating reader.

\begin{table}[!htbp]
\color{black}
\centering\footnotesize
\setlength{\tabcolsep}{4pt}
\caption{Common Task~2 learning-rate screen and final test results.
The middle three columns give mean validation $R^2$ over three tuning
seeds; bold identifies the selected rate. The final column gives test
$R^2$, mean $\pm$ SD over five separate evaluation seeds.}
\label{tab:t2-common-lr}
\begin{tabular}{@{}lccccc@{}}
\toprule
& \multicolumn{3}{c}{Tuning validation $R^2$} & &\\
\cmidrule(lr){2-4}
Model & $10^{-4}$ & $3\times10^{-4}$ & $10^{-3}$ & Selected rate & Test $R^2$\\
\midrule
\PMEVT{} & .4550 & \textbf{.6990} & .6093 & $3\times10^{-4}$ & $.6955\pm.0394$\\
\STEE{} & .3414 & \textbf{.6274} & .5560 & $3\times10^{-4}$ & $.5893\pm.1018$\\
\(\STarch[\EVT,\MLP_{\mathrm{concat}}]\) & .4778 & \textbf{.5863} & .4203 & $3\times10^{-4}$ & $.5254\pm.0910$\\
EVT-Transformer & .3863 & .4799 & \textbf{.5214} & $10^{-3}$ & $.4090\pm.1210$\\
GINE-Transformer+raw & \textbf{.2428} & .2088 & .2315 & $10^{-4}$ & $.1909\pm.0467$\\
PathTransformer & $\mathbf{-.0042}$ & $-.0053$ & $-.0263$ & $10^{-4}$ & $-.0066\pm.0036$\\
ET-chain & $\mathbf{-.0047}$ & $-.0103$ & $-.0246$ & $10^{-4}$ & $-.0088\pm.0052$\\
ET-Twice & $\mathbf{-.0077}$ & $-.0144$ & $-.0378$ & $10^{-4}$ & $-.0091\pm.0103$\\
\(\STarch[\ET,\MLP_{\mathrm{concat}}]\) & $-.0095$ & $-.0093$ & $\mathbf{-.0023}$ & $10^{-3}$ & $-.0095\pm.0064$\\
ET-mix & $\mathbf{-.0050}$ & $-.0151$ & $-.0207$ & $10^{-4}$ & $-.0110\pm.0066$\\
\bottomrule
\end{tabular}

\end{table}

After per-model selection, \PMEVT{} exceeds
\STEE{} on four of five paired seeds. The ET-based models remain near
zero. The comparison including the additive MLP is reported in
\cref{tab:task2-model-comparison}.

\paragraph{Learning-rate selection for the additive MLP.}
\label{app:t2-additive-mlp-lr}

The two models are \(\STarch[\EVT,\MLP_{\mathrm{add}}]\) and
\(\STarch[\ET,\MLP_{\mathrm{add}}]\), with 312,485 and 312,507 parameters.
Both use two graph layers, one additive MLP path layer, four heads, linear
projections and zero dropout. The EVT configuration has width 32 and
graph/path FFN widths 128/1,182. The ET configuration has width 80 and
FFN widths 279/414. Architectures and inputs are fixed before training.

Each fit performs 2,000 AdamW updates, using 512 optimization graphs,
128 validation graphs and 256 test graphs, with four path queries per
graph. The batch size is 64, weight decay is \(10^{-4}\), gradient clipping
is 5, and an 80-update warm-up precedes cosine decay to 5\% of the peak
rate. Validation occurs every 100 updates; the checkpoint is the earliest
minimum of full validation MSE. Training uses float32 Apple MPS and two
CPU threads. Checkpoint validation and final test prediction use CPU.
Target normalization uses the complete 2,048-graph training pool as in
\cref{app:t2-path-alternatives}.

For each model, we screen learning rates
\(\{10^{-4},3\times10^{-4},10^{-3}\}\) with tuning seeds 100--102.
The rate minimizing mean best validation MSE is then used for five
evaluation seeds, 0--4. The selected rates are \(3\times10^{-4}\) for \(\STarch[\EVT,\MLP_{\mathrm{add}}]\)
and \(10^{-4}\) for \(\STarch[\ET,\MLP_{\mathrm{add}}]\). These 18 tuning fits and ten final fits
are distinct trajectories. \Cref{tab:t2-additive-mlp-selected} reports the
test results after validation-based selection.

\begin{table}[htbp]
\centering
\small
\caption{Task 2 with the additive MLP: test scores over five evaluation seeds
(mean $\pm$ sample SD). Rates and checkpoints are selected on validation.}
\label{tab:t2-additive-mlp-selected}
\begin{tabular}{@{}lccc@{}}
\toprule
Model & Selected LR & Test MSE & Test $R^2$\\
\midrule
\(\STarch[\EVT,\MLP_{\mathrm{add}}]\) & \(3\times10^{-4}\) & $0.319\pm0.054$ & $0.661\pm0.058$\\
\(\STarch[\ET,\MLP_{\mathrm{add}}]\) & \(10^{-4}\) & $0.949\pm0.004$ & $-0.009\pm0.004$\\
\bottomrule
\end{tabular}

\end{table}

\paragraph{Resource measurements.}
\Cref{tab:appendix-s4-s5} gives parameter counts and CPU resource
measurements for the models in \cref{tab:t2-fixed-rate-transformer}.

\begin{table}[htbp]
\centering
\setlength{\abovecaptionskip}{0pt}
\caption{Parameter counts and computational measurements for Task 2.
Time and resident set size (RSS) are isolated CPU
training-step measurements; floating-point operation counts (FLOPs) are
per query, in millions (MFLOPs). Fwd/bwd denote forward/backward passes.
Prediction scores are in
\cref{tab:t2-fixed-rate-transformer}.}
\label{tab:appendix-s4-s5}
\scriptsize
\resizebox{\textwidth}{!}{%
\begin{tabular}{@{}lrrrrr@{}}
\toprule
Model & Parameters & Fwd MFLOPs/q & Fwd+bwd MFLOPs/q & ms/update & RSS (MiB)\\
\midrule
PathTransformer & 312,795 & \textbf{4.27} & \textbf{12.75} & \textbf{11.0} & \textbf{404} \\
GINE-Path + raw bypass & 313,581 & 4.56 & 13.34 & 20.7 & 574 \\
ET faithful & 312,717 & 66.36 & 198.41 & 366.7 & 1372 \\
ET direction-mixed & 312,717 & 66.36 & 198.41 & 370.3 & 1358 \\
\(\STarch[\ET,\MLP_{\mathrm{concat}}]\) & 313,247 & 94.13 & 281.67 & 382.7 & 1391 \\
ET-Twice & 312,869 & 75.28 & 225.24 & 394.9 & 1373 \\
EVT-Path & 312,531 & 65.10 & 172.15 & 304.1 & 1413 \\
\STEE{} (balanced) & 312,753 & 48.00 & 140.69 & 313.0 & 1390 \\
\PMEVT{} & 312,811 & 258.54 & 682.95 & 1076.7 & 2695 \\
\bottomrule
\end{tabular}%
}
\end{table}

\FloatBarrier
\subsubsection{Controls on graph encoding}
\paragraph{Graph-stage triangular attention.}
\label{app:t2-graph-ablation}
On the same split, we train \STEE{} and a control deleting only graph-stage
triangular attention jointly for 2,000 updates at a fixed initial learning
rate of $3\times10^{-4}$. Removing this operator lowers mean test $R^2$
from $0.591\pm0.090$ to $-0.003\pm0.003$, with a decrease on all five
paired seeds (0--4). The path processor remains full EVT; widths are fixed
and parameter counts decrease. This end-to-end intervention is separate
from the input-feature and model-specific LR controls in
\cref{app:t2-attribute-controls,app:t2-common-lr}.

\paragraph{Vertex-feature intervention across model families.}
\label{app:t2-attribute-controls}
The eight Gaussian vertex channels contain no information required by the
Task~2 target. We compare Gaussian inputs with all eight channels
set to zero during both training and evaluation. Pair features, graph
topology, paths, labels and the target normalization are held
fixed. This is an input intervention, with unchanged model architectures
and parameter counts.

Each reported model receives 2,000 updates with initial learning rate
$3\times10^{-4}$, batch size~64, 80 warmup updates followed by cosine
decay to 5\% of the initial rate, AdamW weight decay $10^{-4}$ and gradient
clipping at~5. Each Gaussian/zero pair uses the same initialization and
minibatches for seeds 0--4. Checkpoints minimize validation MSE, evaluated
every 100 updates, and are tested on the same 256 held-out graphs
(1,024 queries), following optimization on 512 graphs and validation on
128 graphs. We fit each condition on MPS and check the reported test metrics by
independent CPU evaluation.

\begin{table}[!htbp]
\color{black}
\centering\small
\caption{Task 2 vertex-attribute intervention: test $R^2$, mean $\pm$ SD
over five paired seeds at a fixed initial learning rate of
$3\times10^{-4}$. ET-chain and ET-mix use Transformer
path processors.}
\label{tab:t2-attribute-controls}
\begin{tabular}{@{}lrcc@{}}
\toprule
Model & Parameters & Gaussian vertex features & Zero vertex features\\
\midrule
ET-chain & 312,717 & $-.0142\pm.0113$ & $.8996\pm.0127$\\
ET-mix & 312,717 & $-.0101\pm.0081$ & $.8790\pm.0123$\\
EVT-Transformer & 312,531 & $.5272\pm.0976$ & $.8524\pm.0341$\\
\STEE{} & 312,753 & $.5893\pm.1018$ & $.8377\pm.1030$\\
\(\STarch[\ET,\MLP_{\mathrm{concat}}]\) & 313,247 & $-.0065\pm.0069$ & $.9458\pm.0583$\\
\bottomrule
\end{tabular}

\end{table}

Removing the Gaussian channels improves test $R^2$ on every seed for
the four other models and on four of five seeds for \STEE{}. ET's scores with Gaussian inputs indicate sensitivity to irrelevant
attributes under this protocol. The vertex stream is one architectural
difference that could contribute to EVT's smaller sensitivity.

\paragraph{Vertex-feature intervention with the additive MLP.}
\label{app:t2-vertex-input-ablation}

At the fixed learning rate \(3\times10^{-4}\), we compare the
eight Gaussian vertex channels with the same channels set to zero during
training, validation and test. Pair features, graph structure, queried
paths, targets and split assignments are unchanged. Within each model
and seed, the Gaussian and zero-input fits have identical initialization
and ordered training minibatches; both identities are verified from
their recorded hashes. The intervention preserves every model parameter.

The five EVT Gaussian fits also appear in
\cref{tab:t2-additive-mlp-selected}. For ET, the fixed rate uses a
separate set of five Gaussian fits because validation selects a
different rate. The rate screen and both input conditions together
give 43 distinct training trajectories and 25 final test evaluations.
The fit inventory is validated before test evaluation.

\begin{table}[!htbp]
\centering\small
\caption{Task~2 vertex-feature intervention for the two MLP configurations.
Test \(R^2\) and within-seed changes are reported as mean \(\pm\) sample
SD over five seeds. Gaussian and zero-input fits share the learning rate
\(3\times10^{-4}\), initialization and minibatch order within each model
and seed. For \(\STarch[\EVT,\MLP_{\mathrm{add}}]\), the Gaussian column uses the same five fits as
\cref{tab:t2-additive-mlp-selected}.}
\label{tab:t2-vertex-input-ablation}
\begin{tabular}{@{}lccc@{}}
\toprule
Model & Gaussian $X$ & $X=0$ & Paired $\Delta R^2$\\
\midrule
\(\STarch[\EVT,\MLP_{\mathrm{add}}]\) & $0.661\pm0.058$ & $0.824\pm0.073$ & $+0.163\pm0.102$\\
\(\STarch[\ET,\MLP_{\mathrm{add}}]\) & $-0.026\pm0.018$ & $0.984\pm0.006$ & $+1.011\pm0.017$\\
\bottomrule
\end{tabular}

\end{table}

As reported in \cref{tab:t2-vertex-input-ablation}, zeroing the nuisance
vertex channels improves test \(R^2\) on all five
paired seeds for both models. The effect is especially large for \(\STarch[\ET,\MLP_{\mathrm{add}}]\):
its near-zero score with Gaussian inputs becomes \(0.984\pm0.006\) with
zero inputs. This improvement shows sensitivity to nuisance inputs
at the tested capacity and update budget. \(\STarch[\EVT,\MLP_{\mathrm{add}}]\) also benefits
from this intervention.

\paragraph{Head orientation, readout, and resampling.}
Two-seed pilots (0 and 1) retain 2,000 updates and initial LR
$3\times10^{-4}$ and compare their validation scores at validation-selected checkpoints. The orientation intervention changes only \(\mathrm{ST}[\EVT,\MLP_{\mathrm{add}}]\)'s graph triangular
heads from balanced to chain-only. The readout intervention supplies
both families with $[\bar Z;\bar R_{\rm edge}]$, where
$\bar Z=k^{-1}\sum_{s=1}^k Z_s$ and
$\bar R_{\rm edge}=(k-1)^{-1}\sum_{s=1}^{k-1}R_{s,s+1}$.
The enlarged task-specific layer raises the total model parameter count
from 312,485 to 313,509 in \(\mathrm{ST}[\EVT,\MLP_{\mathrm{add}}]\) and from 312,507 to
318,907 in \(\mathrm{ST}[\ET,\MLP_{\mathrm{add}}]\). The task-specific layers have the same form but capacities remain different.
An additional intervention redraws iid normal $X$ at every training batch,
sharing one draw across all queries of each graph in that batch; validation
and final training-set evaluation retain the original feature draws. Independent
random streams for feature sampling and optimization preserve
the original minibatch and initialization draws.
There are eight orientation/readout fits (including two ET reference reruns)
and four feature-resampling fits, with \(\mathrm{ST}[\EVT,\MLP_{\mathrm{add}}]\) references reused.

\begin{table}[t]
\centering\small
\caption{Exploratory Task 2 diagnostics: validation $R^2$, mean $\pm$ sample
SD over two seeds at a fixed learning rate.}
\label{tab:synthetic-flexibility-diagnostics}
\begin{tabular}{lr}
\toprule
Task 2 exploratory condition & Validation $R^2$ \\
\midrule
\(\mathrm{ST}[\EVT,\MLP_{\mathrm{add}}]\): reference & $0.692 \pm 0.075$ \\
\(\mathrm{ST}[\EVT,\MLP_{\mathrm{add}}]\): chain-only heads & $0.741 \pm 0.009$ \\
\(\mathrm{ST}[\EVT,\MLP_{\mathrm{add}}]\): $\bar Z$ and junction-$R$ readout & $0.585 \pm 0.001$ \\
\(\mathrm{ST}[\EVT,\MLP_{\mathrm{add}}]\): resampled $X$ & $0.805 \pm 0.019$ \\
\(\mathrm{ST}[\ET,\MLP_{\mathrm{add}}]\): rerun reference & $-0.010 \pm 0.006$ \\
\(\mathrm{ST}[\ET,\MLP_{\mathrm{add}}]\): $\bar Z$ and junction-$R$ readout & $-0.006 \pm 0.007$ \\
\(\mathrm{ST}[\ET,\MLP_{\mathrm{add}}]\): resampled $X$ & $-0.006 \pm 0.002$ \\
\bottomrule
\end{tabular}

\end{table}

\(\mathrm{ST}[\EVT,\MLP_{\mathrm{add}}]\) remains effective with chain-only heads, while junction readout and
feature resampling do not recover \(\mathrm{ST}[\ET,\MLP_{\mathrm{add}}]\) performance
(\cref{tab:synthetic-flexibility-diagnostics}). Resampling improves \(\mathrm{ST}[\EVT,\MLP_{\mathrm{add}}]\) on
both seeds. The complete configurations also differ in their vertex stream, endpoint
lifting, diagonal proxies, exchanges, widths and nonlinearities.

\subsubsection{Optimizer-budget sensitivity}
\label{app:s4-budget-sensitivity}

EVT-Path and \STEE{} are trained for \(400,800,1200,1600,\) and \(2000\)
updates, crossed with the same five seeds. Data, model configurations, and
the other Task 2 settings in \cref{app:hyperparameters} remain fixed.
Each horizon is a separate fit with its own cosine schedule and a
validation-selected checkpoint.

The Task~2 comparison and all runs in this budget study use the same graphs,
queries and splits.
The 2,000-update entries reuse the fits reported in
\cref{tab:t2-fixed-rate-transformer}.

\begin{table}[H]
\centering
\setlength{\abovecaptionskip}{0pt}
\caption{Controlled Task 2 optimizer-budget sensitivity. Values are median test
\(R^2\) [min,max] over five seeds. Bold marks the higher median when the
displayed values differ.}
\label{tab:appendix-s4-budget}
\small
\begin{tabular}{@{}rrr@{}}
\toprule
Updates & EVT-Path & $\mathrm{ST}[\EVT,\EVT]$ \\
\midrule
400 & 0.000 [-0.015,0.091] & 0.000 [-0.005,0.023] \\
800 & 0.359 [0.303,0.452] & \textbf{0.421 [0.123,0.534]} \\
1\,200 & 0.466 [0.345,0.588] & \textbf{0.542 [0.377,0.649]} \\
1\,600 & 0.505 [0.395,0.662] & \textbf{0.615 [0.455,0.645]} \\
2\,000 & 0.511 [0.435,0.697] & \textbf{0.637 [0.419,0.661]} \\
\bottomrule
\end{tabular}

\end{table}

Performance is low at 400 updates. \STEE{} has the higher median from 800
updates onward and exceeds EVT-Path on \(2/5,3/5,3/5,3/5\) paired seeds
at 800--2,000 updates, respectively. Score ranges overlap at each horizon.

\paragraph{Optimization diagnostic.}
Both faithful and direction-mixed ET reduce training error while their
validation error remains high.

\paragraph{Bounded ET optimization controls.}
\label{app:t2-et-lr}
The canonical Task 2 comparison uses a shared initial learning rate (LR) of
$3\times10^{-4}$. The following additional controls examine classical and
direction-mixed ET on Task 2, using training and validation data only.
We first screen initial learning rates
$\{10^{-4},3\times10^{-4},10^{-3}\}$ for classical and direction-mixed ET,
each over seeds 0--1 and 2,000 updates: twelve CPU fits and 24,000 updates
in total, with equal search budgets for the two orientations. Both retain
their 312,717 parameters and the Task~2 dataset and 512-graph optimization subset.
AdamW uses weight decay $10^{-4}$, zero dropout, 200 warm-up updates and
cosine decay to 5\% of the initial rate. Training batches contain 64 queries;
evaluation batches contain 32. Checkpoints minimize validation MSE, checked
every 100 updates with no minimum-improvement threshold; every fit completes
the full budget.

We independently recompute the targets and verify normalization,
padding, vertex relabeling, the mean readout, and gradients through both
encoder stages. We then fit the first 16 optimization
graphs (64 queries), with the same architectures and seeds 0--1.
These four small-set fits use 800 updates, AdamW at a constant rate of
$10^{-3}$, zero weight decay and dropout, batches of 16, and clipping at 5.
Checkpoints minimize MSE on all 64 training queries, evaluated every
100 updates. All four training $R^2$ scores exceed $0.99998$.

Mean validation MSE in the LR screen selects $10^{-4}$ for both
orientations. At this rate, we separately increase weight decay to
$10^{-2}$ or set dropout to $0.1$, retaining weight decay $10^{-4}$ in the
latter condition. Dropout applies to all dropout modules, including graph
and path attention, residual and feed-forward branches, and the head.
These eight fits retain the LR screen's other settings and 2,000-update
budget; the four existing fits at $10^{-4}$ supply the reference.
The three controls together use 24 fits and 43,200 updates, divided equally
between the two ET orientations.

\begin{table}[htbp]
\centering
\setlength{\abovecaptionskip}{0pt}
\caption{Task 2 ET controls. Top: validation $R^2$, mean $\pm$ sample SD over
two seeds; bold marks the best setting for each ET variant, including ties
at displayed precision. Base setup uses weight decay $10^{-4}$ and zero dropout.
Bottom: training $R^2$ on the 16-graph subset, reported per seed and rounded
to six decimals. The separate five-seed test scores appear in \cref{tab:t2-fixed-rate-transformer}.}
\label{tab:t2-et-lr}
\small
\begin{tabular}{@{}llcc@{}}
\toprule
\multicolumn{4}{c}{Validation $R^2$: mean $\pm$ SD over seeds 0--1}\\
Setting & Initial LR & Classical ET & Direction-mixed ET\\
\midrule
Base setup & $10^{-4}$ & $\mathbf{-0.0059\pm0.0011}$ & $\mathbf{-0.0065\pm0.0000}$ \\
Base setup & $3\times10^{-4}$ & $-0.0071\pm0.0044$ & $-0.0075\pm0.0026$\\
Base setup & $10^{-3}$ & $-0.0106\pm0.0030$ & $-0.0071\pm0.0020$\\
Weight decay $10^{-2}$ & $10^{-4}$ & $\mathbf{-0.0059\pm0.0011}$ & $\mathbf{-0.0065\pm0.0000}$ \\
Dropout $0.1$ & $10^{-4}$ & $-0.0068\pm0.0001$ & $-0.0068\pm0.0011$\\
\midrule
\multicolumn{4}{c}{Training $R^2$: 16 graphs, 64 queries}\\
Seed & Constant LR & Classical ET & Direction-mixed ET\\
\midrule
0 & $10^{-3}$ & $0.999999$ & $\mathbf{1.000000}$ \\
1 & $10^{-3}$ & $0.999990$ & $\mathbf{1.000000}$ \\
\bottomrule
\end{tabular}

\end{table}

All twenty full-fit validation $R^2$ values remain negative. Increasing
weight decay or dropout brings no improvement over the selected LR setup.
The LR fits end with recorded minibatch MSE between $8.2\times10^{-5}$ and
$0.034$. Together with near-perfect small-set fitting, these checks support
a generalization difficulty within the tested protocol that persists across
the explored rates and regularizers. This sequential validation study covers
the two ET models on Task 2. Their test scores under the common fixed-rate
protocol appear in \cref{tab:t2-fixed-rate-transformer}.
\(\STarch[\ET,\MLP_{\mathrm{concat}}]\) and ET-Twice separately evaluate full-pair path interfaces
(\cref{app:experimental-interfaces}); their different widths and operators
make them comparisons of complete systems.

\FloatBarrier
\subsubsection{Selection and joint access after concatenating-MLP training}
\label{app:t2-frozen-cache}

To isolate the effect of access to off-path pair states, we train a width-44
EVT graph encoder with a concatenating MLP path processor for 2,000 updates
per seed (0--2), then freeze the validation-selected graph encoder. The
two readers share the resulting cache within each seed. We use the same
graphs, queries, 512-graph optimization subset, and target normalization pool as
in the fixed-rate Task 2 comparison (\cref{tab:t2-fixed-rate-transformer}). Encoder training
uses the settings in \cref{app:hyperparameters} with 200 warm-up updates, validation
every 100 updates, evaluation batches of 32, and zero minimum-improvement
threshold. All runs use one CPU thread.

Both readers retain a flattened, ordered selection of all path-vertex and
$P\times P$ states. An additional branch first combines the ordered tuple
$(\widetilde y_{p_su})_{s=1}^{k}$ of cached pair states using a shared MLP,
then sums over $u$.
The selected-only control restricts this sum to distinct path vertices;
the joint reader sums over all graph vertices. The two branches are
concatenated with $\log(1+n)$ and $\log(1+k)$, constant across this dataset,
before scalar prediction. Thus the intervention changes which vertices
contribute to the joint branch, while preserving the full ordered selection.

Each reader has 186,049 trainable parameters: the flattened-selection and
joint branches have widths $64\!\to\!32$, followed by a
$66\!\to\!64\!\to\!1$ readout, with GELU hidden activations.
Readers share initial weights and minibatch draws within each seed, then
train separately for 800 updates. AdamW uses weight decay $10^{-4}$,
gradient clipping at 5, and batches of 64 queries sampled with replacement.
We screen constant learning rates
$\{5\times10^{-5},10^{-4},3\times10^{-4}\}$, selecting checkpoints every
40 updates on validation MSE. Each reader's learning rate minimizes mean
validation MSE across the three seeds; both select $10^{-4}$ before test
evaluation. We fit the readers on the fixed outputs of the separately trained
encoders.

\begin{table}[htbp]
\centering
\setlength{\abovecaptionskip}{0pt}
\caption{Task 2 frozen-cache comparison: test $R^2$ after validation-only
learning-rate selection. The last column gives mean $\pm$ sample standard
deviation over three seeds; the last row contains paired differences.}
\label{tab:t2-frozen-cache}
\small
\begin{tabular}{@{}lcccc@{}}
\toprule
Reader & Seed 0 & Seed 1 & Seed 2 & Mean $\pm$ SD\\
\midrule
$P\times P$ control & \textbf{.465} & \textbf{.509} & .616 & $\mathbf{.530\pm.078}$ \\
Joint $P\times V$ & .461 & .504 & \textbf{.617} & $.528\pm.081$\\
Joint minus control & $-.004$ & $-.005$ & $+.002$ & $-.002\pm.003$\\
\bottomrule
\end{tabular}
\end{table}

The readers achieve similar scores, with a negative paired change on
two seeds and a slightly positive change on the third. These scores
pertain to caches fitted for selected-only prediction. The selected
reader computes off-path contributions before masking, so its cost
includes computations for features excluded from the final score.

\subsubsection{Selection and joint access after additive-MLP training}
\label{app:t2-joint-frozen-readers}

We freeze the graph encoders from the selected Gaussian-input \(\STarch[\EVT,\MLP_{\mathrm{add}}]\)
fits with seeds 0--2. We train only the readers in this study. Both readers have 137,665 parameters and use the same
flattened ordered \(\widetilde X[P],\widetilde Y[P,P]\) branch. Their context branch applies a
shared MLP to the ordered tuple \((\widetilde y_{p_su})_{s\in[k]}\), then sums over
candidate vertices. The selected control uses distinct visited vertices,
and the joint reader uses every valid graph vertex.
Both receive the same explicit graph-size and path-length counts.
The selected implementation computes candidate rows before masking so
the comparison isolates the effect of access on prediction.

The reader grid contains two access modes, three learning rates
\(\{5\times10^{-5},10^{-4},3\times10^{-4}\}\), and three frozen encoders,
for 18 CPU fits. Each fit uses 800 AdamW updates, batches of 64 sampled
with replacement, weight decay \(10^{-4}\), gradient clipping 5, and a
constant learning rate. Paired access modes have identical initialization
and minibatch order. Checkpoints minimize validation MSE, checked every
40 updates. For each access mode, the rate minimizing mean best validation
MSE over the three encoders is selected; both choose \(10^{-4}\).
All 18 fits must finish before the six selected readers are tested.

\begin{table}[htbp]
\centering
\small
\caption{Task 2 frozen-cache readers after additive-MLP training.
Test $R^2$ over three encoders (mean $\pm$ sample SD). Both readers
have 137,665 parameters. Rates and checkpoints use validation only.}
\label{tab:t2-joint-frozen-readers}
\begin{tabular}{@{}lcc@{}}
\toprule
Reader & Selected LR & Test $R^2$\\
\midrule
Selected $P\times P$ & $10^{-4}$ & $0.683\pm0.044$\\
Joint $P\times V$ & $10^{-4}$ & $0.684\pm0.052$\\
\bottomrule
\end{tabular}

\end{table}

The paired joint-minus-selected test \(R^2\) differences are
\(-0.0080,0.0063,0.0042\), giving
\(0.0008\pm0.0077\). The two access modes give similar scores at the
tested capacity and training budget.

\FloatBarrier
\subsubsection{\PMEVT{} readout ablation}
\label{app:t2-marked-readout}

We remove the path Transformer from the query-conditioned reference and
compare an MLP head with an affine head. Both retain the same graph-backbone
architecture: two EVT layers, width 40, four balanced triangular heads,
feed-forward width 183, separate vertex and pair initialization, linear input
projections, and zero dropout. Occurrence marks enter before graph encoding.
For each query, selected path-vertex states are averaged with a padding mask;
the MLP applies $40\!\to\!40\!\to\!1$ linear maps with a GELU hidden
activation, while the affine head applies a single $40\!\to\!1$ map.
Neither variant has a path Transformer, a learned path projection, or an
additional occurrence code at the readout. The MLP variant is the
\PMEVT{} reference in \cref{tab:t2-fixed-rate-mlp-reference}.

The MLP and affine variants are trained separately end to end on seeds
0--4, giving ten fits of exactly 2,000 updates. Within each seed they share
the graph backbone's initial weights, also matching the Transformer
reference, and use the same seeded minibatch schedule. Training updates the graph encoder. Its architecture and initialization
are shared across the variants, and the total parameter counts reflect
the readout sizes (\cref{tab:t2-marked-readout}).

The data, normalization pool and 512-graph optimization subset are unchanged.
AdamW uses LR $3\times10^{-4}$, weight decay $10^{-4}$, clipping at 5,
80 warm-up updates and cosine decay to 5\% of the initial rate. Training
batches contain 64 queries and evaluation batches contain 256. Validation
runs every 100 updates; checkpoints use a minimum improvement of
$10^{-4}$ in validation MSE. The StrikeTwice rate screen in
\cref{app:t2-common-lr} uses a zero threshold. Test predictions
use the selected checkpoint, which may precede update 2,000. The readout comparisons use a fixed learning rate and share the test
split used for the reference readouts.

\begin{table}[htbp]
\centering\small
\caption{Task 2 \PMEVT{} readout ablation: test $R^2$ over five seeds
at 2,000 updates, with the graph-backbone architecture fixed and all weights
trained end to end. Parameter counts include the graph backbone
and the indicated readout.}
\label{tab:t2-marked-readout}
\begin{tabular}{@{}lrcc@{}}
\toprule
Head & Parameters & Median test $R^2$ [IQR] & Mean $\pm$ SD\\
\midrule
Transformer & 312\,811 & $0.700\;[0.664,0.702]$ & $0.688\pm0.046$\\
MLP & 268\,005 & $0.688\;[0.670,0.702]$ & $0.677\pm0.063$\\
Affine & 266\,365 & $0.714\;[0.711,0.718]$ & $0.704\pm0.031$\\
\bottomrule
\end{tabular}

\end{table}

Both simple heads retain strong predictive performance without a path
Transformer. The affine head has the highest observed mean and median,
but the five-seed results do not establish a general advantage of affine
readouts. In particular, this experiment does not compare affine and
nonlinear heads on the Latin-square benchmark. The Latin construction applies
an MLP to each selected vertex together with its occurrence code, then
averages the logits; Task 2 averages the states before applying its scalar
head. The arguments requiring nonlinear occurrence-dependent readouts in
the general theory do not imply failure of an affine head on this specific
scalar target.

\subsubsection{Inference with the pooled MLP readout for \PMEVT{}}
\label{app:task2-trained-cache}

We time the validation-selected checkpoints of
\(\mathrm{ST}[\EVT,\EVT]\), \(\mathrm{ST}[\EVT,\MLP_{\mathrm{add}}]\), and
\PMEVT{} (MLP), with seeds 0--4.
All models use the same 12 test graphs, selected by graph-identity hashes
independently of predictions and timings, with $n=28$ and four test
distinct paths of length $k=7$ per graph. Nested requests contain
$Q\in\{1,2,4\}$ paths, with physical batches $B\in\{1,2,4\}$, $B\leq Q$.
For $Q\in\{8,16,32\}$, we retain the four test paths and select
28 additional distinct valid paths per graph by deterministic hashes,
independently of predictions and timings. These larger requests use nested prefixes of the same 32-path list and
fixed batches $B\in\{1,4\}$. The additional paths serve only for timing
and prediction-consistency checks.
Both request-size panels run on an Apple M4 Max with a 14-core CPU.
Warm CPU FP32 inference uses four computation threads, three warm-ups,
and seven timed repetitions. Request times include initial graph encoding,
selection, path processing, and prediction; loading and input assembly are
excluded. We take the median over repetitions, then average over graphs
within each seed, then over the five seeds. Speedups are ratios of these
overall means. Standard deviations across seed means describe the
observed variation and are not confidence intervals.
For each StrikeTwice model, the comparison holds weights and physical batch size
fixed and changes only whether graph encodings persist across batches.

\Cref{tab:task2-trained-latencies,tab:task2-trained-speedups} report every
measured request size and physical batch size. The two panels group results by path list. Within each cell, all architectures process the same paths
with the same physical batch size. This controls batching, but comparisons
between \PMEVT{} and the reusable StrikeTwice models still include differences
between their complete architectures. The within-model rebuild/reuse
comparisons instead hold weights, inputs and batch size fixed, isolating
the effect of retaining the graph cache.
For example, at $Q=4$, $B=1$, \(\mathrm{ST}[\EVT,\EVT]\),
\(\mathrm{ST}[\EVT,\MLP_{\mathrm{add}}]\), and \PMEVT{} (MLP) take
$17.33$, $10.28$, and $39.67$\,ms per request, respectively.
The corresponding reuse gains within each StrikeTwice model are $2.37\times$ and
$3.23\times$, increasing to $4.08\times$ and $9.48\times$ at
$Q=32$, $B=1$ in the extended-path panel.
When $B=Q$, both schedules encode the graph once and their times are
approximately equal. Checks against saved predictions on the four test paths and ordinary
forward predictions on all paths give a maximum discrepancy of less than
$1.2\cdot10^{-6}$ in standardized target units across schedules and batch
sizes.

\begin{table}[!htbp]
\centering\scriptsize
\setlength{\tabcolsep}{4pt}
\setlength{\abovecaptionskip}{0pt}
\caption{Trained Task 2 models: total CPU request time in milliseconds
(mean $\pm$ descriptive sample SD across five seed means).
$Q$ is the number of distinct paths per graph and $B$ the physical batch size.
Rebuild encodes the graph again for each batch; reuse retains it across
batches. All totals include initial graph encoding. Each seed mean averages
12 graph-level medians of seven timed repetitions. The two panels are
separate timing measurements for the two path lists.}
\label{tab:task2-trained-latencies}
\begin{tabular}{@{}ccrrrrr@{}}
\toprule
 & & \multicolumn{2}{c}{\(\mathrm{ST}[\EVT,\EVT]\)} & \multicolumn{2}{c}{\(\mathrm{ST}[\EVT,\MLP_{\mathrm{add}}]\)} & \PMEVT{}\\
\cmidrule(lr){3-4}\cmidrule(lr){5-6}
$Q$ & $B$ & Rebuild & Reuse & Rebuild & Reuse & (MLP)\\
\midrule
\multicolumn{7}{@{}l}{\textit{Four test paths}}\\
1 & 1 & $10.54\pm0.36$ & $10.57\pm0.39$ & $8.26\pm0.20$ & $8.24\pm0.18$ & $9.92\pm0.40$\\
2 & 1 & $20.76\pm0.76$ & $12.81\pm0.41$ & $16.36\pm0.48$ & $8.85\pm0.26$ & $19.98\pm0.95$\\
2 & 2 & $10.85\pm0.41$ & $10.82\pm0.40$ & $8.51\pm0.19$ & $8.49\pm0.20$ & $17.29\pm0.79$\\
4 & 1 & $41.14\pm1.27$ & $17.33\pm0.47$ & $33.25\pm0.92$ & $10.28\pm0.30$ & $39.67\pm2.06$\\
4 & 2 & $21.47\pm0.51$ & $13.54\pm0.25$ & $16.95\pm0.43$ & $9.33\pm0.26$ & $35.02\pm1.75$\\
4 & 4 & $11.61\pm0.40$ & $11.63\pm0.41$ & $8.60\pm0.24$ & $8.59\pm0.23$ & $28.93\pm1.15$\\
\midrule
\multicolumn{7}{@{}l}{\textit{Extended path list}}\\
8 & 1 & $84.00\pm0.57$ & $26.75\pm0.17$ & $74.85\pm8.66$ & $14.46\pm1.58$ & $91.91\pm10.95$\\
8 & 4 & $23.31\pm0.15$ & $15.31\pm0.32$ & $20.27\pm2.91$ & $11.23\pm1.47$ & $64.02\pm7.87$\\
16 & 1 & $167.11\pm1.92$ & $45.04\pm0.48$ & $152.53\pm19.44$ & $20.65\pm2.91$ & $183.41\pm18.11$\\
16 & 4 & $47.05\pm0.62$ & $22.35\pm0.24$ & $39.81\pm4.80$ & $13.40\pm1.62$ & $128.36\pm14.57$\\
32 & 1 & $333.92\pm2.36$ & $81.86\pm0.45$ & $301.49\pm33.23$ & $31.79\pm3.83$ & $359.10\pm21.65$\\
32 & 4 & $92.91\pm0.83$ & $36.04\pm0.23$ & $79.11\pm9.39$ & $17.81\pm2.23$ & $254.94\pm26.31$\\
\bottomrule
\end{tabular}

\end{table}

\begin{table}[!htbp]
\centering\small
\setlength{\tabcolsep}{6pt}
\setlength{\abovecaptionskip}{0pt}
\caption{Trained Task 2 models: speedups at matched request and batch sizes.
Rebuild/reuse isolates the effect of retaining each StrikeTwice model's graph cache
with weights and inputs held fixed. \PMEVT{}/reuse divides the MLP-readout
marked model's time by each StrikeTwice model's reuse time, comparing complete
architectures at matched batch sizes. Ratios use the unrounded overall mean times in
\cref{tab:task2-trained-latencies}. Values above one favor StrikeTwice reuse.}
\label{tab:task2-trained-speedups}
\begin{tabular}{@{}ccrrrr@{}}
\toprule
 & & \multicolumn{2}{c}{Rebuild / reuse} & \multicolumn{2}{c}{\PMEVT{} / reuse}\\
\cmidrule(lr){3-4}\cmidrule(lr){5-6}
$Q$ & $B$ & \(\mathrm{ST}[\EVT,\EVT]\) & \(\mathrm{ST}[\EVT,\MLP_{\mathrm{add}}]\) & \(\mathrm{ST}[\EVT,\EVT]\) & \(\mathrm{ST}[\EVT,\MLP_{\mathrm{add}}]\)\\
\midrule
\multicolumn{6}{@{}l}{\textit{Four test paths}}\\
1 & 1 & $1.00\times$ & $1.00\times$ & $0.94\times$ & $1.20\times$\\
2 & 1 & $1.62\times$ & $1.85\times$ & $1.56\times$ & $2.26\times$\\
2 & 2 & $1.00\times$ & $1.00\times$ & $1.60\times$ & $2.04\times$\\
4 & 1 & $2.37\times$ & $3.23\times$ & $2.29\times$ & $3.86\times$\\
4 & 2 & $1.59\times$ & $1.82\times$ & $2.59\times$ & $3.75\times$\\
4 & 4 & $1.00\times$ & $1.00\times$ & $2.49\times$ & $3.37\times$\\
\midrule
\multicolumn{6}{@{}l}{\textit{Extended path list}}\\
8 & 1 & $3.14\times$ & $5.18\times$ & $3.44\times$ & $6.36\times$\\
8 & 4 & $1.52\times$ & $1.81\times$ & $4.18\times$ & $5.70\times$\\
16 & 1 & $3.71\times$ & $7.39\times$ & $4.07\times$ & $8.88\times$\\
16 & 4 & $2.11\times$ & $2.97\times$ & $5.74\times$ & $9.58\times$\\
32 & 1 & $4.08\times$ & $9.48\times$ & $4.39\times$ & $11.30\times$\\
32 & 4 & $2.58\times$ & $4.44\times$ & $7.07\times$ & $14.32\times$\\
\bottomrule
\end{tabular}

\end{table}

\subsubsection{Inference with the path Transformer readout for \PMEVT{}}
\label{app:t2-transformer-readout-inference}

For \STEE{}, \(\STarch[\EVT,\MLP_{\mathrm{add}}]\), and \PMEVT{}, we time
the five validation-selected checkpoints per model from
\cref{tab:task2-model-comparison} on the same 12 test graphs as in
\cref{app:task2-trained-cache}. The four-query requests use the original
test paths. The 16-query requests keep those paths and add 12 from the
same deterministic path lists. Every query is processed in a batch of
one. Each StrikeTwice model uses identical weights and queries in both
inference modes. \PMEVT{} encodes the marked graph for every query.

We run CPU float32 inference using four computation threads, with three
warm-ups and seven timed repetitions per condition. Times include initial
graph encoding, path processing, and prediction. Data loading and input
assembly are outside the timed region. We take the median over repetitions
for each graph and seed, then average over graphs and seeds. Each speedup
is the ratio of mean times for the same number of queries.

\begin{table}[htbp]
\centering\small
\caption{Task~2 CPU inference with trained models. Times are milliseconds
to process all queries on one graph. The \PMEVT{} time is reported for four
queries.}
\label{tab:t2-transformer-readout-latencies}
\begin{tabular}{@{}lrrrr@{}}
\toprule
Model & Queries & Encode per query (ms) & Reuse (ms) & Reuse speedup \\
\midrule
\STEE{} & 4 & 44.15 & 18.38 & \(2.40\times\) \\
\(\STarch[\EVT,\MLP_{\mathrm{add}}]\) & 4 & 36.78 & 11.42 & \(3.22\times\) \\
\PMEVT{} & 4 & 42.44 & --- & --- \\
\STEE{} & 16 & 176.64 & 46.78 & \(3.78\times\) \\
\(\STarch[\EVT,\MLP_{\mathrm{add}}]\) & 16 & 147.00 & 19.82 & \(7.42\times\) \\
\bottomrule
\end{tabular}
\end{table}

\FloatBarrier
\subsection{Cache scaling and amortization}
\label{app:cache-scaling}

The timing study fixes \(64\) path queries of length \(k=7\) and
varies \(n\in\{28,56,112\}\) and the number
\(Q\in\{1,4,16,64\}\) of distinct paths per immutable graph.  Balanced
\STEE{} and capacity-matched \PMEVT{} use fixed random weights on an
Apple M4 Max CPU (14 cores, 36 GB RAM), with each worker restricted to four
PyTorch threads.  Every cell aggregates ten fresh processes, each with two
warm-ups and ten timed repetitions. We time \STEE{} from graph encoding
through path processing for all queries in one pass. Separate timings of
cache construction and processing are used for diagnostics. \PMEVT{} encodes
the marked graph for each query. A second \STEE{} measurement processes
paths in the same four-query chunks as \PMEVT{}.
Before timing, we check that both cached variants reproduce ordinary
predictions. We divide each process time by its query count before computing
medians, IQRs and speedup ratios. We record peak process memory (RSS) and
retained-cache bytes separately.

\begin{table}[H]
\centering
\setlength{\abovecaptionskip}{0pt}
\caption{Cache scaling. All times are end-to-end milliseconds/query,
median [IQR] over ten fresh-process medians.  Each \STEE{} total is a directly
timed call covering cache construction and all path predictions.
Native processes cached path queries in batches of up to 64;
symmetric uses chunks of up to four queries for both models.
Speedups divide \PMEVT{} latency by the
corresponding \STEE{} total. Bold marks the lowest time in each row.}
\label{tab:appendix-cache-scaling}
\scriptsize
\resizebox{\textwidth}{!}{%
\begin{tabular}{@{}rrlllrrr@{}}
\toprule
$n$ & $Q$ & $\mathrm{ST}[\EVT,\EVT]$ native & $\mathrm{ST}[\EVT,\EVT]$ symmetric & \PMEVT{} & cache/graph & native & symmetric \\
 & & \multicolumn{3}{c}{milliseconds/query, median [process IQR]} & MiB & \multicolumn{2}{c}{speedup} \\
\midrule
28 & 1 & \textbf{4.480 [4.466,4.543]} & 5.033 [4.976,5.160] & 5.263 [4.872,5.364] & 0.099 & 1.17$\times$ & 1.05$\times$ \\
28 & 4 & \textbf{1.264 [1.172,1.325]} & 1.670 [1.610,1.714] & 4.896 [4.699,5.044] & 0.099 & 3.87$\times$ & 2.93$\times$ \\
28 & 16 & \textbf{0.491 [0.486,0.507]} & 0.978 [0.965,1.003] & 4.579 [4.564,5.102] & 0.099 & 9.32$\times$ & 4.68$\times$ \\
28 & 64 & \textbf{0.304 [0.287,0.346]} & 0.737 [0.701,0.866] & 5.346 [5.158,5.530] & 0.099 & 17.61$\times$ & 7.25$\times$ \\
56 & 1 & \textbf{14.568 [14.528,15.135]} & 14.976 [14.925,15.433] & 17.745 [17.618,18.146] & 0.390 & 1.22$\times$ & 1.18$\times$ \\
56 & 4 & \textbf{3.764 [3.735,3.804]} & 4.200 [4.179,4.234] & 19.330 [18.854,19.755] & 0.390 & 5.14$\times$ & 4.60$\times$ \\
56 & 16 & \textbf{1.113 [1.092,1.118]} & 1.538 [1.522,1.562] & 18.156 [17.853,18.669] & 0.390 & 16.31$\times$ & 11.81$\times$ \\
56 & 64 & \textbf{0.473 [0.466,0.475]} & 0.878 [0.873,0.881] & 17.812 [17.746,17.953] & 0.390 & 37.62$\times$ & 20.29$\times$ \\
112 & 1 & 114.433 [113.796,117.680] & \textbf{113.203 [111.362,116.981]} & 123.610 [122.921,124.604] & 1.545 & 1.08$\times$ & 1.09$\times$ \\
112 & 4 & \textbf{28.547 [28.048,28.998]} & 29.087 [28.555,29.516] & 123.381 [122.531,124.000] & 1.545 & 4.32$\times$ & 4.24$\times$ \\
112 & 16 & \textbf{7.268 [7.229,7.331]} & 7.805 [7.766,7.843] & 123.439 [122.591,123.811] & 1.545 & 16.98$\times$ & 15.81$\times$ \\
112 & 64 & \textbf{2.060 [2.050,2.076]} & 2.602 [2.578,2.622] & 123.588 [122.604,124.256] & 1.545 & 59.98$\times$ & 47.49$\times$ \\
\bottomrule
\end{tabular}
}
\end{table}

Under symmetric chunking, the ratio of median \PMEVT{} to
\STEE{} latency at \(Q=1\) ranges from \(1.05\times\) to \(1.18\times\).
At \(n=28\), however, the fresh-process IQRs overlap substantially, so this
smallest case indicates similar latency. The
\(Q=1\) process-IQRs separate at \(n=56\) and \(n=112\), and cache reuse gives
large gains for \(Q\geq4\).  Native-throughput values from \(Q=4\) onward are
labelled amortization plus batching: at \(n=112\), they reach
\(4.32\times,16.98\times,59.98\times\) for \(Q=4,16,64\), versus
\(4.24\times,15.81\times,47.49\times\) under symmetric chunking.
The symmetric column isolates reuse from
the advantage of larger query batches and preserves a large gain. The exact retained
cache grows from \(0.099\) to \(0.390\) and \(1.545\) MiB per graph as \(n\)
doubles, consistent with dense ordered-pair storage. These CPU measurements
quantify how repeated queries trade cache memory for lower inference time.

\FloatBarrier

\section{ThorAxe-Combo: Benchmark Details}
\label{app:thoraxe-combo}

\subsection{Candidate construction, cohorts, and graph inputs}
\label{app:biology-data}

We use the fixed spliced-exon segments and protein-isoform paths produced by ThorAxe
\citep{zea2021thoraxe}. For each gene--clade group, paths observed in the
designated clade are reserved for candidate eligibility and target annotation.
Paths observed outside that clade define the visible graph and supply two
donor paths. A gene may contribute several clade groups.
The donor paths are split at shared spliced-exons, and candidates recombine the
intervening alternative blocks, each potentially containing several spliced-exons.
Candidate paths and focal positions are determined from these visible donor
paths before reserved-clade annotations are applied.

Candidates recombine at least two differing blocks, each
observed in a protein-isoform path from the reserved clade.
The two focal positions correspond to the first and last differing
blocks, and the number of recombined blocks may exceed two.
The complete candidate path must be absent from the
visible paths. Its label is $1$ if it is observed in the reserved clade
and $0$ otherwise. A negative label records absence from the available
annotations. Biological compatibility remains open.

\begin{table}[H]
\centering
\setlength{\abovecaptionskip}{0pt}
\caption{ThorAxe-Combo cohort drawn from 5,000 source genes,
retaining visible graphs with at most 128 vertices and groups containing
both classes. Counts cover the union of the five outer test folds.}
\label{tab:biology-cohorts}
\small
\begin{tabular}{@{}lr@{}}
\toprule
Property & ThorAxe-Combo\\
\midrule
Eligible genes & 1,072\\
Gene $\times$ clade groups & 1,596\\
Candidates & 16,544\\
Positive / negative candidates & 3,120 / 13,424\\
Median candidates per group & 5\\
Groups with one positive & 899\\
\bottomrule
\end{tabular}

\end{table}

\paragraph{Graph and sequence features.}
The input graph is rebuilt from visible protein-isoform paths, with directed edges
between consecutive spliced-exons. For each spliced-exon, precomputed 960-dimensional
ESM-C 300M embeddings \citep{scale2024esm} are averaged over visible instances
within each species, then across species. Four flags indicate non-boundary,
start, stop, and embedding availability, giving 964 vertex channels.
Seven pair channels encode forward and reverse adjacency, visible-species
and visible-path support fractions, intron presence, normalized log nucleotide
gap, and a self-pair indicator. Support fractions concern adjacent edges.
Graph-based models use an input stem reducing vertex and pair features to 8
and 4 channels. GINE uses the compressed vertex features and directed
adjacency.

\subsection{Models, readouts, and parameter budgets}
\label{app:biology-models}

We evaluate a full path EVT, a dense-attention path processor (TD), and
the additive and concatenating MLP processors defined in
\cref{sec:experiments}. \STEE{} uses triangular pair attention at both
stages. TD keeps triangular graph attention and uses dense self-attention
over the $k^2$ path-pair tokens, retaining the EVT vertex stream and
cross-attentions. Path-pair attention in TD costs $O(k^4d)$, compared with
$O(k^3d)$ in \STEE{}. GINE-Transformer applies directed GINE message passing
followed by a path Transformer.

The graph EVT uses four orientations of triangular attention. All graph
encoders have two layers. Attention-based path processors have two layers
and four heads. We evaluate three readouts for each listed StrikeTwice
model and use the vertex-state readout for GINE-Transformer.

\subsubsection{Concatenating MLP path processor}
After selection, a shared MLP transforms each pair state and a second
MLP reads its row mean concatenated with the vertex state.
\[
R_{st}=\phi(\widehat Y^0_{st}),\qquad
Z_s=\rho\!\left([\widehat X^0_s;k^{-1}\textstyle\sum_t R_{st}]\right),
\]
on the selected, positioned inputs $\widehat X^0,\widehat Y^0$.
The ThorAxe-Combo \(\STarch[\ET,\MLP_{\mathrm{concat}}]\) uses the
endpoint-lifted ET implementation with chain-only triangular attention
at width 80. The corresponding synthetic configuration uses all four
orientations at width 88 (\cref{app:experimental-interfaces}). Both
use ReLU in their graph blocks and GELU in their path MLPs. Their MLP
widths and task heads are task-specific.

\subsubsection{Additive MLP path processor}
\label{app:biology-additive-mlp}
\(\STarch[\EVT,\MLP_{\mathrm{add}}]\) and \(\STarch[\ET,\MLP_{\mathrm{add}}]\) both use a
one-layer MLP path processor following the EVT update order.
In this variant, vertex self-attention,
triangular pair attention, and vertex-to-pair exchange are disabled.
The first aligned FFN updates pairs only. The pair-to-vertex exchange
uses a uniform mean over valid targets in each source row in place of
the full EVT's global pair-to-vertex attention. The final FFN updates
vertices only.
Input projections, active LayerNorms, and residual connections
are retained. Write \((\widehat U,\widehat V)\) for the selected,
tagged cache tensors and \(P_{\mathcal{X}},P_{\mathcal{Y}}\) for this
processor's learned affine input projections, which realize its affine
graph-to-path interface. Omitting dropout and writing
the valid path length as $k$, its path operations are
\[
\widehat Y^0=P_{\mathcal{Y}}(\widehat V),\qquad
R=\widehat Y^0+\mathrm{FFN}_p(\mathrm{LN}_p(\widehat Y^0)),
\]
\[
B_s=P_{\mathcal{X}}(\widehat U_s)+k^{-1}\sum_{t=1}^{k}
\mathrm{LN}_{\mathrm{cross}}(R_{st}),\qquad
Z_s=B_s+\mathrm{FFN}_v(\mathrm{LN}_v(B_s)).
\]
Here \(Z=\widehat X\) and \(R=\widehat Y\) are the
internal path states supplied to the task-specific layer, and \(B_s\) is the
intermediate vertex state.
\(\STarch[\EVT,\MLP_{\mathrm{add}}]\) uses width 32, graph FFN width 128, and path
FFN width 1,182. \(\STarch[\ET,\MLP_{\mathrm{add}}]\) uses width 80, graph FFN width
279, and path FFN width 333. Its graph stage uses an ET-like
configuration with endpoint-lifted pairs, chain-only triangular attention,
and a pairwise ReLU FFN. Vertex outputs are read from the pair diagonal. The path FFNs use GELU in both
models. Thus the two configurations share the path-processing schedule,
with different graph encoders and capacity allocations.
The path processor in \cref{cor:path-decoder-kernel} retains global
attention and permits variable capacity.

\subsubsection{Prediction heads}
These graph-model readouts include focal vertex states, the mean vertex state along the
path, and a learned clade embedding. These vertex states already contain
graph context. The pair-based readouts access pair states directly.
The vertex-state-only readout projects the focal vertex branch to 64 channels.
The other readouts project focal vertices and pair features to 32 channels each.
Writing the focal positions as $(a,c)$,
the focal-pair readout uses $[R_{ac};R_{ca}]$. The consecutive-pair readout uses
$g_R=(k-1)^{-1}\sum_{s=1}^{k-1}R_{s,s+1}$ over the entire valid path.
Consecutive s-exon pairs can span introns or join segments of the same exon.
Pooling covers the full path and excludes padding.

\begin{table}[H]
\centering\small
\setlength{\abovecaptionskip}{0pt}
\caption{Trainable parameter budgets for the consecutive-pair readout.
GINE-Transformer uses a vertex-state readout. Fixed ESM-C embeddings are
inputs to these models.}
\label{tab:biology-capacity}
\begin{tabular}{@{}lrr@{}}
\toprule
Model & Hidden width & Parameters\\
\midrule
\(\STarch[\EVT,\MLP_{\mathrm{add}}]\) & 32 & 351,107\\
\(\STarch[\EVT,\MLP_{\mathrm{concat}}]\) & 32 & 351,079\\
\(\STarch[\ET,\MLP_{\mathrm{add}}]\) & 80 & 349,671\\
\(\STarch[\ET,\MLP_{\mathrm{concat}}]\) & 80 & 349,679\\
GINE-Transformer & 104 & 341,963\\
TD & 32 & 351,079\\
\STEE{} & 32 & 351,375\\
\bottomrule
\end{tabular}
\end{table}

The total parameter counts are similar across systems with different
hidden widths, FLOPs, and memory requirements.
\(\STarch[\EVT,\MLP_{\mathrm{add}}]\) uses 352,067 parameters with vertex
states and 352,195 with focal pairs, compared with 351,107 with consecutive
pairs. The corresponding counts for \(\STarch[\ET,\MLP_{\mathrm{add}}]\)
are 352,071, 352,391, and 349,671.

\paragraph{ESM-C focal-and-mean baseline.}
This baseline reads the two focal 960-dimensional ESM-C embeddings, the
mean ESM-C embedding over the full candidate path, and a learned
16-dimensional clade embedding. Learned projections map the
concatenated focal embeddings and the path mean to 64 channels each.
A $144\!\to\!64\!\to\!1$ classifier produces the prediction.
The projections and classifier use LayerNorm and GELU.
Valid boundary positions and missing embeddings contribute zeros to the
path mean, which excludes padding. The head takes the embeddings and
clade identity directly as input. Its 200,033 trainable
parameters belong to the prediction head, compared with roughly 350,000
for the graph-based models. ESM-C embeddings remain fixed.

\paragraph{Local sequence references.}
\label{app:biology-local-baselines}
Two further references use a path Transformer (two layers, four heads,
width 64, feed-forward width 256, rotary positions) on the ordered 964 raw
vertex channels, with pointwise LayerNorm and a linear projection to width
64. They bypass the graph models' eight-channel input stem. The final head
concatenates the two focal states, their full-path mean, a 64-channel junction
branch, and a 16-channel clade embedding. It uses dimensions
$272\!\to\!64\!\to\!1$ with LayerNorm and GELU.
In PathTransformer, the junction branch is zero. In
PathTransformer+J, a shared affine $7\!\to\!64$ map transforms
the raw attributes of consecutive path pairs, then their masked mean enters
the final head. These junctions undergo no graph propagation. The two
conditions share module shapes and paired initializations, apart from the
additional affine map, and are trained separately. They provide local-input
references with their own parameter budgets.

\subsection{Grouped evaluation, selection, and optimization}
\label{app:biology-evaluation}

Five outer folds evaluate prediction on unseen genes by holding out whole
gene identifiers, keeping all groups and candidates from a gene together.
Within each outer fold and fixed
encoder--readout condition, learning rates
$\{10^{-5},3\!\times\!10^{-5},10^{-4},3\!\times\!10^{-4},10^{-3},3\!\times\!10^{-3}\}$
are compared on three inner partitions and two seeds. Selection maximizes
mean inner macro-AUROC, breaking learning-rate ties toward the smaller value.
Where several heads are available, the main comparison selects the
readout and its learning rate by inner validation in each outer fold
(\cref{tab:thoraxe-main,tab:thoraxe-additional-readouts}). The fixed-head
comparisons in \cref{tab:biology-heads,tab:thoraxe-additive-mlp}
select a learning rate per head. The comparison of the concatenating MLP
and graph-based references uses 2,340 inner fits and 325 final fits
across thirteen encoder--readout conditions, with five final seeds.
The three sequence references use the same candidates, gene partitions,
learning-rate grid, and selection criterion. Each uses a fixed readout
with 180 inner fits and 25 final fits over five seeds. Together, these
references account for 540 inner and 75 final fits.

For each outer fold, approximately 64\% of all genes train the weights,
16\% select the checkpoint, and 20\% form the outer test.

AdamW uses the selected constant learning rate, no warm-up, weight decay
0.01, and global gradient clipping at 1. Each update processes one gene--clade
group, with up to 16 queries. Graph-based models use micro-batches of four
and evaluation chunks of eight. Both sizes are one for the three sequence references.
Training lasts at most 6,000 updates, with validation every 200
updates and early stopping allowed from update 1,000 after eight evaluations
without improvement. Validation macro-AUROC selects the checkpoint, which may
precede update 1,000. Early stopping and sampling can produce different
completed update counts and numbers of passes over the cohorts.

\paragraph{Stopping behavior.}
\label{app:thoraxe-training-sensitivity}
Among the 25 final runs per model, \STEE{} and
the concatenating MLP configurations with EVT and ET stop before the
cap in 23, 20, and 22 runs, respectively. Their median stopping steps
are 2,600, 4,400, and 3,000, and their median selected-checkpoint steps
are 1,000, 2,800, and 1,400.

\paragraph{Metrics and aggregation.}
AUROC and non-interpolated Average Precision (AP) are computed within each
gene--clade group. For each seed, we average over all groups from the five
outer tests with equal group weights, then report the mean and sample standard
deviation of the five seed scores. This gives each group equal weight,
regardless of its size or outer fold, and evaluates each seed separately.
SD measures variability across training seeds. Under a uniformly random ranking,
conditional on each group's candidate and positive counts, expected AP averages
0.5234, against an expected AUROC of 0.5.

\subsection{Reference results and readout comparisons}
\label{app:biology-results}

\begin{table}[H]
\centering
\setlength{\abovecaptionskip}{0pt}
\caption{Encoder--readout comparison: mean outer-test AUROC over five seeds.
The head is fixed in each column and its learning rate selected internally.
All columns include focal vertices, mean path-vertex context, and clade identity.
``+'' identifies the added relational branch. A dash denotes an untested
condition. Bold marks the highest mean in each column.}
\label{tab:biology-heads}
\small
\begin{tabular}{@{}lrrr@{}}
\toprule
Model & Vertex states & + Focal pairs & + Consecutive pairs\\
\midrule
\(\STarch[\EVT,\MLP_{\mathrm{add}}]\) & 0.605 & 0.602 & 0.698\\
\(\STarch[\EVT,\MLP_{\mathrm{concat}}]\) & 0.632 & 0.618 & \textbf{0.702}\\
\(\STarch[\ET,\MLP_{\mathrm{add}}]\) & 0.628 & 0.621 & 0.689\\
\(\STarch[\ET,\MLP_{\mathrm{concat}}]\) & \textbf{0.654} & \textbf{0.630} & 0.690\\
GINE-Transformer & 0.643 & -- & --\\
TD & 0.619 & 0.607 & 0.638\\
$\mathrm{ST}[\EVT,\EVT]$ & 0.620 & 0.612 & 0.635\\
\bottomrule
\end{tabular}

\end{table}

Moving from the vertex-only readout to the consecutive-pair readout
raises mean AUROC for all four MLP configurations. For
\(\STarch[\ET,\MLP_{\mathrm{concat}}]\), the gain is \(0.036\), with gains
in all five fold averages and all five seed averages. The two MLP
processors give closer means with the consecutive-pair readout for
each graph encoder. These separately trained comparisons also vary
vertex projection width, capacity allocation, and selected learning
rate. Inner validation chooses the consecutive-pair readout for all
StrikeTwice configurations in the main comparison and for the two
models in \cref{tab:thoraxe-additional-readouts}.

\begin{table}[H]
\centering\small
\setlength{\abovecaptionskip}{0pt}
\caption{Additive MLP path processors: mean $\pm$ sample SD over five seeds.
Fixed-readout rows select the learning rate by inner validation. Selected
rows choose the learning rate and readout.}
\label{tab:thoraxe-additive-mlp}
\begin{tabular}{@{}lcc@{}}
\toprule
Readout & AUROC & AP\\
\midrule
\multicolumn{3}{@{}l}{\(\STarch[\EVT,\MLP_{\mathrm{add}}]\)}\\
Vertex states & $0.605\pm0.006$ & $0.596\pm0.005$\\
+ Focal pairs & $0.602\pm0.017$ & $0.592\pm0.012$\\
+ Path edges & $0.698\pm0.012$ & $0.672\pm0.012$\\
Internally selected & $0.698\pm0.012$ & $0.672\pm0.012$\\
\midrule
\multicolumn{3}{@{}l}{\(\STarch[\ET,\MLP_{\mathrm{add}}]\)}\\
Vertex states & $0.628\pm0.006$ & $0.614\pm0.005$\\
+ Focal pairs & $0.621\pm0.014$ & $0.609\pm0.010$\\
+ Path edges & $0.689\pm0.031$ & $0.668\pm0.028$\\
Internally selected & $0.689\pm0.031$ & $0.668\pm0.028$\\
\bottomrule
\end{tabular}

\end{table}

\paragraph{Additional configurations with selected readouts.}
\Cref{tab:thoraxe-additional-readouts} reports two further graph-to-path configurations
on the same gene folds as \cref{tab:thoraxe-main}. Both select the
consecutive-pair readout in every outer fold.
\begin{table}[htbp]
\centering\small
\caption{ThorAxe-Combo performance with the concatenating ET--MLP and
dense-attention path processor. Outer-test means and sample SDs over
five training seeds.}
\label{tab:thoraxe-additional-readouts}
\begin{tabular}{@{}lcc@{}}
\toprule
Model & AUROC & AP\\
\midrule
\(\STarch[\ET,\MLP_{\mathrm{concat}}]\) & \(0.690\pm0.012\) & \(0.668\pm0.012\)\\
TD & \(0.638\pm0.018\) & \(0.622\pm0.017\)\\
\bottomrule
\end{tabular}
\end{table}

\paragraph{Path processing across synthetic and biological tasks.}
\label{app:biology-interpretation}
On Task~1, \STEE{} has a higher mean test \(R^2\) than the additive MLP
configurations at 3,200 updates (\cref{tab:task1-model-comparison}). On ThorAxe-Combo,
the readout affects how both MLP configurations compare with \STEE{}.
With an EVT graph encoder and vertex states alone, the additive and
concatenating MLP configurations achieve mean AUROCs of \(0.605\) and
\(0.632\), compared with \(0.620\) for \STEE{}. With consecutive-pair
states, the respective scores are \(0.698\), \(0.702\), and \(0.635\)
(\cref{tab:biology-heads}).

Task 1 requires products of three consecutive edge weights, explicitly testing
composition after path selection. Biological prediction uses sequence
features and observed path support already processed by a relational graph
encoder. A plausible explanation is that the signal available in these
encoded relations benefits more from direct path-edge readout than from
another stage of triangular composition under the evaluated budgets.
This interpretation remains a hypothesis. Full path length and annotation
coverage may also matter: focal positions identify only part of a longer
path, and labels record observed support. Separating these effects would
require targeted controls.

\FloatBarrier
\subsection{Attention within the additive MLP path processor}
\label{app:biology-path-attention}

The four-condition experiment fixes the two-layer, width-32 graph EVT and
the one-layer MLP path processor of \(\STarch[\EVT,\MLP_{\mathrm{add}}]\). It compares
the reference with added vertex self-attention, triangular pair attention,
or both in the path layer. The pair-only and vertex-only FFNs and uniform
source-row exchange remain fixed through all four conditions.
Each model is trained end to end with the path-edge readout. The common
gene splits, six learning rates, three inner partitions, two inner seeds,
and five final seeds give 720 inner and 100 final fits. Learning rates are
selected separately for each condition. The readout is fixed throughout.
Shared maps use paired initializations, and widths are fixed without
compensating for the added parameters. The reference, vertex-attention,
pair-attention, and two-attention conditions have respectively 351,107,
355,395, 356,451, and 360,739 parameters.

The reference is the additive MLP with the consecutive-pair readout.
Adding vertex attention yields mean AUROC \(0.705\pm0.007\) and AP
\(0.683\pm0.004\), compared with \(0.698\pm0.012\) and \(0.672\pm0.012\)
for the reference. Pair attention gives \(0.690\pm0.014\) and
\(0.667\pm0.012\). Adding both gives \(0.687\pm0.020\) and
\(0.664\pm0.018\). These separately trained, fixed-readout conditions
are reported in \cref{tab:biology-path-attention}. Vertex attention has
the highest mean AUROC and AP. Both variants
that add triangular attention score below the additive reference in
mean AUROC and AP. Five-seed SDs describe training variability.

\subsection{Additional graph and two-stage attention controls}
\label{app:biology-flexibility}

We evaluate two sets of attention deletions with the consecutive-pair
readout fixed. The first removes triangular
attention or vertex self-attention from the first EVT of
\(\STarch[\EVT,\MLP_{\mathrm{add}}]\). Its two-layer graph EVT, one-layer
row-aggregation MLP processor, width 32, graph FFN width 128, and path FFN
width 1,182 remain fixed.
Each enabled graph intra-attention has one four-head module per layer.
A graph deletion removes both modules. The second set starts from two
complete EVTs with two layers per stage, width 32, four attention heads,
and FFN width 128. It removes
triangular attention or vertex self-attention in either stage, or both
intra-attentions in the path stage. Both global cross-stream attentions and
both incidence-aware vertex/pair FFN phases remain active.

All conditions fix the path-edge readout, candidate cohort, five outer gene
folds, three inner partitions, two inner seeds, six learning rates, and
five final seeds used above. Learning rates are selected separately by
inner AUROC. Outer folds serve as held-out tests. The first-stage controls
comprise 540 inner and 75 final fits. The two-stage controls comprise
1,080 inner and 150 final fits. The additive MLP comparisons and attention
controls comprise 3,420 inner and 475 final fits. Shared maps have paired
initial weights. The additive reference gives identical seed scores in the
main comparison and its two attention controls. Removing a branch reduces
active parameter counts
at fixed widths and FFN sizes.

\begin{table}[t]
\centering\scriptsize
\caption{Additional ThorAxe-Combo controls: test mean $\pm$ sample SD over
five seeds. All conditions use the fixed path-edge task-specific layer. $\Delta$AUROC is
the mean paired difference from the reference within the same block.
Wins counts positive differences across five paired training seeds. Parameter counts include the
input stem and task-specific layer.}
\label{tab:biology-flexibility}
\begin{tabular}{@{}lrrrrr@{}}
\toprule
Configuration & Parameters & AUROC & AP & $\Delta$AUROC & Wins\\
\midrule
\multicolumn{6}{@{}l}{\(\STarch[\EVT,\MLP_{\mathrm{add}}]\): first-stage controls}\\
Reference & 351,107 & $0.698\pm0.012$ & $0.672\pm0.012$ & --- & ---\\
Graph: no triangular attention & 340,419 & $0.703\pm0.005$ & $0.679\pm0.003$ & $+0.005$ & 4/5\\
Graph: no vertex attention & 342,531 & $0.697\pm0.010$ & $0.674\pm0.011$ & $-0.001$ & 3/5\\
\midrule
\multicolumn{6}{@{}l}{\(\mathrm{ST}[\EVT,\EVT]\): fixed-head controls}\\
Reference & 351,375 & $0.639\pm0.019$ & $0.623\pm0.017$ & --- & ---\\
Graph: no triangular attention & 340,687 & $0.634\pm0.031$ & $0.619\pm0.026$ & $-0.005$ & 2/5\\
Graph: no vertex attention & 342,799 & $0.642\pm0.025$ & $0.625\pm0.021$ & $+0.003$ & 2/5\\
Path: no triangular attention & 340,687 & $0.658\pm0.043$ & $0.641\pm0.035$ & $+0.019$ & 3/5\\
Path: no vertex attention & 342,799 & $0.623\pm0.018$ & $0.613\pm0.015$ & $-0.017$ & 1/5\\
Path: neither intra-attention & 332,111 & $0.647\pm0.030$ & $0.632\pm0.025$ & $+0.007$ & 2/5\\
\bottomrule
\end{tabular}

\end{table}

Removing graph triangular attention from \(\STarch[\EVT,\MLP_{\mathrm{add}}]\)
gives AUROC \(0.703\pm0.005\), versus \(0.698\pm0.012\) for the reference.
Four of five seed differences are positive. Removing graph vertex attention gives
\(0.697\pm0.010\). For \(\mathrm{ST}[\EVT,\EVT]\), removing path triangular
attention gives \(0.658\pm0.043\), versus \(0.639\pm0.019\), with only
three positive seed differences. Removing path vertex attention lowers
the mean to \(0.623\pm0.018\), with four negative seed differences.
Graph-stage restrictions produce small, variable mean changes.
The independent \(\mathrm{ST}[\EVT,\EVT]\) fits with a fixed
consecutive-pair readout have AUROC \(0.639\). The readout-selection experiment gives \(0.635\)
(\cref{tab:thoraxe-main}). The additive MLP scores higher than all
tested two-EVT restrictions on each paired seed. The two-EVT restrictions retain
global exchanges and coupled FFNs.

The effects of attention depend on the configuration and stage at these
training budgets. For the additive MLP with the selected readout,
mean AUROC is \(0.698\) with EVT and \(0.689\) with ET.
EVT scores higher on one of five paired seeds. One low ET score accounts
for much of the difference in means and the larger ET standard deviation.

\begin{table}[t]
\centering\scriptsize
\caption{Learning rates selected independently by inner validation for the
additional fixed-readout ThorAxe-Combo controls. Each entry uses all three
inner partitions and both inner seeds. Ties prefer the smaller rate.}
\label{tab:biology-flexibility-lrs}
\begin{tabular}{@{}lccccc@{}}
\toprule
Configuration & Outer 0 & Outer 1 & Outer 2 & Outer 3 & Outer 4\\
\midrule
\multicolumn{6}{@{}l}{\(\mathrm{ST}[\EVT,\MLP]\)}\\
Reference & $10^{-4}$ & $3\!\times\!10^{-4}$ & $3\!\times\!10^{-5}$ & $10^{-4}$ & $10^{-4}$\\
Graph: no triangular attention & $10^{-4}$ & $10^{-4}$ & $10^{-4}$ & $10^{-4}$ & $10^{-4}$\\
Graph: no vertex attention & $3\!\times\!10^{-4}$ & $10^{-4}$ & $3\!\times\!10^{-4}$ & $3\!\times\!10^{-4}$ & $3\!\times\!10^{-5}$\\
\midrule
\multicolumn{6}{@{}l}{\(\mathrm{ST}[\EVT,\EVT]\)}\\
Reference & $10^{-3}$ & $3\!\times\!10^{-4}$ & $10^{-3}$ & $3\!\times\!10^{-4}$ & $3\!\times\!10^{-4}$\\
Graph: no triangular attention & $10^{-3}$ & $3\!\times\!10^{-4}$ & $3\!\times\!10^{-4}$ & $3\!\times\!10^{-4}$ & $10^{-3}$\\
Graph: no vertex attention & $3\!\times\!10^{-4}$ & $3\!\times\!10^{-4}$ & $3\!\times\!10^{-4}$ & $3\!\times\!10^{-4}$ & $3\!\times\!10^{-4}$\\
Path: no triangular attention & $10^{-4}$ & $3\!\times\!10^{-4}$ & $10^{-4}$ & $3\!\times\!10^{-4}$ & $3\!\times\!10^{-4}$\\
Path: no vertex attention & $10^{-4}$ & $10^{-3}$ & $3\!\times\!10^{-4}$ & $3\!\times\!10^{-4}$ & $10^{-3}$\\
Path: neither intra-attention & $10^{-4}$ & $3\!\times\!10^{-4}$ & $10^{-4}$ & $3\!\times\!10^{-4}$ & $3\!\times\!10^{-4}$\\
\bottomrule
\end{tabular}

\end{table}

The metric verification and available data for these comparisons
are reported in \cref{app:biology-provenance,app:biology-scope}.

\FloatBarrier
\subsection{Path marking and graph reuse}
\label{app:thoraxe-marked-application}
\PMEVT{} encodes the marked graph separately for each query and applies a
path Transformer to the selected vertex states. We use the six-rate
nested selection protocol in \cref{app:biology-evaluation}, with three
inner folds and two inner training seeds for each of five outer gene
splits. The 180 inner fits select rates for 25 final fits. The
vertex-only readout yields mean AUROC $0.6362\pm0.0136$ on held-out
genes, compared with $0.6354\pm0.0233$ for \STEE{} with the
consecutive-pair readout. The SD is computed over five seed aggregates,
each pooling groups across all outer splits. Total wall time averages 38.56
versus 13.76 minutes per final fit ($2.80\times$). These times include
data loading, model setup, training, validation and final prediction.
\PMEVT{} uses training and evaluation query chunks of one, compared with
four and eight for \STEE{}, and has 315,055 versus 351,375 parameters.
The ratio measures end-to-end cost for two systems with different readouts,
query chunk sizes, and stopping times.

\paragraph{ThorAxe-Combo inference timing.}
The A100 timing study in \cref{sec:biology} uses the trained \PMEVT{}
checkpoint for seed~0 in each outer fold and the same initialized
\STEE{} weights across folds. Both process all 16,544 candidates in
batches of up to eight. We take the median of three synchronized
repetitions per gene--clade group and divide the sum of these times by
the total query count. \STEE{} takes \(3.339\) ms per query when encoding
once per batch and \(2.538\) ms when retaining the cache across batches.
\PMEVT{} takes \(3.577\) ms per query, with graph encoding conditioned on
each candidate.

\FloatBarrier
\subsection{Vertex-input projection width}
\label{app:biology-node-projection}

For \STEE{}, we compare projections from 964 vertex-input channels to
either 8 or 32 channels. Both variants use a width-32 encoder, path-edge
readout, and four-channel pair-input projection. The 32-channel projection
adds 23,928 parameters. Both variants are trained independently. The pilot
uses inner folds 0 and 1 of outer fold 0, two training seeds, and learning rates
$\{10^{-5},10^{-4},3\!\times\!10^{-4},10^{-3}\}$ per configuration,
for 32 fits. The update ceiling and checkpoint-selection and early-stopping
rules follow \cref{app:biology-evaluation}.
For each projection width, we select the learning rate with the
highest mean validation macro-AUROC across the two folds and two seeds.
All pilot scores come from inner validation.

\begin{table}[H]
\centering
\setlength{\abovecaptionskip}{0pt}
\caption{ThorAxe-Combo vertex-input projection pilot for \STEE{}.
Mean validation macro-AUROC over two inner folds and two seeds at the
selected learning rate of $3\!\times\!10^{-4}$ for both widths.
Bold marks the higher mean.
These are exploratory validation scores, distinct from the outer-test
results in \cref{tab:thoraxe-main,tab:thoraxe-additional-readouts}.}
\label{tab:biology-node-projection}
\small
\begin{tabular}{@{}lcc@{}}
\toprule
Model & Projection to 8 channels & Projection to 32 channels\\
\midrule
$\mathrm{ST}[\EVT,\EVT]$ & 0.613 & \textbf{0.623} \\
\bottomrule
\end{tabular}

\end{table}

For \STEE{}, the 32-channel projection raises mean inner-validation
macro-AUROC from \(0.613\) to \(0.623\) (\cref{tab:biology-node-projection}).
The per-fit validation scores are available for this comparison.

\subsubsection{Input-width pilot with the additive MLP}
We train \(\STarch[\EVT,\MLP_{\mathrm{add}}]\) with a 964-to-8 or 964-to-32 vertex-input
projection. Both variants use a four-channel pair-input projection, a
width-32 graph encoder, and the same path processor and junction readout.
The two models have 351,107 and 375,035 trainable parameters, respectively.
For each width and each of four learning rates, we trained on inner
splits 0 and 1 of outer fold~0 using training seeds 0 and 1, for 32 fits.
Each fit used the checkpoint with the highest validation macro-AUROC.
The validation sets contained 3,984 and 5,044 candidates in 407 and 422
gene--clade groups, respectively. We computed AUROC and average precision
within each group, averaged groups equally within each fit, then averaged
the four fold--seed scores for each width and rate.

\begin{table}[H]
\centering\small
\caption{ThorAxe-Combo inner-validation scores for \(\STarch[\EVT,\MLP_{\mathrm{add}}]\) by
vertex-input width and learning rate. Bold marks the rate selected
independently for each width by mean macro-AUROC. \(\Delta\) is the mean
paired difference (width 32 minus width 8).}
\label{tab:thoraxe-additive-node-width}
\begin{tabular}{@{}lrrr@{\hspace{1.8em}}rrr@{}}
\toprule
 & \multicolumn{3}{c}{Macro-AUROC} & \multicolumn{3}{c}{Macro-AP}\\
\cmidrule(lr){2-4}\cmidrule(l){5-7}
Learning rate & Width 8 & Width 32 & \(\Delta\) & Width 8 & Width 32 & \(\Delta\)\\
\midrule
\(10^{-5}\)           & 0.6775 & 0.6728 & \(-0.0046\) & 0.6562 & 0.6507 & \(-0.0055\)\\
\(\mathbf{10^{-4}}\)  & 0.7082 & 0.7031 & \(-0.0052\) & 0.6839 & 0.6779 & \(-0.0060\)\\
\(3\times 10^{-4}\)   & 0.6844 & 0.6777 & \(-0.0067\) & 0.6632 & 0.6567 & \(-0.0065\)\\
\(10^{-3}\)           & 0.6565 & 0.6087 & \(-0.0478\) & 0.6308 & 0.5971 & \(-0.0337\)\\
\bottomrule
\end{tabular}

\end{table}

Both widths select \(10^{-4}\). At this rate, widening reduces mean
macro-AUROC from \(0.7082\) to \(0.7031\) and mean macro-AP from
\(0.6839\) to \(0.6779\). Width 8 leads on both validation folds for seed 0
and width 32 leads on both folds for seed 1.
The mean paired differences (32 minus 8) are
\(-0.0052\) in macro-AUROC and \(-0.0060\) in macro-AP, with sample
standard deviations of \(0.0103\) and \(0.0109\) across the four pairs.
We retain the eight-channel projection for the reference model because
it has the higher mean on both metrics at every tested rate and uses
23,928 fewer parameters. Selection and comparison use the same
inner-validation folds. The held-out-gene test scores from the main
nested evaluation are reported in \cref{tab:thoraxe-main}.

\subsection{Evaluation scope and evidence}
\label{app:biology-scope}

The evaluation concerns unseen genes within the eligible population.
All four reserved clades---birds, Laurasiatheria, primates, and rodents---also
occur among training genes, and homologous families may recur.
ThorAxe segmentation and alignments are fixed upstream. Excluding a clade
rebuilds the visible graph and features from the remaining paths.
Candidate eligibility uses reserved-clade support annotations.

\paragraph{Comparability and verification.}
\label{app:biology-provenance}
The model comparisons use the same candidate cohort and gene splits,
with selection confined to inner validation. We checked candidate
identities, labels, gene splits, clade mappings, learning-rate grids,
training seeds, and budgets for the additive MLP comparisons and their
attention controls. We reconstructed model selection from 3,420 inner
fits. The reported metrics for 475 final fits were recomputed from saved
per-candidate predictions and 151,620 outer-test gene--clade metric rows,
with equal weights across groups.

The evaluation records also cover 325 graph-model fits and 540 inner
and 75 final fits for the three sequence references. The recorded
inner-validation scores and outer-test group metrics support
reconstruction of the reported selection decisions and numerical summaries.

\subsection{Input and evidence availability}
\label{app:biology-software}
The candidate cohort and the 20 outer and inner gene partitions are fixed.
The full biological inputs, including pretrained ESM-C representations,
are distributed separately through Figshare. Benchmark preparation uses
inputs from ThorAxe, Ensembl, and ESM-C. The accompanying evidence includes
group metrics and selection records. Exact prediction replay requires
verification against the embedding snapshot used for training and access to
the corresponding predictor weights. Those weights and complete
per-candidate predictions are unavailable in the supplement.

\FloatBarrier
\section{Atlas: Data, Training, and Cache Reuse}
\label{app:atlas}

Atlas evaluates round-trip-time prediction along observed network paths.
We describe the data and checkpoint selection, then compare training
budgets, readers on fixed graph states, and the execution cost of cache
reuse.

\subsection{Data and temporal evaluation}
\label{app:atlas-validation}

RIPE Atlas region-mesh measurements supply observed responding-interface
paths and round-trip times (RTTs) \citep{ripencc2015atlas,ripeatlasmeshes}.
Each region/month/destination-address graph retains up to 256 interfaces,
and paths have up to 32 occurrences. Snapshot topology and transition
counts use unlabelled routes from that snapshot. The prior at each
responding interface is the median of its valid February--March 2026 RTT
observations, pooled across probes and targets. For an interface without
observations in those months, we use the median of all valid RTTs in the
same period. The inputs also include snapshot context features. Models
predict corrections to the prior (\cref{app:profile-input-access}).
PathTransformer combines each path
vertex with locally projected incoming and outgoing path-edge attributes
before its two self-attention blocks. The StrikeTwice models encode the
full graph and reuse its states across paths.

Training uses April--July. We select learning rates and checkpoints on a
fixed cohort of 4,096 August queries. The full August partition contains
30,838 paths and 218,474 labelled occurrences in 701 probe--region--month
blocks. The September test contains 30,419 paths and 213,556 labelled
occurrences in 734 blocks. Probe--region--month blocks are the resampling
units. We evaluate RTT prediction on known snapshot topology. Missing
replies are masked. Responding interfaces can come from different packet
routes, and differences in RTT combine outward and return travel times.

\subsection{Temporal comparison across model capacities}
\label{app:application-protocol}

AdamW uses weight decay \(0.01\), gradient-norm clipping at 1, and a constant
learning rate. Effective batches contain up to 16 paths. Training minimizes
unclipped absolute error averaged over observed targets within each query
and then equally across valid queries, divided by the fixed 10-ms residual
scale. Masks exclude missing targets and padding. We validate every 500
updates, with early stopping after eight checks without improvement and a
minimum of 2,000 updates.

PathTransformer and \STEE{} receive a three-rate screen at width 64 with
seed 0 and 2,000 updates, followed by further validation-based architecture
and learning-rate selection. We train the selected configurations at seeds
0--5 with an 8,000-update cap. The two \(\MLP_{\mathrm{add}}\)
configurations use width 64, two graph layers, four graph-attention heads,
graph FFN width 128, one residual pair/vertex MLP path block, and zero
dropout. The screen for each additive model uses rates
\(10^{-4},3\times10^{-4},10^{-3}\), seed 0 and 2,000 updates. It selects
\(10^{-4}\) for \(\STarch[\EVT,\MLP_{\mathrm{add}}]\) and
\(3\times10^{-4}\) for \(\STarch[\ET,\MLP_{\mathrm{add}}]\). We train
each selected additive model at seeds 1--5. All final fits reach 8,000
updates. The models have different parameter counts and search budgets
(\cref{tab:application-configurations}).

The additive MLP updates the selected pair states, aggregates them by
source row, and updates the corresponding path-vertex states. The final
head reads each updated vertex state together with its normalized
occurrence index. The ET graph encoder uses chain-only triangular
attention and diagonal pair states as vertex proxies. The EVT encoder
couples vertex and pair states and uses four graph-attention orientations.

\begin{table}[htbp]
\centering\small\setlength{\tabcolsep}{4pt}
\begin{tabular}{@{}lrrrrr@{}}\toprule
Model & Width & \shortstack{Layers\\G/P} & \shortstack{Heads\\G/P} & LR & \shortstack{Params\\(k)}\\\midrule
PathTransformer & 64 & --/2 & --/4 & \(3\times10^{-4}\) & 101.2\\
\(\STarch[\EVT,\MLP_{\mathrm{add}}]\) & 64 & 2/1 & 4/-- & \(10^{-4}\) & 436.0\\
\(\STarch[\ET,\MLP_{\mathrm{add}}]\) & 64 & 2/1 & 4/-- & \(3\times10^{-4}\) & 135.2\\
\STEE{} & 64 & 2/2 & 4/8 & \(3\times10^{-4}\) & 778.9\\
\bottomrule\end{tabular}

\caption{Atlas model configurations in the comparison across capacities.
G/P denotes graph/path layers or heads. Parameter counts are in thousands.}
\label{tab:application-configurations}
\end{table}

\subsubsection{Validation performance and training throughput}

At the common 3,000-update validation checkpoint, selected for this
descriptive comparison, mean MAE is \(11.948\pm0.068\)\,ms for
PathTransformer and \(11.899\pm0.127\)\,ms for \STEE{}. The six fits
at seeds 0--2 reach 8,000 updates. At that endpoint, \STEE{} has lower
validation error in all three paired seeds and about 3.5\% lower mean
error than PathTransformer (\cref{tab:atlas-validation-long}). Its
warm training throughput is lower.

\begin{table}[htbp]
\centering\small\setlength{\tabcolsep}{4pt}
\begin{tabular}{@{}lrrr@{}}
\toprule
Model & \shortstack{8k final\\MAE (ms)$\downarrow$} & \shortstack{Best checkpoint\\MAE (ms)$\downarrow$} & \shortstack{Warm train\\queries/s}\\
\midrule
PathTransformer & 11.606 $\pm$ 0.180 & 11.366 $\pm$ 0.127 & \textbf{295.9} \\
$\mathrm{ST}[\EVT,\EVT]$ & \textbf{11.202 $\pm$ 0.264} & \textbf{11.004 $\pm$ 0.060} & 44.7\\
\bottomrule
\end{tabular}

\caption{Atlas validation, mean \(\pm\) sample SD over seeds 0--2.
Results are shown at 8,000 updates and at each validation-selected
checkpoint. Throughput measures warm training queries per second.}
\label{tab:atlas-validation-long}
\end{table}

For PathTransformer and \STEE{} checkpoints selected at seeds 0--2,
we average query errors over the three fitted seeds and resample 688
whole probe blocks in 4,000 paired bootstrap draws. The resulting
estimate is a \(0.362\)\,ms reduction for \STEE{} with a conditional
95\% interval of \([0.150,0.649]\)\,ms. The interval describes block
resampling on the validation set, conditional on the fitted models.

\subsubsection{September error and variation across seeds}
\label{app:atlas-confirmation}

We verify that the saved predictions on the 4,096-query August cohort
replay exactly on scored entries for the six selected PathTransformer
and \STEE{} fits at seeds 0--2. Architectures, rates and checkpoint
rules are fixed for September evaluation. Seeds 3--5 use the same
settings with separate initializations and sampling. The comparison
pools seeds 1--5, with seed 0 reserved for development
(\cref{tab:atlas-temporal-test,tab:atlas-seeds}). The selected checkpoint
is the 8,000-update endpoint in eight of the ten additive MLP fits.
At seed 4, the selected checkpoints are at 6,000 updates for
\(\STarch[\EVT,\MLP_{\mathrm{add}}]\) and 7,500 updates for
\(\STarch[\ET,\MLP_{\mathrm{add}}]\).

\begin{table}[htbp]
\centering\small
\begin{tabular}{@{}lcc@{}}\toprule
System & Selected & 8k endpoint\\\midrule
PathTransformer & $11.201\pm0.089$ & $11.234\pm0.128$\\
\(\STarch[\EVT,\MLP_{\mathrm{add}}]\) & $11.245\pm0.047$ & $11.245\pm0.047$\\
\(\STarch[\ET,\MLP_{\mathrm{add}}]\) & $11.210\pm0.038$ & $11.220\pm0.036$\\
\(\mathrm{ST}[\EVT,\EVT]\) & $\mathbf{10.940\pm0.102}$ & $\mathbf{10.953\pm0.118}$\\
\bottomrule\end{tabular}

\caption{Atlas September test across model capacities. Unclipped query-macro
MAE in ms, mean \(\pm\) sample SD over seeds 1--5. Selected denotes the
checkpoint chosen on the fixed August cohort. The fixed 8,000-update
endpoint appears separately. The February--March RTT prior alone yields
12.809\,ms MAE.}
\label{tab:atlas-temporal-test}
\end{table}

\begin{table}[htbp]
\centering\small
\begin{tabular}{@{}lrrrr@{}}\toprule
Seed & PathTransformer & \(\STarch[\EVT,\MLP_{\mathrm{add}}]\) & \(\STarch[\ET,\MLP_{\mathrm{add}}]\) & \STEE{}\\\midrule
0 (development) & 11.1079 & -- & -- & \textbf{10.8768}\\
1 & 11.2580 & 11.2345 & 11.2543 & \textbf{11.0263}\\
2 & 11.2351 & 11.2639 & 11.2392 & \textbf{10.9261}\\
3 & 11.0732 & 11.1711 & 11.1629 & \textbf{10.8284}\\
4 & 11.2918 & 11.2992 & 11.1826 & \textbf{11.0625}\\
5 & 11.1480 & 11.2548 & 11.2107 & \textbf{10.8585}\\
\bottomrule\end{tabular}

\caption{September MAE at validation-selected checkpoints. The comparison
pools seeds 1--5. Seed 0 participates in the selection of model settings and
is shown where a final fit is available.}
\label{tab:atlas-seeds}
\end{table}

\begin{table}[htbp]
\centering\small
\begin{tabular}{@{}lrrr@{}}\toprule
Seed & PathTransformer & \(\STarch[\EVT,\MLP_{\mathrm{add}}]\) & \(\STarch[\ET,\MLP_{\mathrm{add}}]\)\\\midrule
1 & $0.232$ & $0.208$ & $0.228$\\
2 & $0.309$ & $0.338$ & $0.313$\\
3 & $0.245$ & $0.343$ & $0.334$\\
4 & $0.229$ & $0.237$ & $0.120$\\
5 & $0.289$ & $0.396$ & $0.352$\\
\midrule Mean & $0.261$ & $0.304$ & $0.270$\\
\bottomrule\end{tabular}

\caption{Paired September MAE reductions (ms) for \STEE{}.
Each entry subtracts the error of \STEE{} from that of the column's
model at the same seed. Positive values favor \STEE{}. We compute
differences and their mean across seeds 1--5 before rounding.}
\label{tab:atlas-paired-gains}
\end{table}

\STEE{} achieves lower MAE than each alternative on all five seeds.
Its paired difference against PathTransformer is
\(-0.2609\pm0.0362\)\,ms (mean \(\pm\) sample SD), which describes
variation over initializations on the same September test set. The
block-bootstrap interval above captures variation across August blocks
for the fitted models at seeds 0--2.
\(\STarch[\EVT,\MLP_{\mathrm{add}}]\) and
\(\STarch[\ET,\MLP_{\mathrm{add}}]\) have similar mean errors,
\(11.245\pm0.047\)\,ms and \(11.210\pm0.038\)\,ms, respectively.
The two additive models use 435,977 and 135,233 parameters and run
for 40.92 and 21.95 minutes per final fit on average, including
compilation, validation, and checkpoint overhead. These comparisons
describe complete configurations at a fixed number of updates.

We verify the 40 saved evaluation panels for the additive MLP
configurations with an independent float64 calculation, including five
seeds, selected and final checkpoints, and both full evaluation partitions.
Query identities, targets, priors, masks and padding match the stored
records. The maximum absolute MAE discrepancy is below
\(1.9\cdot10^{-8}\)\,ms.

\FloatBarrier
\subsection{Capacity and learning-rate controls}
\label{app:atlas-common-comparison}

The comparison in \cref{tab:atlas-main} uses a common three-rate
screen (\(10^{-4},3\times10^{-4},10^{-3}\)) on seed 0 for 2,000 updates,
followed by evaluation over five final seeds (1--5) at the 8,000-update
budget. Model selection uses the fixed 4,096-query August cohort and all
models use the same preprocessing and September evaluation cohort.
PathTransformer and both MLP configurations use capacities close to
\STEE{} (\cref{tab:atlas-common-comparison}).

RouteNet \citep{rusek2019routenet} retains path--link message passing and
predicts RTT at each observed position along one query path. The inputs
describe the sampled path and snapshot topology. We compare a ReLU head
with an official-style head containing two SELU layers, dropout, and direct
concatenation of the path state. Both heads use the same search and budget
as the other models in this comparison.

\begin{table}[htbp]
\centering\small
\caption{Atlas parameter counts and learning rates selected by the common
August screen. September MAE is reported in \cref{tab:atlas-main}.
\PMEVT{} uses a query-marked graph encoder and a path Transformer.}
\label{tab:atlas-common-comparison}
\begin{tabular}{@{}lrc@{}}
\toprule
Model & Parameters & Selected rate\\
\midrule
\STEE{} & 778,897 & $10^{-4}$\\
\(\STarch[\EVT,\MLP_{\mathrm{add}}]\) & 778,859 & $3\times10^{-4}$\\
\(\STarch[\EVT,\MLP_{\mathrm{concat}}]\) & 778,897 & $3\times10^{-4}$\\
PathTransformer & 793,729 & $10^{-4}$\\
\PMEVT{} & 456,905 & $3\times10^{-4}$\\
RouteNet-ReLU & 779,927 & $3\times10^{-4}$\\
RouteNet-SELU & 781,174 & $3\times10^{-4}$\\
\bottomrule
\end{tabular}

\end{table}

At the common-screen rate, \STEE{} improves on PathTransformer on all
five seeds and has lower mean MAE than both RouteNet adaptations. The
2,000-update August screen favors \(10^{-4}\) over \(3\times10^{-4}\)
for \STEE{} by \(0.012\)\,ms. In a separate 8,000-update control,
\(3\times10^{-4}\) improves August validation on all five seeds and
yields a September mean of \(11.011\pm0.145\)\,ms, compared with
\(11.089\pm0.093\)\,ms at the common-screen rate. We report both in
\cref{tab:atlas-main}.

\subsubsection{Training budget and checkpoint selection}
\label{app:atlas-long-budget}

The budget comparison trains \STEE{}, \(\STarch[\EVT,\MLP_{\mathrm{concat}}]\),
and PathTransformer for 24,000 updates with paired seeds 1--5. We use
\(3\times10^{-4}\) for \STEE{} based on August validation in the 8,000-update
rate control. The concatenating MLP configuration has 778,897 parameters
and retains its screen-selected rate of \(3\times10^{-4}\).
PathTransformer retains \(10^{-4}\). AdamW uses weight decay \(0.01\),
gradient clipping at 1 and effective batch size 16 (microbatch size 8).
We validate every 500 updates on the fixed August cohort. Each fit reaches
24,000 updates, and for each budget we select the checkpoint with the lowest
validation MAE within that prefix of the trajectory
(\cref{tab:atlas-main}).

\STEE{} and PathTransformer improve from 8,000 to 24,000 updates on every
seed, with lower MAE for \STEE{} on all five paired seeds at 24,000.
The five RouteNet-SELU fits continue at the same rate up to 24,000
updates. Three stop between 22,000 and 23,500 updates and two reach the
cap. Their mean test MAE is \(10.886\pm0.110\)\,ms, and \STEE{} has
lower errors on all five paired seeds.

At 24,000 updates on an A100-SXM4-40GB in FP32, mean cumulative wall times
per seed are 8.43\,h for \STEE{} and 21.58\,min for PathTransformer.
They include preparation, training, scheduled validation and checkpoint
writes through the final validation event. Learning-rate search, recovery
writes after validation, and September evaluation are outside the timing
window.

\subsubsection{Additive MLP at matched capacity}
\label{app:atlas-additive-mlp-protocol}

\(\STarch[\EVT,\MLP_{\mathrm{add}}]\) in \cref{tab:atlas-main}
uses 1,457 hidden units in each of its two path-stage feed-forward
networks, giving 778,859 parameters versus 778,897 for \STEE{}.
The three-rate August screen uses seed 0 and 2,000 updates and selects
\(3\times10^{-4}\). All five final seeds (1--5) reach 24,000 updates
at this constant rate. We select the best August checkpoint within each
budget. September MAE decreases on all five seeds between 8,000 and
16,000 updates (\(10.821\pm0.078\)\,ms at 16,000), and again at 24,000.
\Cref{app:atlas-additive-a100} reports the inference latency experiment
using the selected checkpoint from seed 1.

\subsubsection{Training cost of encoding a marked path}
\label{app:marked-applications}

\PMEVT{} encodes the graph separately for each query and applies a path
Transformer to selected vertex states. At the common screen rates and
8,000 updates, its mean test MAE is close to that of \STEE{}
(\cref{tab:atlas-main}). Training and validation take an average
of 9.22\,h per seed for \PMEVT{} and 2.76\,h for \STEE{}
(\(3.34\times\)). Both use effective batch size 16, with microbatches of
two and eight paths, respectively.

\FloatBarrier
\subsection{Readers on fixed graph states}

\subsubsection{Selection and joint reading on identical frozen caches}
\label{app:atlas-readers}
We freeze the EVT graph encoders from end-to-end RTT fits of
\(\STarch[\EVT,\MLP_{\mathrm{concat}}]\) at seeds 0--2. The encoders use width 64, two four-orientation graph
layers, four heads, learning rate $10^{-4}$, and the optimizer settings above.
Their selected checkpoints are updates 6,500, 8,000, and 7,500. Both readers
consume the same fingerprinted vertex and pair tensors; their weights start
from scratch while graph encoders remain fixed.

Each reader has 133,065 trainable parameters and two branches. The first
reads the ordered $P\times P$ selection. The second combines relations to
all ordered path occurrences for each candidate vertex before aggregating
across candidates. Selection restricts candidates to the distinct visited
vertices; joint reading uses every valid graph vertex. Both directed pair
states, vertex states, occurrence positions, and vertex-equality indicators
are supplied, with padding masks and explicit candidate counts. Thus the
selection control has the same additional nonlinear branch as the joint
reader. Both readers are newly trained for this comparison.

The common search uses three learning rates ($10^{-4},3\cdot10^{-4},10^{-3}$)
and two weight decays ($0,0.01$). The 36 fits use 3,000 updates each.
Mean endpoint validation MAE over the three encoders selects
$3\cdot10^{-4}$ and $0.01$ for both readers. Two new reader initializations
(31 and 32) per encoder then run for 8,000 updates, producing 12 fits with
matched sampling. The primary checkpoint minimizes MAE on the fixed August
subset. After checkpoint selection, we score the full August and
September partitions. The September partition is shared with the main
Atlas comparison.

\begin{table}[htbp]
\centering\small\setlength{\tabcolsep}{4pt}
\begin{tabular}{@{}lccc@{}}\toprule
Reader & August selected & September selected & September 8k\\\midrule
Selection ($P\times P$) & $\mathbf{10.706\pm0.059}$ & $\mathbf{10.956\pm0.045}$ & $\mathbf{10.953\pm0.036}$\\
Joint ($P\times V$) & $10.733\pm0.044$ & $11.050\pm0.054$ & $10.997\pm0.049$\\
\bottomrule\end{tabular}

\caption{Frozen Atlas caches: query-macro MAE (ms). We average two reader
initializations within each encoder, then report mean $\pm$ SD across the
three encoders. Selected and 8k distinguish validation selection from the
fixed endpoint. The encoder checkpoints were selected for RTT prediction
before fitting these readers.}
\label{tab:atlas-readers}
\end{table}

Selection gives lower mean MAE at the tested budget
(\cref{tab:atlas-readers}). The difference between joint reading and
selection at the selected checkpoints is $+0.0935$\,ms, with
conditional 95\% block-bootstrap interval $[-0.0021,0.2050]$\,ms. At 8k
the difference is $+0.0436$\,ms ($[0.0061,0.0992]$\,ms). We use 10,000
resamples of the 734 test blocks after averaging fitted readers. These
intervals describe variation across test blocks conditional on the fitted
readers. We verify that both readers see identical cached states, that
gradients reach their trainable branches, and that sampling is paired.
Eleven of the 12 reader fits still improve validation near the end of the
8,000-update budget. This comparison measures how the readers use the
tested finite graph representations.

\subsubsection{Path-stage triangular attention on a frozen encoder}
\label{app:atlas-triangle-control}
We freeze one pretrained \STEE{} graph encoder and compare fresh full-EVT
readers with deletion of only triangular path attention (EVT-) and with the
same deletion plus wider FFNs. The latter approximately matches the
full reader's output-connected parameter count. All remaining attention,
normalization, exchange and readout operations are retained. For each
width and reader, we screen \(10^{-4},3\times10^{-4},10^{-3}\) for 8,000
updates with tuning seed~0. Final fits use three paired seeds
(1--3), identical cached representations and minibatches, and the
constant selected rate. The 2,000-update results select a checkpoint
within that prefix of an 8,000-update fit; their rates were selected by
the 8,000-update screen. August determines selection throughout.

The frozen encoder is the graph backbone from the \STEE{} fit at seed~1
and constant learning rate $10^{-4}$. August validation selects the
checkpoint at update 8,000.
It has width~64, two graph layers, four graph-attention heads and
377,416 frozen parameters. We fix this checkpoint before fitting the
readers, which use their own decoder and readout weights.

\Cref{tab:atlas-frozen-config} specifies the fresh two-layer,
eight-head readers. Counts include input projections and the prediction
head, and exclude the frozen encoder. Registered counts include every
reader parameter; output-connected counts exclude the terminal pair
FFN and normalization, and the unused pair pre-normalizations in EVT-,
which do not affect the vertex readout. FFN compensation targets the
output-connected count. The selected rate is used for all three final
seeds and both budget prefixes.

\begin{table}[!htbp]
\color{black}
\centering\footnotesize
\setlength{\tabcolsep}{4pt}
\caption{Atlas frozen-reader configurations. FFN is the feed-forward
hidden width. Registered and output-connected counts refer to reader
parameters only; LR is selected on the August cohort.}
\label{tab:atlas-frozen-config}
\begin{tabular}{@{}rlrrrr@{}}
\toprule
Width & Reader & FFN & Registered & Output-connected & LR\\
\midrule
32 & Full EVT & 64 & 107,561 & 99,015 & $3\times10^{-4}$\\
32 & EVT- & 64 & 97,001 & 88,327 & $3\times10^{-4}$\\
32 & EVT- + wider FFNs & 78 & 109,713 & 99,219 & $10^{-3}$\\
\midrule
64 & Full EVT & 128 & 401,481 & 368,007 & $3\times10^{-4}$\\
64 & EVT- & 128 & 359,881 & 326,151 & $3\times10^{-4}$\\
64 & EVT- + wider FFNs & 155 & 408,589 & 367,893 & $3\times10^{-4}$\\
\bottomrule
\end{tabular}
\end{table}

\begin{table}[!htbp]
\color{black}
\centering\small
\caption{Atlas frozen-reader control: September MAE (ms), mean
$\pm$ SD over three paired reader seeds, conditional on one frozen
\STEE{} encoder.}
\label{tab:atlas-frozen-readers}
\begin{tabular}{@{}rrccc@{}}
\toprule
Width & Updates & Full EVT & EVT- & EVT- + wider FFNs\\
\midrule
32 & 2,000 & $\mathbf{11.233\pm.052}$ & $11.258\pm.039$ & $11.249\pm.025$\\
32 & 8,000 & $\mathbf{10.678\pm.082}$ & $10.886\pm.047$ & $10.750\pm.063$\\
64 & 2,000 & $\mathbf{11.187\pm.011}$ & $11.237\pm.026$ & $11.235\pm.012$\\
64 & 8,000 & $\mathbf{10.610\pm.070}$ & $10.684\pm.071$ & $10.728\pm.027$\\
\bottomrule
\end{tabular}

\end{table}

Full EVT has lower mean error in every setting. It improves on direct
deletion on all three seeds in each setting. Against FFN compensation,
it wins two of three seeds at width~32 and all three at width~64.
This comparison isolates the contribution of triangular attention to
the path reader on the fixed representations.

\FloatBarrier
\subsection{Inference latency with cached encodings}
\label{app:atlas-reuse}

We compare inference strategies using the same weights and query inputs
within each model and the same physical batch size. Warm synchronized
TPU v6e timings use FP32 and include the first graph encoding. We compile
before timing and exclude host transfers from device latency. Prediction
discrepancies across modes are below \(0.001\)\,ms.

\subsubsection{Full EVT on TPU}
\label{app:atlas-reuse-full-evt}

For the selected seed-0 \STEE{} model, we compare independent graph
encoding per query, sharing within each batch, and persistent reuse
across every batch. We use \(B\in\{1,4,8\}\) and
\(Q\in\{1,4,16,32\}\) where \(B\leq Q\). Nine graphs support all query
counts and have 182--256 vertices, 33--48 paths and path lengths 3--16.
Three additional graphs with paths of length 18--19 provide low-reuse
controls at \(Q=1\). We pad all graphs to 256 vertices and take seven
synchronized warm repetitions per case and mode. The table reports
medians across graphs of their within-case median times, and medians of
the paired case ratios for speedups.

\begin{table}[htbp]
\centering\small\setlength{\tabcolsep}{4pt}
\begin{tabular}{@{}lrrrrrr@{}}\toprule
Model & $B$ & Independent & Within batch & Persistent & I/P & W/P\\\midrule
$\mathrm{ST}[\EVT,\EVT]$ & 1 & 753.10 & 753.52 & \textbf{401.56} & 1.88$\times$ & 1.88$\times$\\
$\mathrm{ST}[\EVT,\EVT]$ & 4 & 194.41 & 188.86 & \textbf{108.88} & 1.79$\times$ & 1.73$\times$\\
$\mathrm{ST}[\EVT,\EVT]$ & 8 & 99.44 & 95.61 & \textbf{60.67} & 1.64$\times$ & 1.58$\times$\\
\bottomrule\end{tabular}

\caption{Atlas inference on nine graphs with \(Q=32\) distinct paths.
Warm device times (ms) include graph encoding and path processing.
Each batch size \(B\) is held fixed across modes.
I/P and W/P compare independent and within-batch encoding with
persistent reuse, respectively. Bold marks the lowest time for each
batch size. Each graph cache stores 16.0625 MiB of FP32 tensors.}
\label{tab:atlas-cache-reuse}
\end{table}

At \(B=8\), persistent reuse takes \(1.896\)\,ms per path for \STEE{}.
Including input transfer and output collection gives a synchronized
request latency of \(79.78\)\,ms for \(Q=32\). The path-only
PathTransformer takes \(21.26\)\,ms at the same shape. A cold-process
probe for \STEE{} takes \(27.09\)\,s, mostly for compilation, and records
an allocator peak of \(72.83\) MiB. The allocator peak includes
working allocations in addition to the cache tensors. Replaying the
4,096-query validation cohort reproduces the stored TPU predictions.

\subsubsection{Additive MLP configurations on TPU}

For \(\STarch[\EVT,\MLP_{\mathrm{add}}]\) and
\(\STarch[\ET,\MLP_{\mathrm{add}}]\) in \cref{tab:application-configurations},
we evaluate the selected seed-1 checkpoints on the same three validation
graphs with 216, 224 and 222 vertices. These graphs are padded to 256
vertices and use 16-position path buckets. Nested requests contain
\(Q\in\{1,4,16,32\}\) distinct paths, with \(B=1\) and also \(B=8\)
for \(Q=16,32\). We compare one encoding per batch with one graph cache
per request, with three warmups and alternating mode order in seven
paired repetitions.

At \(Q=32\) and \(B=8\), reuse reduces request time from 60.156 to
25.923\,ms for \(\STarch[\EVT,\MLP_{\mathrm{add}}]\)
(\(2.32\times\)) and from 35.035 to 19.048\,ms for
\(\STarch[\ET,\MLP_{\mathrm{add}}]\) (\(1.84\times\)). At \(B=1\),
the speedups are \(3.99\times\) and \(2.53\times\), respectively.
We average the three graph-level median times before taking their ratio.
At \(Q=1\), both modes encode the graph once. Each cache has a logical FP32
tensor payload of 16.0625 MiB at 256 vertices. The additive MLP and full
EVT profiles use different graph cohorts, seeds and timing summaries.
Differentiable training has separate recomputation costs
(\cref{app:cache-scaling}).

\subsubsection{Capacity-matched additive MLP on A100}
\label{app:atlas-additive-a100}

We profile the seed-1 checkpoint selected within 24,000 updates in
\cref{app:atlas-additive-mlp-protocol} on nine August graph groups.
On an A100-40GB, we time 32 queries in four batches of eight. We compare
one encoding per batch with one encoding shared across batches. Both
timings include the initial encoding. The median of paired graph-wise
speedups is \(3.94\times\), with identical predicted RTTs.

\FloatBarrier
\subsection{Memory, time and cache amortization at larger graph sizes}
\label{app:large-graph-profiling}
Row/channel tiling evaluates the bilinear triangular operator without
materializing all scores at once
(\cref{app:flash-triatt,prop:flash-triatt}, \citealp{dao2022flash}).
Activation checkpointing \citep{chen2016sublinear} reduces training
memory by recomputing activations in backward. The arithmetic cost
remains cubic and the cache size quadratic. We profile the Atlas
\STEE{} configuration (778,897 parameters,
width~64, two graph and two path layers), holding architecture,
initialized weights, inputs and FP32 precision fixed. Only graph-encoder
execution changes. The \emph{plain} mode uses the native implementation.
The \emph{light} mode uses triangular tiles of 32 source rows and four
value channels, cross-attention query tiles of 128 vertex rows and 4,096
pair rows, and per-attention-tile checkpointing. The \emph{aggressive}
mode uses triangular tiles of eight source rows and two value channels,
cross-attention query tiles of 32 vertex rows and 1,024 pair rows, and
whole-layer checkpointing. Each softmax includes all keys and bridge vertices.
Before profiling, output, loss and all parameter-gradient parity checks
pass at $n=8$ in FP64 and $n=64$ in FP32.

\begin{table}[!htbp]
\color{black}
\centering\small
\caption{Full-size Atlas \STEE{} on synthetic graph shapes: peak allocated
GPU memory in GiB (mean forward-plus-backward time in seconds). Each
request contains one graph and 16 length-16 paths. OOM denotes an observed
failure under a common 35~GiB allocation cap.}
\label{tab:atlas-large-graphs}
\begin{tabular}{@{}lccccc@{}}
\toprule
Execution & $n=64$ & $n=128$ & $n=256$ & $n=512$ & $n=1{,}024$\\
\midrule
Plain & 0.40 (0.091) & 2.38 (0.108) & 17.95 (0.726) & OOM & OOM\\
Light & 0.25 (0.122) & 0.46 (0.158) & 1.62 (0.572) & 6.29 (4.840) & 24.92 (38.442)\\
Aggressive & 0.19 (0.394) & 0.23 (0.724) & 0.70 (1.695) & 2.59 (9.020) & 10.16 (71.448)\\
\bottomrule
\end{tabular}

\end{table}

Measurements use an A100-SXM4-40GB, PyTorch~2.9.1/CUDA~12.9, FP32 without
TF32, two warmups and ten timed repetitions. They include graph encoding,
all path predictions, a synthetic MSE loss and backward through both
stages. Initialization, transfers, gradient clearing and optimizer steps
are outside the timed region. We report the maximum peak allocation over
repetitions, including resident weights and inputs. At 1,024 vertices,
aggressive execution reduces peak memory by a factor of 2.45 relative to
the light configuration and takes \(1.86\times\) as long.
The experiment measures execution cost on controlled synthetic graph shapes.

For inference, we separately measure encoding time $E$ and cached-query
time $D$ for one trained \STEE{} checkpoint, using aggressive tiling,
length-eight paths and batch size~1. Activation checkpointing is disabled.
There is one synthetic graph per size. Encoding uses ten fresh cache
constructions after two warmups. Decoding uses 320 queries after 64
separate warmup queries, with synchronized GPU timing. Cached-query time
includes selection, projections and the prediction head. Initialization
and transfers are excluded.

\begin{table}[!htbp]
\color{black}
\centering\small
\caption{Inference cost and estimated amortization for the full-size
Atlas \STEE{}. With $E$ and $D$ expressed in the same units,
$\widehat Q_{10\%}=\lceil9E/D\rceil$ estimates the number of queries at
which initial encoding contributes at most 10\% of $E+QD$.}
\label{tab:atlas-cache-amortization}
\begin{tabular}{@{}rrrr@{}}
\toprule
Vertices & Encoding $E$ (s) & Cached query $D$ (ms) & $\widehat Q_{10\%}$\\
\midrule
256 & 0.322 & 17.508 & 166\\
512 & 2.175 & 17.706 & 1,106\\
1,024 & 17.354 & 17.458 & 8,946\\
\bottomrule
\end{tabular}

\end{table}

Encoding grows steeply with graph size, whereas decoding selected
length-eight paths remains about 17.5--17.7~ms. The additive estimate
$T(Q)\approx E+QD$ was checked against 30 directly timed settings, with
maximum absolute relative error of 5.52\%. The 10\% thresholds
are estimates based on the additive model. Reuse can save computation
before encoding falls below 10\% of total time. The training table reports
peak allocations across the forward and backward passes. The inference
table describes the time spent building and reading a fixed cache.

\subsection{Input and evidence availability}
\label{app:atlas-software}
The Atlas benchmark is reconstructed from 256 public RIPE Atlas
measurement archives. Checksums identify the source archives and the
reconstructed data, including path order, masks, priors, labels, and temporal
splits.

Numerical records cover validation and temporal test scores, frozen-reader
scores and conditional bootstrap intervals, and paired inference timings.
Verifying predictions from saved checkpoints requires the original weights
and the corresponding validation predictions for the fixed 4,096-query
August cohort.

\FloatBarrier

\section{EVT Operators and Normalization}
\label{app:operators}

This appendix specifies the operations used by the proofs: normalization,
MLP updates at each vertex or pair, tensor reshaping, and both attention operators.
The collection \(\theta_{\mathrm{EVT}}\) in \cref{eq:evt-core} comprises
the trainable attention, MLP, and normalization parameters of all layers
of the EVT core. Affine input/output interfaces and task-specific heads
have separate parameters.
We suppress layer, branch, head, and parameter subscripts: every displayed
occurrence of \(\Norm,\Att,\TriAtt,\MLP_{\mathcal{X}}\), and \(\MLP_{\mathcal{Y}}\) owns its
trainable parameters unless sharing is explicitly stated.  Parameters are
shared over the relevant vertex or pair indices, not across the two EVT
stages.  The maps \(\Flat\) and \(\Unflat\) are parameter-free.

\subsection{Overview of the EVT updates}
\label{app:evt-overview}
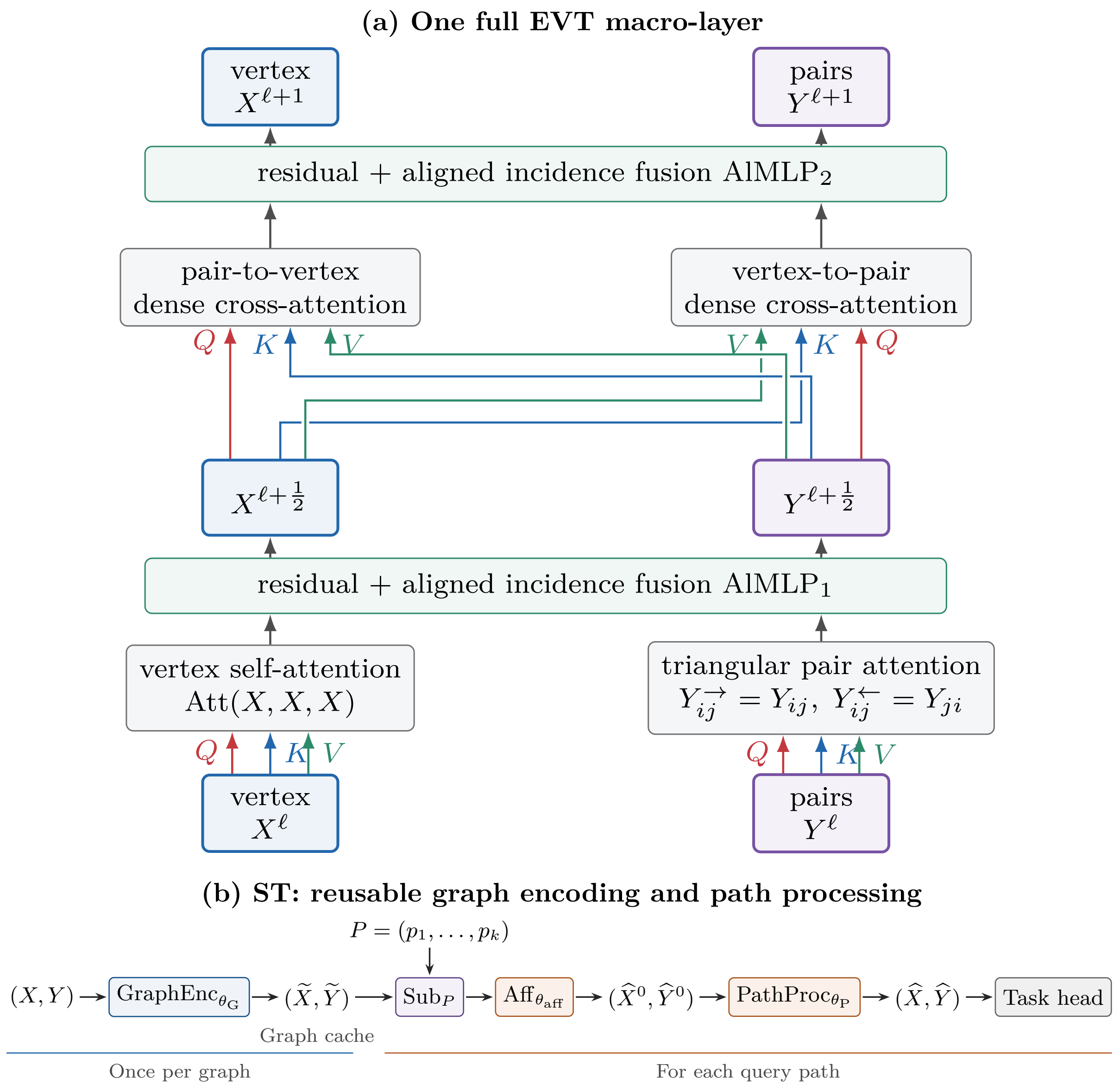
\begin{figure}[tbp]
\centering
\textbf{(a) One full EVT macro-layer}\par\smallskip
\resizebox{.78\linewidth}{!}{%
\begin{tikzpicture}[
 x=1cm,y=1cm,font=\scriptsize,
 xstate/.style={draw=graphblue,fill=graphblue!8,rounded corners=2pt,
   thick,minimum width=1.35cm,minimum height=.75cm,align=center},
 ystate/.style={draw=pairpurple,fill=pairpurple!8,rounded corners=2pt,
   thick,minimum width=1.35cm,minimum height=.75cm,align=center},
 op/.style={draw=black!55,fill=softgray,rounded corners=2pt,
   minimum width=2.55cm,minimum height=.68cm,align=center},
 fuse/.style={draw=pairgreen,fill=pairgreen!7,rounded corners=2pt,
   minimum width=8.0cm,minimum height=.55cm,align=center},
 flow/.style={-{Latex[length=1.7mm]},semithick,draw=black!70},
 qarr/.style={-{Latex[length=1.7mm]},semithick,draw=queryred},
 karr/.style={-{Latex[length=1.7mm]},semithick,draw=graphblue},
 varr/.style={-{Latex[length=1.7mm]},semithick,draw=pairgreen},
 crossing/.style={preaction={draw=white,line width=2.2pt,-}}
]
 \node[xstate] (x0) at (2.25,0) {vertex\\\(X^\ell\)};
 \node[ystate] (y0) at (7.75,0) {pairs\\\(Y^\ell\)};
 \node[op] (self) at (2.25,1.25)
   {vertex self-attention\\\(\Att(X,X,X)\)};
 \node[op] (tri) at (7.75,1.25)
   {triangular pair attention\\\(Y^\rightarrow_{ij}=Y_{ij},\;Y^\leftarrow_{ij}=Y_{ji}\)};
 \draw[qarr] ([xshift=-.38cm]x0.north)
   --node[midway,left,text=queryred]{\(Q\)} ([xshift=-.38cm]self.south);
 \draw[karr] (x0.north)
   --node[midway,right,text=graphblue]{\(K\)} (self.south);
 \draw[varr] ([xshift=.38cm]x0.north)
   --node[midway,right,text=pairgreen]{\(V\)} ([xshift=.38cm]self.south);
 \draw[qarr] ([xshift=-.38cm]y0.north)
   --node[midway,left,text=queryred]{\(Q\)} ([xshift=-.38cm]tri.south);
 \draw[karr] (y0.north)
   --node[midway,right,text=graphblue]{\(K\)} (tri.south);
 \draw[varr] ([xshift=.38cm]y0.north)
   --node[midway,right,text=pairgreen]{\(V\)} ([xshift=.38cm]tri.south);

 \node[fuse] (fuse1) at (5,2.27)
   {residual + aligned incidence fusion \(\AlMLP_1\)};
 \draw[flow] (self.north)--(self.north |- fuse1.south);
 \draw[flow] (tri.north)--(tri.north |- fuse1.south);
 \node[xstate] (xh) at (2.25,3.15) {\(X^{\ell+\frac12}\)};
 \node[ystate] (yh) at (7.75,3.15) {\(Y^{\ell+\frac12}\)};
 \draw[flow] (xh.south |- fuse1.north)--(xh.south);
 \draw[flow] (yh.south |- fuse1.north)--(yh.south);

 \node[op] (crossx) at (2.25,5.25)
   {pair-to-vertex\\dense cross-attention};
 \node[op] (crossy) at (7.75,5.25)
   {vertex-to-pair\\dense cross-attention};
 \draw[qarr] ([xshift=-.40cm]xh.north)
   --node[pos=.88,left,text=queryred]{\(Q\)} ([xshift=-.40cm]crossx.south);
 \draw[qarr] ([xshift=.40cm]yh.north)
   --node[pos=.88,right,text=queryred]{\(Q\)} ([xshift=.40cm]crossy.south);

 \coordinate (xk) at (2.35,3.90);
 \coordinate (xv) at (2.60,4.12);
 \coordinate (yk) at (7.65,4.36);
 \coordinate (yv) at (7.40,4.58);
 \coordinate (crossxk) at ([xshift=.20cm]crossx.south);
 \coordinate (crossxv) at ([xshift=.60cm]crossx.south);
 \coordinate (crossyk) at ([xshift=-.20cm]crossy.south);
 \coordinate (crossyv) at ([xshift=-.60cm]crossy.south);
 \draw[karr] ([xshift=.10cm]xh.north)--(xk)
   --(xk -| crossyk)--(crossyk);
 \draw[varr,crossing] ([xshift=.35cm]xh.north)--(xv)
   --(xv -| crossyv)--(crossyv);
 \draw[karr,crossing] ([xshift=-.10cm]yh.north)--(yk)
   --(yk -| crossxk)--(crossxk);
 \draw[varr,crossing] ([xshift=-.35cm]yh.north)--(yv)
   --(yv -| crossxv)--(crossxv);
 \node[anchor=east,text=graphblue,inner sep=1pt]
   at ([xshift=-2pt,yshift=-5pt]crossxk) {\(K\)};
 \node[anchor=west,text=pairgreen,inner sep=1pt]
   at ([xshift=2pt,yshift=-5pt]crossxv) {\(V\)};
 \node[anchor=west,text=graphblue,inner sep=1pt]
   at ([xshift=2pt,yshift=-5pt]crossyk) {\(K\)};
 \node[anchor=east,text=pairgreen,inner sep=1pt]
   at ([xshift=-2pt,yshift=-5pt]crossyv) {\(V\)};

 \node[fuse] (fuse2) at (5,6.38)
   {residual + aligned incidence fusion \(\AlMLP_2\)};
 \draw[flow] (crossx.north)--(crossx.north |- fuse2.south);
 \draw[flow] (crossy.north)--(crossy.north |- fuse2.south);
 \node[xstate] (x1) at (2.25,7.25) {vertex\\\(X^{\ell+1}\)};
 \node[ystate] (y1) at (7.75,7.25) {pairs\\\(Y^{\ell+1}\)};
 \draw[flow] (x1.south |- fuse2.north)--(x1.south);
 \draw[flow] (y1.south |- fuse2.north)--(y1.south);
\end{tikzpicture}}
\par\medskip
\textbf{(b) ST: reusable graph encoding and path processing}\par\smallskip
\resizebox{.98\linewidth}{!}{%
\begin{tikzpicture}[
  font=\footnotesize,node distance=4.4mm,
  state/.style={inner sep=1pt},
  operator/.style={rounded corners=2pt,inner xsep=3pt,
    inner ysep=3pt,minimum height=5.5mm,line width=.5pt},
  graph op/.style={operator,draw=graphblue!80!black,fill=graphblue!7},
  selector/.style={operator,draw=pairpurple!80!black,fill=pairpurple!8},
  path op/.style={operator,draw=pathorange!75!black,fill=pathorange!10},
  head/.style={operator,draw=black!60,fill=black!6},
  label text/.style={font=\scriptsize,text=black!72,inner sep=0pt},
  flow/.style={-{Stealth[length=1.6mm,width=1.2mm]},line width=.65pt,
    draw=black!85,shorten <=.8pt,shorten >=.8pt}
]
  \node[state] (graph) {\((X,Y)\)};
  \node[graph op,right=of graph] (encoder) {\(\GraphEnc_{\theta_{\mathrm G}}\)};
  \node[state,right=of encoder] (cache) {\((\widetilde X,\widetilde Y)\)};
  \node[selector,right=6mm of cache] (select) {\(\Sub_P\)};
  \node[path op,right=of select] (bridge) {\(\Aff_{\theta_{\mathrm{aff}}}\)};
  \node[state,right=of bridge] (selected) {\((\widehat X^0,\widehat Y^0)\)};
  \node[path op,right=of selected] (processor) {\(\PathProc_{\theta_{\mathrm P}}\)};
  \node[state,right=of processor] (states) {\((\widehat X,\widehat Y)\)};
  \node[head,right=of states] (head) {Task head};
  \node[state,above=4.5mm of select] (query) {\(P=(p_1,\ldots,p_k)\)};

  \foreach \from/\to in {graph/encoder,encoder/cache,cache/select,
                        select/bridge,bridge/selected,selected/processor,
                        processor/states,states/head}
    \draw[flow] (\from.east)--(\to.west);
  \draw[flow] (query.south)--(select.north);

  \node[label text,below=2mm of cache] {Graph cache};
  \coordinate (graph rule) at ($(graph.west)+(0,-8mm)$);
  \coordinate (query rule) at ($(select.west)+(-1.5mm,-8mm)$);
  \draw[graphblue,line width=.5pt]
    (graph rule)--(cache.east |- graph rule)
    node[midway,below=1.5mm,label text] {Once per graph};
  \draw[pathorange!80!black,line width=.5pt]
    (query rule)--(head.east |- query rule)
    node[midway,below=1.5mm,label text] {For each query path};
\end{tikzpicture}}
\caption{Detailed EVT and StrikeTwice architectures.
\textbf{(a)} Vertex (blue) and pair (purple) streams update in parallel,
fuse, exchange information by cross-attention, and fuse again.
Red/blue/green arrows denote queries/keys/values; each sublayer is residual.
\textbf{(b)} The graph encoder computes the cache once. Each query enters
at \(\Sub_P\), which selects the ordered path states and adds position
codes and repeated-vertex indicators. The affine bridge sets the
path-stage widths of \(\widehat X^0,\widehat Y^0\) before the path
processor. EVT can instantiate both networks independently.}
\label{fig:cascade}
\end{figure}

Vertex and pair states exchange information through attention and
shared MLP updates, as shown in the detailed appendix diagram
(\cref{fig:cascade}a). Each macro-layer consists of two steps.
The first step updates vertices by self-attention and pairs by triangular
attention in parallel, using the oriented views defined in \cref{sec:architecture}.
Write \(n\) for the number of indices processed by EVT; replace \(n\) by
\(k\) when EVT processes the selected occurrence graph.
For one head, \(W_Q,W_K\) are learned query/key projections of width \(d_a\),
and \(W_L,W_R\) are learned value projections of a common width.
Tokens are column vectors in these formulas, \(\odot\) is the element-wise
product, and \(\operatorname{softmax}_r(a)=e^{a_r}/\sum_{u=1}^n e^{a_u}\).
For orientations \((\alpha,\beta)\in\{\rightarrow,\leftarrow\}^2\),
write \(\bar y_{ij}=\Norm(y_{ij})\), with branch-specific normalization
parameters and head indices on the projections suppressed. Head \(h\) returns
\begin{equation}
 \TriAtt_h^{\alpha,\beta}(Y)_{ij}
 =\sum_{r=1}^n \operatorname{softmax}_r\!\left(
   \frac{\langle W_Q\bar y^\alpha_{ir},W_K\bar y^\beta_{rj}\rangle}{\sqrt{d_a}}
   \right)
   \bigl(W_L\bar y^\alpha_{ir}\odot W_R\bar y^\beta_{rj}\bigr),
 \label{eq:oriented-triatt-appendix}
\end{equation}
Each head has independent projections and its own softmax over the intermediate vertex \(r\). Its value combines both pair states through an element-wise product.
The head output has its value width; \(\TriAtt\) denotes their concatenation
followed by a projection to the receiving pair-state width
(\cref{eq:tri-detail}).
If both value projections extract the same scalar normalized channel
\(A_{ij}=w^\top\bar y_{ij}\), zero logits yield \(n^{-1}A^2\), \(n^{-1}AA^\top\), \(n^{-1}A^\top A\), or
\(n^{-1}(A^\top)^2\): a directed chain, a common target, a common source, or a reverse chain, respectively.
The first follows Edge Transformers \citep{bergen2021}; the middle
two match the bilinear cores of AlphaFold's triangle multiplication
\citep{jumper2021alphafold}.
EVT uses these orientations in content-dependent attention heads,
retaining cubic dense complexity at fixed total width
(\cref{eq:four-pair-contractions}).

The aligned residual increments in \cref{eq:almlp}, with their
normalization wrappers made explicit, are
\begin{equation}
 \AlMLP(X,Y):=\left(
   (\MLP_{\mathcal{X}}(\Norm(x_i\concat y_{ii})))_i,
   (\MLP_{\mathcal{Y}}(\Norm(y_{ij}\concat x_i\concat x_j\concat\one_{\{i=j\}})))_{ij}
 \right).
 \label{eq:almlp-appendix}
\end{equation}
Each MLP is a biased Linear--GELU--Linear map. One MLP is shared over
vertices and the other over ordered pairs; they normalize their complete concatenated
inputs, including the equality indicator. The update of \(x_i\) reads
\((x_i,y_{ii})\); that of \(y_{ij}\) reads \((y_{ij},x_i,x_j,\one_{\{i=j\}})\).
Direct reverse-pair input is omitted from the aligned MLP because oriented
triangular heads already provide access to transposed pair states.
The second step applies dense cross-attention in both directions.
Each vertex attends to all pair states, flattened into a list,
and each pair attends to all vertex states. Both cross-attentions
read the states produced by the first step and are followed by
another \(\AlMLP\) update with separate parameters.
Every internal attention and MLP update is residual.
The complete equations are given in \cref{app:macro-layer-order}.

EVT uses pre-normalization with GELU MLPs.
The theoretical results cover both LayerNorm and RMSNorm;
the reported EVT experiments use LayerNorm (\cref{sec:experiments}).
The EVT core preserves its vertex and pair widths. The theory adds
separate outer affine input and output maps, allowing representation
widths to differ from raw feature and prediction dimensions. At the
path-stage input, position and repeated-vertex indicators are injected
through affine stems and mixed into the states. The family in
\cref{def:striketwice-family} varies the finite capacity of this
architecture; each experiment fixes a capacity budget.

\subsection{Domains, symmetry, and prediction interfaces}
\label{app:domains}

\paragraph{Learning on graphs with ordered paths.}
Vertex names may change without changing the prediction. Path positions
specify the order of the query and remain fixed when vertices are renamed.

For the theory, fix maximum sizes \(N,K\geq1\), raw input widths
\(d_{\mathcal{X}}^{\mathrm{in}},d_{\mathcal{Y}}^{\mathrm{in}}\geq1\), and output widths
\(d_{\mathcal{X}}^{\mathrm{out}},d_{\mathcal{Y}}^{\mathrm{out}}\geq1\). For rooted predictions, \(q\geq1\) denotes a single
output-vector width. For
\(1\leq n\leq N\) and \(1\leq k\leq K\), let
\begin{equation}
 \mathcal D_{n,k}
 :=\R^{n\times d_{\mathcal{X}}^{\mathrm{in}}}\times\R^{n\times n\times d_{\mathcal{Y}}^{\mathrm{in}}}\times[n]^k,
 \qquad
 \mathcal D_{\leq N,\leq K}:=\bigsqcup_{n=1}^{N}\bigsqcup_{k=1}^{K}
 \mathcal D_{n,k}.
 \label{eq:domain}
\end{equation}
Each component has the Euclidean topology on features and the discrete
topology on \([n]^k\); \(\mathcal D_{\leq N,\leq K}\) has the finite disjoint-union
topology.
An input \(u=(X,Y,P)\in\mathcal D_{n,k}\) consists of vertex features
\(X=(x_i)\), dense ordered-pair features \(Y=(y_{ij})\), and an ordered
index sequence \(P=(p_1,\ldots,p_k)\), which we call a path.  Repetitions
are allowed, and the architecture is defined for every \(P\in[n]^k\): it
does not require consecutive indices to be adjacent.  In typical graph
applications, \(P\) is nevertheless restricted to a valid walk.  When
\(y_{ij}\) contains an edge indicator \(A_{ij}\), this is the subdomain
defined by \(A_{p_sp_{s+1}}=1\) for every consecutive pair; all results below
restrict automatically to that subdomain.

Here and throughout, ``edge'' in Edge--Vertex Transformer refers to the
dense, relabeling-equivariant ordered-pair index set \([n]^2\), including
diagonal and nonadjacent pairs, rather than only to the edges of a sparse
graph.  If an input graph has adjacency matrix \(A\), then \(A_{ij}\) is a
coordinate of \(y_{ij}\), together with any pair or edge attributes; a
declared null value represents edge-only attributes when \(A_{ij}=0\).
Outputs belong to
\begin{equation}
  \mathcal O_K:=\bigsqcup_{k=1}^{K}
  \left(\R^{k\times d_{\mathcal{X}}^{\mathrm{out}}}\times\R^{k\times k\times d_{\mathcal{Y}}^{\mathrm{out}}}\right),
  \label{eq:output-domain}
\end{equation}
with the finite disjoint-union topology.  Every architecture below is one
map on \(\mathcal D_{\leq N,\leq K}\): the same parameters act on every size component.

The theory surrounds each architecture core by
shared affine input and output maps. The input map acts on the graph
features and leaves \(P\) unchanged; the output map reads the two terminal
representation streams separately, with coordinate projections among its
allowed choices. For StrikeTwice, writing
\( (\widehat X,\widehat Y)=\STarch_{\theta_{\mathrm{ST}}}
(\Aff_{\theta_{\mathrm{in}}}(X,Y),P)\), the prediction is
\begin{equation}
f_\theta(X,Y,P)
=\Aff_{\theta_{\mathrm{out}}}(\widehat X,\widehat Y)
=\left((\pi_{\mathcal{X}}(\widehat x_s))_s,
 (\pi_{\mathcal{Y}}(\widehat y_{st}))_{s,t}\right).
\label{eq:path-output}
\end{equation}
Here \(\pi_{\mathcal{X}}:\R^{q_{\mathcal{X}}}\to
\R^{d_{\mathcal{X}}^{\mathrm{out}}}\) and
\(\pi_{\mathcal{Y}}:\R^{q_{\mathcal{Y}}}\to
\R^{d_{\mathcal{Y}}^{\mathrm{out}}}\) are affine, and
\(\theta=(\theta_{\mathrm{in}},\theta_{\mathrm{ST}},
\theta_{\mathrm{out}})\); the core parameter tuple includes graph,
selection-affine, and path parameters. These outer maps are not part of
the dimension-preserving EVT core. They define the prescribed vertex--pair
predictions. They can serve directly as task-specific layers; additional
pooling or nonlinear task-specific layers used in experiments are not part
of this family. The proofs first construct arbitrary
vertex-output coordinates and then extend the construction to joint outputs
in \cref{app:joint-output-extension}.
For a domain subset \(S\), compatible predictions are
\begin{equation}
 \mathcal C_\Psi(S):=
 \left\{f\in\mathcal C(S;\mathcal O_K):
 \begin{array}{l}
 f\text{ is length preserving, and}\\[-1pt]
 \ker(\Psi_{N,K}|_S)\subseteq\ker(f)
 \end{array}\right\}
 \label{eq:compatible-targets}
\end{equation}

Let \(\mathfrak C_N\) be the common space of depth-\(R_N\) formal colors.
The profile of \cref{def:profile} takes values in
\(\bigsqcup_{n=1}^{N}\bigsqcup_{k=1}^{K}
(\{n\}\times\{k\}\times\mathfrak C_N^{k\times k})\).
For length-preserving \(f\) and \(h\), the uniform norm is
\begin{equation}
 \|f-h\|_{\infty,S}:=\sup_{u\in S}
 \sqrt{\|f_{\mathcal{X}}(u)-h_{\mathcal{X}}(u)\|_{\mathrm F}^2+
 \|f_{\mathcal{Y}}(u)-h_{\mathcal{Y}}(u)\|_{\mathrm F}^2}.
 \label{eq:uniform-norm}
\end{equation}

For a graph relabeling \(\sigma\in\mathfrak S_n\), define
\begin{equation}
  (\sigma\mathbin\cdot X)_{\sigma(i)}=x_i,\qquad
  (\sigma\mathbin\cdot Y)_{\sigma(i)\sigma(j)}=y_{ij},\qquad
  (\sigma P)_s=\sigma(p_s).
  \label{eq:graph-action}
\end{equation}
The target symmetry is
\begin{equation}
 f(\sigma\mathbin\cdot X,\sigma\mathbin\cdot Y,\sigma P)=f(X,Y,P).
 \label{eq:invariance}
\end{equation}
Permuting path positions is not a symmetry unless separately imposed.

\paragraph{Flattening convention.}
For \(Y\in\R^{n\times n\times d}\) and \(Y^{\mathrm{flat}}\in\R^{n^2\times d}\), fix
\(\rho_n(i,j)=(i-1)n+j\) and define
\begin{equation}
 [\Flat(Y)]_{\rho_n(i,j)}=y_{ij},
 \qquad
 [\Unflat(Y^{\mathrm{flat}})]_{ij}=Y^{\mathrm{flat}}_{\rho_n(i,j)}.
 \label{eq:flat-unflat}
\end{equation}

\paragraph{Position codes.}
For the selection operator \(\Sub_P\) in StrikeTwice, it is enough to
choose any injective absolute-position map
\(\phi:[K]\to\R^{d_\phi}\) and to retain endpoint order through
\(\phi(s)\concat\phi(t)\). For bounded-size theory we use the canonical
one-hot choice:
\begin{equation}
 \phi(s)=(\one_{\{r=s\}})_{r=1}^{K}\in\R^K.
 \label{eq:positions}
\end{equation}
Because \([K]\) is finite, any other fixed injective code can be decoded
by a pointwise GELU MLP (\cref{lem:gelu-finite-interpolation}). With
sufficient state width and an attention-disabled preparation block, this
gives the same variable-capacity approximation class. A learned lookup
must have distinct entries to retain this guarantee. A symmetric pair code
would generally lose orientation. PathMarked sums the codes before graph
encoding, so it requires distinct sums for distinct occurrence sets;
injectivity of individual codes alone is insufficient. Its theoretical
family therefore uses the one-hot choice in \cref{eq:positions}.

\paragraph{Encoder and task-specific layers.}
\label{app:practical-interface}
The encoder uses complete ordered selection and a learned affine
graph-to-path bridge, distinct from the outer raw-feature embedding.
Task-specific prediction interfaces are detailed in \cref{app:model-input-access}.
With the canonical one-hot absolute code and the optional
signed relative feature
\(\rho_K(t-s)=(t-s)/\max\{1,K-1\}\), it initializes the path stage as
\begin{align}
 \widehat x^0_s
   &=A_{\mathcal{X}}^{\mathrm{bridge}}
     [\widetilde x_{p_s}\concat\phi(s)]+b_{\mathcal{X}}^{\mathrm{bridge}},
 \label{eq:practical-pullback-x}\\
 \widehat y^0_{st}
   &=A_{\mathcal{Y}}^{\mathrm{bridge}}
     [\widetilde y_{p_sp_t}\concat\phi(s)\concat\phi(t)\concat\rho_K(t-s)
       \concat\one_{\{p_s=p_t\}}]+b_{\mathcal{Y}}^{\mathrm{bridge}}.
 \label{eq:practical-pullback-y}
\end{align}
Here \(A_{\mathcal X}^{\mathrm{bridge}}\) has shape
\(q_{\mathcal X}\times(d_{\mathcal X}+K)\),
\(A_{\mathcal Y}^{\mathrm{bridge}}\) has shape
\(q_{\mathcal Y}\times(d_{\mathcal Y}+2K+2)\), and the biases have
widths \(q_{\mathcal X},q_{\mathcal Y}\). Without the optional relative
feature, the pair input width is \(d_{\mathcal Y}+2K+1\), exactly as in
\cref{eq:tokenwise-affine}. Since \(\rho_K(t-s)\) is a fixed linear
function of the two one-hot codes, its affine contribution can be absorbed
into the bridge without changing the theoretical family.
For padded batches, a selected pair is valid exactly when both occurrences
are valid. The appended bit $\one_{\{p_s=p_t\}}$ records vertex equality;
the path-stage MLP separately receives $\one_{\{s=t\}}$, which records
occurrence equality. For example, in $P=(u,v,u)$ the first bit is one at
$(s,t)=(1,3)$ and the second is zero. Invalid occurrences and pairs are
zeroed after projections and updates, and masked means divide by the
number of valid occurrences.
The positional codes and vertex-equality bit in
\cref{eq:practical-pullback-y} are injected once and then treated as
ordinary learned coordinates. These operations preserve graph relabeling
invariance and path order sensitivity. The family includes information-preserving
choices of the projections as capacity increases, without requiring every
trained projection to be injective (\cref{def:striketwice-family}).

The graph and path encoders produce the states
\begin{equation}
 \operatorname{Enc}_\theta(X,Y,P)=(\widetilde X,\widetilde Y,Z,R),
 \qquad Z:=\widehat X,\quad R:=\widehat Y,
 \label{eq:practical-encoder-output}
\end{equation}
which task-specific layers map to predictions. For scalar regression with
\(\mathrm{ST}[\EVT,\EVT]\),
let \(k_{\max}\)
be the padded path length and \(m_s\in\{0,1\}\) the validity mask.
The experiments use the exact masked mean
\begin{equation}
 u=\frac{\sum_{s=1}^{k_{\max}}m_s Z_s}{\sum_{s=1}^{k_{\max}}m_s},
 \qquad \widehat y=\MLP_{\mathrm{task}}(u),
 \label{eq:scalar-task-head}
\end{equation}
Structured tasks may apply shared rowwise task-specific layers to \(Z_s\) and use the adjacent pair state
\(J_s=R_{s,s+1}\) for junction predictions. These task-specific layers read the
encoder outputs. The joint affine interface in \cref{eq:path-output} fixes the
model output for the expressivity statements. Scalar tasks use
\cref{eq:scalar-task-head}.
For any continuous profile-compatible scalar target, the construction can
write the same scalar approximation at every occurrence, so mean pooling
preserves it and an affine task-specific layer selects it. Pair or
ordinal task-specific layers are described with their experiments.

\subsection{Tokenwise normalization}
\label{app:token-normalization}

EVT applies tokenwise feature normalization before every attention
sublayer and to the complete concatenated input of each aligned MLP.
For \(z\in\R^d\), let \(\gamma,\beta\in\R^d\) be learned scale and shift
vectors, \(\varepsilon>0\) a fixed stabilizer, and \(\one\in\R^d\)
the all-ones vector.
We consider two normalization choices. RMSNorm is defined by
\begin{equation}
 \RMSNorm_{\gamma,\varepsilon}(z)
 :=\gamma\odot\frac{z}{\sqrt{d^{-1}\|z\|_2^2+\varepsilon}}.
 \label{eq:rmsnorm-detail}
\end{equation}
LayerNorm is defined by
\begin{equation}
 \LN_{\gamma,\beta,\varepsilon}(z)
 =\gamma\odot
   \frac{z-\bar z\one}
        {\sqrt{d^{-1}\|z-\bar z\one\|_2^2+\varepsilon}}
   +\beta,
 \qquad
 \bar z=d^{-1}\one^\top z,
 \label{eq:layernorm-detail}
\end{equation}
We write \(\Norm\) for either normalization, with one choice fixed throughout
each model. The theoretical results hold separately for both choices;
the reported EVT experiments use LayerNorm.
The selector injects its position and
graph-equality features once into ordinary learned states. Attention
normalizes each entire token, and each aligned MLP normalizes its entire
concatenated local input. No tag is exempt from normalization.
The proof reserves constant and tag
coordinates by zeroing the corresponding residual output rows; this is a
choice of parameters within the architecture, not an additional operation.

\subsection{Pointwise MLPs and aligned fusion}

Each pointwise MLP has two learned affine maps and the elementwise GELU
activation \(g(t)=t\Phi(t)\), where \(\Phi\) is the standard Gaussian
distribution function \citep{hendrycks2016gelu}:
\begin{equation}
 \MLP_\theta(z)=W_2g(W_1z+b_1)+b_2.
 \label{eq:mlp-detail}
\end{equation}
For the local inputs displayed below, write
\begin{equation}
 \MLP(z_1,\ldots,z_r;\tau)
 :=\MLP_\theta\!\left(\Norm([z_1\concat\cdots\concat z_r\concat\tau])\right),
 \label{eq:mlp-arguments}
\end{equation}
where the semicolon denotes a supplied local covariate, such as the
equality bit, which is included in that same normalization. Absent
arguments are omitted.
Each MLP in a residual block returns the width of its receiver stream.
For vertex and pair widths \(d_{\mathcal X},d_{\mathcal Y}\), the aligned
vertex MLP maps \(\R^{d_{\mathcal X}+d_{\mathcal Y}}\) to
\(\R^{d_{\mathcal X}}\), while the aligned pair MLP maps
\(\R^{d_{\mathcal Y}+2d_{\mathcal X}+1}\) to \(\R^{d_{\mathcal Y}}\).
The additional scalar is the equality indicator; the hidden widths
of the two MLPs may differ from their input and output widths.
In \(\AlMLP\), the vertex and pair MLPs are independent;
the two \(\AlMLP\) occurrences in one EVT
macro-layer are also independent unless parameters are explicitly tied.

\subsection{Flattening pair states}

The mutually inverse reshapes in \cref{eq:flat-unflat} use lexicographic order
only to enumerate tokens, without adding positional information. Attention
is equivariant to query-row permutations and invariant to simultaneous
key/value-row permutations.

\subsection{Exact EVT macro-layer order}
\label{app:macro-layer-order}

With \(\ell\) the macro-layer index, the two steps described in
\cref{app:evt-overview} are, explicitly,
\begin{align}
 \bar X^\ell&=X^\ell+\Att(X^\ell,X^\ell,X^\ell),&
 \bar Y^\ell&=Y^\ell+\TriAtt(Y^\ell),
 \label{eq:within-att}\\
 (X^{\ell+\frac12},Y^{\ell+\frac12})
   &=(\bar X^\ell,\bar Y^\ell)+\AlMLP(\bar X^\ell,\bar Y^\ell),
 \label{eq:within-mlp}
\end{align}
followed by
\begin{align}
 \bar X^{\ell+\frac12}
   &=X^{\ell+\frac12}+
     \Att(X^{\ell+\frac12},\Flat Y^{\ell+\frac12},
                         \Flat Y^{\ell+\frac12}),
 \label{eq:y-to-x}\\
 \bar Y^{\ell+\frac12}
   &=Y^{\ell+\frac12}+
     \Unflat\Att(\Flat Y^{\ell+\frac12},X^{\ell+\frac12},
                                      X^{\ell+\frac12}),
 \label{eq:x-to-y}\\
 (X^{\ell+1},Y^{\ell+1})
   &=(\bar X^{\ell+\frac12},\bar Y^{\ell+\frac12})
     +\AlMLP(\bar X^{\ell+\frac12},\bar Y^{\ell+\frac12}).
 \label{eq:cross-mlp}
\end{align}
The aligned updates use \cref{eq:almlp}, with independent parameters at each step.

\subsection{Ordinary attention}

Each displayed attention occurrence owns independent projections.  For one
head, let \(q_a,k_b,v_b\) be receiver, key, and value tokens. Set
\(\widetilde q_a=\Norm(q_a)\),
\(\widetilde k_b=\Norm(k_b)\), and
\(\widetilde v_b=\Norm(v_b)\), with the learned normalization parameters
of the corresponding branch. Here \(Q=(q_a),K=(k_b),V=(v_b)\)
denote token arrays, not the query count, path bound, or graph vertex set.
The matrices \(W_Q,W_K\) project to query/key width \(d_a\), and \(W_V\)
projects values to width \(d_v\). One head \(h\) returns
\begin{equation}
 \Att_h(Q,K,V)_a
 =\sum_b
 \frac{\exp(\langle W_Q\widetilde q_a,W_K\widetilde k_b\rangle/\sqrt{d_a})}
      {\sum_c\exp(\langle W_Q\widetilde q_a,W_K\widetilde k_c\rangle/\sqrt{d_a})}
 W_V\widetilde v_b,
 \label{eq:attention-detail}
\end{equation}
For multiple heads, \(\Att(Q,K,V)_a=W_O[\Att_1(Q,K,V)_a\concat\cdots\concat
\Att_H(Q,K,V)_a]\). If the head value widths are \(d_v^h\) and the
receiver width is \(d_{\mathrm{rec}}\), then
\(W_O\in\R^{d_{\mathrm{rec}}\times\sum_h d_v^h}\).
The operator \(\Att\) returns only this increment; residual addition is
explicit in \cref{app:macro-layer-order}. Normalization follows
\cref{app:token-normalization}.

\subsection{Triangular attention}

For \(n\times n\) graph pair states \(Y=(y_{ij})\), normalize each pair token once as
\(y^{\mathrm{norm}}_{ij}=\Norm(y_{ij})\). Following the oriented-view
convention in \cref{sec:architecture}, set
\begin{equation}
 y^{\mathrm{norm},\rightarrow}_{ij}:=y^{\mathrm{norm}}_{ij},
 \qquad
 y^{\mathrm{norm},\leftarrow}_{ij}:=y^{\mathrm{norm}}_{ji}.
 \label{eq:normalized-oriented-views}
\end{equation}
Each of the \(H\) heads \(h\in[H]\) independently fixes a direction pair
\((\alpha_h,\beta_h)\in
\{\rightarrow,\leftarrow\}^2\) and defines
\begin{align}
 q^h_{ir}&=W^h_Q y^{\mathrm{norm},\alpha_h}_{ir},&
 k^h_{rj}&=W^h_K y^{\mathrm{norm},\beta_h}_{rj},\nonumber\\
 \ell^h_{irj}&=\langle q^h_{ir},k^h_{rj}\rangle/\sqrt{d_a^h},&
 v^h_{irj}&=(W^h_L y^{\mathrm{norm},\alpha_h}_{ir})
              \odot(W^h_R y^{\mathrm{norm},\beta_h}_{rj}),
 \label{eq:tri-head}
\end{align}
where \(W_Q^h,W_K^h\in\R^{d_a^h\times d_{\mathcal Y}}\) and
\(W_L^h,W_R^h\in\R^{d_v^h\times d_{\mathcal Y}}\).
Then
\begin{equation}
 T^h_{ij}=\sum_{r=1}^{n}
 \frac{e^{\ell^h_{irj}}}{\sum_{u=1}^{n}e^{\ell^h_{iuj}}}
 v^h_{irj},
 \qquad
 \TriAtt(Y)_{ij}=W_O^\triangle[T^1_{ij}\concat\cdots\concat T^H_{ij}].
 \label{eq:tri-detail}
\end{equation}
The output matrix has shape
\(W_O^\triangle\in\R^{d_{\mathcal Y}\times\sum_h d_v^h}\),
so the increment has the width required by the residual addition.
We write \(\TriAtt^{\alpha,\beta}\) when all heads in a bank use the
same pair and \(\TriAtt\) for the general head-wise mixture.  Because every
head owns its denominator, the four direction classes are never forced to
compete inside one softmax; their outputs interact only through
\(W_O^\triangle\) and subsequent pointwise maps.  A layer may omit a class or
assign several heads to it.
For each intermediate index \(r\), the value combines the same two oriented
pair states used in the score. This joint value can encode the counts of
paired colors used in the \(2\)-FWL proof. With zero logits, a value depending
on only one of these states summarizes only that state's distribution;
it does not record which colors occur together. Values based on only one
state, with nonconstant scores depending on both states, require a separate
expressivity analysis.
Like \(\Att\), \(\TriAtt\) returns an increment and does not include the
explicit residual connection.

\section{Detailed Expressivity Analysis}
\label{app:proofs}

We state the EVT--ET comparison and define the StrikeTwice family.
Auxiliary residual computations provide exact color codes and compact
approximants. A simulation theorem realizes them to arbitrary accuracy with
RMSNorm or LayerNorm and GELU. The proofs then recover the selected profile and
establish the kernel and approximation statements for StrikeTwice itself.

\paragraph{Supporting graph-stage results.}
Extending the ET analysis of \citet{muller2024}, direction-mixed EVT and
direction-mixed ET with endpoint features share rooted distinguishing power
and compact approximation closure (\cref{prop:evt-et}). Reversed-pair access
strictly extends the forward-tokenized chain-only restriction on directed
inputs (\cref{prop:directional-strictness}). Complementary depth comparisons for the auxiliary
computations are given in \cref{prop:finite-depth}.

\subsection{Information kernels and capacity}
\label{app:theory-interpretation}

\paragraph{Kernels as information boundaries.}
For a map \(f\), let
\(\ker(f):=\{(u,v):f(u)=f(v)\}\).  For a family \(\mathcal F\) of maps on a
common domain, define its common kernel by
\begin{equation}
 \Ker(\mathcal F):=\bigcap_{f\in\mathcal F}\ker(f)
 =\{(u,v):f(u)=f(v)\ \text{for every }f\in\mathcal F\}.
 \label{eq:family-kernel}
\end{equation}
This equivalence relation consists exactly of the pairs that no member of the
family can distinguish.  An individual network may identify additional
inputs; the common kernel records what the entire family can never separate.
In the present length-preserving metric setting,
every uniform limit of members of \(\mathcal F\) must therefore take the same
value on inputs that no family member distinguishes.  Kernel equality with an explicit profile is the necessary
information-level part of a universality theorem. StrikeTwice and
PathMarked use the ordinary kernel; graph-shaped EVT outputs use the
equivariant kernel of rooted evaluations (\cref{eq:equivariant-kernel}).
In each case the corresponding kernel compatibility is also sufficient for
uniform approximation on compact sets
(\cref{thm:evt-expressivity,thm:main-expressivity,thm:pathmarked-expressivity}).
For outputs in \(\mathcal O_K\),
equality is equality in the disjoint union, so different output lengths are
distinct.

\paragraph{Why access to reversed pair states matters.}
If every head is restricted to the compositional variant
\(\TriAtt^{\rightarrow,\rightarrow}\), while pointwise pair maps continue to
receive only \(y_{ij}\), the corresponding refinement starts instead from
\begin{equation}
 c^{0,\mathrm{pure}}_{ij}
 =(y_{ij},x_i,x_j,\one_{\{i=j\}}).
 \label{eq:pure-2fwl-init}
\end{equation}
For directed inputs this loses information: at rooted graph-stage pair
readouts, the common kernel of the family allowing access to reversed pair states is
strictly smaller than that of its pure restriction.  In particular, on an explicit
four-vertex directed example, two rooted pairs have the same stable pure
color although the corresponding entries of \(YY^\top\) differ.  No member
of either pure rooted ET/EVT family can therefore approximate even the
coordinate map \((Y;i,j)\mapsto(YY^\top)_{ij}\) with uniform error below
\(1/2\) on that finite compact set. By contrast, an oriented auxiliary
computation forms \(n^{-1}YY^\top\) and recovers its scale from the
bounded-size cardinality anchor. The simulation theorem transfers its
separation and compact approximants to the normalized architecture.
On the binary witness, a constant-variance affine lift also gives an
exact normalized realization (\cref{prop:directional-strictness}).
The strict kernel statement and witness are proved in
\cref{app:directional-strictness}.
The comparison concerns rooted graph-stage states, before path selection,
with arbitrary head allocations.

\paragraph{Capacity and depth.}
Kernel equality guarantees that some finite family member separates each
pair of distinct profiles. For each compact \(S\), compatible target \(f\),
and \(\varepsilon>0\), \cref{thm:main-expressivity} supplies a finite
network whose resources may depend on \(S,f,\varepsilon,N,K\).
The theorem gives no quantitative bound on total capacity or guarantee of
learning. The explicit construction uses the sufficient refinement bound
\(R_N=2N^2\) before the additional setup, occurrence-decoding, gating,
broadcast, and readout blocks. For the synthetic graph size \(N=28\), this
bound alone is \(1{,}568\) refinement stages, whereas the synthetic models use
two graph and two path EVT macro-layers.

\paragraph{Consequences for EVT.}
The graph-stage proof also shows that direction-mixed EVT and the
endpoint-lifted, direction-mixed pair-only ET have the same rooted common
kernel---namely, the one induced by stable \(2\)-FWL with bidirectional
initialization---and the same uniform closure on every compact set.  The native
vertex stream therefore changes finite-depth and computational organization,
not the unlimited-capacity information boundary.  The precise rooted
statement and complementary shallow comparisons are in
\cref{prop:evt-et,prop:finite-depth}.

The second triangular operator is also a resource choice. A positioned path
processor using pair MLPs and global
pair-to-vertex attention can recover the complete selected profile without
path triangular attention (\cref{cor:path-decoder-kernel}). This preserves
the limiting function class but may require many auxiliary state coordinates.

\paragraph{Ordered selection and bounded sizes.}
Beyond the graph-level EVT--ET characterization, the theorem propagates the
expressivity analysis through an ordered \(P\times P\) selection and a second
relational network. Shared parameters across bounded sizes are already
covered by graph approximation frameworks
\citep{keriven2019universal,azizian2021}. Here, for fixed \(N,K\), one parameter
set acts on every component and, for each target, compact set, and accuracy,
one network approximates simultaneously across all represented
\(n\leq N\) and \(k\leq K\).
Common refinement depths and cardinality anchors handle bounded sizes
without padding. A shared vector encoding realizes the finitely many colors
occurring in each two-input comparison or finite input set
(\cref{app:profile-proof}). On compact sets, uniform approximation uses a
finite bank of separating models and a shared neural readout
(\cref{app:universality-proof}).

\begin{remark}[Path-conditioned global refinement]
\Cref{app:marked-model} compares reusable selection with global refinement
conditioned on occurrence marks, whose graph states depend on the query.
\end{remark}

\subsection{EVT--ET expressivity comparison}
\label{app:evt-et-comparison}

Define the rooted pair and vertex domains
\begin{equation}
 \mathcal R_N^{\mathcal{Y}}:=\bigsqcup_{n=1}^N
 \bigl(\R^{n\times d_{\mathcal{X}}^{\mathrm{in}}}\times\R^{n\times n\times d_{\mathcal{Y}}^{\mathrm{in}}}\times[n]^2\bigr),
 \qquad
 \mathcal R_N^{\mathcal{X}}:=\bigsqcup_{n=1}^N
 \bigl(\R^{n\times d_{\mathcal{X}}^{\mathrm{in}}}\times\R^{n\times n\times d_{\mathcal{Y}}^{\mathrm{in}}}\times[n]\bigr),
 \label{eq:rooted-domains}
\end{equation}
with their finite disjoint-union topologies.  For \(L\geq0\), all four
families below consist of maps to the same fixed space \(\R^q\), with one
parameter set shared over every \(n\leq N\).  The pair-only families are
\begin{equation}
 \begin{aligned}
 \mathcal E^{\mathcal{Y}}_{N,L}&:=\{(X,Y;i,j)\mapsto\pi(e^L_{ij})\},&
 \mathcal E^{\mathcal{X}}_{N,L}&:=\{(X,Y;i)\mapsto\pi(e^L_{ii})\},
 \end{aligned}
 \label{eq:rooted-et-families}
\end{equation}
where \(e^0\) is the endpoint tokenizer in \cref{eq:endpoint-lift}, each
counted ET layer is one direction-mixed triangular-attention residual followed
by one shared pairwise-MLP residual, and \(\pi\) selects designated
coordinates.  Each head is independently assigned its direction pair.  The EVT
families are
\begin{equation}
 \begin{aligned}
 \mathcal V^{\mathcal{Y}}_{N,L}&:=\{(X,Y;i,j)\mapsto\pi_{\mathcal{Y}}(y^L_{ij})\},&
 \mathcal V^{\mathcal{X}}_{N,L}&:=\{(X,Y;i)\mapsto\pi_{\mathcal{X}}(x^L_i)\},
 \end{aligned}
 \label{eq:rooted-evt-families}
\end{equation}
where \((X^L,Y^L)\) follows exactly \(L\) macro-layers
\cref{eq:within-att,eq:within-mlp,eq:y-to-x,eq:x-to-y,eq:cross-mlp}.
The raw affine lifts and final coordinate projections are interfaces and do
not count as layers.  In particular, the ET tokenizer may fuse endpoint
features before its first counted layer, whereas an EVT macro-layer used to
write that fused token into pair workspace is counted.

For each rooted family, the lifts, MLPs, head dimensions, direction choices,
normalization, and residual updates follow the architectural conventions of
\cref{def:striketwice-family}, with the input tokenizers and layer counts
defined above.
Unsubscripted families mean the arbitrary-depth unions
\(\mathcal E_N^\bullet=\bigcup_{L\geq0}\mathcal E_{N,L}^\bullet\) and
\(\mathcal V_N^\bullet=\bigcup_{L\geq0}\mathcal V_{N,L}^\bullet\).

The vertex stream does not raise the stable graph-information ceiling, but it
can expose useful operations at a shallower or cheaper typed schedule.  To
compare it with a pair-only direction-mixed Edge Transformer, initialize ET
by the pointwise endpoint lift used in prior ET analyses
\citep{bergen2021,muller2024}:
\begin{equation}
 e^0_{ij}=\operatorname{Lift}(y_{ij},x_i,x_j,\one_{\{i=j\}},1),
 \label{eq:endpoint-lift}
\end{equation}
where \(\operatorname{Lift}\) is a learned affine map, allowed to be
injective at sufficient width. An ET layer applies tokenwise pre-normalization
before triangular attention and before its shared GELU pair MLP.
Both rooted ET and EVT use the normalization and activation conventions of
\cref{def:striketwice-family}. Endpoint initialization changes how early
features become available; the exact schedule calculations below are
carried out on auxiliary computations before applying the simulation
of \cref{thm:ln-gelu-simulation}.

On a rooted pair input, set \(\gamma_N^{\mathcal{Y}}(X,Y;i,j)=c^{R_N}_{ij}\); on a
rooted vertex input, set \(\gamma_N^{\mathcal{X}}(X,Y;i)=c^{R_N}_{ii}\).

\begin{proposition}[Rooted EVT--ET kernel and closure equivalence]
\label{prop:evt-et}
For the rooted families above,
\begin{equation}
 \Ker(\mathcal V_N^{\mathcal{Y}})=\Ker(\mathcal E_N^{\mathcal{Y}})=\ker(\gamma_N^{\mathcal{Y}}),
 \qquad
 \Ker(\mathcal V_N^{\mathcal{X}})=\Ker(\mathcal E_N^{\mathcal{X}})=\ker(\gamma_N^{\mathcal{X}}).
 \label{eq:evt-et-kernel}
\end{equation}
Moreover, if \(S\) is a compact set of rooted pair or rooted vertex inputs
and \(\gamma\) is the corresponding code, then the appropriate ET and EVT
uniform closures both equal
\begin{equation}
 \mathcal C_\gamma(S)
 :=\{f\in\mathcal C(S;\R^q):\ker(\gamma|_S)\subseteq\ker(f)\}.
 \label{eq:rooted-closure}
\end{equation}
\end{proposition}
\begin{proof}[Proof sketch]
The transpose-closure and stable-reconstruction lemma below shows that one
pair color determines both endpoint colors, every oriented multiset over intermediate vertices,
and the global pair-color multiset.  Hence pair and vertex states factor
through \(c^\star_{ij}\) and \(c^\star_{ii}\).  Conversely, an oriented setup
head exposes the reverse atomic state, after which triangular attention and
finite-set MLP interpolation realize each \(2\)-FWL refinement in the
auxiliary computations. Finite feature-bank compilation and Stone--Weierstrass
produce auxiliary compact approximants. The simulation in
\cref{thm:ln-gelu-simulation} transfers separators and approximants to
normalized GELU ET and EVT. The detailed construction, including one network shared
over all \(n\leq N\), is given in \cref{app:evt-et-proof}.
\end{proof}

\paragraph{Auxiliary schedule comparison.}
For the next proposition only, the superscript \(\mathrm{aux}\) denotes
\cref{def:auxiliary-computations}: unnormalized residual computations with
ReLU pointwise maps. This isolates an exact algebraic comparison of access
schedules. It does not assert the same fixed-depth inclusions for arbitrary
normalized GELU networks. Each constructed auxiliary target nevertheless
admits a normalized approximation at the same depth by
\cref{thm:ln-gelu-simulation}.
To isolate direct pair-to-vertex access, let
\(\mathcal V^{\mathcal{X},\mathrm{red},\mathrm{aux}}_{N,1}\) use the following reduced typed
schedule: one triangular pair residual and its pair MLP, followed by one
global \(Y\!\to X\) attention and its vertex MLP.  Vertex self-attention,
the first vertex MLP, \(X\!\to Y\) attention, and the final pair MLP are
disabled.  This is not a full EVT macro-layer.  For the reduced comparison, the affine endpoint lift \cref{eq:endpoint-lift} is common
tokenization and costs no ET layer; one ET layer means exactly one
triangular-attention residual followed by one pairwise-MLP residual.  A
reduced depth counts one execution of the displayed typed schedule.

\begin{proposition}[Depth comparison for auxiliary computations]
\label{prop:finite-depth}
For the auxiliary computations with the indicated layer counts and every
compact set \(S\) of rooted vertex inputs,
\begin{align}
 \mathcal E^{\mathcal{Y},\mathrm{aux}}_{N,L}\subseteq\mathcal V^{\mathcal{Y},\mathrm{aux}}_{N,L+1},
 \qquad
 \mathcal V^{\mathcal{X},\mathrm{red},\mathrm{aux}}_{N,1}|_S
 \subseteq\overline{\mathcal E^{\mathcal{X},\mathrm{aux}}_{N,3}|_S}^{\|\cdot\|_\infty}.
 \label{eq:reduced-evt-et3}
\end{align}
The depth-three bound is sharp for this reduced schedule: for \(N\geq7\),
there is a two-point finite domain of rooted binary graphs on which a reduced
one-layer EVT target has positive distance from every two-layer ET.  On a
second two-point rooted domain, a two-layer ET target separates the inputs
whereas every reduced one-layer EVT agrees.  Thus
\(\mathcal V^{\mathcal{X},\mathrm{red},\mathrm{aux}}_{N,1}\) and \(\mathcal E^{\mathcal{X},\mathrm{aux}}_{N,2}\) are
incomparable after restriction to suitable finite domains.
\end{proposition}
\begin{proof}[Proof sketch]
The first inclusion uses one attention-disabled EVT setup layer and then
disables the extra branches.  For the second, the three ET layers respectively
compute the reduced pair update, row moments, and global moments at the
diagonal root.  A uniform separable approximation of the softmax kernel then
recovers the pair-to-vertex update.  Triangle density and two rooted leaves of
a broom graph give the two strict directions.  The layer schedule, uniform
error argument, and finite domains are explicit in \cref{app:finite-depth}.
In particular, no three-layer compilation of a full EVT macro-layer is
claimed.
\end{proof}

Within these auxiliary computations, ordinary vertex self-attention is not the source of the gap: two triangular
heads with reversed query/value roles reproduce both endpoint updates exactly
in one ET layer
(\cref{app:vertex-attention}).  EVT instead avoids storing and recomputing
those shared endpoint results across all \(n^2\) pair tokens.

\subsection{Equivariant EVT kernels and approximation}
\label{app:evt-array-proof}

We prove \cref{thm:evt-expressivity} for the graph-shaped predictions of
\(f_\theta=\Aff_{\theta_{\mathrm{out}}}\circ
\EVT_{\theta_{\mathrm{EVT}}}\circ\Aff_{\theta_{\mathrm{in}}}\),
not merely for one selected coordinate. The EVT core preserves its freely
chosen representation widths. The input domain is
\[
 \mathcal G_{\leq N}=\bigsqcup_{n=1}^N
 \left(\R^{n\times d_{\mathcal{X}}^{\mathrm{in}}}
 \times\R^{n\times n\times d_{\mathcal{Y}}^{\mathrm{in}}}\right),
\]
and a graph of size \(n\) receives an output in
\(\R^{n\times d_{\mathcal{X}}^{\mathrm{out}}}\times
\R^{n\times n\times d_{\mathcal{Y}}^{\mathrm{out}}}\).
Equality of complete arrays includes equality of their shapes.
For any \(T\subseteq\mathcal G_{\leq N}\), write
\(T_n=T\cap\mathcal G_n\) and define the indexed domains
\[
 \widetilde T_{\mathcal X}:=\bigsqcup_{n=1}^N T_n\times[n],
 \qquad
 \widetilde T_{\mathcal Y}:=\bigsqcup_{n=1}^N T_n\times[n]^2.
\]
The two components of \(\ker^{\mathrm{eq}}(f|_T)\) are ordinary
equivalence relations on \(\widetilde T_{\mathcal X}\) and
\(\widetilde T_{\mathcal Y}\), respectively; no comparison is made
between vertex and pair outputs. For the color profile they are the
kernels of
\(\Psi_{\mathrm{EVT},\mathcal X}^{\bullet}(G;i)=c^\star_{ii}\) and
\(\Psi_{\mathrm{EVT},\mathcal Y}^{\bullet}(G;i,j)=c^\star_{ij}\).
The indices only select output locations: they are not supplied to the
encoder and do not mark its input. If \(T\) is compact, both indexed
domains are finite disjoint unions of compact sets.

\begin{proof}[Proof of \cref{thm:evt-expressivity}]
\textbf{The equivariant kernel.}
The factorization in \cref{eq:evt-factorization} holds across all graphs
and all output locations: the vertex output is a function of
\(c^\star_{ii}\), and the pair output is a function of
\(c^\star_{ij}\). Shared affine output maps preserve this property.
Consequently
\[
 \ker^{\mathrm{eq}}(\Psi_{\mathrm{EVT}})
 \subseteq\Ker^{\mathrm{eq}}(\mathcal H_{\mathrm{EVT}}).
\]
Conversely, take two vertex-rooted inputs with different diagonal colors,
or two pair-rooted inputs with different pair colors. The corresponding
rooted-kernel identity in \cref{prop:evt-et} supplies a normalized EVT
computation with different outputs on those inputs. A scalar coordinate
already separates them. Use the shared affine output map to place that
coordinate in the first prediction channel of the corresponding stream,
and set all other output channels to zero. This is a member of
\(\mathcal H_{\mathrm{EVT}}\), because both prescribed output widths
are positive and its internal widths are free. It separates the two
rooted inputs while using the same parameters at all graph sizes.
This proves the reverse inclusion in both components, and hence the
asserted equivariant-kernel equality.

\textbf{Necessity of kernel compatibility.}
The first inclusion shows that every member of
\(\mathcal H_{\mathrm{EVT}}|_S\) satisfies
\cref{eq:evt-entrywise-compatible}.
All its operations are continuous on each size component; the positive
normalization stabilizer ensures this also for RMSNorm and LayerNorm.
Uniform limits remain continuous and preserve the two entrywise
equalities. Therefore
\[
 \overline{\mathcal H_{\mathrm{EVT}}|_S}
 \subseteq\mathcal C^{\mathrm{eq}}_{\Psi_{\mathrm{EVT}}}(S).
\]

\textbf{Simultaneous approximation of both arrays.}
Fix \(f\in\mathcal C^{\mathrm{eq}}_{\Psi_{\mathrm{EVT}}}(S)\) and
\(\varepsilon>0\). On the compact indexed domains
\(\widetilde S_{\mathcal X}\) and \(\widetilde S_{\mathcal Y}\),
the targets \(f_{\mathcal X}^{\bullet}\) and
\(f_{\mathcal Y}^{\bullet}\) are continuous and constant on the fibers
of the diagonal and pair colors,
respectively, by \cref{eq:evt-entrywise-compatible}.
The auxiliary Stone--Weierstrass construction in the proof of
\cref{prop:evt-et}, detailed in \cref{app:evt-et-proof}, therefore gives
auxiliary EVT computations approximating these two targets uniformly to
any prescribed Euclidean error \(\delta>0\).

Apply \cref{lem:parallel-compilation} to these two auxiliary computations,
padding their depths by identity layers and retaining their designated
outputs in separate groups of coordinates. This gives one auxiliary EVT
whose vertex and pair arrays \(A_{\mathcal{X}},A_{\mathcal{Y}}\)
simultaneously satisfy
\[
 \sup_{G\in S}
 \left\|\bigl(A_{\mathcal{X}}(G),A_{\mathcal{Y}}(G)\bigr)-f(G)
 \right\|_{\mathrm F}
 \leq\sqrt{N+N^2}\,\delta.
\]
The two feature dimensions need not coincide: the Euclidean error
\(\delta\) bounds each complete vertex vector or pair vector.
Choose \(\delta<\varepsilon/(2\sqrt{N+N^2})\).
The entire-array encoder version of \cref{thm:ln-gelu-simulation} then
approximates this single auxiliary computation to joint uniform error
less than \(\varepsilon/2\), with either the prescribed pre-RMSNorm or
pre-LayerNorm. The triangle inequality yields a prediction in
\(\mathcal H_{\mathrm{EVT}}\) within
\(\varepsilon\) of \(f\).

Parallel compilation is performed before normalization is introduced;
simply concatenating two already normalized networks would change their
normalization statistics and would not justify this argument. The
simulator uses finite widened states, fixed along each residual stack,
surrounded by the shared outer affine input and output maps. Thus one finite parameter
set approximates all locations and all sizes represented in \(S\).
This proves the reverse inclusion and the asserted closure equality.
\end{proof}

\begin{remark}[Equivariant kernels, equivariance, and whole-array kernels]
On a relabeling-stable compact set \(S\), the colors satisfy
\(c^\star_{\sigma(i)\sigma(j)}(\sigma G)=c^\star_{ij}(G)\).
Applying \cref{eq:evt-entrywise-compatible} with \(G'=\sigma G\)
therefore gives permutation equivariance of both output arrays.
The converse is false: equivariance alone does not impose all equalities
between locations having the same stable color.

Likewise, the ordinary condition
\(\ker(\Psi_{\mathrm{EVT}}|_S)\subseteq\ker(f)\) is insufficient for
EVT approximation. The complete indexed color array retains every initial
color, hence every raw feature \(x_i,y_{ij}\), so its whole-array kernel
is the diagonal relation on \(\mathcal G_{\leq N}\). The common
whole-array kernel of \(\mathcal H_{\mathrm{EVT}}\) is also this relation:
different input sizes give different output shapes, and any difference
in a raw feature coordinate can be copied to the corresponding output
stream by shared affine input and output maps with every residual branch
disabled. Thus
\(\Ker(\mathcal H_{\mathrm{EVT}})=\ker(\Psi_{\mathrm{EVT}})\)
still holds, but that ordinary kernel places no restriction on a
continuous graph-shaped target. The equivariant kernel retains the
additional comparisons between locations and graphs needed for
approximation.
For an explicit equivariant counterexample, use the rook and Shrikhande
graphs from \cref{app:marked-small-witnesses}, with constant vertex
features and binary adjacency pair features. Let \(S\) contain all
relabelings of both graphs. This is a finite compact relabeling-stable
set. Define \(f_{\mathcal{X}}\) to be zero on every rook-graph vertex
and the first standard basis vector of \(\R^{d_{\mathcal{X}}^{\mathrm{out}}}\) on
every Shrikhande-graph vertex, and set \(f_{\mathcal{Y}}=0\).
This target is continuous, equivariant, and compatible with the
whole-array kernel. However, all diagonal stable colors agree across the
two graphs, as verified in \cref{app:marked-small-witnesses}. Every prediction in
\(\mathcal H_{\mathrm{EVT}}\)
must consequently return the same vertex vector in both cases and has
uniform error at least \(1/2\). The equivariant-kernel condition in
\cref{thm:evt-expressivity} excludes precisely this obstruction.
\end{remark}

\subsection{StrikeTwice at variable capacity}
\label{app:theory-scope}

\begin{definition}[StrikeTwice architecture family]
\label{def:striketwice-family}
The family \(\mathcal H_{N,K}\) consists of maps
\(\mathcal D_{\leq N,\leq K}\to\mathcal O_K\) obtained from
the core in \cref{eq:graph-stage,eq:tokenwise-affine,eq:path-stage}, surrounded
by the affine maps in \cref{eq:path-output}. Thus
\(f_\theta=\Aff_{\theta_{\mathrm{out}}}\circ
\STarch_{\theta_{\mathrm{ST}}}\circ
\Aff_{\theta_{\mathrm{in}}}\), with the last map leaving the path
unchanged. The following choices specify the family.
\begin{enumerate}
 \item Graph and path stages have independent parameters and arbitrary
 finite macro-depths, state widths, head counts, and query/key and value
 dimensions. Each triangular head may use any of the four orientations.
 Parameters are shared over indices and all admissible sizes within a stage.
 Every macro-layer follows exactly
 \cref{eq:within-att,eq:within-mlp,eq:y-to-x,eq:x-to-y,eq:cross-mlp}.
 \item The outer input map lifts raw token widths
 \(d_{\mathcal{X}}^{\mathrm{in}},d_{\mathcal{Y}}^{\mathrm{in}}\)
 to free graph widths \(d_{\mathcal{X}},d_{\mathcal{Y}}\), reading only
 \(x_i\) for a vertex and \(y_{ij}\) for a pair. The graph EVT preserves
 these widths, so its cache has the same dimensions. After \(\Sub_P\),
 the core bridge \(\Aff_{\theta_{\mathrm{aff}}}\) maps token widths
 \(d_{\mathcal{X}}+K,d_{\mathcal{Y}}+2K+1\) to independently chosen
 path widths \(q_{\mathcal{X}},q_{\mathcal{Y}}\), which the path EVT
 preserves. These affine maps may compress information; at
 sufficiently large widths they also admit information-preserving choices
 and additional constant or zero coordinates. Stems and coordinate readouts
 do not count as macro-layers. The aligned pair MLP reads \(y_{ij},x_i,x_j\)
 and the index-equality bit, never \(y_{ji}\) directly; reverse access uses
 oriented triangular heads.
 \item Fix either RMSNorm or LayerNorm throughout a model. Attention uses
 tokenwise pre-normalization; each aligned MLP normalizes its entire
 concatenated local input, followed by two biased affine maps with one
 hidden GELU layer. Normalization has learned gains (and LayerNorm biases)
 and a fixed positive stabilizer. Each branch owns its
 parameters and can be disabled by zeroing its residual output. There is
 no final state normalization, and dropout is inactive at evaluation.
 \item Position and equality features are injected into ordinary states.
 No coordinates are required to stay fixed. Shared affine output interfaces
 map the final vertex and pair states to \(\R^{d_{\mathcal{X}}^{\mathrm{out}}}\) and \(\R^{d_{\mathcal{Y}}^{\mathrm{out}}}\),
 respectively; selecting designated state coordinates is allowed.
 This outer output map is separate from the EVT core and is not residual.
 Task-specific layers are specified in
 \cref{app:practical-interface}.
\end{enumerate}
Widths, depths, head counts, and MLP widths are finite for each model but
unbounded across the family. State widths remain fixed through identity
residual additions within a stage; the outer affine maps and the core
graph-to-path bridge allow the raw, graph, path, and prediction widths
to differ. No terminal graph projection is required: any such affine
projection followed by selection can be absorbed into the bridge,
which also reads the unchanged occurrence tags. Head counts and projection
dimensions may differ across layers and attention banks. All four orientations are available;
a bank may omit some. Usual width-divisibility constraints can be imposed
without changing the results, as shown in \cref{app:unified-simulation}.
One parameter set acts simultaneously on every bounded size component.
Experimental budgets fix the capacity of these same encoders and attach
the task-specific layers described with each experiment.
\end{definition}

\paragraph{Auxiliary computations for the proofs.}
Exact color codes and parallel groups of coordinates are convenient for
constructing approximants. The following notation describes intermediate
computations that will be simulated inside StrikeTwice by
\cref{thm:ln-gelu-simulation}; it does not change the architecture definition.
\begin{definition}[Auxiliary residual computations]
\label{def:auxiliary-computations}
Use the same indexing, attention formulas, and residual schedules, but read
unnormalized states in the attention projections and pointwise maps, and
allow biased ReLU MLPs of arbitrary finite width and depth. Affine input
lifts and affine selectors can retain all source coordinates, position and
equality tags, and any finite constant or zero workspace. Residual output
rows may be zero so that these coordinates stay fixed.
Write \(\mathcal H^{\mathrm{aux}}_{N,K}\) for the resulting cascade
computations, and add the superscript \(\mathrm{aux}\) to the rooted
\(\mathcal E\) and \(\mathcal V\) notation when these auxiliary
computations are used. Their interfaces and final coordinate projections
are affine and do not count as layers. The raw pair lift reads only
\(y_{ij}\); the ET endpoint lift is \cref{eq:endpoint-lift}.
\end{definition}
The exact algebra below is carried out on these auxiliary computations.
Its separators and compact approximants are transferred to the architecture
of \cref{def:striketwice-family} by the simulation theorem, preserving the
number of macro-layers for approximation. Exact finite interpolation uses
an additional block when stated in \cref{cor:unified-finite-interpolation}.

The path-profile proof invokes \cref{prop:evt-et} as a graph-stage realization
result: every first-stage EVT factors through stable \(2\)-FWL colors, and one
can be chosen to separate any finite collection of such colors.  This is the
only graph-stage expressivity fact required below.

\subsection{Auxiliary compilation primitives}
\label{app:compilation}

\begin{lemma}[Compilation of auxiliary computations]
\label{lem:parallel-compilation}
The auxiliary computations of \cref{def:auxiliary-computations} have
the following properties.
\begin{enumerate}
 \item They can create constant and zero auxiliary coordinates, copy or
 protect designated coordinates exactly, disable any branch, pad depth with
 identity layers, and interpolate an arbitrary map exactly on a finite set of
 distinct token codes.
 \item Any finite collection of networks from one of these families can be
 compiled into one wider network whose disjoint groups of coordinates reproduce all
 their designated states exactly.  For cascades, graph-stage and path-stage
 depths are padded independently.
 \item On a compact input set, one appended attention-disabled pointwise ReLU
 block can uniformly approximate any continuous function of the combined
 coordinates stored at the same rooted token.
\end{enumerate}
\end{lemma}
\begin{proof}
A biased ReLU MLP represents affine maps exactly by the identity
\(z=\operatorname{ReLU}(z)-\operatorname{ReLU}(-z)\).  It therefore writes
constants, copies coordinates, and leaves protected output rows at zero.  To
interpolate distinct codes \(z_1,\ldots,z_t\), choose a linear functional
whose values on these codes are pairwise distinct and interpolate the desired
outputs by a univariate continuous piecewise-affine function; its hinge
representation is a finite biased ReLU network. Zeroing an attention output
matrix, or both an MLP's final matrix and final bias, disables that residual.
Zeroing every residual gives a legal identity layer.

For parallel compilation, first pad the component schedules to their maximum
depths with these identity layers.  Concatenate their input lifts and hidden
states, assign each original attention head to its own group of coordinates, and make every
output projection and pointwise MLP block diagonal.  Thus each softmax is
formed only from the queries and keys of its original network; neither residuals
nor nonlinearities mix the coordinate groups.  If an implementation requires a common head
dimension \(D\), zero-pad a head of original query dimension \(d_a\) and
multiply its query projection by \(\sqrt{D/d_a}\).  Its score remains
\(\langle q,k\rangle/\sqrt{d_a}\) after division by \(\sqrt D\); value
coordinates are merely zero-padded.  Hidden ReLU depths are equalized by
identity layers on their nonnegative hidden activations.  This constructs the
joint state exactly, including when the original networks have different
widths, head counts, and, for a cascade, graph and path depths.

Finally, the image of a compact input set under the combined state map is compact.  Extend each
coordinate of a continuous postcomposition from that image to an enclosing
Euclidean box by the Tietze extension theorem, and apply the ReLU universal
approximation theorem there.  The resulting pointwise map is written into
fresh zero coordinates by a final disabled-attention residual block.  The
exact direct sum is the architecture-specific closure needed by the usual
compact relative-universality strategy for invariant and equivariant networks
\citep{keriven2019universal,maron2019universality,azizian2021}.
\end{proof}

\subsection{Simulation with normalization and GELU}
\label{app:unified-simulation}

The constructions in the expressivity proofs use the auxiliary computations
of \cref{def:auxiliary-computations} to perform exact finite coding and
parallel compilation. This subsection implements their outputs, to arbitrary
uniform accuracy, in the StrikeTwice architecture of
\cref{def:striketwice-family}. ReLU gates and unnormalized intermediate
registers are proof devices. Every resulting model uses the chosen pre-normalization,
one-hidden-layer GELU aligned MLPs, the outer affine input/output maps,
and the core graph-to-path affine bridge.
The proof also covers the rooted EVT and endpoint-lifted ET encoders.

\paragraph{Linear recovery from normalized states.}
Let $s\in\R^d$ contain a complete auxiliary token, including its constants,
workspace, and fixed tags. For a state width $D\geq d+3$ and $A>0$, define
\begin{equation}
 E_A(s)=[s;0;A;-A;0_{D-d-3}].
 \label{eq:unified-encoding}
\end{equation}
These are ordinary state coordinates created by an affine stem. LayerNorm
has its prescribed stabilizer $\epsilon>0$; choose unit gain and zero bias.

\begin{lemma}[Linear recovery with constant anchors]
\label{lem:unified-anchors}
For $r\geq1$, let
\[
 Z_A(v)=[E_A(s_1);\ldots;E_A(s_r);b],\qquad
 v=[s_1;\ldots;s_r;b],\qquad M=rD+\dim b,
\]
where $b$ is a bounded local covariate, possibly absent. Choose the index
$a_0$ of one displayed zero coordinate. Let $\Pi$ extract each coordinate
of $v$ after subtracting coordinate $a_0$, and set
\[
 L_{r,A,M}=\sqrt{2rA^2/M+\epsilon}\,\Pi.
\]
Then
\begin{align}
 L_{r,A,M}\LN_\epsilon(Z_A(v))&=t_A(v)v,
 \label{eq:unified-linear-recovery}\\
 t_A(v)&=\left(1+
 \frac{\|v\|_2^2-(\one^\top v)^2/M}{2rA^2+M\epsilon}
 \right)^{-1/2}.
 \label{eq:unified-scale}
\end{align}
Uniformly over $\|v\|_2\leq B$,
\begin{equation}
 \|L_{r,A,M}\LN_\epsilon(Z_A(v))-v\|_2
 \leq\frac{B^3}{4rA^2+2M\epsilon}.
 \label{eq:unified-anchor-error}
\end{equation}
The decoding map depends on the local input type, not on indices or sizes.
\end{lemma}

\begin{proof}
The anchors sum to zero, and hence
\[
 \one^\top Z_A(v)=\one^\top v,\qquad
 \|Z_A(v)\|_2^2=2rA^2+\|v\|_2^2.
\]
The variance used by LayerNorm is therefore
$2rA^2/M+\|v\|_2^2/M-(\one^\top v)^2/M^2$.
Subtracting the normalized zero coordinate cancels the mean exactly.
Multiplication by $\sqrt{2rA^2/M+\epsilon}$ gives
\cref{eq:unified-linear-recovery,eq:unified-scale}.
Since $\dim v\leq M$, Cauchy--Schwarz gives
$0\leq \|v\|_2^2-(\one^\top v)^2/M\leq\|v\|_2^2$.
Apply $0\leq1-(1+x)^{-1/2}\leq x/2$ for $x\geq0$ to obtain
\cref{eq:unified-anchor-error}.
\end{proof}

For attention this lemma uses $r=1$, $M=D$, and no $b$. For an aligned
vertex MLP it uses $r=2$, $M=2D$, and no $b$. For an aligned pair MLP
it uses $r=3$, $b=\delta_{ij}$, and $M=3D+1$; at the path stage
$\delta_{st}$ is occurrence equality. Thus the proof covers LayerNorm of
the \emph{entire} concatenation, including the equality bit. A pairwise
ET MLP uses the $r=1$ case.

The injectivity of $h\mapsto\LN([h;0;1])$ alone would not suffice:
its inverse uses a ratio, which a linear attention projection cannot in
general evaluate. Here a fixed linear map recovers the payload with a
uniform $O(A^{-2})$ error on bounded sets. Constant mean and variance are
not required.

\begin{lemma}[Linear recovery with RMSNorm]
\label{lem:unified-rms-anchors}
For the same anchored tuple \(Z_A(v)\), let \(\Pi_0\) select the
payload coordinates directly and set
\(L^{\mathrm{RMS}}_{r,A,M}=\sqrt{2rA^2/M+\epsilon}\,\Pi_0\).
With unit RMSNorm gain,
\[
 L^{\mathrm{RMS}}_{r,A,M}\RMSNorm_\epsilon(Z_A(v))
 =\left(1+\frac{\|v\|_2^2}{2rA^2+M\epsilon}\right)^{-1/2}v.
\]
Its error on \(\|v\|_2\leq B\) is at most
\(B^3/(4rA^2+2M\epsilon)\).
\end{lemma}
\begin{proof}
The squared norm of the tuple is \(2rA^2+\|v\|_2^2\).
Substitution in RMSNorm gives the identity; apply
\(0\leq1-(1+x)^{-1/2}\leq x/2\) for the bound.
\end{proof}
In the following simulation, write \(\Lambda_{r,A,M}\) for
\(L_{r,A,M}\) with LayerNorm and for \(L^{\mathrm{RMS}}_{r,A,M}\)
with RMSNorm, and let \(t_A\) denote the corresponding scalar factor.
Both decoders are linear maps and
therefore can be absorbed into attention projections and the first
affine map of each MLP. The proof does not rely only on injectivity
of normalized states.

\paragraph{GELU MLPs and immutable coordinates.}
Write $g(t)=t\Phi(t)$ for GELU, where $\Phi$ is the standard Gaussian
distribution function \citep{hendrycks2016gelu}, and let $g_a(t)=g(at)/a$.

\begin{lemma}[Shallow GELU approximation]
\label{lem:unified-gelu-approximation}
For $a>0$ and $t\in\R$,
\begin{equation}
 |g_a(t)-\operatorname{ReLU}(t)|
 =|t|\Phi(-a|t|)\leq\frac{1}{a\sqrt{2\pi}}.
 \label{eq:unified-gelu-error}
\end{equation}
Biased Linear--GELU--Linear MLPs of arbitrary finite hidden width are
uniformly dense in the continuous maps on any compact Euclidean box.
\end{lemma}

\begin{proof}
The identity follows from $\Phi(-t)=1-\Phi(t)$, and the bound follows
from $x\Phi(-x)\leq(2\pi)^{-1/2}$ for $x\geq0$.
The nonpolynomial-activation approximation theorem gives density of biased
one-hidden-layer ReLU MLPs \citep{leshno1993}. For a fixed approximant
$c+\sum_j a_j\operatorname{ReLU}(w_j^\top v+b_j)$, replacing the
hinges by $g_a$ changes its output norm by at most
$\sum_j\|a_j\|_2/(a\sqrt{2\pi})$.
The input and output rescalings are absorbed into the two affine maps.
Vector-valued targets use a finite union of hidden units.
\end{proof}

Let $J:\R^d\to\R^D$ append zeros on the auxiliary coordinates. Then
\begin{equation}
 E_A(s)+J\Delta=E_A(s+\Delta).
 \label{eq:unified-residual}
\end{equation}
Following each attention output projection or MLP output affine map by
$J$ preserves the zero reference, anchors, and padding exactly. If an
auxiliary tag must remain fixed, its logical output row is also zero.
These are parameter choices of an ordinary residual block. Branches and
unused heads can be disabled exactly by zero output maps.

\begin{theorem}[Uniform simulation in StrikeTwice]
\label{thm:ln-gelu-simulation}
Fix $N,K\geq1$ and an auxiliary computation $H$ of
\cref{def:auxiliary-computations}, with finite capacity, affine raw stems
and affine selector interfaces, and any fixed finite numbers of output
coordinates per occurrence and per ordered occurrence pair.
For every compact $S\subset\mathcal D_{\leq N,\leq K}$ and every
$\eta>0$, some StrikeTwice model $\widetilde H$ in
\cref{def:striketwice-family} satisfies
\begin{equation}
 \sup_{u\in S}\|\widetilde H(u)-H(u)\|_{\mathrm F}<\eta.
 \label{eq:unified-simulation}
\end{equation}
One parameter set serves all size components. The graph and path
macro-depths can equal those of $H$. The simulator has finite width and
one-hidden-layer GELU MLPs; its parameters and widths may depend on $S$
and $\eta$.

The same statement holds for rooted EVT encoders and for endpoint-lifted
ET encoders, preserving their respective counted depths and approximating
the entire terminal arrays after fixed coordinate projections. An ET
layer here consists of one triangular-attention residual followed by one
pairwise-MLP residual, both with the chosen pre-normalization in the simulator. Any
disabled residual branches or restrictions on active head orientations
in $H$ can be retained. If an implementation requires a positive depth,
one zero-increment identity layer can represent a depth-zero stage.
\end{theorem}

\begin{proof}
\emph{Fixed logical neighborhoods.}
Pad the finitely many complete auxiliary stream and stage types to a
common logical dimension $d$. Treat simultaneous residual updates as one
transition, preserving the prescribed schedule. On each discrete
size/path component, the auxiliary stems, attentions, ReLU MLPs, and
selector are continuous. The finitely many images of $S$ are compact.
Choose fixed compact logical boxes containing these images and their
radius-one neighborhoods, as well as all local MLP tuples reached there.
The boxes can be chosen centered at zero.

\emph{Outer maps and the graph-to-path bridge.}
In the outer input map, replace an auxiliary raw stem $I$ by $E_A\circ I$. Its raw input access
is unchanged. At the selector, the first $d$ coordinates of every
simulated graph token are its unscaled logical state. The learned affine
path stem extracts the required coordinates, applies the auxiliary
affine interface, copies the position and graph-equality tags, and
appends fresh anchors. On encoded states this produces exactly the
encoded auxiliary path input. An optional relative feature
$(t-s)/\max\{1,K-1\}$ is affine in the canonical one-hot position
codes and can be absorbed into the same stem.
The stem need only preserve the reachable affine subspace of encoded
tokens. In particular, mapping a $D$-dimensional token plus tags to a
$D$-dimensional state need not be globally injective: its free logical
coordinates and tags fit within $d$, and the remaining coordinates are
fixed anchors or unused padding. Equation~\eqref{eq:unified-residual}
keeps approximate states in this same encoded subspace.

\emph{Attention.}
Use unit normalization gain and, for LayerNorm, zero bias. For each auxiliary token
projection $W$, use $W\Lambda_{1,A,D}$, retaining any affine bias if present.
By \cref{lem:unified-anchors,lem:unified-rms-anchors}, using the decoder
for the chosen normalization, its output on $E_A(s)$ converges
uniformly to $Ws$ on the fixed logical boxes. This applies to queries,
keys, and values, including both factors of a triangular Hadamard
value product. The head orientation and its own softmax denominator
are retained. Dot products, Hadamard products, finite softmaxes, and
output maps are continuous on the resulting compact projected sets,
so their logical residual increments converge uniformly to the
auxiliary increments. The output map writes them through $J$.

For a common head width $d'=D/H_0$, zero-pad all auxiliary projected
vectors. If an auxiliary query/key width is $d_a$, multiply its query
projection by $\sqrt{d'/d_a}$, so that
\[
 \frac{\langle\sqrt{d'/d_a}\,q,k\rangle}{\sqrt{d'}}
 =\frac{\langle q,k\rangle}{\sqrt{d_a}}.
\]
Choose $H_0\in4\mathbb N$ large enough for every ordinary head bank
and with $H_0/4$ at least the largest auxiliary number of heads of any
one triangular orientation. Choose $D$, divisible by $H_0$, so that
$D/H_0$ accommodates every query/key and value dimension. A balanced
bank with $H_0/4$ heads per orientation can contain all required heads;
unused heads have zero outputs. This balanced common allocation is
only a sufficient choice for the construction. The architecture allows
different widths, head counts, and orientation allocations between
banks and stages. When preserving an orientation restriction, only
the permitted heads have nonzero outputs; unused heads can also be
assigned an allowed orientation without changing the computation.

\emph{Aligned MLPs.}
For each auxiliary local increment $f(v)$, choose a fixed shallow GELU
MLP $G(v)$ that approximates $f$ on the appropriate larger logical box,
using \cref{lem:unified-gelu-approximation}. Replacing its first affine
matrix $B_1$ by $B_1\Lambda_{r,A,M}$ makes its input after the full
concatenation normalization equal to $t_A(v)v$, with the corresponding
factor from \cref{lem:unified-anchors,lem:unified-rms-anchors}. Its output is written as
$JG(t_A(v)v)$, with all protected output rows zero. On the fixed box,
\[
 \|G(t_A(v)v)-f(v)\|
 \leq \|G(t_A(v)v)-G(v)\|+\|G(v)-f(v)\|.
\]
The second term is as small as desired by the choice of $G$; for this
fixed $G$, the first tends uniformly to zero as $A\to\infty$.
This also covers arbitrary-depth auxiliary ReLU MLPs, since their
increments are continuous maps on Euclidean space.

The order of choices is relevant: first fix the logical boxes and
GELU approximants, then choose a common state width $D$, and finally
choose $A$. If the implementation fixes the hidden width at $4D$, or
another fixed positive multiple of $D$, enlarge $D$ to accommodate all
chosen hidden widths and pad unused units with zero outputs. The
linear maps in \cref{lem:unified-anchors,lem:unified-rms-anchors} are then formed for this
fixed $D$. No increase in the MLP depth or in the macro-depth is needed.

\emph{Error accumulation.}
Let $F_j$ denote the $j$th auxiliary transition in logical array
coordinates, including the selector. On the chosen neighborhoods it
has a finite Lipschitz constant $L_j$: ReLU MLPs are Lipschitz and
attention is smooth on bounded sets. Take a maximum over the finitely
many discrete components. Let $\widetilde F_j$ be the simulated logical
transition and choose a uniform bound
$\|\widetilde F_j(z)-F_j(z)\|_{\mathrm F}\leq\delta_j$ there.
The preceding local estimates yield such bounds for full arrays since
$n\leq N$ and $k\leq K$. They make all $\delta_j$ arbitrarily small
with finitely many approximants and one sufficiently large $A$.
The reached error obeys
\begin{equation}
 e_{j+1}\leq\delta_j+L_j e_j,\qquad e_0=0.
 \label{eq:unified-error-recurrence}
\end{equation}
Only the auxiliary transition constants $L_j$ are used. Choosing the
$\delta_j$ so that every partial recurrence is below one and the final
readout error is below $\eta$ proves inductively that the approximate
states remain inside the chosen neighborhoods. The output projection
reads the logical coordinates directly. All widths, weights, and $A$
are finite and independent of the particular input and its size.

The same argument without the selector gives the rooted and full-array
EVT statements. For ET, its affine endpoint lift becomes $E_A\circ I$;
the triangular attention construction is unchanged, and its local
pairwise MLP uses $r=1$. The proof then follows its original sequence
of counted layers. Zero auxiliary increments are implemented by zero
output maps throughout, which proves the restriction assertions.
\end{proof}

\paragraph{How the simulation is used.}
RMSNorm, LayerNorm and GELU are local continuous functions, so they preserve
the stable-color factorization used for the upper bounds. For the
converse kernel inclusion, an auxiliary separator on two inputs can
be simulated with error below one third of its output separation.
For compact density, an auxiliary approximant with error below
$\eta/2$ is simulated to within $\eta/2$.
This transfers the separation and approximation constructions without
claiming exact parallel compilation of normalized models: normalization
would couple their coordinate groups. Exact finite outputs require
the following additional argument.

\begin{lemma}[Exact finite GELU interpolation]
\label{lem:gelu-finite-interpolation}
For distinct $z_1,\ldots,z_m\in\R^p$ and arbitrary
$w_1,\ldots,w_m\in\R^q$, a biased one-hidden-layer GELU MLP
maps $z_i$ exactly to $w_i$ for every $i$.
\end{lemma}

\begin{proof}
Choose a linear functional giving distinct scalars
$t_1<\cdots<t_m$ on the points, after reordering. The evaluation matrix
of $1,\operatorname{ReLU}(t-t_1),\ldots,
\operatorname{ReLU}(t-t_{m-1})$ is lower triangular, with diagonal
$1,t_2-t_1,\ldots,t_m-t_{m-1}$. Replace each hinge by $g_a$.
By \cref{eq:unified-gelu-error} the new matrix converges to this
invertible matrix, and hence is invertible for sufficiently large
finite $a$. Solve the evaluation system for the output coefficients,
coordinatewise. The output bias supplies the constant column; for
$m=1$, a constant output suffices.
\end{proof}

\begin{corollary}[Exact finite outputs from separated codes]
\label{cor:unified-finite-interpolation}
Let $S$ be finite and let an auxiliary cascade produce distinct
terminal vertex codes for every distinct $(\Psi_{N,K}(u),s)$
represented in $S$. If its graph/path macro-depths are $(L_G,L_P)$,
every profile-compatible target on $S$ is exactly represented by a
StrikeTwice model of macro-depths $(L_G,L_P+1)$, apart from any
mandatory identity padding. Disabled path-attention branches can
remain disabled.
\end{corollary}

\begin{proof}
Reserve additional zero logical coordinates for the desired output.
Apply \cref{thm:ln-gelu-simulation} accurately enough to retain all
distinct codes. The simulator keeps its anchors exact. The stable-color
factorization ensures that equal profile/position classes still have
identical terminal states.

Append one path macro-layer with all attention outputs zero. Its first
aligned vertex MLP receives $\nu=\Norm([E_A(\widehat x_s);E_A(\widehat y_{ss})])$.
This normalization is injective on anchored tuples: for any physical
coordinate $z_j$ of the unnormalized tuple,
\begin{equation}
 z_j=A\frac{\nu_j-\nu_{a_0}}{\nu_{+A}-\nu_{a_0}},
 \label{eq:unified-anchor-inverse}
\end{equation}
whose denominator is positive. Distinct vertex codes therefore give
distinct normalized MLP inputs. Apply
\cref{lem:gelu-finite-interpolation} on this finite input set to write
the prescribed vertex targets into the reserved coordinates. Simultaneously,
the aligned pair MLP reads the two endpoint codes and its own pair state.
The endpoint codes determine the profile and the ordered positions \((s,t)\);
its anchored normalized input is injective on these local tuples by the
same ratio formula. A second application of
\cref{lem:gelu-finite-interpolation} writes the prescribed pair targets
into fresh zero pair coordinates. Set all other output rows and the
remaining sublayers to zero. Both coordinate readouts are exact.
\end{proof}

Thus approximation preserves the auxiliary macro-depth, whereas the
finite exact-output construction uses one additional path macro-layer.
Neither result asserts equality of internal trajectories or exact
inclusions between function sets at prescribed depths. Capacity and
parameter magnitudes may grow with the compact set and requested
accuracy; the result imposes no guarantee at a fixed experimental budget.

\paragraph{Marked encoders and occurrence-tagged readouts.}
The same architecture can encode the marked inputs from
\cref{eq:path-mark,eq:marked-init}. Its vertex readout must retain
the receiving occurrence: it is a shared GELU MLP on
$[\widetilde x_{p_s};\phi(s)]$ for EVT, or
$[\widetilde e_{p_sp_s};\phi(s)]$ for ET.
The pair readout additionally receives both endpoint states and both
occurrence tags, as specified in \cref{app:marked-model}.

\begin{corollary}[Simulation with a marked readout]
\label{cor:unified-marked-transfer}
Every auxiliary marked EVT or ET computation followed by a biased
ReLU occurrence-tagged vertex--pair readout can be uniformly approximated on a
compact domain by its normalized GELU encoder with the same counted
depth and two shared one-hidden-layer GELU readouts, when its affine stem
acts on the already marked features. For the before-mark affine interface
of \(\PMarch(\EVT)\), at most one additional setup macro-layer suffices
(\cref{app:marked-before-lift}). On a finite domain,
if the auxiliary encoder separates the selected marked diagonal
colors, every marked-profile-compatible target can be represented
exactly by a sufficiently accurate simulation of this encoder and
GELU occurrence-tagged readouts. This terminal interpolation requires no
encoder layer beyond those used for the simulation and any input setup.
\end{corollary}

\begin{proof}
Marked inputs associated with a compact domain form a compact set:
the marks are fixed on each discrete path component. With an affine stem
on those marked features, apply the encoder statement of
\cref{thm:ln-gelu-simulation}, which preserves its counted depth.
For the before-mark interface, first use the auxiliary setup in
\cref{app:marked-before-lift}; the same anchor-recovery argument then
simulates that setup and the original encoder in at most one additional
macro-layer. The first affine
map of each readout can extract the logical payload directly and retain
the unchanged occurrence tags, ignoring anchors and padding.
Let $R$ denote the two auxiliary ReLU readouts and $L_R$ a common Lipschitz constant
on a fixed compact box containing its reached tuples and a
radius-one neighborhood. Choose GELU MLPs $G$ with error at most
$\delta$ relative to $R$ throughout that box. If the encoder payload
error is at most $e<1/\sqrt3$ at each selected token, a pair-readout
tuple, which concatenates at most three token payloads, has error at
most $\sqrt3e<1$ and stays in that neighborhood; its tags are exact.
Comparing first $G$
with $R$ at the approximate tuple and then $R$ at the exact tuple
gives joint output error at most $\sqrt{K+K^2}(\delta+L_R e)$,
absorbing the fixed number of readout input tokens into $L_R$.
Both terms can be made arbitrarily small.

For finite exact outputs, simulate a color-separating encoder closely
enough to retain its distinct codes. By
\cref{lem:marked-rooted-recovery}, a selected marked diagonal color
determines the marked profile. Together with $\phi(s)$ and, for pairs,
$\phi(t)$, the finite readout inputs distinguish every profile/position
class whose target may differ. Equal classes have equal readout inputs by
marked stable-color factorization. Apply
\cref{lem:gelu-finite-interpolation} separately to the two readouts, so that
interpolation itself adds no encoder layer.
\end{proof}

The occurrence tag is essential for this prediction statement. For
$P=(1,1)$ on one vertex, an untagged shared vertex readout gives $(a,a)$
and cannot represent the compatible singleton-domain target $(0,1)$.
The mark records all visits to a vertex; $\phi(s)$ identifies which
visit receives the current prediction.

\paragraph{Scalar task-specific layers.}
The occurrence-output results also support masked-mean scalar task-specific layers.
For a continuous scalar profile-compatible target $f$, use the
occurrence target $h_{\mathcal{X}}(u)_s=(f(u),0,\ldots,0)$ with zero pair output and average its first
coordinate. Averaging does not amplify the approximation error.
A GELU MLP can perform an exact affine readout since
$g(t)-g(-t)=t$. These are choices of encoder and task-specific layer parameters at
variable capacity; application-specific fixed adapters and trained
finite budgets retain their stated specifications.

\subsection{Proof of the stable EVT--ET equivalence}
\label{app:evt-et-proof}

\begin{lemma}[Uniform joint stabilization]
\label{lem:joint-stabilization}
Consider any two graph inputs of sizes \(n,m\leq N\), and compute their
formal colors \(c^r\) and \(\widetilde c^r\) by
\cref{eq:2fwl-step}.  On the disjoint union
\(\Omega=[n]^2\sqcup[m]^2\), let \(\mathcal P_r\) be the partition whose
blocks are equality classes of depth-\(r\) colors.  Then
\(\mathcal P_{r+1}\) refines \(\mathcal P_r\), and
\begin{equation}
 \mathcal P_{R_N}=\mathcal P_{R_N+1},
 \qquad R_N=2N^2.
 \label{eq:joint-stabilization}
\end{equation}
Consequently, equality of two terminal colors, possibly from different
graphs, implies equality of their terminal multiset over intermediate vertices:
\begin{equation}
 c^{R_N}_{ij}=\widetilde c^{R_N}_{i'j'}
 \quad\Longrightarrow\quad
 \multiset{(c^{R_N}_{iu},c^{R_N}_{uj}):u\in[n]}
 =
 \multiset{(\widetilde c^{R_N}_{i'v},
             \widetilde c^{R_N}_{vj'}):v\in[m]}.
 \label{eq:terminal-bridge-well-defined}
\end{equation}
\end{lemma}
\begin{proof}
Because \(c^{r+1}_{ij}\) retains \(c^r_{ij}\) as its first component,
equality at round \(r+1\) implies equality at round \(r\); hence
\(\mathcal P_{r+1}\) refines \(\mathcal P_r\).  The set \(\Omega\) has
\(n^2+m^2\leq2N^2\) elements.  Every strict refinement increases the number
of blocks by at least one, so there can be at most
\(n^2+m^2-1\) strict refinements.  Once
\(\mathcal P_{r+1}=\mathcal P_r\), no later split is possible: on each
\(\mathcal P_r\)-block the round-\(r+1\) color is then constant, so replacing
every round-\(r\) color in a multiset over intermediate vertices by its round-\(r+1\) color is a
deterministic relabeling of partition classes.  Thus the partition remains
stationary.  It follows that it is stationary by round
\(n^2+m^2\), and therefore by the common round \(R_N=2N^2\).

If the colors on the left of
\cref{eq:terminal-bridge-well-defined} are equal, their positions lie in one
block of \(\mathcal P_{R_N}\).  Stationarity places them in one block of
\(\mathcal P_{R_N+1}\) as well.  Equality of the literal round-\(R_N+1\)
tuples then gives equality of their second components, which are exactly the
two displayed multisets over intermediate vertices.
\end{proof}

\begin{lemma}[Transpose closure and stable-color reconstruction]
\label{lem:stable-reconstruction}
For a terminal pair color \(a=c^\star_{ij}\), the color \(a\) determines its
reversed color \(c^\star_{ji}\), its terminal multiset over intermediate vertices, the two
endpoint diagonal colors, and the global multisets of terminal pair and
diagonal colors.
\end{lemma}
\begin{proof}
First define an involution on formal colors.  At initialization,
\begin{equation}
 \tau_0(y_{ij},y_{ji},x_i,x_j,\delta)
 :=(y_{ji},y_{ij},x_j,x_i,\delta).
 \label{eq:color-transpose-zero}
\end{equation}
If a round-\(r+1\) color is written as \((c,M)\), where
\(M=\multiset{(a,b)}\), set
\begin{equation}
 \tau_{r+1}(c,M)
 :=\left(\tau_r(c),
   \multiset{(\tau_r(b),\tau_r(a)):(a,b)\in M}\right).
 \label{eq:color-transpose-step}
\end{equation}
Induction in \cref{eq:2fwl-step} gives
\(c^r_{ji}=\tau_r(c^r_{ij})\): for each intermediate vertex, transposition
reverses both ordered pairs and exchanges their positions in the tuple.
Thus the terminal color determines its reversed color.

Let
\begin{equation}
 M(a):=\multiset{(c^\star_{ir},c^\star_{rj}):r\in[n]}.
 \label{eq:stable-bridge}
\end{equation}
By \cref{lem:joint-stabilization}, \(M(a)\) is well defined: the same formal
color in any other graph instance has the same multiset over intermediate vertices.  Its first
and second projections give the row and column color multisets.  More
explicitly, let \(\tau_\star:=\tau_{R_N}\),
\(\epsilon(\rightarrow)=0\), and \(\epsilon(\leftarrow)=1\).  The multiset
of paired colors read by direction pair \((\alpha,\beta)\) is
\begin{equation}
 M^{\alpha,\beta}(a)
 =\multiset{(
   \tau_\star^{\epsilon(\alpha)}(b_1),
   \tau_\star^{\epsilon(\beta)}(b_2)):
   (b_1,b_2)\in M(a)}.
 \label{eq:oriented-stable-bridge}
\end{equation}
Thus all four variants remain functions of one canonical multiset over
intermediate vertices, with the two colors in each tuple associated with the same vertex.
In \(M(a)\), exactly one tuple has a diagonal first component, which
identifies \(c^\star_{ii}\), and exactly one has a diagonal second component,
which identifies \(c^\star_{jj}\).  The global pair-color multiset is also
recovered: the first projection of \(M(c^\star_{ij})\) gives
\(\multiset{c^\star_{ir}:r\in[n]}\); for every recovered color
\(a=c^\star_{ir}\), the second projection of the well-defined \(M(a)\) is
\(\multiset{c^\star_{ur}:u\in[n]}\).  Their union with multiplicity over
\(r\) is the global pair-color multiset, and its diagonal submultiset is then
also known.
\end{proof}

Consequently, ordinary vertex attention, all four oriented triangular
variants, both cross-attentions, and both aligned MLPs are functions of the
receiver stable color and stable global multisets.  Indeed, the reversed pair
state at \((r,i)\) is a function of
\(c^\star_{ri}=\tau_{R_N}(c^\star_{ir})\); the same argument applies when
the other pair state read by the head is reversed.
RMSNorm, LayerNorm and GELU act only on these tokenwise or aligned local inputs,
so they preserve this factorization. Induction over layers proves
\begin{equation}
 x_i^\ell=a_\ell(c^\star_{ii}),
 \qquad y_{ij}^\ell=b_\ell(c^\star_{ij}).
 \label{eq:evt-factorization}
\end{equation}

For separation, work first with the auxiliary computations of
\cref{def:auxiliary-computations}. On a finite pair of inputs or finite set
\(S\), first make the
bidirectional initial color locally available without changing
\(\AlMLP\).  In EVT, one attention-disabled setup macro-layer uses the
original aligned pair MLP to store the forward atomic token
\begin{equation}
 a_{ij}:=(y_{ij},x_i,x_j,\one_{\{i=j\}},1)
 \label{eq:forward-atomic-token}
\end{equation}
in protected pair workspace.  Pair-only ET already has this token through
\cref{eq:endpoint-lift}.  In the next triangular layer, at receiver
\((i,j)\), a zero-logit \(\TriAtt^{\leftarrow,\rightarrow}\) head reads
the state at \((u,i)\) and multiplies it by the diagonal indicator from
\((u,j)\), retaining only the term with \(u=j\):
\begin{equation}
 g_{ij}=\frac1n\sum_{u=1}^n a_{ui}\one_{\{u=j\}}
       =\frac1n a_{ji}.
 \label{eq:reverse-copy-head}
\end{equation}
The Hadamard value in \cref{eq:tri-head} implements the displayed gate, and
the head writes into fresh workspace.  The following original aligned pair
MLP sees \((a_{ij},g_{ij})\).  Its last coordinate is \(1/n\), so finite-set
interpolation recovers \(a_{ji}\) and maps every
initial color occurring in \(S\) exactly to a linearly independent basis
code \(\eta_0(c)\).  The ET pairwise MLP performs the same step.  Uniformly on
compact sets, division by \(1/n\geq1/N\) is a continuous pointwise operation
and can be approximated arbitrarily well.  Thus reverse access is realized by
an oriented setup head without changing the pointwise pair interface.

At round \(r\), suppose the state carries the corresponding basis code
\(\eta_r(c)\).  In a triangular head choose zero queries and keys, and
projections
\(W_L,W_R\) such that
\begin{equation}
 [W_L\eta_r(a)]_{a'b'}=\one_{\{a'=a\}},\qquad
 [W_R\eta_r(b)]_{a'b'}=\one_{\{b'=b\}},\qquad
 W_L\eta_r(a)\odot W_R\eta_r(b)=e_{(a,b)}.
 \label{eq:pair-basis}
\end{equation}
Then the head returns
\begin{equation}
 h^r_{ij}=\frac1n\sum_{u=1}^{n}
 e_{(c^r_{iu},c^r_{uj})}.
 \label{eq:bridge-histogram}
\end{equation}
The normalized histogram is injective jointly over all \(n\leq N\).  Indeed,
let \(D(c)\in\{0,1\}\) denote the initial diagonal-bit component retained by
the color \(c\).  Then \(D(c^r_{iu})=1\) singles out exactly \(u=i\), so
\begin{equation}
 \lambda(h):=\sum_{(a,b):D(a)=1}h_{ab}=\frac1n,
 \qquad
 \operatorname{mult}(a,b)=\frac{h_{ab}}{\lambda(h)}.
 \label{eq:cardinality-anchor}
\end{equation}
A finite-set ReLU MLP maps \((\eta_r(c^r_{ij}),h^r_{ij})\) to
\(\eta_{r+1}(c^{r+1}_{ij})\).  One shared vector encoding handles the union of all
colors appearing in the chosen pair or finite set, across all its size
components.  The resulting network still uses one parameter set on the
whole bounded domain.  This proves separation for
the rooted ET pair family; diagonal selection gives the rooted vertex family,
while in EVT the first aligned pair update writes the code into pair
workspace. With the intervening cross-attentions disabled, the second
aligned update copies its diagonal instance into the vertex stream and
preserves the pair code. Thus the copy reads the updated pair state without
changing the macro-depth.
These auxiliary codes have a positive separation margin on the finite
comparison. Apply \cref{thm:ln-gelu-simulation} with error less than half
that margin: one normalized GELU encoder still distinguishes every
required pair. Together with \cref{eq:evt-factorization}, this proves the
kernel part of \cref{prop:evt-et}.

Let \(S\) be a compact rooted input set.  By the kernel equality, the
relation \(R:=\ker(\gamma|_S)\) is also an intersection of kernels of
continuous network maps, hence is closed.  The quotient \(Q=S/R\) is
therefore compact Hausdorff.  Auxiliary network coordinates descend to continuous
functions on \(Q\) and separate its points by their kernel equality. Let
\(\mathcal A\) be their scalar uniform closure.  By
\cref{lem:parallel-compilation}, constants belong to \(\mathcal A\), and any
two representatives can be evaluated exactly in separate coordinates at the same token before a
terminal MLP approximates their sum or product.  Hence \(\mathcal A\) is a
closed unital separating subalgebra of \(\mathcal C(Q)\).  Stone--Weierstrass gives
\(\mathcal A=\mathcal C(Q)\), and auxiliary parallel compilation handles the \(q\)
coordinates. Applying \cref{thm:ln-gelu-simulation} to each resulting
approximant gives \cref{eq:rooted-closure} for normalized GELU ET and EVT;
the opposite inclusion follows from continuity and factorization.
This quotient argument follows
the standard relative-universality viewpoint for invariant networks
\citep{keriven2019universal,maron2019universality,azizian2021}; the exact
parallel construction is supplied here by \cref{lem:parallel-compilation}.

\subsection{Depth comparison for auxiliary computations}
\label{app:finite-depth}

This subsection proves \cref{prop:finite-depth} for the auxiliary
computations of \cref{def:auxiliary-computations}. Exact copies and raw
scalar products refer to these proof computations. The normalized
architecture approximates every constructed computation at the same
macro-depth; no converse simulation at fixed depth is used or asserted.
Throughout this subsection, the affine endpoint lift is depth zero, every
numbered ET layer contains one triangular-attention residual and its
pairwise-MLP residual, and the reduced one-layer family follows exactly the
pair-update--then--\(Y\!\to X\) schedule stated before
\cref{prop:finite-depth}.  No unnumbered setup MLP is used.
Both lower-bound domains use symmetric adjacency tensors and the same
ordered endpoint tokenizer.  Their bidirectional atomic color is therefore a
deterministic relabeling of the forward color.  The finite-round version of
the transpose recursion
\cref{eq:color-transpose-zero,eq:color-transpose-step} then makes every
oriented multiset over intermediate vertices at round \(r\) a function of the canonical
compositional multiset over intermediate vertices.  By induction, an \(L\)-layer
direction-mixed ET state still factors through the round-\(L\) color used in
the lower bounds.  The reduced broom comparison has only one triangular
pair layer, for which raw symmetry already suffices.

\subsubsection{A three-layer upper bound for a reduced EVT readout}

Fix a reduced readout and one of its pair-to-vertex heads; finitely many heads
are evaluated in parallel using separate groups of state coordinates. Let \(z_{uv}\) denote the pair state
after the reduced triangular residual and pair MLP, and
\begin{equation}
 q_i=W_Qx_i,\qquad k_{uv}=W_Kz_{uv},\qquad v_{uv}=W_Vz_{uv},\qquad
 \kappa(q,k)=\exp(\langle q,k\rangle/\sqrt{d_a}).
 \label{eq:reduced-cross-coordinates}
\end{equation}
The exact cross-attention head is \(A_i=N_i/D_i\), where
\begin{equation}
 D_i=\frac1{n^2}\sum_{u,v}\kappa(q_i,k_{uv}),\qquad
 N_i=\frac1{n^2}\sum_{u,v}\kappa(q_i,k_{uv})v_{uv}.
 \label{eq:exact-global-softmax}
\end{equation}
The factor \(n^{-2}\) cancels in the quotient.

All queries, keys, and values reached from a compact rooted domain \(S\),
over its finitely many size components, lie in common compact sets
\(Q_S,K_S,V_S\).  Hence for any \(\eta>0\) there are continuous scalar
functions \(a_h,b_h\) such that
\begin{equation}
 \sup_{Q_S\times K_S}\left|
 \kappa(q,k)-\sum_{h=1}^Ha_h(q)b_h(k)\right|<\eta.
 \label{eq:separable-kernel}
\end{equation}
Define
\begin{equation}
 B_h=\frac1{n^2}\sum_{u,v}b_h(k_{uv}),\qquad
 C_h=\frac1{n^2}\sum_{u,v}b_h(k_{uv})v_{uv}.
 \label{eq:separable-moments}
\end{equation}
The following schedule computes approximations of these moments.  By
\cref{lem:parallel-compilation}, every quantity listed as protected occupies
fresh residual coordinates.  The triangular value is an elementwise product
of two projected pair states; choosing one projection to return the
constant-one channel passes the other projected state unchanged.

\begin{table}[H]
\centering
\small
\begin{tabularx}{\linewidth}{@{}>{\centering\arraybackslash}p{.07\linewidth}
 >{\raggedright\arraybackslash}p{.28\linewidth}
 >{\raggedright\arraybackslash}p{.37\linewidth}Y@{}}
\toprule
ET layer & Triangular heads & Pairwise MLP & Protected after the layer \\
\midrule
1 & Reproduce the reduced triangular pair residual in separate coordinates.
& Compose its exact pair update with ReLU approximants of
  \(b_h(k_{uv})\) and \(b_h(k_{uv})v_{uv}\) for every \(h\).
& \(z_{uv},k_{uv},v_{uv}\), endpoint data and \(q_u\), the root data
  \((x_u,z_{uu})\) on diagonal tokens, moments, and \(1\). \\
2 & With zero scores, read the moment coordinates from the state at
  \((u,r)\) and \(1\) from the state at \((r,j)\).  At every \((u,j)\) this gives the
  row means \(n^{-1}\sum_r b_h(k_{ur})\) and
  \(n^{-1}\sum_r b_h(k_{ur})v_{ur}\).
& Copy the row means into fresh channels; leave the query and root coordinates
  unchanged.
& Row moments, \(q_u\), endpoint data, and the diagonal root data. \\
3 & With zero scores, read \(1\) from \((i,r)\) and the row moments indexed
  by \(r\) from \((r,j)\).  Their averages are \(B_h,C_h\) at every
  \((i,j)\).
& At \((i,i)\), approximate the quotient in
  \cref{eq:separable-softmax} and the complete reduced vertex residual and
  readout.
& The designated diagonal output. \\
\bottomrule
\end{tabularx}
\caption{Three ET layers compiling the reduced pair-update--then--\(Y\!\to
X\) schedule.  ``Protected'' means that subsequent attention output
projections and residual MLP increments write zero into those coordinates.}
\label{tab:reduced-et-schedule}
\end{table}

The moment maps act on the updated pair state \(z_{uv}\), not on the
input to its pair MLP. The first auxiliary MLP therefore composes that
exact pair update with ReLU approximants of the moment maps, retaining
\(z_{uv}\) in identity channels. Finite parallel compilation and depth
padding realize their joint output. Auxiliary MLP depth is unrestricted
(\cref{def:auxiliary-computations}), so these compositions require no
additional counted ET layer. The last MLP sees \(q_i\), the
preserved reduced root state, and all global moments, and approximates
\begin{equation}
 \frac{\sum_h a_h(q_i)C_h}{\sum_h a_h(q_i)B_h}.
 \label{eq:separable-softmax}
\end{equation}
It can therefore also apply the reduced vertex MLP in the same pointwise
block.

For completeness, let \(M_V=\sup_{V_S}\|v\|\),
\(d_0=\min_{Q_S\times K_S}\kappa>0\), and
\(A_0=\sum_h\sup_{Q_S}|a_h|\).  If the first MLP approximates each stored
moment integrand within \(\delta\), the two normalized aggregations do not
amplify that error.  Consequently the final denominator and numerator differ
from \(D_i,N_i\) by at most \(\eta+A_0\delta\) and
\(M_V\eta+A_0\delta\), respectively, uniformly in \(n\leq N\) and in
\(S\).  Choose \(\eta+A_0\delta<d_0/2\); the approximate denominator is then
at least \(d_0/2\), so the quotient is uniformly stable.  Uniform
approximation of the final continuous pointwise map uses the remaining error
budget.  More explicitly, for a requested error \(\varepsilon\), first choose
\(\eta,\delta\) so that uniform continuity of the quotient and reduced
readout makes their propagated error below \(\varepsilon/2\), then choose the
last MLP approximation below \(\varepsilon/2\).  This proves
\cref{eq:reduced-evt-et3}.  Width and head count may
depend on \(S\) and the requested accuracy, but the three ET layers in
\cref{tab:reduced-et-schedule} do not; the result is neither an exact
parameter identity nor a compilation of a full EVT macro-layer.

\subsubsection{Vertex attention inside triangular attention}
\label{app:vertex-attention}

For one vertex head
\begin{equation}
 A_i(X)=\sum_t\operatorname{softmax}_t
 \left(\langle W_Qx_i,W_Kx_t\rangle/\sqrt{d_a}\right)W_Vx_t,
 \label{eq:vertex-head}
\end{equation}
the endpoint lift supplies \(x_i,x_t\) and a constant.  In the first
endpoint-simulation head choose
\begin{equation}
 q_{it}=W_Qx_i,\quad k_{tj}=W_Kx_t,\quad
 v^{\mathrm L}_{it}=1,\quad v^{\mathrm R}_{tj}=W_Vx_t.
 \label{eq:endpoint-simulation-head}
\end{equation}
Its output is \(A_i(X)\), independent of \(j\).  Reversing the query/key and
value roles means using a second head with
\begin{equation}
 q_{it}=W_Kx_t,\quad k_{tj}=W_Qx_j,\quad
 v^{\mathrm L}_{it}=W_Vx_t,\quad v^{\mathrm R}_{tj}=1.
 \label{eq:endpoint-simulation-head-reverse}
\end{equation}
Its output is \(A_j(X)\), independent of \(i\).  The pair-only ET therefore implements both
endpoint updates in one triangular layer, with two groups of heads but no depth depending on
\(n\).  Its standard execution nevertheless forms \(n^3\) scores indexed by
the two endpoints and an intermediate vertex, and
stores the endpoint results redundantly over \(n^2\) pairs, whereas native
vertex attention forms \(n^2\) scores and stores \(n\) vertex states.

\subsubsection{Sharpness and the reverse shallow gap}

Taking a scalar designated output, we use the same explicit binary encoding
in both witnesses.  For a simple undirected graph \(G\), set every vertex
feature to \(x_i=1\), set
\(y_{ij}=A^G_{ij}\), and retain the diagonal bit supplied by
\cref{eq:endpoint-lift}.  Thus diagonal, edge, and nonedge tokens have the
three distinct coordinate pairs \((A_{ij},\one_{\{i=j\}})=(0,1),(1,0),(0,0)\);
the attention heads below read these coordinates directly, so no setup layer
or finite-code MLP is hidden in the depth count.  We define \(\mathsf I(G,r)\)
for this input with designated vertex root \(r\).

For the first domain, let
\begin{equation}
 G_0=\{r\}\sqcup C_6,\qquad
 G_1=\{r\}\sqcup(C_3\sqcup C_3),\qquad
 S_\triangle=\{\mathsf I(G_0,r),\mathsf I(G_1,r)\}.
 \label{eq:triangle-witness}
\end{equation}
Both graphs have seven vertices.  A uniform triangular head followed by its
pair MLP computes
\begin{equation}
 h_{ij}=\frac1n\sum_u A_{iu}A_{uj},
 \qquad z_{ij}=A_{ij}h_{ij}.
 \label{eq:triangle-local}
\end{equation}
The pair MLP gates by the binary variable \(A_{ij}\) exactly: for
\(0\leq h\leq1\),
\(Ah=\operatorname{ReLU}(h)-\operatorname{ReLU}(h-A)\).  Uniform
pair-to-vertex attention then broadcasts, at every root,
\begin{equation}
 T_i(A)=\frac1{n^2}\sum_{i',j'}z_{i'j'}
       =\frac{\operatorname{tr}(A^3)}{n^3}.
 \label{eq:triangle-density}
\end{equation}
Uniform \(Y\!\to X\) attention gives \cref{eq:triangle-density}, hence
\(T|_{S_\triangle}\in
\mathcal V^{\mathcal{X},\mathrm{red},\mathrm{aux}}_{N,1}|_{S_\triangle}\).

For either graph and every nonroot \(u\), the one-round colors
\((c^1_{ru},c^1_{ur})\) depend only on the facts that \(r\) is isolated and
\(u\) has degree two.  The root term is also identical.  Therefore the
multiset over intermediate vertices defining \(c^2_{rr}\) agree, so
\(c^2_{rr}(G_0)=c^2_{rr}(G_1)\).  Induction over the two ET layers shows that
every diagonal two-layer ET readout factors through this color and hence
agrees on \(S_\triangle\).  Yet \(G_0\) has no triangle while \(G_1\) has
two, so \(T_r(G_0)=0\) and \(T_r(G_1)=12/7^3\).  Consequently
\begin{equation}
 \inf_{E\in\mathcal E^{\mathcal{X},\mathrm{aux}}_{N,2}}
 \|T-E\|_{\infty,S_\triangle}\geq \frac{6}{7^3}.
 \label{eq:triangle-gap}
\end{equation}

For the reverse domain, let the connected broom be
\(G_{\mathrm B}=(V_{\mathrm B},E_{\mathrm B})\), with
\(V_{\mathrm B}=\{a,b,c,v,w\}\) and
\begin{equation}
 E_{\mathrm B}=\{\{v,a\},\{v,b\},\{v,w\},\{w,c\}\}.
 \label{eq:broom-graph}
\end{equation}
Set
\(S_{\mathrm B}=\{\mathsf I(G_{\mathrm B},a),
\mathsf I(G_{\mathrm B},c)\}\).  The leaf roots
\(a,c\) have the same one-round diagonal color.  Moreover, the unrooted input
is the same in both elements of \(S_{\mathrm B}\), so its global multiset of
one-round pair colors is identical.  In any reduced readout, the pair phase
is a function of the one-round pair color; the constant vertex query then
turns global \(Y\!\to X\) attention into a function of that global multiset;
and the last vertex MLP additionally sees only the rooted self-pair
state.  Every member of \(\mathcal V^{\mathcal{X},\mathrm{red},\mathrm{aux}}_{N,1}\) therefore has
the same value at \(a\) and \(c\).

By contrast, the target
\begin{equation}
 F_i=\sum_jA_{ij}\deg(j)
 \label{eq:broom-target}
\end{equation}
belongs to \(\mathcal E^{\mathcal{X},\mathrm{aux}}_{N,2}|_{S_{\mathrm B}}\) under the stated depth
convention.  Indeed, a zero-score head in layer one stores
\(n^{-1}\sum_rA_{ur}=\deg(u)/n\) at every pair \((u,v)\).  At the diagonal
\((i,i)\), a zero-score layer-two head multiplies \(A_{ij}\) from the state
at \((i,j)\) by \(\deg(j)/n\) from the state at \((j,i)\), and averages
over \(j\), yielding \(F_i/n^2\).  Since \(n=5\) on \(S_{\mathrm B}\), the
same layer's final linear MLP multiplies by \(25\).  Thus no preliminary
coding layer is used.  Finally, \(F_a=\deg(v)=3\) and
\(F_c=\deg(w)=2\), and hence
\begin{equation}
 F|_{S_{\mathrm B}}\in
 \mathcal E^{\mathcal{X},\mathrm{aux}}_{N,2}|_{S_{\mathrm B}}
 \setminus\mathcal V^{\mathcal{X},\mathrm{red},\mathrm{aux}}_{N,1}|_{S_{\mathrm B}},
 \label{eq:broom-gap}
\end{equation}
proving the reverse non-inclusion.

The two finite domains therefore isolate opposite shallow advantages without
asserting an ordering between full macro-layer classes.  Together with
\cref{tab:reduced-et-schedule}, they prove the sharpness and incomparability
claims in \cref{prop:finite-depth}.

\subsection{Proof of the path-profile theorem}
\label{app:profile-proof}

We first prove the vertex-output statement at any finite output width
\(q\). In this subsection and the following vertex-output construction,
\(\mathcal H^{\mathrm v}_{N,K}\) and
\(\mathcal C_\Psi^{\mathrm v}(S)\) denote the corresponding family and
compatible targets in \(\bigsqcup_{k\leq K}\R^{k\times q}\).
We write \(\mathcal H^{\mathrm{aux,v}}_{N,K}\) for its auxiliary
counterpart with the same vertex-only output convention.
Joint outputs are treated in \cref{app:joint-output-extension}.

We first construct auxiliary codes showing that positioned occurrence
processing can recover the entire selected matrix. The construction uses
\cref{def:auxiliary-computations}; it will then be simulated by StrikeTwice.  Define the symbolic initial occurrence color
\begin{equation}
 d^0_{st}:=(\cstar_{p_sp_t},\phi(s),\phi(t),\one_{\{s=t\}}).
 \label{eq:occurrence-color}
\end{equation}
On a two-input comparison or finite set, a preparatory macro-layer disables
attention and uses the biased aligned pair MLP to write a basis code
\(\eta_0(d^0_{st})\) into pair workspace.  Subsequent triangular layers
realize the same symbolic \(2\)-FWL recursion on the occurrence set.

\begin{lemma}[Two-round recovery of the positioned selection]
\label{lem:rooted-recovery}
For injective \(\phi\), every color \(d^2_{st}\) determines the complete
indexed matrix \((d^0_{uv})_{u,v\in[k]}\) and the root indices \((s,t)\).
\end{lemma}
\begin{proof}
The multiset over intermediate vertices in \(d^2_{st}\) contains
\((d^1_{su},d^1_{ut})\) for every \(u\).  Each first component retains
\(\phi(u)\), so the multiset is indexed and yields every \(d^1_{su}\).
For fixed \(u\), the multiset over intermediate vertices inside \(d^1_{su}\) contains
\((d^0_{sv},d^0_{vu})\) for every \(v\).  Its retained \(\phi(v)\) indexes
these entries and recovers \(d^0_{vu}\) for all \(v\).  Varying \(u\)
recovers the full matrix.  Retained root tags recover \(s,t\).
\end{proof}

By \cref{prop:evt-et}, every graph-stage EVT vertex and pair state factors
through its rooted stable \(2\)-FWL color.  Hence equal \(\Psi_{N,K}\) makes
all selected first-stage states equal, and deterministic second-stage updates
preserve equality. Conversely, the auxiliary finite simultaneous realization
in \cref{app:evt-et-proof} supplies one graph-stage computation that assigns basis codes
to every stable graph color in the two compared inputs.  The selected
\(d^0\) arrays then differ exactly when the profiles differ.
The occurrence version of \cref{eq:cardinality-anchor}, using the occurrence
diagonal predicate, recovers \(k\) from a uniform triangular head.  Two
refinement layers realize the code of \cref{lem:rooted-recovery}.  The
aligned vertex MLP reads the self-pair state \(d^2_{ss}\), copies its
profile code into \(\widehat x_s\), and writes a separating scalar into a designated
vertex-output coordinate.  Different \(k\) already gives different output
components.

The same construction works with one parameter set over all bounded sizes:
the graph code uses the common terminal depth \(R_N=2N^2\) justified by
\cref{lem:joint-stabilization}, while the positioned occurrence processor uses
exactly the two refinements of \cref{lem:rooted-recovery}, uniformly for every
\(k\leq K\).  Finite-set MLPs are defined on the union of all codes occurring
in the comparison. No padding or explicit size embedding is needed.
At the graph stage, the diagonal predicate supplies the \(1/n\) anchor
used by this construction (\cref{eq:cardinality-anchor}). At the path stage,
the mean \(k^{-1}\sum_{s=1}^k\phi(s)\) of the one-hot position tags has
exactly \(k\) active coordinates, each equal to \(1/k\), and thus determines
\(k\) without the occurrence-diagonal predicate
(\cref{eq:positioned-broadcast}).
For each two-input comparison of equal length, the auxiliary outputs have a
positive separation margin. Simulate the complete cascade with less than
half this margin using \cref{thm:ln-gelu-simulation}. The resulting
StrikeTwice model separates the inputs. Together with the upper factorization,
this proves \(\Ker(\mathcal H^{\mathrm v}_{N,K})=\ker(\Psi_{N,K})\); the same identity
also holds for the auxiliary computations.

\subsection{Universality proof}
\label{app:universality-proof}

For finite \(S\), the compact-density statement strengthens to exact
interpolation:
\begin{equation}
 \{H|_S:H\in\mathcal H^{\mathrm v}_{N,K}\}=\mathcal C_\Psi^{\mathrm v}(S).
 \label{eq:finite-universality}
\end{equation}
Indeed, the auxiliary finite simultaneous realization from
\cref{app:evt-et-proof,lem:rooted-recovery} gives every occurrence an
injective code for \((\Psi_{N,K}(u),s)\) on \(S\). Simulate these codes
with an error below one third of their minimum nonzero separation when
there is more than one code, reserving fresh
zero output coordinates. By \cref{cor:unified-finite-interpolation}, one
additional attention-disabled path macro-layer interpolates the prescribed
rows exactly using a GELU MLP on the anchored normalized local tuples.
For a single code the target is constant and the output bias suffices.
Necessity follows from the kernel identity. Exact finite interpolation
therefore belongs to the same normalized GELU family; it does not assert
exact reproduction of the auxiliary circuit at the same depth.

For compact \(S\), construct an auxiliary vertex-output approximant first.
Fix \(f\in\mathcal C_\Psi^{\mathrm v}(S)\) and \(\varepsilon>0\).
For each length \(k\), set
\begin{equation}
 E_{\varepsilon,k}
 :=\{(u,v)\in S_k^2:
       \|f(u)-f(v)\|_{\mathrm F}\geq\varepsilon/2\},
 \qquad S_k:=S\cap\bigsqcup_{n\leq N}\mathcal D_{n,k}.
 \label{eq:separator-set}
\end{equation}
The finite union \(E_\varepsilon=\bigcup_kE_{\varepsilon,k}\) is compact and
disjoint from \(\Ker(\mathcal H^{\mathrm{aux,v}}_{N,K})\). The open separator sets
\(\{(u,v):H(u)\neq H(v)\}\), \(H\in\mathcal H^{\mathrm{aux,v}}_{N,K}\), cover it; select a
finite subcover \(H_1,\ldots,H_m\), and append the zero output \(H_0\) so
the joint array shape records length.  Let \(A=(H_0,\ldots,H_m)\).  Among inputs
with the same value of \(A\), the outputs of \(f\) differ by less than
\(\varepsilon/2\) in Frobenius norm.

The image \(A(S)\) is compact.  For each \(a\in A(S)\), choose a
representative \(u_a\in A^{-1}(a)\), and set
\begin{equation}
 U_a:=\{u\in S_{k(a)}:\|f(u)-f(u_a)\|_{\mathrm F}<\varepsilon/2\},
 \qquad
 V_a:=A(S)\setminus A(S\setminus U_a).
 \label{eq:fiber-neighborhood}
\end{equation}
Here \(k(a)\) is the length recorded by the joint array shape.
The bound on the variation of \(f\) within each preimage gives \(A^{-1}(a)\subset U_a\).  Since
\(S\setminus U_a\) is compact, \(V_a\) is an open neighborhood of \(a\),
and \(A^{-1}(V_a)\subset U_a\). On each open-and-closed length component
\(A(S_k)\), choose a finite partition of unity subordinate to selected
\(V_{a_j}\) with \(k(a_j)=k\). These partitions define the continuous,
length-preserving map
\(R(b)=\sum_{j:k(a_j)=k(b)}\rho_j(b)f(u_{a_j})\) on \(A(S)\).
Every sum thus combines matrices of the same size, and
\begin{equation}
 \|f-R\circ A\|_{\infty,S}<\varepsilon/2.
 \label{eq:approx-factor}
\end{equation}
By \cref{lem:parallel-compilation}, one wider auxiliary cascade evaluates the finite
computations \(H_j\) exactly in disjoint groups of coordinates after padding the graph and
path depths independently. Define
\(A_s=(H_0(u)_s,\ldots,H_m(u)_s)\).  Compactness gives an \(M\geq1\) such
that every coordinate \(z\) of every \(A_s\), over \(u\in S\), satisfies
\(|z|\leq M\).  For \(g\in\{0,1\}\), the biased ReLU gate
\begin{equation}
 G_M(z,g):=\operatorname{ReLU}(z+Mg)
          -\operatorname{ReLU}(z-Mg)-Mg
 \label{eq:exact-binary-gate}
\end{equation}
satisfies \(G_M(z,0)=0\) and \(G_M(z,1)=z\) throughout this interval.  A
terminal path MLP applies this gate coordinatewise with
\(g=[\phi(s)]_t\), thereby placing \((1,A_s)\) exactly in block \(t=s\) of
a fixed \(K\)-block workspace and zero in every other block.  No product
primitive is being assumed.  A zero-query, zero-key vertex-attention head
then broadcasts
\begin{equation}
 B_k(A):=\frac1k\sum_{s=1}^k \phi(s)\otimes(1,A_s)
 \label{eq:positioned-broadcast}
\end{equation}
to every occurrence.  The leading occupancy coordinate in each block
identifies the active positions and the factor \(1/k\), so \(B_k\) is an
injective continuous embedding of the finite disjoint union of joint output
arrays: its nonzero blocks recover \(k\), and division of each data block by
its occupancy recovers \(A_s\).  A shared terminal MLP sees
\((B_k(A),\phi(s))\) and uniformly
approximates \(R(A)_s\).  More precisely, on the compact row-code set
\begin{equation}
 \mathcal B:=\{(B_{k(u)}(A(u)),\phi(s)):u\in S,\ s\in[k(u)]\},
 \label{eq:row-code-set}
\end{equation}
define \(g(B_{k(u)}(A(u)),\phi(s)):=R(A(u))_s\).  Injectivity of \(B_k\)
makes \(g\) well defined and continuous.  Extend its coordinates by Tietze
to a compact Euclidean neighborhood of \(\mathcal B\), and approximate that
extension by the terminal MLP with Euclidean error below
\(\varepsilon/(2\sqrt K)\) per output row. The resulting matrix error is
at most \(\sqrt{k(u)}\,\varepsilon/(2\sqrt K)\leq\varepsilon/2\)
in Frobenius norm. Together with \cref{eq:approx-factor}, the total error
is below \(\varepsilon\) for the auxiliary computation. Using this
construction with tolerance \(\varepsilon/2\) and then
\cref{thm:ln-gelu-simulation} with tolerance \(\varepsilon/2\) gives the
required StrikeTwice approximant. Every StrikeTwice map is continuous,
length preserving, and profile compatible, as are its uniform limits.
This gives the reverse inclusion for vertex outputs.

\paragraph{Joint vertex and pair outputs.}
\label{app:joint-output-extension}
Fix \(f=(f_{\mathcal{X}},f_{\mathcal{Y}})\in\mathcal C_\Psi(S)\) and
\(\varepsilon>0\). At each occurrence \(s\), let \(G(u)_s\) be the
vector consisting of \(f_{\mathcal{X}}(u)_s\) followed by the complete row
\((f_{\mathcal{Y}}(u)_{st})_{t=1}^{k(u)}\), padded by zeros to \(K\) pair entries.
This defines a continuous, profile-compatible vertex target of width
\(d_{\mathcal{X}}^{\mathrm{out}}+Kd_{\mathcal{Y}}^{\mathrm{out}}\): the size components are open and closed.
Apply the auxiliary vertex-output construction above to obtain
\(\widetilde G\) with uniform Frobenius error below \(\varepsilon/2\),
retaining the output rows and position tags in internal state coordinates.
A terminal attention-disabled aligned MLP copies the first
\(d_{\mathcal{X}}^{\mathrm{out}}\) coordinates of \(\widetilde G(u)_s\) into
designated vertex coordinates. Its pair branch reads the remaining
row blocks held at \(s\) and the position tag \(\phi(t)\) held at \(t\), and
selects the \(t\)-th block into designated pair coordinates. On the
bounded states the gates of \cref{eq:exact-binary-gate} implement this
selection exactly for auxiliary computations. Both branches read the
same pre-update states; their internal widths remain unchanged. Only the
subsequent shared affine output interfaces select the prescribed
prediction coordinates, so the final vertex width does not restrict
the pair readout. Selecting these disjoint blocks discards only padding, so
the joint error remains below \(\varepsilon/2\). Apply the simulation
theorem with tolerance \(\varepsilon/2\) to obtain the desired normalized
joint-output approximant.

The anchored finite-interpolation argument of
\cref{app:unified-simulation} also applies to both streams. Finite targets can
be interpolated exactly by the two independent terminal GELU MLPs on
the finitely many anchored local tuples. Conversely, every pair output
still factors through the selected profile. The vertex-output separating
constructions remain available with zero pair outputs. Consequently both
the kernel identity and compact universality hold for the joint-output
family, completing \cref{thm:main-expressivity}. This extension uses no
path triangular attention beyond what the vertex construction requires,
so it also applies to \cref{cor:path-decoder-kernel}.

\subsection{A positioned path processor without path triangular attention}
\label{app:path-decoder}

Let \(\mathcal H^{\mathrm{proc}}_{N,K}\subseteq\mathcal H_{N,K}\) be the
subfamily in which every triangular-attention increment in the path stage is
identically zero. All other operations and conventions remain those of
\cref{def:striketwice-family}.

\begin{corollary}[Profile-preserving positioned path processor]
\label{cor:path-decoder-kernel}
On the full bounded-size domain,
\begin{equation}
 \Ker(\mathcal H^{\mathrm{proc}}_{N,K})=\ker(\Psi_{N,K}).
 \label{eq:path-decoder-kernel}
\end{equation}
For every compact \(S\subset\mathcal D_{\leq N,\leq K}\),
\begin{equation}
 \overline{\{H|_S:H\in\mathcal H^{\mathrm{proc}}_{N,K}\}}
 ^{\|\cdot\|_{\infty,S}}
 =\mathcal C_\Psi(S).
 \label{eq:path-decoder-closure}
\end{equation}
For finite \(S\), every target in \(\mathcal C_\Psi(S)\) can be
interpolated exactly with at most two path-stage macro-layers after a
suitable graph-stage encoder.
\end{corollary}

\begin{proof}
Fix a finite set \(S\). Work first with auxiliary computations. The
simultaneous realization in \cref{app:evt-et-proof} supplies one
graph-stage computation carrying a distinct basis code \(\eta(c)\in\R^d\) for every terminal graph color occurring in
\(S\).  An injective choice of the affine selector preserves these codes together with the
ordered tags \(\phi(s),\phi(t)\).  With within-stream attention disabled, the first
aligned pair MLP of a path macro-layer writes into fresh workspace the
\(K^2\)-block vector
\begin{equation}
 U_{st}=\phi(s)\otimes\phi(t)\otimes
       \bigl(1,\eta(\cstar_{p_sp_t})\bigr).
 \label{eq:path-decoder-blocks}
\end{equation}
This is a prescribed map on finitely many distinct local codes and tags,
so \cref{lem:parallel-compilation} realizes it exactly by a shared ReLU
MLP.  All unused residual outputs are zero.

The following global \(Y\!\to X\) attention in the same macro-layer reads
these half-step pair states, as specified in \cref{eq:y-to-x}.  Set its
queries and keys to zero and let its value projection select \(U_{st}\).
Writing into fresh vertex coordinates, it broadcasts to every occurrence
\begin{equation}
 B=\frac1{k^2}\sum_{s,t=1}^k
       \phi(s)\otimes\phi(t)\otimes
       \bigl(1,\eta(\cstar_{p_sp_t})\bigr).
 \label{eq:indexed-pair-broadcast}
\end{equation}
Each active block has occupancy \(1/k^2\geq1/K^2\), while inactive blocks
are zero.  Thus \(B\) determines \(k\) and the complete indexed matrix of
codes.  Each terminal color also determines \(n\), since it retains the
cardinality of its first-round multiset over intermediate vertices.  Hence \(B\) determines
\(\Psi_{N,K}\) on this finite set.  The last aligned vertex MLP receives
\(B\) and the protected root tag \(\phi(s)\); finite interpolation writes
any prescribed compatible vertex row into the output coordinates. The
pair branch reads both positioned endpoints after this same broadcast,
and hence can also interpolate the corresponding pair target.
No path triangular-attention head is used in this auxiliary construction.
It also supplies a terminal code for \((\Psi_{N,K}(u),s)\), by retaining
\(B\) and \(\phi(s)\). Simulate this one-path-layer code construction
using \cref{thm:ln-gelu-simulation}, which keeps the path triangular
branches zero. Apply \cref{cor:unified-finite-interpolation} to add one
attention-disabled GELU interpolation block. This proves exact finite
interpolation with two path macro-layers in StrikeTwice.

For the kernel equality, a sufficiently accurate simulation of the
auxiliary separator on two inputs already separates unequal profiles of the
same length.  Different lengths already give different output components.
The reverse kernel inclusion follows from
\(\mathcal H^{\mathrm{proc}}_{N,K}\subseteq\mathcal H_{N,K}\) and
\cref{thm:main-expressivity}, proving \cref{eq:path-decoder-kernel}.

For compact \(S\), carry out the compact construction of
\cref{app:universality-proof} on auxiliary computations with zero path
triangular increments. Their kernel identity supplies a finite set of
auxiliary separators for \cref{eq:separator-set}; the
construction of neighborhoods of the preimages, followed by a partition of unity, gives
\cref{eq:approx-factor}.  Constants, identity padding, and exact parallel
channel compilation preserve the condition that every path triangular
increment is zero.  Finally, the positioned readout in
\cref{eq:exact-binary-gate,eq:positioned-broadcast,eq:row-code-set} uses only
path MLPs and vertex self-attention, which remain available.  It therefore
produces auxiliary vertex-output approximants with no path triangular
attention. Apply the row-packing construction of
\cref{app:joint-output-extension}; its extra aligned MLP also uses no
triangular attention, yielding the joint-output approximants.
Apply \cref{thm:ln-gelu-simulation}, retaining these zero branches, to
obtain arbitrarily accurate approximations within
\(\mathcal H^{\mathrm{proc}}_{N,K}\). The opposite closure inclusion
follows from the subfamily inclusion and \cref{thm:main-expressivity}.
\end{proof}

The two-macro-layer bound is a sufficient finite-set construction after a
suitable graph encoder. Its first path layer approximately reproduces
auxiliary profile codes; its second writes the exact finite outputs.
The compact construction may require additional path layers to combine
features and apply the terminal readout. No minimal-depth claim is made.

\paragraph{Auxiliary width and experimental budgets.}
The explicit auxiliary state in \cref{eq:path-decoder-blocks} has
\(K^2(1+d)\) coordinates, where \(d\) is the dimension of the color code
\(\eta\) chosen for the finite set \(S\), rather than the experimental hidden width.
For illustration, setting \(K=7\) and \(d=32\) gives
\(7^2(1+32)=1{,}617\) auxiliary coordinates, about \(50.5\) times width 32.
This counts only the displayed workspace, in addition to retained states and
tags; the interpolation MLPs also require capacity.
It is a sufficient allocation for this construction, not a minimal-width
lower bound for all positioned path processors.
The evaluated Task 1 \(\mathrm{ST}[\EVT,\EVT]\) and \(\mathrm{ST}[\EVT,\EVT_{\mathrm D}]\) models use width 32, FFN widths 128 and 130,
and 312,753 and 312,457 parameters, respectively
(\cref{app:t1-mechanism}); they do not instantiate this auxiliary workspace.

\section{Query Marking and Reusable Selection}
\label{app:marked-model}
\label{app:access-profiles}

We compare \(\PMarch(\EVT)\) with the reusable StrikeTwice family
\(\mathcal H_{N,K}\), using the same terminal depth \(R_N=2N^2\).
Occurrence marks and marked initial colors are
\begin{align}
 m_P(i)&:=\sum_{s\in[k]:p_s=i}\phi(s)\in\R^K,
 \label{eq:path-mark}\\
 \bar c^0_{ij}&=(y_{ij},y_{ji},x_i,x_j,\one_{\{i=j\}},m_P(i),m_P(j)).
 \label{eq:marked-init}
\end{align}
Write \(\bar c^\star_{ij}:=\bar c^{R_N}_{ij}\), and define
\begin{equation}
 \Psi^{\mathrm{mark}}_{N,K}(X,Y,P)
 :=\left(n,k,(\bar c^\star_{p_sp_s})_{s\in[k]}\right).
 \label{eq:marked-profile}
\end{equation}
For \(S\subset\mathcal D_{\leq N,\leq K}\), its compatible class is
\begin{equation}
 \mathcal C_{\Psi^{\mathrm{mark}}}(S)
 :=\{f\in\mathcal C(S;\mathcal O_K):
 f\text{ is length preserving, }\ker(\Psi^{\mathrm{mark}}_{N,K}|_S)
 \subseteq\ker(f)\}.
 \label{eq:marked-compatible-functions}
\end{equation}

The family \(\mathcal H^{\mathrm{mark}}_{N,K}\) surrounds the PathMarked
core with the same outer affine maps as \cref{def:striketwice-family}.
Its input map lifts raw features to widths
\(d_{\mathcal{X}},d_{\mathcal{Y}}\) before marking; the marked EVT core
preserves widths \(d_{\mathcal{X}}+K,d_{\mathcal{Y}}+2K\).
To state the information profile independently of this learned lift, use
the raw marked inputs
\(X^{\mathrm{mark}}=((x_i\concat m_P(i)))_{i\in[n]}\) and
\(Y^{\mathrm{mark}}=((y_{ij}\concat m_P(i)\concat m_P(j)))_{ij}\).
The duplicated endpoint marks carry no information beyond \(\bar c^0_{ij}\).
Thus EVT initialization on these raw marked inputs and \cref{eq:marked-init} differ only
by a fixed injective recoding, as do their subsequent profiles; their
equality classes and kernels coincide.
The residual stack has no terminal normalization or separate terminal
graph projection. A learned input lift may discard information; the
family also includes injective lifts, so this does not change the common
kernel. The auxiliary proof below may use affine lifts after raw marking;
the implementation of those constructions with the prescribed before-mark
outer lift is justified immediately afterward.
The comparison family \(\mathcal H^{\mathrm{mark,ET}}_{N,K}\) uses the
endpoint-initialized, direction-mixed ET of \cref{app:evt-et-comparison},
with both endpoint marks included in its stem.
Both use the chosen pre-normalization, biased GELU MLPs, arbitrary finite widths and
depths, and shared biased GELU readouts. The vertex map reads the selected vertex
state (or ET diagonal state) and occurrence tag \(\phi(s)\); the pair map
reads the selected pair and both selected endpoint states, together with
\(\phi(s),\phi(t)\) and the repeated-vertex bit supplied by \(\Sub_P\).
The core readouts produce free widths \(q_{\mathcal{X}},q_{\mathcal{Y}}\);
the outer output map gives the prescribed widths
\(d_{\mathcal{X}}^{\mathrm{out}},d_{\mathcal{Y}}^{\mathrm{out}}\).
Encoder and readout parameters are shared over all
\(n\leq N\) and \(k\leq K\). Graph states depend on \(P\).
Each terminal two-layer MLP ends with a biased affine map to its output
width and uses neither normalization nor residual addition.
The outer output map can be absorbed into these MLPs' final affine maps.

\paragraph{Lifting before marking.}
\label{app:marked-before-lift}
The placement of the input lift does not alter the variable-capacity
kernel or closure. Choose a sufficiently large physical width \(D\),
set \(d_{\mathcal{X}}=D-K\) and \(d_{\mathcal{Y}}=D-2K\), and let the
raw affine lifts store the raw features, zero workspace, and constant
anchors \(0,A,-A\). After marking, both streams have width \(D\).
Their tokens are fixed coordinate permutations of the anchored states
\(E_A(s)\) from \cref{eq:unified-encoding}, with the unscaled marks
included in the logical payload \(s\). With unit gains, RMSNorm and
LayerNorm commute with these coordinate permutations; permuting the
linear recovery maps in \cref{lem:unified-anchors,lem:unified-rms-anchors}
therefore recovers raw features and marks with the same uniform
\(O(A^{-2})\) error on compact sets. Blockwise coordinate permutations
give the same statement for concatenated aligned-MLP inputs.

At the auxiliary level, one attention-disabled setup macro-layer writes
any required affine function of the marked raw tokens into fresh zero
workspace, using the increment ``desired state minus current state.''
Subsequent blocks run the original auxiliary construction in that
workspace and preserve the reference features, marks, and anchors by
zero residual output rows. The normalization simulation reproduces this
construction to arbitrary compact accuracy, and the final tagged MLPs
ignore the reference coordinates. Distinct finite color codes remain
separated under sufficiently accurate simulation, after which the
unnormalized terminal GELU readouts interpolate the desired outputs.
This justifies the proof constructions below without a learned lift
after \(\Mark_P\). A post-mark-lift auxiliary encoder of depth \(L\)
therefore has a before-mark-lift simulator of depth at most \(L+1\),
with additional finite width. This is not a fixed-depth equivalence or an identification
with an archived implementation.

\begin{lemma}[Marked recovery and forgetting]
\label{lem:marked-rooted-recovery}
A terminal marked color \(\bar c^\star_{ij}\) determines
\[
 \bigl(\bar c^\star_{ip_t},\bar c^\star_{p_tj}\bigr)_{t\in[k]}.
\]
A selected diagonal therefore determines \(n,k\), its indexed row of
selected pair colors, and all selected diagonal colors, including
repeated occurrences. At every depth, the unmarked color is a function
of the marked color.
\end{lemma}
\begin{proof}
By \cref{lem:joint-stabilization}, applied to marked inputs,
\(\bar c^\star_{ij}\) determines
\(\multiset{(\bar c^\star_{iv},\bar c^\star_{vj}):v\in[n]}\).
Exactly one entry has a nonzero \(t\)-th coordinate in the intermediate
vertex's mark: \(v=p_t\). Its two components give the claimed colors.
For \(i=j=p_s\), each recovered \(\bar c^\star_{p_sp_t}\) determines
its endpoint diagonals by \cref{lem:stable-reconstruction}.
Applying this recovery to those diagonals gives the complete ordered
matrix of selected marked pair colors.
Repeated occurrences select the same vertex; the external tag
\(\phi(s)\) distinguishes their output positions.
The first-round multiset cardinality records \(n\), and the union of
intermediate mark supports is \([k]\), recording \(k\).
Finally, dropping marks defines a map \(f_0\); applying \(f_r\) to the
receiver and both components of each multiset entry defines \(f_{r+1}\).
Induction gives \(c^r_{ij}=f_r(\bar c^r_{ij})\).
The same construction forgets any chosen subset of mark coordinates.
\end{proof}

\subsection{Neural characterization of the marked profile}
\label{app:marked-approximation}

\begin{corollary}[Marked kernel and relative universality]
\label{prop:marked-strict}
For \(N,K\geq1\),
\begin{equation}
 \Ker(\mathcal H^{\mathrm{mark}}_{N,K})
 =\Ker(\mathcal H^{\mathrm{mark,ET}}_{N,K})
 =\ker(\Psi^{\mathrm{mark}}_{N,K}).
 \label{eq:marked-strict}
\end{equation}
For every compact \(S\subset\mathcal D_{\leq N,\leq K}\), both marked
families have uniform closure \(\mathcal C_{\Psi^{\mathrm{mark}}}(S)\).
For finite \(S\), both realize every member of this class exactly.
\end{corollary}
\begin{proof}
\textbf{Soundness and finite realization.}
Rooted-color factorization from \cref{prop:evt-et}, applied to the
marked features, shows that selected vertex states factor through marked
diagonal colors and selected pair states through marked pair colors.
\Cref{lem:marked-rooted-recovery} recovers all these colors from the profile.
RMSNorm, LayerNorm and GELU are local shared maps,
so the tagged readout gives equal outputs on equal marked profiles.
Conversely, the finite simultaneous color construction in
\cref{app:evt-et-proof} produces auxiliary ReLU computations assigning
distinct codes to the finitely many marked diagonal colors.
By \cref{lem:marked-rooted-recovery}, each selected diagonal code
recovers the complete marked profile.
Apply \cref{thm:ln-gelu-simulation} to these auxiliary encoders on the
finite marked input set. When there are distinct codes, choose the
error smaller than one third of their minimum distance; the one-code
case needs no separation margin. This gives normalized,
GELU encoders whose selected states still distinguish every pair of
different marked profiles. Appending \(\phi(s)\) also distinguishes
output occurrences, including repeated visits to the same vertex.
The two shared GELU readouts interpolate the prescribed vertex and pair
outputs exactly on these finite sets of tagged local states by
\cref{lem:gelu-finite-interpolation}. For a pair output, the first endpoint
already identifies the whole profile, and the two tags identify \((s,t)\).
This proves finite interpolation for both families; applying the same
argument to a two-input set gives the kernel identity. The interpolation
adds no encoder layer beyond the simulation and possible input setup
above, since the tagged readout is already part of each model.

\textbf{Compact approximation.}
Fix compact \(S\), \(f\in\mathcal C_{\Psi^{\mathrm{mark}}}(S)\), and
\(\varepsilon>0\). For each \(s\in[K]\), let
\(S_s:=\{u\in S:k(u)\geq s\}\); this set is compact because length
components are open and closed in the finite disjoint-union topology.
Discard empty \(S_s\), and map each remaining one to rooted marked inputs:
\begin{equation}
 T_s(X,Y,P):=(X^{\mathrm{mark}},Y^{\mathrm{mark}};p_s).
 \label{eq:marked-root-embedding}
\end{equation}
The map \(T_s\) is continuous and injective: the marks recover every
position of \(P\), including repeats, while the original features are
retained. It is thus a homeomorphism onto its compact image.
Let \(G(u)_s\in\R^{d_{\mathcal{X}}^{\mathrm{out}}+Kd_{\mathcal{Y}}^{\mathrm{out}}}\) concatenate \(f_{\mathcal{X}}(u)_s\)
and the row \((f_{\mathcal{Y}}(u)_{st})_{t\leq k(u)}\), zero-padded to \(K\) entries.
Define \(f_s(T_s(u)):=G(u)_s\).
This is a continuous rooted prediction. Equal marked diagonal colors
recover equal complete marked profiles by
\cref{lem:marked-rooted-recovery}, so \(f_s\) is compatible with the
rooted color on \(T_s(S_s)\).

Use the auxiliary rooted approximants constructed in the proof of
\cref{prop:evt-et} to approximate each \(f_s\) with row error below
\(\varepsilon/(2\sqrt K)\). Compile these at most \(K\) auxiliary
encoders in parallel, padding their depths with identity layers
(\cref{lem:parallel-compilation}). They all read the same marked graph;
each encoder's output coordinates are available at every selected root.
On the resulting compact set, the binary ReLU gate of
\cref{eq:exact-binary-gate}, controlled by \([\phi(s)]_a\), selects
the output group indexed by \(a=s\). This defines one auxiliary
encoder with a shared occurrence-tagged ReLU readout for \(G\).
Its vertex map selects the first \(d_{\mathcal{X}}^{\mathrm{out}}\) coordinates; its pair map
uses the tag \(\phi(t)\) to select the \(t\)-th pair block from row
\(s\), using the same exact bounded ReLU gates. Denote these two shared
readouts collectively by \(R\). Their joint array error is below
\(\varepsilon/2\), since removing padding does not increase the norm.

Apply \cref{thm:ln-gelu-simulation} to the combined auxiliary encoder
on the compact set of marked inputs, and approximate \(R\) by
two shared GELU MLPs on fixed compact neighborhoods of their tagged local states
(\cref{app:unified-simulation}). These maps read the reproduced logical
coordinates and ignore the simulator's auxiliary anchors and padding. If the encoder error at each selected
state is at most \(e\), the per-output MLP approximation error is at most
\(\delta\), and \(L_R\) is a Lipschitz constant of \(R\), the additional
array error is at most
\(\sqrt{K+K^2}(\delta+L_Re)\), absorbing the fixed number of input
states into \(L_R\). Choosing \(e,\delta>0\) so that this is
below \(\varepsilon/2\) gives one normalized GELU model whose uniform
error is below \(\varepsilon\). Its parameters are shared over every
size component of \(S\). The exact parallel compilation was used only
for auxiliary computations; normalization is allowed to mix all coordinates
in the resulting model.
Uniform limits remain continuous, length preserving, and constant on
marked profile classes, proving the reverse closure inclusion.
\end{proof}

\subsection{Proof of the query-marking comparison}
\label{app:pathmarked-hierarchy-proof}

\begin{proof}[Proof of \cref{thm:pathmarked-expressivity}]
By \cref{lem:marked-rooted-recovery}, the marked profile determines all
selected marked pair colors. Forgetting marks recovers
\((\cstar_{p_sp_t})_{s,t\in[k]}\), while \(n,k\) are already included.
Thus \(\Psi_{N,K}=r\circ\Psi^{\mathrm{mark}}_{N,K}\) for a recovery map
\(r\), and
\(\ker(\Psi^{\mathrm{mark}}_{N,K})\subseteq\ker(\Psi_{N,K})\).
The neural kernel identities follow from \cref{prop:marked-strict} and
\cref{thm:main-expressivity}.
For every \(S\), a continuous length-preserving function constant on
selected profile classes is also constant on marked profile classes.
Hence \(\mathcal C_\Psi(S)\subseteq\mathcal C_{\Psi^{\mathrm{mark}}}(S)\);
this step does not require continuity of \(r\).
For compact \(S\), the neural closure identities are supplied by
\cref{prop:marked-strict,thm:main-expressivity}.

\medskip
\noindent\textbf{EVT with marks only at the vertex inputs.}
Consider the same EVT family and tagged readouts with marks appended only
to vertices. On the resulting raw features \((X^{\mathrm{mark}},Y)\),
the initial \(2\)-FWL color contains the two endpoint marks through
\(x_i\concat m_P(i)\) and \(x_j\concat m_P(j)\). Regrouping its entries
gives \cref{eq:marked-init}, so the stable profile has the same equality
classes. The rooted EVT factorization and
\cref{lem:marked-rooted-recovery} therefore give the same upper bound
on distinguishing power and uniform approximation.

For the converse, apply the rooted EVT construction to
\((X^{\mathrm{mark}},Y;p_s)\). In its first setup layer, the attentions are
disabled and the aligned pair MLP reads both marked endpoint states
(\cref{eq:almlp}). It writes the atomic token of
\cref{eq:forward-atomic-token}, now including \(m_P(i)\) and \(m_P(j)\),
into reserved pair coordinates. This copy is exact in the auxiliary
computation and makes both marks available to subsequent triangular
layers.

For the normalized GELU simulator, choose \(d_{\mathcal X}=D-K\) and
\(d_{\mathcal Y}=D\) before marking, with \(D\) large enough for the
anchors and workspace. As in \cref{app:marked-before-lift}, any affine
lift of the marked inputs required by an auxiliary encoder can be
written into workspace by an attention-disabled setup layer.
\Cref{thm:ln-gelu-simulation} approximates these setup operations
and the subsequent encoder jointly and uniformly on compact sets.
The finite separation margins and the compact approximation argument
with occurrence-tagged readouts in \cref{app:marked-approximation}
then give the same kernel and uniform closure. The comparison with
StrikeTwice, including strictness, follows from these equalities.
\end{proof}

\subsection{Smaller strict witnesses}
\label{app:marked-small-witnesses}

The Latin-square construction in \cref{sec:cache-access} connects
strictness to the certified dataset. Smaller witnesses also give
strictness for \(N\geq16\), even with a single marked position.
On \(\mathbb Z_4^2\), the rook graph \(R\) joins distinct vertices sharing
a coordinate; the Shrikhande graph \(S\) has adjacency differences
\(\{\pm(1,0),\pm(0,1),\pm(1,1)\}\), modulo four.
Use the same constant vertex features and binary adjacency in both.
Their strongly regular parameters \((16,6,2,2)\) give the same three
unmarked colors at every round: diagonal \(\mathrm D\), adjacent
\(\mathrm A\), and nonadjacent \(\mathrm N\).

\paragraph{One marked position.}
Take \(P=(r)\), \(r=(0,0)\), so the selected profiles agree.
Write \(b_u=A_{ru}\) and \(B_{uv}=b_uA_{uv}b_v\).
The neighbors of \(r\) induce two disjoint triangles in \(R\) and a
six-cycle in \(S\), giving \(\operatorname{tr}(B_R^3)=12\) and
\(\operatorname{tr}(B_S^3)=0\).
The marked intermediate vertex makes \(B_{uv}\) a function of
\(\bar c^1_{uv}\). Then \(\bar c^2_{uv}\) determines
\(h_{uv}=B_{uv}\sum_w B_{uw}B_{wv}\), \(\bar c^3_{ru}\) determines
\(\sum_v h_{vu}\), and \(\bar c^4_{rr}\) determines
\(\sum_{u,v}h_{vu}=\operatorname{tr}(B^3)\).
Thus the marked diagonals differ.
This single-mark gain is also reported by
\citet[Appendix~G, Table~7]{zhou2023relational}.

\paragraph{Three-position simple paths.}
Choose the triangles
\begin{equation}
 P_R=((0,0),(0,1),(0,2)),\qquad
 P_S=((0,0),(0,1),(1,1)).
 \label{eq:marked-witness}
\end{equation}
Their selected profiles agree, whereas their common-neighbor intersections
are \(\{(0,3)\}\) and \(\varnothing\), respectively.
Marked refinement recovers the three adjacency indicators to each vertex
in one round and counts joint neighbors in the next, separating the
marked profiles exactly as in \cref{app:latin-marked-proof}.
The open paths
\begin{equation}
 Q_R=((0,1),(0,0),(1,0)),\qquad
 Q_S=((0,1),(0,0),(0,3))
 \label{eq:marked-open-witness}
\end{equation}
also have equal selected profiles. Projecting their marks onto the central
position recovers the one-position witness above, so their marked central
diagonals differ. Both examples use three-position simple paths.

\begin{corollary}[Strict marking on sixteen-vertex graphs]
\label{cor:marked-small-strict}
For \(N\geq16\) and \(K\geq1\),
\(\Ker(\mathcal H^{\mathrm{mark}}_{N,K})
\subsetneq\Ker(\mathcal H_{N,K})\).
The compatible continuous approximation classes are strictly nested on
the one-position witness set for every \(K\geq1\), and on the
three-position witness sets for \(K\geq3\).
\end{corollary}
\begin{proof}
The one-position witness has equal selected profiles and different marked
profiles. Apply the kernel and finite-interpolation statements of
\cref{prop:marked-strict,thm:main-expressivity}.
\end{proof}

\paragraph{Encoding cost.}
Occurrence-identifying marks are required by these characterizations;
binary membership flags do not encode path order or repeated positions.
At fixed depth, dense \(\PMarch(\EVT)\) uses \(O(n^3d)\) work per query.
StrikeTwice encodes the graph once and uses \(O(k^3d)\) path-stage work
per query, up to channel-mixing terms.
These costs describe dense implementations at fixed depth and width.

\section{Selection, Joint Cache Reading, and Marking}
\label{app:cache-access-hierarchy}
\subsection{Proof of the expressivity hierarchy}
\label{app:access-hierarchy-proof}

For a profile \(A\), let \(\mathcal F_A\) be the family of predictions
\(f=g\circ A\) with the prescribed occurrence-indexed shape, where \(g\)
is any function on the image of \(A\).

\begin{proof}[Proof of \cref{prop:access-profile-hierarchy}]
\textbf{Reconstruction.}
The unique tuple in \(\Gamma\) with a diagonal color in coordinate \(t\)
is the one for \(u=p_t\); it recovers the \(t\)-th column of \(\Psi\).
This also handles repeated occurrences.
Conversely, a marked diagonal at \(p_1\) determines the multiset of colors
\(b=\bar c^\star_{p_1u}\). Applying \cref{lem:marked-rooted-recovery}
to each \(b\), then forgetting marks, recovers
\((\cstar_{p_1u},\ldots,\cstar_{p_ku})\), preserving the same \(u\)
across coordinates and its multiplicity. Thus \(\Psi^{\mathrm{mark}}\)
determines \(\Gamma\).
For \(k=1\), the terminal multiset at \(\cstar_{p_1p_1}\) recovers
\(\Gamma\) by its first projection. For \(k=2\), use the multiset at
\(\cstar_{p_1p_2}\) and transpose its second component by
\cref{lem:stable-reconstruction}. Hence \(\Psi\) also determines
\(\Gamma\) in these cases.

\textbf{Function families.}
The reconstructions give maps \(r_1,r_2\) such that
\(\Psi=r_1\circ\Gamma\) and
\(\Gamma=r_2\circ\Psi^{\mathrm{mark}}\).
Composing an arbitrary reader with these maps proves
\(\mathcal F_\Psi\subseteq\mathcal F_\Gamma
\subseteq\mathcal F_{\Psi^{\mathrm{mark}}}\).
Equivalently, since these profiles record the output length,
\(\Ker(\mathcal F_A)=\ker(A)\): unequal same-length profiles can be
separated by assigning distinct constant output arrays, and unequal
lengths already give distinct output components. The reverse
reconstruction for \(k\leq2\) gives equality of the selected and joint
families on that subdomain. The same kernel implications also nest
their continuous compatible targets on every compact subset; no
continuity of the reconstruction maps is needed for this formulation.

\textbf{Shared witnesses.}
On \(\mathbb Z_4^2\), the rook graph \(R\) joins distinct vertices sharing
a coordinate; the Shrikhande graph \(S\) has adjacency differences
\(\mathcal A=\{\pm(1,0),\pm(0,1),\pm(1,1)\}\), modulo four.
Use constant vertex features and adjacency pair features. Both graphs have
strongly regular parameters \((16,6,2,2)\), which determine every paired
relation count. Inductively, their unmarked colors have the same three
classes at every round: diagonal \(\mathrm D\), adjacent \(\mathrm A\),
and nonadjacent \(\mathrm N\).

\textbf{One marked position.}
Take \(P=(r)\), \(r=(0,0)\), so the unmarked profiles agree.
Write \(b_u=A_{ru}\) and \(B_{uv}=b_uA_{uv}b_v\). The neighbors of \(r\)
induce two disjoint triangles in \(R\) and a six-cycle in \(S\), giving
\(\operatorname{tr}(B_R^3)=12\) and \(\operatorname{tr}(B_S^3)=0\).
The unique marked intermediate vertex makes \(B_{uv}\) a function of
\(\bar c^1_{uv}\). Then \(\bar c^2_{uv}\) determines
\(q_{uv}=B_{uv}\sum_w B_{uw}B_{wv}\), \(\bar c^3_{ru}\) determines
\(\sum_v q_{vu}\), and \(\bar c^4_{rr}\) determines
\(\sum_{u,v}q_{vu}=\operatorname{tr}(B^3)\).
Thus the marked diagonals differ. This single-mark gain on rook--Shrikhande
is already reported by \citet[Appendix~G, Table~7]{zhou2023relational}.

\textbf{Three-position paths.}
First choose the triangles
\begin{equation}
 P_R=((0,0),(0,1),(0,2)),\qquad
 P_S=((0,0),(0,1),(1,1)).
 \label{eq:cache-selection-joint-witness}
\end{equation}
Their \(\Psi\) profiles agree. Their common-neighbor intersections are
\(\{(0,3)\}\) and \(\varnothing\), respectively, by the adjacency rules.
Hence the multiplicity of \(\mathrm{AAA}\) in \(\Gamma\) differs.
Marked refinement also detects this difference in two rounds: each
\(\bar c^1_{p_su}\) identifies the marked intermediates and recovers all
three adjacency indicators to \(u\); the next diagonal refinement counts
those with all three indicators equal to one.

For the other strict inclusion, take the open paths
\begin{equation}
 Q_R=((0,1),(0,0),(1,0)),\qquad
 Q_S=((0,1),(0,0),(0,3)).
 \label{eq:joint-access-marked-witness}
\end{equation}
Their endpoints are nonadjacent and neither triple has a common neighbor.
The shared degrees, pairwise intersection counts and zero triple
intersection give the same complete \(\Gamma\) histogram:
\begin{center}
\small
\begin{tabular}{@{}lrrrrrr@{}}
\toprule
Pattern & DAN, ADA, NAD & AAN, NAA & ANA & ANN, NNA & NNN & AAA, NAN\\
Multiplicity of each & 1 & 2 & 1 & 2 & 4 & 0\\
\bottomrule
\end{tabular}
\end{center}
Projecting marks onto the central occurrence gives the one-position
witness above, so the marked central diagonals differ by
\cref{lem:marked-rooted-recovery}. Both separating pairs use simple
three-position paths.
Each unequal-profile witness yields a lookup table available to the
finer interface and not the coarser one, proving strictness of the
function-family inclusions as well as of the kernel inclusions.
On the finite witness sets these lookup tables are continuous, so
strictness also holds for the corresponding continuous target classes
on suitable compact subsets.
\end{proof}

\paragraph{Aggregation control and scope.}
For any \(k\geq2\), the indexed two-row histograms
\[
 \bigl(\multiset{(\cstar_{p_su},\cstar_{p_tu}):u\in[n]}\bigr)_{s<t}
\]
have the same kernel as \(\Psi\): each follows from
\(\cstar_{p_sp_t}\) by stable reconstruction and transposition, while its
diagonal entries recover both endpoint diagonals and both pair orientations.
The experiment in \cref{sec:cache-access,app:certified-cache-access}
tests the gain of joint three-row reading over these separate summaries.
The comparison with marked refinement is established by the proof above.

The profiles are compared after stabilization. A binary membership mark cannot replace occurrence
marks. Joint reading accesses \(kn\) pair states instead of \(k^2\), while
preserving the unmarked cache. At fixed depth, the dense marked implementation
uses \(O(n^3d)\) work per query; the reusable cascade encodes the graph once
and uses \(O(k^3d)\) path-stage work per query. These are implementation
costs, not computational lower bounds.

\section{Directional Attention: Strictness, Position, and Memory}
\label{app:extensions}

We first give the contraction identities and strict comparison behind the
directional operator used in the main architecture.  The remaining optional
extensions address positional bias and activation memory: EVT--RoPE changes
the path-stage positional parameterization while preserving its information
boundary, and streaming over the intermediate index evaluates every oriented
triangular head with less activation memory.

\subsection{Directional triangular attention and strictness}
\label{app:directional-strictness}

The aligned pair map remains exactly \cref{eq:almlp}: its receiver
\((i,j)\) reads \(y_{ij}\), not \(y_{ji}\).  Direction is introduced only
through the two indexed views defined in \cref{sec:architecture}.  These views
remain jointly equivariant because transposition commutes with graph
relabeling:
\begin{equation}
 (\sigma\mathbin\cdot Y)^{\leftarrow}_{ij}
 =y_{\sigma^{-1}(j),\sigma^{-1}(i)}
 =(\sigma\mathbin\cdot Y^{\leftarrow})_{ij}.
 \label{eq:transpose-equivariance}
\end{equation}
Thus head-wise direction choices preserve relabeling equivariance while
keeping the two ordered pair states distinct.

\paragraph{Chain and common-endpoint contractions.}
The benefit can be seen with uniform attention weights. Let
\(A\in\R^{m\times m}\) be one scalar channel obtained by a shared linear
projection of the normalized pair states. Use this same projection on both
value branches, set
\(A^{\rightarrow}=A\) and \(A^{\leftarrow}=A^\top\), and set the triangular
logits to zero.  With identity scalar value projections and all \(m\)
intermediate indices valid, a head returns
\(m^{-1}\mathcal T_{\alpha\beta}(A)\), where
\begin{equation}
 \mathcal T_{\alpha\beta}(A)_{ij}
 :=\sum_{r=1}^{m}A^{\alpha}_{ir}A^{\beta}_{rj},
 \qquad \alpha,\beta\in\{\rightarrow,\leftarrow\}.
 \label{eq:four-pair-contractions}
\end{equation}
The four choices are
\begin{center}
\small
\begin{tabularx}{\linewidth}{@{}c>{\raggedright\arraybackslash}p{.34\linewidth}Y@{}}
\toprule
views $(\alpha,\beta)$ & indexed contraction & matrix form and motif\\
\midrule
\(\rightarrow\rightarrow\) & \(\sum_r A_{ir}A_{rj}\)
 & \(A^2\): directed chain \(i\to r\to j\)\\
\(\rightarrow\leftarrow\) & \(\sum_r A_{ir}A_{jr}\)
 & \(AA^\top\): common target \(i\to r\leftarrow j\)\\
\(\leftarrow\rightarrow\) & \(\sum_r A_{ri}A_{rj}\)
 & \(A^\top A\): common source \(i\leftarrow r\to j\)\\
\(\leftarrow\leftarrow\) & \(\sum_r A_{ri}A_{jr}\)
 & \((A^\top)^2\): reverse directed chain\\
\bottomrule
\end{tabularx}
\end{center}
A balanced four-head configuration assigns one head to each orientation,
while the chain-only variant assigns all four heads to the forward chain
in the first row. With padding, uniform attention divides by
the number of valid intermediate indices, rather than the padded size.
The middle two are the bilinear cores of AlphaFold's outgoing and incoming
triangle multiplication, up to its projections, gates, and post-sum
processing \citep{jumper2021alphafold,abramson2024alphafold3}.  AlphaFold's
triangle-attention blocks and the triplet architectures reviewed in
\cref{app:related} share some of these index geometries but use different
scores or values.  EVT assigns the four choices head-wise inside the same
attention layer, whose value depends jointly on both oriented pair states.
With nonzero logits it produces content-dependent,
normalized analogues rather than fixed matrix products.

Softmax produces the average in \cref{eq:four-pair-contractions}, whereas the
displayed matrix products are sums.  For fixed \(m\), the factor \(m\) is
absorbed exactly by the output projection. To recover raw-input matrix
products uniformly over \(m\leq N\), the auxiliary construction uses
a diagonal head to supply \(s=1/m\) from \cref{eq:cardinality-anchor},
then approximates \((u,s)\mapsto u/s\) on compact sets where \(s\geq1/N\).
\Cref{thm:ln-gelu-simulation} transfers this computation to pre-normalized
and GELU. With masks, the source computation needs a valid diagonal
anchor or an explicit count of valid intermediate indices.

\paragraph{Strictness relative to the forward-tokenized compositional restriction.}
Let \(\mathcal E_N^{\mathcal{Y},\mathrm{pure}}\) and
\(\mathcal V_N^{\mathcal{Y},\mathrm{pure}}\) be the rooted ET/EVT families obtained by
restricting every triangular head to \((\rightarrow,\rightarrow)\),
with the same raw stems and ET endpoint tokenizer as above. Their color
comparison uses the forward initialization \cref{eq:pure-2fwl-init}. Write
\(\gamma_N^{\mathcal{Y},\mathrm{pure}}(X,Y;i,j)=c^{R_N,\mathrm{pure}}_{ij}\) for the
corresponding stable rooted color.

\begin{proposition}[Strict directional gain]
\label{prop:directional-strictness}
For unrestricted directed pair inputs and \(N\geq4\),
\begin{equation}
 \Ker(\mathcal E_N^{\mathcal{Y}})=\Ker(\mathcal V_N^{\mathcal{Y}})=\ker(\gamma_N^{\mathcal{Y}})
 \subsetneq
 \ker(\gamma_N^{\mathcal{Y},\mathrm{pure}})
 =\Ker(\mathcal E_N^{\mathcal{Y},\mathrm{pure}})
 =\Ker(\mathcal V_N^{\mathcal{Y},\mathrm{pure}}).
 \label{eq:directional-kernel-strict}
\end{equation}
Moreover, the pure families cannot uniformly approximate the rooted
coordinate map \((A;i,j)\mapsto(AA^\top)_{ij}\) on the finite witness below
with error smaller than \(1/2\), whereas one
\((\rightarrow,\leftarrow)\) head realizes it up to the recoverable factor
\(1/m\).
\end{proposition}
\begin{proof}
The mixed families contain their pure restrictions, giving the non-strict
kernel inclusion.  For the pure kernel equality, omit the transpose part of
\cref{lem:stable-reconstruction}: the terminal canonical multiset over intermediate vertices
still recovers both endpoint colors and the required global multisets.
Consequently, the factorization argument of \cref{eq:evt-factorization}
shows that every pure ET or EVT state is a function of its stable pure
color.  Conversely, the forward setup and zero-logit
\((\rightarrow,\rightarrow)\) basis-histogram construction
\cref{eq:pair-basis,eq:bridge-histogram} realize every pure refinement
in auxiliary coordinates. Their finite separators transfer by
\cref{thm:ln-gelu-simulation}, keeping only the compositional orientation
active. Thus
\(\Ker(\mathcal E_N^{\mathcal{Y},\mathrm{pure}})
=\Ker(\mathcal V_N^{\mathcal{Y},\mathrm{pure}})
=\ker(\gamma_N^{\mathcal{Y},\mathrm{pure}})\).

For strictness, let vertex features be constant and consider
\begin{equation}
 A=
 \begin{pmatrix}
 0&1&1&0\\
 1&0&0&1\\
 0&0&0&1\\
 0&0&1&0
 \end{pmatrix}.
 \label{eq:direction-witness}
\end{equation}
Starting from \((A_{ij},\one_{\{i=j\}})\), compositional \(2\)-FWL reaches the
stable class matrix
\begin{equation}
 \begin{pmatrix}
 0&1&2&3\\
 1&0&3&2\\
 4&4&5&6\\
 4&4&6&5
 \end{pmatrix}.
 \label{eq:direction-witness-colors}
\end{equation}
Indeed, direct enumeration of the ordered pairs of initial colors over all
intermediate vertices gives these seven classes after one refinement.  If
\(H_a\) denotes the multiset over intermediate vertices of any entry labeled
\(a\), then
\begin{equation}
\begin{aligned}
 H_0&=\multiset{(0,0),(1,1),(2,4),(3,4)},\\
 H_1&=\multiset{(0,1),(1,0),(2,4),(3,4)},\\
 H_2&=\multiset{(0,2),(1,3),(2,5),(3,6)},\\
 H_3&=\multiset{(0,3),(1,2),(2,6),(3,5)},\\
 H_4&=\multiset{(4,0),(4,1),(5,4),(6,4)},\\
 H_5&=\multiset{(4,2),(4,3),(5,5),(6,6)},\\
 H_6&=\multiset{(4,2),(4,3),(5,6),(6,5)}.
\end{aligned}
\label{eq:direction-witness-histograms}
\end{equation}
These multisets are constant within each displayed class and pairwise
distinct, so one further refinement preserves exactly the same partition;
it is stable.
Hence rooted pairs \((3,1)\) and \((3,2)\) are indistinguishable to every
pure graph-stage ET or EVT, while
\begin{equation}
 (AA^\top)_{31}=0,
 \qquad
 (AA^\top)_{32}=1.
 \label{eq:direction-witness-target}
\end{equation}
Every such pure pair readout therefore has entrywise uniform error at
least \(1/2\) on the rooted two-point set
\(\{(A;3,1),(A;3,2)\}\) for
\((A;i,j)\mapsto(AA^\top)_{ij}\).  The numerical class labels
in \cref{eq:direction-witness-colors} are arbitrary; only their equality
pattern matters. For an exact normalized realization on this binary
witness, choose the affine pair lift \([A_{ij};1-A_{ij};0]\), padded by
zeros if needed. Its coordinate sum and squared norm are both one for
every pair, so its LayerNorm mean and variance, and its RMSNorm squared
norm, are constant. Subtracting
the normalized zero coordinate and rescaling recovers \(A_{ij}\) by a
linear value projection. For endpoint-lifted ET, ignore the additional
endpoint inputs in this parameter choice. A zero-logit
\((\rightarrow,\leftarrow)\) head therefore returns
\(\tfrac14AA^\top\) exactly; its output projection multiplies by four
and writes to a fresh zero coordinate. All later increments are zero.  The
mixed family contains the pure family by assigning every head
\((\rightarrow,\rightarrow)\), so this
witness makes the inclusion strict.  The approximation lower bound and mixed
realization follow from
\cref{eq:direction-witness-target,eq:four-pair-contractions,eq:cardinality-anchor}.
\end{proof}
The obstruction disappears on symmetric inputs, where \(A^\top=A\), and more
generally whenever the reverse atomic pair type is determined by the forward
one.  This rooted witness does not by itself prove strictness after a
restricted \(P\times P\) selection; that additionally depends on which ordered
occurrences the path selects.

\paragraph{Numerical check on the directed witness.}
Assigning one head to each orientation gives EVT access to
all four directed pair contractions. On the witness of
\cref{prop:directional-strictness}, chain-only EVT leaves the two rooted-pair
states equal up to numerical precision (maximum gap \(2.2\times10^{-16}\)
over four random initializations), whereas balanced EVT separates them in
every run (minimum gap \(5.2\times10^{-2}\)). This one-layer probe checks
the predicted separation on the directed witness.

\subsection{Endpoint-aware EVT--RoPE}
\label{app:evt-rope}

Only the path EVT has semantic coordinates: vertex tokens are occurrences
and pair tokens have two ordered occurrence endpoints.  We therefore apply
RoPE to the second EVT, never to arbitrary graph labels in the cached stage.

For even rotary width \(d_r\), let
\begin{equation}
 R_a=\bigoplus_{h=1}^{d_r/2}
 \begin{bmatrix}
  \cos(\omega_ha)&-\sin(\omega_ha)\\
  \sin(\omega_ha)& \cos(\omega_ha)
 \end{bmatrix},
 \qquad R_a^\top R_b=R_{b-a}.
 \label{eq:rope-rotation}
\end{equation}
Give \(\widehat x_s\) one port \(\mathsf V\) at \(s\), and \(\widehat y_{st}\) a source port
\(\mathsf S\) at \(s\) and target port \(\mathsf T\) at \(t\).  For an
oriented view, these are the endpoints of the pair state actually read:
\begin{equation}
 (\operatorname{pos}_{\mathsf S}(\widehat y^\alpha_{st}),
  \operatorname{pos}_{\mathsf T}(\widehat y^\alpha_{st}))
 =\begin{cases}(s,t),&\alpha=\rightarrow,\\
                (t,s),&\alpha=\leftarrow.
   \end{cases}
 \label{eq:oriented-rope-ports}
\end{equation}
For attention inputs \(z\in\R^{d_z}\) and \(z'\in\R^{d_{z'}}\),
each attention head and type has learned rotary projections for each
port pairing \((a,b)\):
\(q_z^{ab}:=W_Q^{ab}z\in\R^{d_r}\) and
\(k_{z'}^{ab}:=W_K^{ab}z'\in\R^{d_r}\), with
\(W_Q^{ab}\in\R^{d_r\times d_z}\) and
\(W_K^{ab}\in\R^{d_r\times d_{z'}}\).
These projections have independent parameters for each head, attention
type and port pairing; head and type indices are omitted.
For a set \(\mathcal A\) of query--key port pairings, add
\begin{equation}
 \ell^{\mathrm{rot}}_{\mathcal A}(z,z')
 =\frac1{\sqrt{|\mathcal A|d_r}}
 \sum_{(a,b)\in\mathcal A}
 \left\langle
  R_{\operatorname{pos}_a(z)}q_z^{ab},
  R_{\operatorname{pos}_b(z')}k_{z'}^{ab}
 \right\rangle
 \label{eq:evt-rope}
\end{equation}
to the plain content score, with a trainable gate.  The attention types use:
\begin{center}
\small
\begin{tabularx}{\linewidth}{@{}>{\raggedright\arraybackslash}p{.23\linewidth}
 >{\raggedright\arraybackslash}p{.48\linewidth}Y@{}}
\toprule
attention & port pairings \(\mathcal A\) & offsets\\
\midrule
\(\widehat x_s\leftarrow\widehat x_u\) & \((\mathsf V,\mathsf V)\) & \(u-s\)\\
\(\widehat x_s\leftarrow\widehat y_{uv}\) & \((\mathsf V,\mathsf S),(\mathsf V,\mathsf T)\)
 & \(u-s,v-s\)\\
\(\widehat y_{st}\leftarrow\widehat x_u\) & \((\mathsf S,\mathsf V),(\mathsf T,\mathsf V)\)
 & \(u-s,u-t\)\\
\(\widehat y^\alpha_{sr},\widehat y^\beta_{rt}\) in
\(\TriAtt^{\alpha,\beta}\) &
 \((\mathsf S,\mathsf S),(\mathsf T,\mathsf T),
   (\mathsf S,\mathsf T),(\mathsf T,\mathsf S)\)
 & differences between the actual oriented endpoints in
   \cref{eq:oriented-rope-ports}\\
\bottomrule
\end{tabularx}
\end{center}
For \((\alpha,\beta)=(\rightarrow,\rightarrow)\), the four triangular offsets
reduce to \(r-s,t-r,t-s,0\); the other variants follow by swapping the
corresponding source and target positions.  Only queries and keys are rotated;
values and residual states stay in a common frame.  Source and target
projections remain untied to preserve orientation.  This is the endpoint
analogue of ordinary RoPE \citep{su2021roformer}.

The exact theorem keeps the selector's injective absolute tags \(\phi(s)\) and
\(\phi(s)\concat\phi(t)\). RoPE is
an additional relative bias, not a replacement: constant values do not reveal
absolute positions through rotated logits alone.  Let
\(\mathcal H^{\mathrm{RoPE}}_{N,K}\) extend the family of
\cref{def:striketwice-family} by adding \cref{eq:evt-rope} to every attention
in the second EVT. The selector and its freely learned affine bridge are
unchanged. Rotary positions are deterministic occurrence-index metadata;
they need not be preserved in hidden feature channels. Every rotary gate
can be set exactly to zero, which recovers each original network with
the same parameters and therefore gives
\(\mathcal H_{N,K}\subseteq\mathcal H^{\mathrm{RoPE}}_{N,K}\), hence
\[
 \Ker(\mathcal H^{\mathrm{RoPE}}_{N,K})
 \subseteq \Ker(\mathcal H_{N,K})=\ker(\Psi_{N,K}).
\]
Conversely, the rotary matrices are deterministic functions of the
occurrence indices. Induction through the second stage therefore shows
that every RoPE state remains a function of the positioned selected states,
so every member is profile compatible and
\[
 \ker(\Psi_{N,K})
 \subseteq\Ker(\mathcal H^{\mathrm{RoPE}}_{N,K}).
\]
Thus the kernels are equal.  On every compact
\(S\subset\mathcal D_{\leq N,\leq K}\), family containment and
\cref{thm:main-expressivity} give
\[
 \begin{aligned}
 \mathcal C_\Psi(S)
 &=\overline{\{H|_S:H\in\mathcal H_{N,K}\}}^{\|\cdot\|_{\infty,S}}\\
 &\subseteq
 \overline{\{H|_S:H\in\mathcal H^{\mathrm{RoPE}}_{N,K}\}}
 ^{\|\cdot\|_{\infty,S}}.
 \end{aligned}
\]
Every RoPE map is continuous and profile compatible, and these properties
are preserved by uniform limits; hence the reverse inclusion holds as well.
EVT--RoPE consequently has the same proved common kernel and relative
uniform closure, but a different parameterization.

\subsection{Flash-style triangular attention}
\label{app:flash-triatt}

Triangular attention is cubic in arithmetic but need not materialize cubic
activations.  Streaming over blocks of the intermediate index evaluates the
exact same map in the spirit of FlashAttention \citep{dao2022flash}.

Suppress batch indices, fix one head with direction pair
\((\alpha,\beta)\), and let \(m=n\) or \(m=k\).  Its projected oriented
views are \(Q^\alpha_{ir},V^{\mathrm L,\alpha}_{ir}\) and
\(K^\beta_{rj},V^{\mathrm R,\beta}_{rj}\), with query/key dimension
\(d_a\) and value dimension \(d_v\) per head. We define
\begin{align}
 S_{irj}&=\langle Q^\alpha_{ir},K^\beta_{rj}\rangle/\sqrt{d_a},&
 P_{irj}&=\frac{e^{S_{irj}}}{\sum_u e^{S_{iuj}}},\nonumber\\
 V^\triangle_{irj}&=V^{\mathrm L,\alpha}_{ir}
                    \odot V^{\mathrm R,\beta}_{rj},&
 O_{ij}&=\sum_rP_{irj}V^\triangle_{irj}.
 \label{eq:flash-map}
\end{align}

\begin{proposition}[Streaming triangular attention]
\label{prop:flash-triatt}
Partition the reduction axis indexed by \(r\) into blocks of size at most
\(b\).  In exact
arithmetic, the forward map \cref{eq:flash-map} can be evaluated without a
globally materialized tensor indexed by \((i,r,j)\).  For batch size \(B\)
and \(H\) heads, its attention-core working activation memory, measured in
stored scalar entries, is
\begin{equation}
 O\!\left(BHm^2(d_a+d_v)+BHbm^2(d_a+d_v)\right),
 \label{eq:flash-memory}
\end{equation}
where the first term is quadratic state resident in high-bandwidth memory
(HBM) and the second is a conservative bound on temporary storage for the
current block of intermediate indices.  At fixed numerical precision, the
same asymptotic bound holds in bytes.  Thus hardware-bounded \(b\) gives
\(O(BHm^2(d_a+d_v))\) memory.  The arithmetic remains
\(\Theta(BHm^3(d_a+d_v))\).  Reverse-mode training with tile recomputation
admits the same asymptotic attention-core working-memory bound.
\end{proposition}
\begin{proof}
For each output \((i,j)\), retain a running
maximum \(\mu\), normalizer \(z\), and vector accumulator \(a\), initialized
as \(\mu=-\infty\), \(z=0\), and \(a=0\).  For a new block \(U\), set
\begin{align}
 \mu^+&=\max\{\mu,\max_{r\in U}S_{irj}\},&
 \rho&=e^{\mu-\mu^+},\nonumber\\
 z^+&=\rho z+\sum_{r\in U}e^{S_{irj}-\mu^+},&
 a^+&=\rho a+\sum_{r\in U}e^{S_{irj}-\mu^+}V^\triangle_{irj}.
 \label{eq:online-softmax}
\end{align}
Induction over blocks gives \(O_{ij}=a/z\).  Only the current block of scores
and joint values, together with quadratic output accumulators, is live;
every intermediate index is still visited for every output pair.  Globally
resident oriented projections
\(Q^\alpha,K^\beta,V^{\mathrm L,\alpha},V^{\mathrm R,\beta}\), outputs,
and softmax statistics give the first term in
\cref{eq:flash-memory}.  The temporary block, with at most \(b\) intermediate
indices for all output pairs, gives the second; a physical kernel further
tiles the \((i,j)\) axes, so only its output tile occupies SRAM or registers
and no cubic tensor is written to HBM.

For training, save the four projections, \(O\), and
\(\operatorname{LSE}_{ij}=\mu_{ij}+\log z_{ij}\), all quadratic in \(m\),
and recompute each score and probability tile during the backward pass.  With
upstream \(G_{ij}\), set
\begin{equation}
 D_{irj}=P_{irj}
 \bigl(\langle G_{ij},V^\triangle_{irj}\rangle
       -\langle G_{ij},O_{ij}\rangle\bigr).
 \label{eq:flash-dscore}
\end{equation}
Then
\begin{align}
 \partial Q^\alpha_{ir}&=d_a^{-1/2}\sum_jD_{irj}K^\beta_{rj},&
 \partial K^\beta_{rj}&=d_a^{-1/2}\sum_iD_{irj}Q^\alpha_{ir},\nonumber\\
 \partial V^{\mathrm L,\alpha}_{ir}
 &=\sum_jP_{irj}G_{ij}\odot V^{\mathrm R,\beta}_{rj},&
 \partial V^{\mathrm R,\beta}_{rj}
 &=\sum_iP_{irj}G_{ij}\odot V^{\mathrm L,\alpha}_{ir}.
 \label{eq:flash-backward}
\end{align}
Separate passes accumulate the left derivatives by \((i,r)\)
and the right derivatives by \((r,j)\), assigning each derivative entry to
one accumulation pass.  For a transposed view, the final
scatter simply swaps the two physical tensor indices; it does not alter the
live-memory order.  These gradient accumulators are quadratic; the recomputed
\(P,D\) tiles are released before advancing, so no cubic tensor is saved and
global atomic updates can be avoided by accumulating each entry in its
assigned pass.  The bounds in
\cref{prop:flash-triatt} cover
only the attention core.  They exclude parameter and optimizer state, as well
as residual and non-attention MLP activations; projection tensors, outputs,
LSE values, and their attention-core gradients are included.
\end{proof}

Unlike standard scaled dot-product attention, the value depends jointly on
both oriented pair states in \cref{eq:flash-map}. The proved claim is an exact
quadratic attention-core working-memory schedule under hardware-bounded
tiling; cubic arithmetic remains. No empirical kernel speedup is claimed.

\section{Extended Related Work}
\label{app:related}

The central distinction in this paper is between the information available
to a model and the computation used to expose it.  We position both aspects
relative to graph Transformers, path models, and conditional architectures.

\paragraph{Graph Transformers and higher-order expressivity.}
Graphormer, GraphGPS, TokenGT, and edge-augmented Transformers inject graph
structure through attention biases, hybrid message passing, or vertex/edge
tokens \citep{ying2021graphormer,rampasek2022graphgps,kim2022tokengt,
hussain2021egt}.  Higher-order variants instead move part or all of the
computation from vertices to ordered pairs or triplets.

\paragraph{Triangular and triplet-based pair interactions.}
Non-softmax pair layers already aggregate two pair states through a third index.
PPGN interleaves pointwise MLPs with featurewise matrix multiplication, while
PPGN++ also supplies the transposed pair tensor; coordinate transposition is
part of the general basis of second-order equivariant linear maps
\citep{maron2019ppgn,puny2023equivariant,maron2019equivariant}.  AlphaFold's
outgoing and incoming triangle-multiplication blocks use the complementary
common-target and common-source contractions, and Uni-Mol+ combines both in
one pair update
\citep{jumper2021alphafold,abramson2024alphafold3,lu2024unimolplus}.

Softmax-normalized variants make different choices within the same family of
three-index geometries.  Edge Transformers (ETs) use the compositional pair states
\((i,r)\) and \((r,j)\) \citep{bergen2021}; their graph-distinguishing
boundary is analyzed through \(2\)-FWL \citep{muller2024}.  Separately,
\citet{kozachinskiy2025strassen} prove lower bounds for one-layer standard-softmax
Transformers on several compositional tasks and, for quotient binary-relation
composition, for one-layer mixed standard/triangular-attention Transformers
with subpolynomial head counts, embedding dimension, and output-MLP size;
these task-specific bounds are not ET-specific. AlphaFold's
starting- and ending-vertex triangle-attention blocks use complementary
common-endpoint incidence patterns.  In each block, the receiver supplies the
query, one of the other two pair states supplies key and value, and the
remaining pair state supplies a scalar bias
\citep{jumper2021alphafold,abramson2024alphafold3}.  Triplet Edge Attention
scores the receiver \((i,j)\) jointly with \((i,r)\), \((j,r)\), and
vertex/global features, then aggregates a value from \((i,r)\), using the
common-target geometry
\citep{jung2023tripletedge}.  The Triplet Graph Transformer couples
complementary common-target and common-source patterns, using
\((i,r),(j,r)\) and \((r,i),(r,j)\), respectively.  In its attention form,
one of these pair states supplies the value while the other modulates its
score and gate; its cheaper aggregation uses softmax-normalized,
sigmoid-gated analogues of the corresponding bilinear contractions
\citep{hussain2024triplet}. In these attention variants, each value comes
from one pair state; EVT computes values jointly from both.

Our four oriented variants provide common notation for the chain,
common-target, common-source, and reverse-chain index patterns, without
identifying the architectures' complete score and value maps.  EVT
distributes these variants across heads and couples the resulting pair states
to vertices.  Relative to the forward-tokenized compositional ET family
formalized in this paper, access to reversed pair states yields a strictly finer rooted-pair
common kernel on directed inputs.  The path theorem then characterizes the
profile obtained by selecting the resulting stable colors at path
occurrences
(\cref{sec:architecture,sec:theory,app:evt-et-comparison,app:directional-strictness}).

\paragraph{Coupled vertex--relation representations.}
Architectures with vertex and edge states are common in relational reasoning,
molecular modeling, and routing.  RouteNet, for example, couples paths and
links, while edge-aware graph Transformers update several entity types
\citep{rusek2019routenet,hussain2021egt}. EVT retains all ordered pairs and
supports native vertex--pair communication. TT applies this typed
block on both the ambient graph and the selected occurrence graph; other
instances vary the path processor while retaining the same reusable interface.

\paragraph{Graph-conditioned sequences and paths.}
Cai and Lam's graph-to-sequence Transformer, WikiGraphs, and CERES condition
sequence generation or language modeling on graph representations \citep{cai2020graph2seq,
wang2021wikigraphs,feng2022ceres}.  PathNN, pathGCN, and Gretel use paths as
aggregation objects or prediction targets \citep{michel2023pathnn,
eliasof2022pathgcn,cordonnier2019gretel}.  Our path is an ordered index
sequence that may repeat vertices and indexes the output itself; in graph
applications it is typically adjacency compliant.  The complete pair
selection preserves occurrence-level relations while keeping the upstream
graph memory independent of the path.

\paragraph{Target labeling and subgraph expressivity.}
The labeling trick conditions graph representations on target sets and
ordered structures \citep{wang2025labeling}. The tuple labels of
\citet{zhou2023relational} record every occurrence position, including
repeated vertices; their framework also aggregates over tuples to obtain
graph representations. \citet{xu2024higherorder} compare tuple representations
obtained by varying the number of marks and the order of the network.
Subgraph WL analyses characterize the effects of marking and pooling order
\citep{zhang2023subgraph}, while \citet{feng2023rethinking} extend FWL through
joint tuple aggregation during refinement. The gain of marking and its
per-query recomputation cost are established \citep{wang2025labeling}; the
single-mark rook--Shrikhande separation is also known
\citep[Appendix~G]{zhou2023relational}.

Building on these constructions, we characterize an ordered graph-to-path
interface and compare three precise information profiles: complete
\(P\times P\) selection, invariant joint \(P\times V\) reading of the same
stable unmarked pair cache, and occurrence-marked refinement. To our
knowledge, this exact comparison has not previously been established for
the ordered pair-cache interface considered here. It identifies which
distinctions can be recovered by changing the readout while preserving
graph reuse, and which require richer information. The assumption of maximally expressive vertex representations in \citet[Theorem~21]{wang2025labeling} is stronger than a
\(2\)-FWL bound; that completeness result alone does not give our
ET/EVT characterization.

\paragraph{Genomic context and cache-then-query models.}
DNA language models learn from linear sequences
\citep{nguyen2023hyenadna,benegas2023gpn,dallatorre2025nucleotide}, whereas
pangenomes and pantranscriptomes represent multiple biological sequences in
one graph \citep{garrison2018variation,liao2023pangenome,
sibbesen2023pantranscriptome}.  The many-paths-per-graph regime makes a
cache-then-query decomposition natural.  It is analogous to an
encoder--decoder system in which the source memory is reusable, but our
decoder is relational and aligned with both selected vertices and pairs.

\paragraph{Symmetry and relative universality.}
Graph-network universality and relative approximation are established
\citep{keriven2019universal,maron2019universality,azizian2021}, including
shared parameters across bounded graph sizes. Transformer approximation
with positional information is also studied by
\citet{yun2020transformeruniversality}; its \(L^p\) guarantees do not supply
our uniform-norm conclusion. We identify the relevant information explicitly
as a selected \(2\)-FWL profile. The architecture-specific work establishes
exact internal feature-bank compilation and a shared readout over bounded
path lengths, allowing the relative approximation argument to apply.

\end{document}